\documentclass[10pt,draft,doublespacing]{book}
\usepackage{makeidx}
\usepackage{multicol}
\usepackage{multirow}
\usepackage{footmisc}

\usepackage{amsmath}
\usepackage{amssymb}
\usepackage{algorithm}
\usepackage{algorithmic}
\usepackage{array}
\usepackage[usenames]{color}
\usepackage[demo]{graphicx}
\usepackage{setspace}
\usepackage{subfigure}
\usepackage{caption}
\usepackage{tabularx}
\usepackage{url}
\usepackage{fancyhdr}
\usepackage{titlesec}
\definecolor{mycolor1}{rgb}{0.5, 0.5, 0.5}

\renewcommand{\sectionmark}[1]{\markright{\normalfont \sffamily \S\thesection}}
\renewcommand{\thechapter}{\arabic{chapter}}
\titleformat{\chapter}[display]
{\bfseries\Large  \thispagestyle{empty} } {\centering
\textcolor{mycolor1}{\Huge\thechapter}} {1ex} {\centering}

\renewcommand{\thepart}{\sffamily \Roman{part}}
\titleformat{\part}[display]
{\bfseries\Huge \thispagestyle{empty} } {\centering
\textcolor{mycolor1}{\thepart}} {1ex} {\sffamily \centering}

\newcommand{\abstract}[1]{{\Large \centering \bf #1\\\vspace{5pt}}}

\newcolumntype{V}{>{$\vcenter\bgroup\hbox\bgroup}c<{\egroup\egroup$}}

\newcommand{\diag}[0]{\textrm{diag}}
\newcommand{\sign}[0]{\textrm{sign}}
\def\Hline{\noalign{\hrule height 4\arrayrulewidth}}

\begin{document}
\title{\bf \huge \hspace{-19pt} Automatic Face Recognition from Video\\~\\
\large  \color{red}
Images removed to fit arXive's size requirements \\ For complete version see \url{https://www.researchgate.net/profile/Ognjen_Arandjelovic/}}

\author{
\begin{tabular}{c}
\LARGE Ognjen Arandjelovi\'c\vspace{10pt}\\
\LARGE Trinity College\\
\vspace{130pt}\\
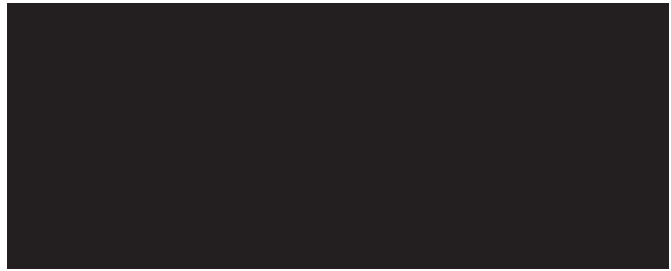
\includegraphics[width=250pt]{clogo1.ps}\\\\\\
\Large This dissertation is submitted for\\
\Large the degree of Doctor of Philosophy
\end{tabular}
}

\date{}


\maketitle \fancyhead[RO,LE]{} \fancyhead[RE,LO]{}
\newpage

\cleardoublepage

\onehalfspacing

\graphicspath{{./00pre/}}
\abstract{Abstract}

\begin{spacing}{1.65}

The objective of this work is to automatically recognize faces from
video sequences in a realistic, unconstrained setup in which
illumination conditions are extreme and greatly changing, viewpoint
and user motion pattern have a wide variability, and video input is
of low quality. At the centre of focus are face appearance
manifolds: this thesis presents a significant advance of their
understanding and application in the sphere of face recognition. The
two main contributions are the \textit{Generic Shape-Illumination
Manifold} recognition algorithm and the \textit{Anisotropic Manifold
Space} clustering.

The \textit{Generic Shape-Illumination Manifold} algorithm shows how
video sequences of unconstrained head motion can be reliably
compared in the presence of greatly changing imaging conditions. Our
approach consists of combining \textit{a priori} domain-specific
knowledge in the form of a photometric model of image formation,
with a statistical model of generic face appearance variation. One
of the key ideas is the \emph{reillumination} algorithm which takes
two sequences of faces and produces a third, synthetic one, that
contains the same poses as the first in the illumination of the
second.

The \textit{Anisotropic Manifold Space} clustering algorithm is
proposed to automatically determine the cast of a feature-length
film, without any dataset-specific training information. The method
is based on modelling coherence of dissimilarities between
appearance manifolds: it is shown how inter- and intra-personal
similarities can be exploited by mapping each manifold into a single
point in the \textit{manifold space}. This concept allows for a
useful interpretation of classical clustering approaches, which
highlights their limitations. A superior method is proposed that
uses a mixture-based generative model to hierarchically grow class
boundaries corresponding to different individuals.

The \textit{Generic Shape-Illumination Manifold} is evaluated on a
large data corpus acquired in real-world conditions and its
performance is shown to greatly exceed that of state-of-the-art
methods in the literature and the best performing commercial
software. Empirical evaluation of the \textit{Anisotropic Manifold
Space} clustering on a popular situation comedy is also described
with excellent preliminary results.

\end{spacing}

\cleardoublepage

\tableofcontents

\listoffigures

\listoftables

\singlespacing
\newpage
\doublespacing \onehalfspacing

\fancyhead[RO,LE]{\textcolor{Gray}\leftmark}
\fancyhead[RE,LO]{\textcolor{Gray}{\S\thesection}}
\fancyfoot[LE,RO]{\hrule\vspace{10pt}~\thepage~} \fancyfoot[CE,CO]{}

\part{Preliminaries}

\graphicspath{{./01intro/}}
\chapter{Introduction}
\label{Chp: Intro}
\begin{center}
  \vspace{-20pt}
  \footnotesize
  \framebox{\includegraphics[width=0.85\textwidth]{title_img.eps}}\\
  Michelangelo. \textit{Genesis}\\
  1509-1512, Fresco\\
  Sistine Chapel, Vatican
\end{center}

\cleardoublepage

This chapter sets up the main premises of the thesis. We start by
defining the problem of automatic face recognition and explain why
this is an extremely challenging task, followed by an overview of
its practical significance. An argument for the advantages of face
recognition, in the context of other biometrics, is presented.
Finally, the main assumptions and claims of the thesis are stated,
followed by a synopsis of the remaining chapters.

\section{Problem statement}
This thesis addresses the problem of automatic face recognition from
video in realistic imaging conditions. While in some previous work
the term ``face recognition'' has been used for any face-based
biometric identification, we will operationally define face
recognition as classification of persons by their identity using
images acquired in the visible electromagnetic spectrum.

\begin{figure}
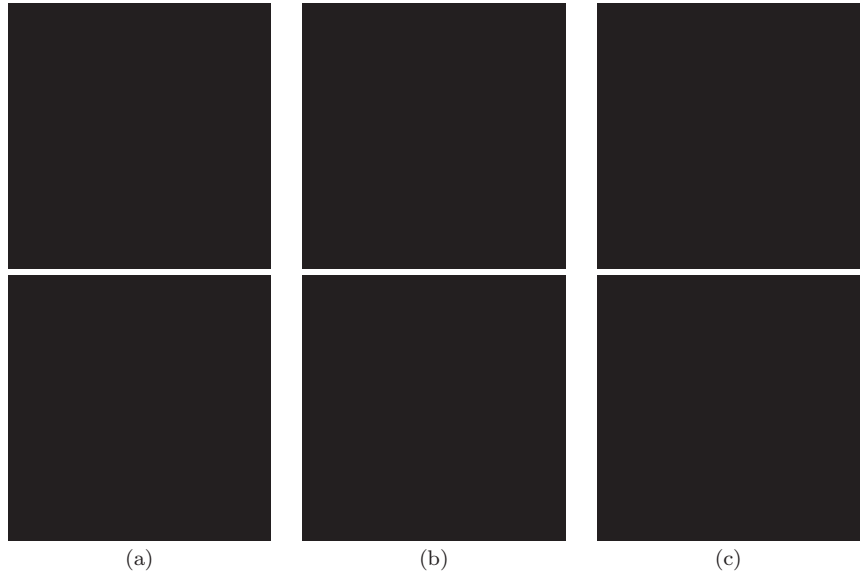

  \centering
  \footnotesize
  \begin{tabular}{ccc}
    \includegraphics[width=0.25\textwidth]{CVPR2005_lighting_01.ps} &
    \includegraphics[width=0.25\textwidth]{CVPR2005_pose_01.ps} &
    \includegraphics[width=0.25\textwidth]{CVPR2005_expression_01.ps}\\
    \includegraphics[width=0.25\textwidth]{CVPR2005_lighting_02.ps} &
    \includegraphics[width=0.25\textwidth]{CVPR2005_pose_02.ps} &
    \includegraphics[width=0.25\textwidth]{CVPR2005_expression_02.ps}\\
    (a) & (b) & (c) \\
  \end{tabular}
  \caption[Effects of imaging conditions.]{\it The effects of imaging conditions -- illumination (a),
            pose (b) and expression (c) -- on the appearance of a face
            are dramatic and present the main difficulty to automatic
            face recognition methods.}
            \label{Fig: Challenges}
\end{figure}

Humans seemingly effortlessly recognize faces on a daily basis, yet
the same task has so far proved to be of formidable difficulty to
automatic methods \cite{BBC2004, Bost2002, ChanSeetBurt+2003,
ZhaoChelPhilRose2004}. A number of factors other than one's identity
influence the way an imaged face appears. Lighting conditions, and
especially light angle, can drastically change the appearance of a
face. Facial expressions, including closed or partially closed eyes,
also complicate the problem, just as head pose and scale changes do.
Background clutter and partial occlusions, be they artifacts in
front of the face (such as glasses), or resulting from hair style
change, growing a beard or a moustache all pose problems to
automatic methods. Invariance to the aforementioned factors is a
major research challenge, see Figure~\ref{Fig: Challenges}.

\section{Applications}
\begin{figure}
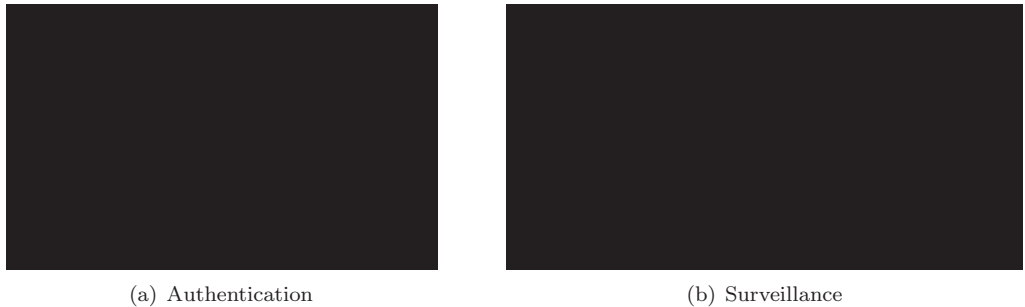

  \centering
  \subfigure[Authentication]{\includegraphics[width=0.41\textwidth]{auth.eps}}
  \hspace{0.05\textwidth}
  \subfigure[Surveillance]{\includegraphics[width=0.49\textwidth]{cctv.eps}}
  \caption[Security applications.]{ \it The two most common paradigms for security applications of
                automatic face recognition are (a) authentication and (b)
                surveillance. It is important to note the drastically different
                data acquisition conditions. In the authentication setup,
                the imaging conditions are typically milder, more control
                over the illumination setup can be exercised and the user
                can be asked to cooperate to a degree, for example by
                performing head motion. In surveillance environment, the
                viewpoint and illumination are largely uncontrolled and
                often extreme, face scale can have a large range and image
                quality is poor.  }
  \label{Fig: Security}
\end{figure}

The most popularized use of automatic face recognition is in a broad
range of security applications. These can be typically categorized
under either (i) voluntary authentication, such as for the purpose
of accessing a building or a computer system, or for passport
control, or (ii) surveillance, for example for identifying known
criminals at airports or offenders from CCTV footage, see
Figure~\ref{Fig: Security}.

In addition to its security uses, the rapid technological
development we are experiencing has created a range of novel
promising applications for face recognition. Mobile devices, such as
PDAs and mobile phones with cameras, together with freely available
software for video-conferencing over the Internet, are examples of
technologies that manufacturers are trying to make ``aware'' of
their environment for the purpose of easier, more intuitive
interaction with the human user. Cheap and readily available imaging
devices, such as cameras and camcorders, and storage equipment (such
as DVDs, flash memory and HDDs) have also created a problem in
organizing large volumes of visual data. Given that humans (and
faces) are often at the centre of interest in images and videos,
face recognition can be employed for content-based retrieval and
organization, or even synthesis of imagery, see Figure~\ref{Fig:
Org}.

\begin{figure}
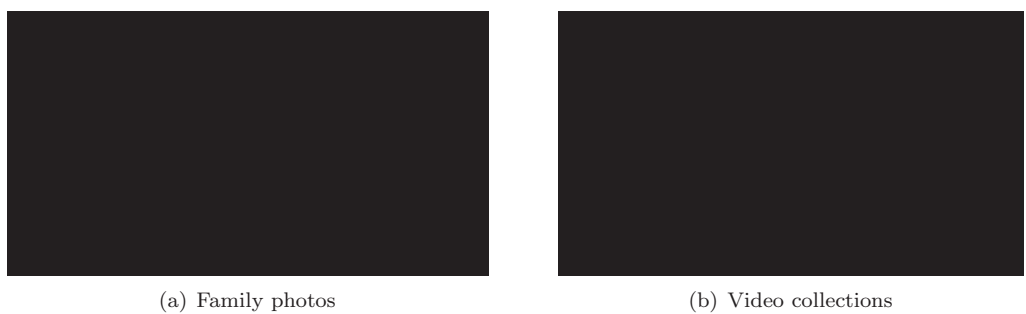

  \centering
  \subfigure[Family photos]{\includegraphics[width=0.458\textwidth]{photo1.eps}}
  \hspace{0.05\textwidth}
  \subfigure[Video collections]{\includegraphics[width=0.442\textwidth]{frame1.eps}}
  \caption[Content-based data organization applications.]
          { \it Data organization applications of face recognition are rapidly
                gaining in importance. Automatic organization and retrieval of
                (a) amateur photographs or (b) video collections are some of the
                examples. Face recognition is in both cases extremely difficult,
                with large and uncontrolled variations in pose, illumination and
                expression, further complicated by image clutter and frequent
                partial occlusions of faces. }
  \label{Fig: Org}
  \vspace{6pt}\hrule
\end{figure}

\begin{figure*}
  \centering
  \rotatebox{270}{\includegraphics[width=0.5\textwidth]{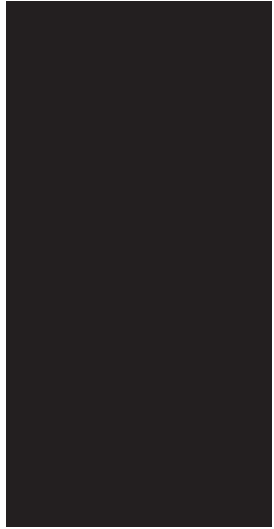}}

  \caption[Face recognition revenues.]{ \it Total face recognition market revenues in the period
                2002--2007 (\$m) \cite{Int_}. }
  \label{Fig: Revenue}
  \vspace{6pt}\hrule
\end{figure*}

The increasing commercial interest in face recognition technology is
well witnessed by the trend of the relevant market revenues, as
shown in Figure~\ref{Fig: Revenue}.

\section{A case for face recognition from video}
Over the years, a great number of biometrics have been proposed and
found their use for the task of human identification. Examples are
fingerprint \cite{JainHongPank+1997, MaltMaioJain+2003}, iris
\cite{Daug1992, WildAsmuGreeHsu1994} or retinal scan-based methods.
Some of these have achieved impressively high identification rates
\cite{Grae2003} (e.g.\ retinal scan $\sim10^{-7}$ error rate
\cite{NCSC_}).

However, face recognition has a few distinct advantages. In many
cases, face information may be the only cue available, such as in
the increasingly important content-based multimedia retrieval
applications \cite{AranCipo2006c, AranZiss2005, AranZiss2006,
LeSatoHoul2006, SiviZiss2003}. In others, such as in some
surveillance environments, bad imaging conditions render any single
cue insufficient and a fusion of several may be needed (e.g.\ see
\cite{BrunFala1995, BrunFalaPogg+1995, SingGyaoBebi+2004}). Even for
access-control applications, when more reliable cues are available
\cite{NCSC_}, face recognition has the attractive property of being
very intuitive to humans as well as non-invasive, making it readily
acceptable by wider audiences. Finally, face recognition does not
require user cooperation.

\paragraph{Video.} The nature of many practical applications is such that more than
a single image of a face is available. In surveillance, for example,
the face can be tracked to provide a temporal sequence of a moving
face. For access-control use of face recognition the user may be
assumed to be cooperative and hence be instructed to move in front
of a fixed camera. This is important as a number of technical
advantages of using video exist: person-specific dynamics can be
learnt, or more effective face representations be obtained (e.g.\
super-resolution images or a 3D face model) than in the single-shot
recognition setup. Regardless of the way in which multiple images of
a face are acquired, this abundance of information can be used to
achieve greater robustness of face recognition by resolving some of
the inherent ambiguities (shape, texture, illumination etc.) of
single-shot recognition.

\section{Basic premises and synopsis}
The first major premise of work in this thesis is:
\begin{description}
  \item[Premise 1] \textit{Neither purely discriminative nor purely generative
    approaches are very successful for face recognition in realistic
    imaging conditions.}
\end{description}
Hence, the development of an algorithm that in a principle manner
combines (i) a generative model of well-understood stages of image
formation with, (ii) data-driven machine learning, is one of our
aims. Secondly:
\begin{description}
  \item[Premise 2] \textit{Face appearance manifolds provide a powerful way of
    representing video sequences (or sets) of faces and allow for a
    unified treatment of illumination, pose and face motion pattern
    changes.}
\end{description}

Thus, the structure of this work is as follows. In Chapter~\ref{Chp:
Review} we review the existing literature on face recognition and
highlight the limitations of state-of-the-art methods, motivating
the aforementioned premises. Chapter~\ref{Chp: MDD} introduces the
notion of appearance manifolds and proposes a solution to the
simplest formulation of the recognition problem addressed in this
thesis. The subsequent chapters build up on this work, relaxing
assumptions about the data from which recognition is performed,
culminating with the \textit{Generic Shape-Illumination} method in
Chapter~\ref{Chp: gSIM}. The two chapters that follow apply the
introduced concepts on the problem of face-driven content-based
video retrieval and propose a novel framework for making further
use of the available data. A summary of the thesis and its major
conclusions are presented in Chapter~\ref{Chp: Conc}.

\subsection{List of publications}\label{SS: Pubs}
The following publications have directly resulted from the work
described in this thesis:

\begin{description}
    \item[Journal publications]
        \hspace{10pt}
        \begin{enumerate}
          \item O. Arandjelovi{\'c} and R. Cipolla. An information-theoretic approach to face recognition
                          from face motion manifolds. \textit{Image and Vision Computing (special issue
                          on Face Processing in Video Sequences)}, 24(6):639--647, 2006.

          \item M. Johnson, G. Brostow, J. Shotton, O. Arandjelovi\'c and R. Cipolla. Semantic photo
                          synthesis. \textit{Computer Graphics Forum}, 3(25):407--413, 2006.

          \item O. Arandjelovi\'c and R. Cipolla. Incremental learning of temporally-coherent Gaussian
                          mixture models. \textit{Society of Manufacturing Engineers (SME) Technical Papers}, 2, 2006.

          \item T-K. Kim, O. Arandjelovi\'c and R. Cipolla. Boosted manifold principal angles for image
                          set-based recognition. \textit{Pattern Recognition}, 40(9):pages 2475--2484, 2007.

          \item O. Arandjelovi\'c and R. Cipolla. A pose-wise linear illumination manifold model for face
              recognition using video. \textit{Computer Vision and Image Understanding}, 113(1):113--125, 2009.
              
          \item O. Arandjelovi\'c and R. Cipolla. A methodology for rapid illumination-invariant face recognition
              using image processing filters. \textit{Computer Vision and Image Understanding}, 113(2):159--171, 2009.

          \item O. Arandjelovi\'c, R. Hammoud and R. Cipolla. Thermal and reflectance based personal
              identification methodology in challenging variable illuminations. \textit{Pattern Recognition}, 43(5):1801--1813, 2010.
    
          \item O. Arandjelovi\'c and R. Cipolla. Achieving robust face recognition from video by combining a weak photometric model and a learnt generic face invariant. \textit{Pattern Recognition}, 46(1):9--23, January 2013.
        \end{enumerate}
    \item[Conference proceedings]
          \hspace{10pt}
        \begin{enumerate}
          \item O. Arandjelovi{\'c} and R. Cipolla. Face recognition from face motion manifolds using
                  robust kernel resistor-average distance. In \textit{Proc. IEEE Workshop on
                  Face Processing in Video}, 5:88, 2004.

          \item O. Arandjelovi{\'c} and R. Cipolla.  An illumination invariant face recognition system for
                          access control using video. In \textit{Proc. British Machine Vision Conference},
                          pages 537--546, 2004.

          \item O. Arandjelovi{\'c}, G. Shakhnarovich, J. Fisher, R. Cipolla, and T. Darrell.
                          Face recognition with image sets using manifold density divergence. In
                          \textit{Proc. IEEE Conference on Computer Vision and Pattern Recognition},
                          1:581--588, 2005.

          \item O. Arandjelovi{\'c} and A. Zisserman. Automatic face recognition for film character
                          retrieval in feature-length films. In
                          \textit{Proc. IEEE Conference on Computer Vision and Pattern Recognition},
                          1:860--867, 2005.

          \item T-K. Kim, O. Arandjelovi{\'c} and R. Cipolla. Learning over sets using boosted manifold
                          principal angles (BoMPA). In \textit{Proc. British Machine Vision Conference},
                          2:779--788, 2005.

          \item O. Arandjelovi\'c and R. Cipolla. Incremental learning of temporally-coherent Gaussian
                          mixture models. In \textit{Proc. British Machine Vision Conference}, 2:759--768, 2005.

          \item O. Arandjelovi\'c and R. Cipolla. A new look at filtering techniques for illumination
                          invariance in automatic face recognition. In \textit{Proc. IEEE Conference on Automatic
                          Face and Gesture Recognition}, pages 449--454, 2006.

          \item O. Arandjelovi\'c and R. Cipolla. Face recognition from video using the generic
                          shape-illumination manifold. In \textit{Proc. European Conference on Computer
                          Vision}, 4:27--40, 2006.

          \item O. Arandjelovi\'c and R. Cipolla. Automatic cast listing in feature-length films with
                          anisotropic manifold space. In \textit{Proc. IEEE Conference on Computer Vision
                          Pattern Recognition}, 2:1513--1520, 2006.

          \item G. Brostow, M. Johnson, J. Shotton, O. Arandjelovi\'c and R. Cipolla. Semantic photo
                          synthesis. In \textit{Proc. Eurographics}, 2006.

          \item O. Arandjelovi\'c, R. Hammoud and R. Cipolla. Multi-sensory face biometric fusion (for
                          personal identification). In \textit{Proc. IEEE Workshop on Object Tracking and
                          Classification Beyond the Visible Spectrum}, pages 128--135, 2006.

          \item O. Arandjelovi\'c and R. Cipolla. Face set classification using maximally probable mutual
                          modes.  In \textit{Proc. IAPR International Conference on Pattern Recognition},
                          pages 511--514, 2006.

          \item O. Arandjelovi\'c, R. Hammoud and R. Cipolla. On face recognition by fusing visual and
                          thermal face biometrics. In \textit{Proc. IEEE International Conference on
                          Advanced Video and Signal Based Surveillance}, pages 50--56, 2006.
    \end{enumerate}
    \item[Book chapters]
        \hspace{10pt}
        \begin{enumerate}
          \item O. Arandjelovi\'c and A. Zisserman. On film character retrieval in feature-length films.
                          Chapter in \textit{Interactive video: Algorithms and Technologies}, Springer-Verlag,
                          ISBN 978-3-540-33214-5, 2006.

          \item O. Arandjelovi\'c, R. Hammoud and R. Cipolla. Towards person authentication by fusing visual
                          and thermal face biometrics. Chapter in \textit{Multi-Sensory Multi-Modal Face
                          Biometrics for Personal Identification}, Springer-Verlag, ISBN 978-3-540-49344-0, 2007.

          \item O. Arandjelovi\'c, R. Hammoud and R. Cipolla. A person authentication system based on
                          visual and thermal face biometrics. Chapter in \textit{Object Tracking and
                          Classification Beyond the Visible Spectrum}, Springer-Verlag, ISBN 978-3-540-49344-0, 2007.

          \item O. Arandjelovi\'c and R. Cipolla. Achieving Illumination Invariance using Image
                          Chapter in \textit{Face Recognition}, Advanced Robotic Systems, ISBN 978-3-902613-03-5, 2007.
        \end{enumerate}

\end{description}

\graphicspath{{./02rev/}}
\chapter{Background}
\label{Chp: Review}
\begin{center}
  \footnotesize
  \vspace{-20pt}
  \framebox{\includegraphics[width=0.85\textwidth]{title_img.eps}}\\
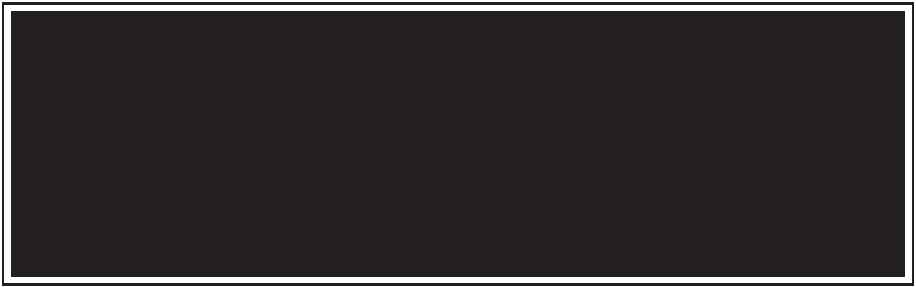
  Paul Gauguin. \textit{The Swineherd}\\
  1888, Oil on canvas, 74 x 93 cm\\
  Los Angeles County Museum of Art, Los Angeles
\end{center}

\cleardoublepage

Important practical applications of automatic face recognition have
made it a very popular research area of computer vision. This is
evidenced by a vast number of face recognition algorithms developed
over the last three decades and, in recent years, the emergence of a
number of commercial face recognition systems. This chapter: (i)
presents an account of the face detection and recognition
literature, highlighting the limitations of the state-of-the-art,
(ii) explains the performance measures used to gauge the
effectiveness of the proposed algorithms, and (iii) describes the
data sets on which algorithms in this thesis were evaluated.

\section{Introduction}
\begin{figure*}[!t]
  \centering
  \includegraphics[width=0.9\textwidth]{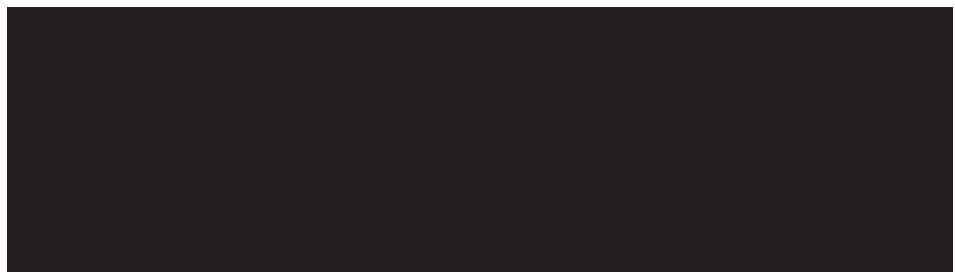}
  \caption[Face recognition system components.]
  { \it The main stages of an automatic face recognition system.
        (i) A face detector is used to detect (localize in space and scale)
        faces in a cluttered scene; this is followed by (ii) extraction
        of features used to represent faces and, finally (iii) extracted
        features are compared to those stored in a training database to
        associate novel faces to known individuals. }
  \label{Fig: Steps}
  \vspace{6pt}\hrule
\end{figure*}

At the coarsest level, the task of automatic, or computer-based,
face recognition inherently involves three stages: (i)
detection/localization of faces in images, (ii) feature extraction
and (iii) actual recognition, as shown in Figure~\ref{Fig: Steps}.
The focus of this thesis, and consequently the literature review, is
on the last of the three tasks. However, it is important to
understand the difficulties of the two preceding steps. Any
inaccuracy injected in these stages impacts the data a system must
deal with in the recognition stage. Additionally, the usefulness of
the overall system in practice ultimately depends on the performance
of the entire cascade. For this reason we first turn our attention
to the face detection literature and review some of the most
influential approaches.

\section{Face detection}
\begin{figure*}
  \centering
  \includegraphics[width=0.8\textwidth]{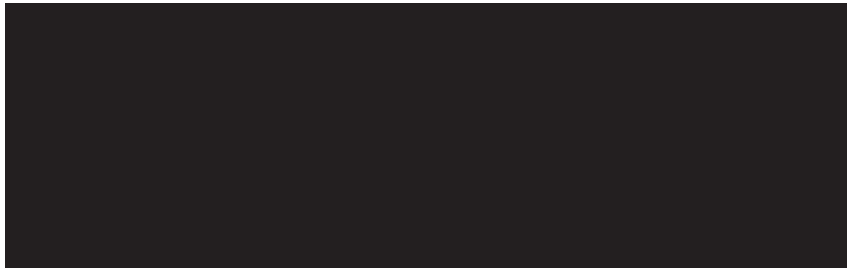}
  \caption[Face detection example.]{ \it An input image and the
            result using the popular Viola-Jones face detector
            \cite{ViolJone2004}. Detected faces are shown with
            green square bounding boxes, showing their location
            and scale. One missed detection (i.e.\ a false negative)
            is shown with a red bounding box. There are no false
            detections (i.e.\ false positives). }
  \label{Fig: DetEx}
\end{figure*}

Unlike recognition which concerns itself with discrimination of
objects in a category, the task of detection is that of discerning
the entire category. Specifically, face detection deals with the
problem of determining the presence of and localizing faces in
images. Much like face recognition, this is complicated by in-plane
rotation of faces, occlusion, expression changes, pose (out-of-plane
rotation) and illumination, see Figure~\ref{Fig: DetEx}.

Face detection technology is fairly mature and a number of reliable
face detectors have been built. Here we summarize state-of-the-art
approaches -- for a higher level of detail the reader may find it
useful to refer to a general purpose review \cite{Hjel2001,
YangAhujKrieg2002}.

\begin{table}
  \Large
  \caption[Detection approaches overview.]{ \it An overview of face detection approaches.\vspace{10pt} }
  \begin{tabular*}{1.00\textwidth}{@{\extracolsep{\fill}}l|l|l|l|l|l|l}
    \Hline
    \Huge & \multicolumn{6}{c}{\textbf{\normalsize Face detection approaches}} \\
    \hline
    ~\bf \small Input data     & \multicolumn{3}{c|}{\small Still images} & \multicolumn{3}{c}{\small Sequences} \\
    \hline
    ~\bf \small Approach       & \multicolumn{3}{c|}{\small Ensemble} & \multicolumn{3}{c}{\small Cascade} \\
    \hline
    ~\bf \small Cues & \small Greyscale & \small Colour & \small Motion & \small Depth & \multicolumn{2}{c}{\small Other}  \\
    \hline
    ~\bf \small Representation & \multicolumn{2}{c|}{\small Holistic} & \multicolumn{2}{c|}{\small Feature-based} & \multicolumn{2}{c}{\small Hybrid} \\
    \hline
    ~\bf \small Search         & \multicolumn{2}{c|}{\small Greedy} & \multicolumn{2}{c|}{\small Exhaustive} & \multicolumn{2}{c}{\small Focus of attention} \\
    \Hline
  \end{tabular*}
  \label{Tab: Overview 1}
\end{table}

\paragraph{State-of-the-art methods.}
Most of the current state-of-the-art face detection methods are
holistic in nature, as opposed to part-based. While part-based (also
know as constellation of features) approaches were proposed for
their advantage of exhibiting greater viewpoint robustness
\cite{HeisPoggPont2000}, they have largely been abandoned for
complex, cluttered scenes in favour of multiple view-specific
detectors that treat the face as a whole. Henceforth this should be
assumed when talking about holistic methods, unless otherwise
stated. One such successful algorithm was proposed by Rowley
\textit{et al.} \cite{RowlBaluKana1998}. An input image is scanned
at multiple scales with a neural network classifier which is fed
filtered appearance of the current patch, see Figure~\ref{Fig:
NNdet}. Sung and Poggio \cite{SungPogg1998} also employ a
Multi-Layer Perceptron (MLP), but in a statistical framework,
learning ``face'' and ``non-face'' appearances as Gaussian mixtures
embedded in the 361-dimensional image space ($19 \times 19$ pixels).
Classification is then performed based on the ``difference'' vector
between the appearance of the current patch and the learnt
statistical models. Much like in \cite{RowlBaluKana1998}, an
exhaustive search over the location/scale space is performed. The
method of Schneiderman and Kanade \cite{Schneiderman2000} moves away
from greyscale appearance, proposing to use histograms of wavelet
coefficients instead. An extension to video sequence-based
detections was proposed by Mikolajczyk \textit{et al.} in
\cite{MikoChouSchm2001} -- a dramatic reduction of the search space
between consecutive frames was achieved by propagating temporal
information using the Condensation tracking algorithm \cite{IsarBlak1998}.

\begin{figure*}
  \centering
  \includegraphics[width=1.0\textwidth]{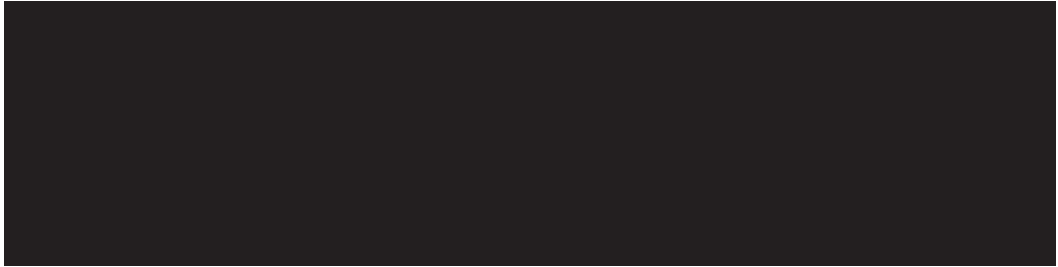}
  \caption[Neural network face detection.]{ \it A diagram of the
    method of Rowley \textit{et al.} \cite{RowlBaluKana1998}. This
    approach is representative of the group of neural network-based face
    detection approaches: (i) input image is scanned at different scales
    and locations, (ii) features are extracted from the current window and
    are (iii) fed into an MLP classifier. }
  \label{Fig: NNdet}
\end{figure*}

\begin{table}
  \Large
  \caption[State-of-the-art detection performance.]
  { \it A performance summary of state-of-the-art face detection algorithms. Shown is the \% of detected
        faces (i.e.\ true positives), followed by a number of incorrect detections (i.e.\ false positives).\vspace{10pt} }
  \begin{tabularx}{1.00\textwidth}{l|XlX|XlX}
    \Hline
      \normalsize \bf \multirow{3}{*}{Method} & \multicolumn{6}{c}{\normalsize \bf Data set} \\
         & \multicolumn{3}{c|}{\normalsize CMU} & \multicolumn{3}{c}{\normalsize MIT} \\
         & \multicolumn{3}{c|}{\normalsize (130 images, 507 faces)} & \multicolumn{3}{c}{\normalsize (23 images, 57 faces)} \\
    \hline
      \small F\'{e}raud \textit{et al.}
          \cite{FeraBernVial+2001}       && \small 86.0\%~/~8  &&& \small             &\\
      \small Garcia-Delakis
          \cite{GarcDela2004}            && \small 90.3\%~/~8  &&& \small 90.1\%~/~7  &\\
      \small Li \textit{et al.}
          \cite{LiZhuZhan+2002}          && \small 90.2\%~/~31 &&& \small             &\\
      \small Rowley \textit{et al.}
          \cite{RowlBaluKana1998}        && \small 86.2\%~/~23 &&& \small 84.5\%~/~8  &\\
      \small Sung-Poggio
          \cite{SungPogg1998}            && \small             &&& \small 79.9\%~/~5  &\\
      \small Schneiderman-Kanade
          \cite{Schneiderman2000}        && \small 90.5\%~/~33 &&& \small 91.1\%~/~12 &\\
      \small Viola-Jones
          \cite{ViolJone2004}            && \small 85.2\%~/~5  &&& \small 77.8\%~/~31 &\\
    \Hline
  \end{tabularx}
  \label{Tab: DetPerf}
\end{table}

While achieving good accuracy -- see Table~\ref{Tab: DetPerf} -- the
described early approaches suffer from tremendous computational
overhead. More recent methods have focused on online detector
efficiency, with attention-based rather than exhaustive search over
an entire image. The key observation is that the number of face
patches in a typical image is usually much smaller than the number
of non-faces. A hierarchial search that quickly eliminate many
unpromising candidates was proposed in \cite{FeraBernVial+2001}.
Simplest and fastest filters are applied first, greatly reducing the
workload of the subsequent, gradually slower and more complex
classifiers. The same principle using support Vector Machines was
employed by Romdhani \textit{et al.} \cite{RomdTorrSchoBlak2004}. In
\cite{FeraBernVial+2001} F\'{e}raud \textit{et al.} used a variety
of cues including colour and motion-based filters. A cascaded
approach was also employed in the breakthrough method of Viola and
Jones \cite{ViolJone2004}. This detector, including a number of
extensions proposed thereafter, is the fastest one currently
available (the authors report a speedup of a factor of 15 over
\cite{RowlBaluKana1998}). This is achieved by several means: (i) the
attention cascade mentioned previously reduces the number of
computationally heavy operations, (ii) it is based on boosting fast
weak learners \cite{FreuScha1995}, and (iii) the proposed integral
image representation eliminates repeated computation of Haar
feature-responses. Improvements to the original detector have since
been proposed, e.g.\ by using a pyramidal structure
\cite{HuanAiLiLao2005, LiZhuZhan+2002} for multi-view detection and
rotation invariance \cite{HuanAiLiLao2005, WuAiHuanLao2004}, or
joint low-level features \cite{MitaKaneHori2005} for reducing the
number of false positive detections.

\section{Face recognition}
There are many criterions by which one may decide to cluster face
recognition algorithms, depending on the focus of discussion, see
Table~\ref{Tab: Overview 1}. For us it will be useful to start by
talking about the type of data available as input and the conditions
in which such data was acquired. As was mentioned in the previous
chapter, the main sources of changes in one's appearance are the
illumination, head pose, image quality, facial expressions and
occlusions. In \emph{controlled} imaging conditions, some of or all
these variables are fixed so as to simplify recognition, and this is
known \textit{a priori}. This is a possible scenario in the
acquisition of passport photographs, for example.

\begin{table}
  \Large
  \caption[Recognition approaches overview.]{ \it An overview of face recognition approaches. \vspace{10pt} }
  \begin{tabular*}{1.00\textwidth}{@{\extracolsep{\fill}}l|l|l|l|l|l|l}
    \Hline
    \Huge & \multicolumn{6}{c}{\bf \normalsize Face recognition approaches} \\
    \hline
    ~\bf \small Acquisition
    conditions     & \multicolumn{2}{c|}{\small Controlled} & \multicolumn{2}{c|}{\small ``Loosely'' controlled} & \multicolumn{2}{c}{\small Extreme} \\
    \hline
    ~\bf \small Input data     & \multicolumn{2}{c|}{\small Still images} & \multicolumn{2}{c|}{\small Image sets} & \multicolumn{2}{c}{\small Sequences} \\
    \hline
    ~\bf \small Modality & \multicolumn{2}{c|}{\small Optical data} & \multicolumn{2}{c|}{\small Other (IR, range etc.)} & \multicolumn{2}{c}{\small Hybrid} \\
    \hline
    ~\bf \small Representation & \multicolumn{2}{c|}{\small Holistic} & \multicolumn{2}{c|}{\small Feature-based} & \multicolumn{2}{c}{\small Hybrid} \\
    \hline
    ~\bf \small Approach       & \multicolumn{3}{c|}{\small Appearance-based} & \multicolumn{3}{c}{\small Model-based} \\
    \Hline
  \end{tabular*}
  \label{Tab: Overview 1}
\end{table}

Historically, the first attempts at automatic face recognition date
back to the early 1970s and were able to cope with some success with
this problem setup only. These pioneering methods relied on
predefined geometric features for recognition \cite{Kell1970,
Kana1973}. Distances \cite{BrunPogg1993a} or angles between
locations of characteristic facial features (such as the eyes, the
nose etc.) were used to discriminate between individuals, typically
using a Bayes classifier \cite{BrunPogg1993a}. In \cite{Kana1973}
Kanade reported correct identification of only 15 out of 20
individuals under controlled pose. Later Goldstein \textit{et al.}
\cite{GoldHarmLesk1972} and, Kaya and Kobayashi \cite{KayaKoba1972}
(also see work by Bledsoe \textit{et al.} \cite{Bled1966,
ChanBled1965}) showed geometric features to be sufficiently
discriminatory if facial features are manually selected.

Most obviously, geometric feature-based methods are inherently very
sensitive to head pose variation, or equivalently, the camera angle.
Additionally, these methods also suffer from sensitivity to noise in
the stage of localization of facial features. While geometric
features themselves are insensitive to illumination changes, the
difficulty of their extraction is especially prominent in extreme
lighting conditions and when the image resolution is low.

\subsection{Statistical, appearance-based methods}
In sharp contrast to the geometric, feature-based algorithms are
appearance-based methods. These revived research interest in face
recognition in the early 1990s and to this day are dominant in
number in the literature. Appearance-based methods, as their name
suggests, perform recognition directly from the way faces appear in
images, interpreting them as ordered collections of pixels. Faces
are typically represented as vectors in the $D$-dimensional
\emph{image space}, where $D$ is the number of image pixels (and
hence usually large). Discrimination between individuals is then
performed by employing statistical models that explain inter- and/or
intra- personal appearance changes.

\begin{figure*}[!t]
  \centering
  \subfigure[Conceptual drawing]{\includegraphics[width=0.6\textwidth]{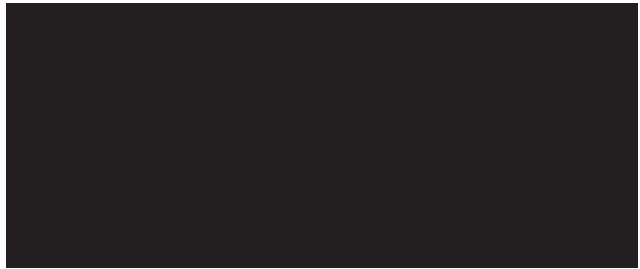}}
  \subfigure[Face space basis as images]{\includegraphics[width=0.9\textwidth]{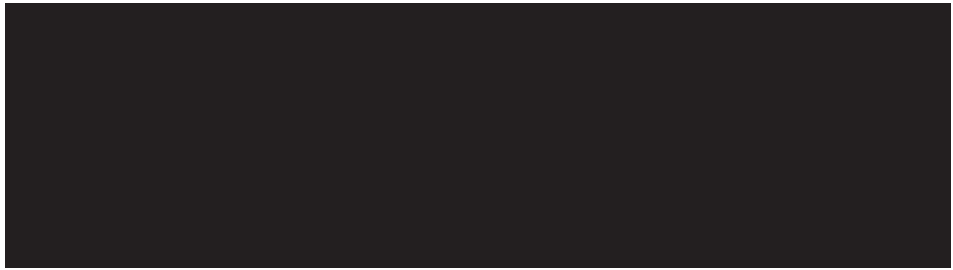}}
  \caption[Eigenfaces and PCA.]{ \it (a) A conceptual picture of the Eigenfaces method.
            All data is projected and then classified in the linear subspace corresponding
            to the dominant modes of variation across the entire data set, estimated using
            PCA. (b) The first 10 dominant modes of the CamFace data, shown as images.
            These are easily interpreted as corresponding to the main modes of
            illumination-affected appearance changes (brigher/darker face, strong light
            from the left/right etc.) and pose. }
  \label{Fig: PCA}
  \vspace{6pt}\hrule
\end{figure*}

The \emph{Eigenfaces} algorithm of Turk and Pentland
\cite{TurkPent1991, TurkPent1991a} is the most famous algorithm of
this group. Its approach was motivated by previous work by Kohonen
\cite{Koho1977} on auto-associative memory matrices for storage and
retrieval of face images, and by Kirby and Sirovich
\cite{SiroKirb1987, KirbSiro1990}. It uses Principal Component
Analysis (PCA) to construct the so-called \emph{face space} -- a
space of dimension $d \ll D$ that explains appearance variations of
human faces, see Figure~\ref{Fig: PCA}. Recognition is performed by
projecting all data onto the face space and classifying a novel face
to the closest class. The most common norms used in the literature
are the Euclidean (also known as $L_2$), $L_1$ and Mahalanobis
\cite{DrapBaekBartBeve2003, BeveSheDrap2001}, in part dictated by
the availability of training data.

By learning what appearance variations one can expect across the
corpus of \emph{all} human faces and then by effectively
\emph{reconstructing} any novel data using the underlying subspace
model, the main advantage of Eigenfaces is that of suppressing noise
in data \cite{WangTang2003}. By interlacing the subspace projection
with RANSAC \cite{FiscBoll1981}, occlusion detection and removal are
also possible \cite{BlacJeps1998, AranZiss2005}. However, inter- and
intra-personal appearance variations are not learnt separately.
Hence, the method is recognized as more suitable for detection and
compression tasks \cite{MoghPent1995, KingXu1997} than recognition.

A Bayesian extension of Eigenfaces, proposed by Moghaddam \textit{et
al.} \cite{MoghWahiPent1998}, improves on the original method by
learning the mean intra-personal subspace. Recognition decision is
again cast using the assumption that appearance for each person
follows a Gaussian distribution, with also Gaussian, but isotropic
image noise.

\begin{figure*}[!t]
  \centering
  \subfigure[Conceptual drawing]{\includegraphics[width=0.65\textwidth]{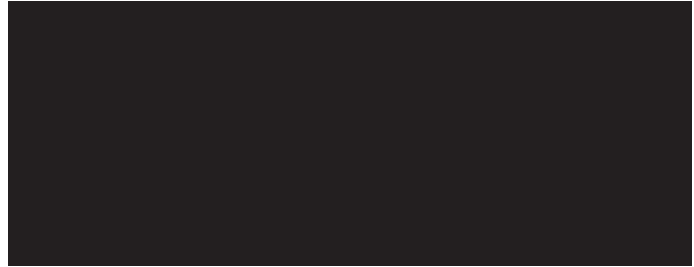}}
  \subfigure[LDA basis as images]{\includegraphics[width=0.9\textwidth]{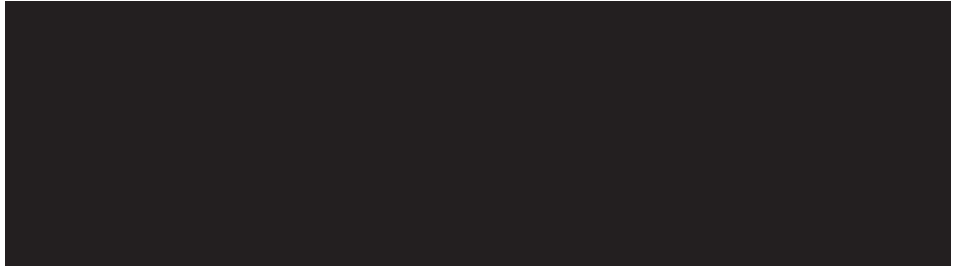}}
  \caption[Fisherfaces and LDA.]{ \it (a) A conceptual picture of the Fisherfaces method.
            All data is projected and then classified in the linear subspace that best
            separates classes, which are assumed to be isotropic and Gaussian.
            (b) The first 10 most discriminative modes of the CamFace data, shown as images.
            To the human eye these are not as meaningful as face space basis, see
            Figure~\ref{Fig: PCA}.  }
  \label{Fig: LDA}
  \vspace{6pt}\hrule
\end{figure*}

To address the lack of discriminative power of Eigenfaces, another
appearance-based subspace method was proposed -- the
\emph{Fisherfaces} \cite{Yamb2000, ZhaoChelRosePhil2000}, named
after Fisher's Linear Discriminant Analysis (LDA) that it employs.
Under the assumption of isotropically Gaussian class data, LDA
constructs the optimally discriminating subspace in terms of
maximizing the between to within class scatter, as shown in
Figure~\ref{Fig: LDA}. Given sufficient training data, Fisherfaces
typically perform better than Eigenfaces \cite{Yamb2000, Weng1993}
with further invariance to lighting conditions when applied to
Fourier transformed images \cite{AkamSasaFukuSuen1991}. One of the
weaknesses of Fisherfaces is that the estimate of the optimal
projection subspace is sensitive to a particular choice of training
images \cite{BeveSheDrap2001}. This finding is important as it
highlights the need for more extensive use of domain specific
information. It is therefore not surprising that limited
improvements were achieved by applying other purely statistical
techniques on the face recognition task: Independent Component
Analysis (ICA) \cite{BaekDrapBeveShe2002, DrapBaekBartBeve2003,
BartLadeSejn1998, BartMoveSejn2002}; Singular Value Decomposition
(SVD) \cite{PresTeukVettFlan1992}; Canonical Correlation Analysis
(CCA) \cite{SunHengJin+2005}; Non-negative Matrix Factorization
(NNMF) \cite{WangJiaHu+2005}. Simple linear extrapolation
techniques, such as the Nearest Feature Line (NFL) \cite{LiLu1999},
Nearest Feature Plane (NFS) or Nearest Feature Space (NFS) also
failed to achieve significant performance increase using holistic
appearance.

\begin{figure}[!t]
  \centering
  \includegraphics[width=0.8\textwidth]{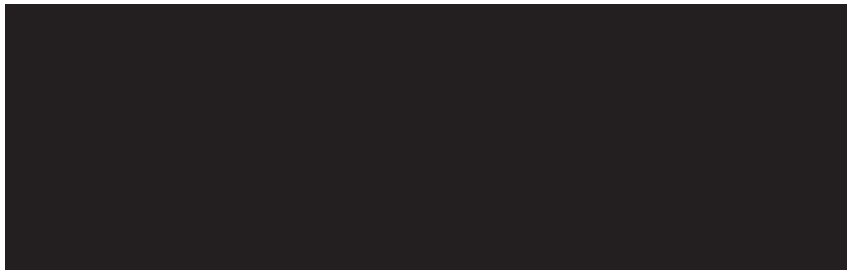}
  \caption[Bottleneck neural network.]{ \it The feed-forward bottleneck
            neural network structure, used to implement nonlinear
            projection of faces. Functionality-wise, two distinct
            parts of the network can be identified: (i) layers 1-3
            perform compression of data by exploiting the inherently
            low-dimensional nature of facial appearance variations;
            (ii) layers 4-5 perform classification of the projected
            data. }
  \label{Fig: NN}
  \vspace{6pt}\hrule
\end{figure}

\paragraph{Nonlinear approaches.}
Promising results of the early subspace methods and new research
challenges they uncovered, motivated a large number of related
approaches. Some of these focused on relaxing the crude assumption
that the appearance of a face conforms to a Gaussian distribution.
The most popular direction employed the kernel approach
\cite{SchoSmolMull1999,SchoSmol2002} with methods such as
\emph{Kernel Eigenfaces} \cite{YangAhujKrieg2000, Yang2002},
\emph{Kernel Fisherfaces} \cite{YangAhujKrieg2000}, \emph{Kernel
Principal Angles} \cite{WolfShas2003}, \emph{Kernel RAD}
\cite{AranCipo2004, AranCipo2006}, \emph{Kernel ICA}
\cite{YangGaoZhan+2005} and others (e.g.\ see
\cite{ZhouKrueChel2003}). As an alternative to Kernel Eigenfaces, a
multi-layer perceptron (MLP) neural network with a bottleneck
architecture \cite{deMeCott1993, Malt1998}, shown in
Figure~\ref{Fig: NN}, was proposed to implement nonlinear PCA
projection of faces \cite{MoghPent2002}, but has since been largely
abandoned due to the difficulty to optimally train
\cite{ChenDesaZhan1997}\footnote{The reader may be interested in the
following recent paper that proposes an automatic way of
initializing the network weights so that they are close to a good
solution \cite{HintSala2006}.}. The recently proposed Isomap
\cite{TeneSilvLang2000} and Locally Linear Embedding (LLE)
\cite{RoweSaul2001} algorithms were successfully applied to
unfolding nonlinear appearance manifolds \cite{BaiYinShi+2005,
KimKitt2005, PangLiuYu2006, Yang2002a}, as were piece-wise linear
models \cite{AranShakFish+2005, LeeHoYangKrie2003,
PentMoghStar1994}.

\paragraph{Local feature-based methods.}
While the above-mentioned methods improve on the linear subspace
approaches by more accurately modelling appearance variations
\emph{seen} in training, they fail to significantly improve on the
limited ability of the original methods in \emph{generalizing}
appearance to unseen imaging conditions (i.e.\ illumination, pose
and so on) \cite{BaekDrapBeveShe2002, Barr1998, GrosShiCohn2001,
Nefi1996, ShasLeviAvid2002}.

Local feature-based methods were proposed as an alternative to
holistic appearance algorithms, as a way of achieving a higher
degree of viewpoint invariance. Due to the smoothness of faces, a
local surface patch is nearly planar and its appearance changes can
be expected to be better approximated by a linear subspace than
those of an entire face. Furthermore, their more limited spatial
extent and the consequent lower subspace dimensionality have both
computational benefits and are less likely to suffer from the
so-called \emph{curse of dimensionality}.

Trivial extensions such as \emph{Eigenfeatures}
\cite{AbdiValeEdel1998, PentMoghStar1994} and \emph{Eigeneyes}
\cite{CampFeriCesa2000} demonstrated this, achieving recognition
rates better than those of the original Eigenfaces
\cite{CampFeriCesa2000}. Even better results were obtained using
hybrid methods i.e.\ combinations of holistic and local patch-based
appearances \cite{PentMoghStar1994}.

An influential group of local features-based methods are the
\emph{Local Feature Analysis} (LFA) (or elastic graph matching)
algorithms, the most acclaimed of these being \emph{Elastic Bunch
Graph Matching} (EBGM) \cite{ArcaCampLanz2003, Bolm2003,
PeneAtic1996, WiskFellKrug+1997, WiskFellKrug+1999}\footnote{The
reader should note that LFA-based algorithms are sometimes
categorized as model-based. In this thesis the term model-based is
used for algorithms that explain the entire observed face
appearance. }. LFA methods have proven to be amongst the most
successful in the literature \cite{HeoAbidPaik+2003} and are
employed in commercial systems such as FaceIt$^{\circledR}$ by
Identix \cite{Iden} (the best performing commercial face recognition
software in the 2003 Face Recognition Vendor Test
\cite{PhilGrotMichBlac+2003}).

\begin{figure}[!t]
  \centering
  \includegraphics[width=0.7\textwidth]{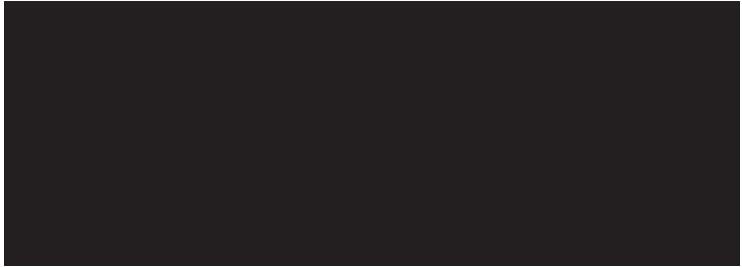}
  \caption[Elastic Graph Matching.]{ \it An elastic graph
            successfully fitted to an input image. Its nodes
            correspond to fiducial points, which are used to
            characterize local, discriminative appearance.
            Graph topology (i.e.\ fiducial point adjacency
            constraints) is typically used only in the fitting
            stage, but is discarded for recognition as it is
            greatly distorted by viewpoint variations.  }
  \label{Fig: EGM}
  \vspace{6pt}\hrule
\end{figure}

The common underlying idea behind these methods is that they combine
in a unified manner holistic with local features, and appearance
information with geometric structure. Each face is represented by a
graph overlayed on the appearance image. Its nodes correspond to the
locations of features used to describe local face appearance, while
its edges constrain the holistic geometry by representing feature
adjacency, as shown in Figure~\ref{Fig: EGM}. To establish the
similarity between two faces, their elastic graph representations
are compared by measuring the distortion between their topological
configurations and the similarities of feature appearances.

\begin{figure*}[!t]
  \centering
  \includegraphics[width=0.8\textwidth]{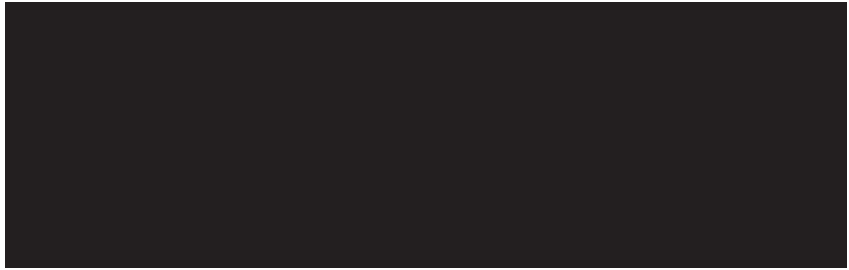}
  \caption[Gabor wavelets.]
    { \it A 2-dimensional Gabor wavelet, shown as a surface and
          a corresponding intensity image. The wavelet closely
          resembles the response of receptive fields of the visual
          cortex are provides a trade-off between spatial and
          frequency localization of the signal (i.e.\ appearance). }
  \label{Fig: Gabor}
  \vspace{6pt}\hrule
\end{figure*}

Various LFA methods primarily differ: (i) in how local feature
appearances are represented and (ii) in the way two elastic graphs
are compared. In Elastic Bunch Graph Matching \cite{Bolm2003,
Seni1999, WiskFellKrug+1997, WiskFellKrug+1999}, for example, Gabor
wavelet \cite{Gabo1946} jets are used to characterize local
appearance. In part, their use is attractive in this context because
the functional form of Gabor wavelets closely resembles the response
of receptive fields of the visual cortex \cite{Daug1980, Daug1988,
JonePalm1987, Marc1980}, see Figure~\ref{Fig: Gabor}. They also
provide powerful means of extracting \emph{local} frequency
information, which has been widely employed in various branches of
computer vision for characterizing texture \cite{HongKaplSmit2004,
Lee1996, MuneGaneArum+2005, PunLee2004}. Local responses to
multi-scale morphological operators (dilation and erosion) were also
successfully used as fiducial point descriptors
\cite{KotrTefaPita1998, KotrTefaPita2000, KotrTefaPita2000a,
KotrTefaPita2000b}.

Unlike any of the previous methods, LFA algorithms generalize well
in the presence of facial expression changes \cite{PhilVard1995,
WiskFell1999}. On the other hand, much like the early
geometric-feature based methods, significant viewpoint changes pose
problems in both the graph fitting stage, as well as in recognition,
as the projected topological layout of fiducial features changes
dramatically with out-of-plane head rotation. Furthermore, both
wavelet-based and morphological response-based descriptors show
little invariance to illumination changes, causing a sharp
performance decay in realistic imaging conditions (an Equal Error
Rate of 25-35\% was reported in \cite{KotrTefaPita2000b}) and,
importantly for the work presented in this thesis, with low
resolution images \cite{SterPnevPoly2006}.

\paragraph{Appearance-based methods -- a summary.}
In closing, purely appearance-based recognition approaches can
achieve good generalization to unseen (i) poses and (ii) facial
expressions by using local or hybrid local and holistic features.
However, they all poorly generalize in the presence of large
illumination changes.

\subsection{Model-based methods}\label{CH2: Models}
The success of LFA in recognition across pose and expression can be
attributed to the shift away from purely statistical pattern
classification to the use of models that exploit \textit{a priori}
knowledge about the very specific instance of classification that
face recognition is. Model-based methods take this approach further.
They formulate models of image formation with the intention of
recovering (i) mainly person-specific (e.g.\ face albedo or shape)
and (ii) extrinsic, nuisance variables (e.g.\ illumination
direction, or head yaw). The key challenge lies in coming up with
models for which the parameter estimation problem is not ambiguous
or ill-conditioned.

\begin{figure*}[!t]
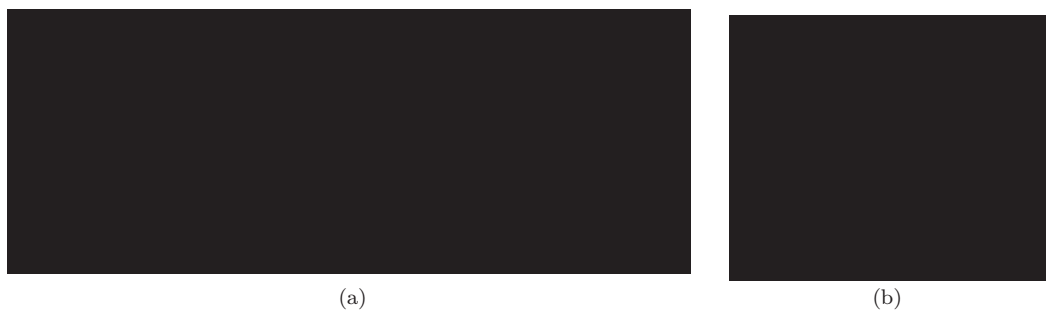

  \centering
  \footnotesize
  \begin{tabular}{VV}
    \includegraphics[width=0.65\textwidth]{lambertian.eps} \vspace{5pt} &
    \includegraphics[width=0.30\textwidth]{lambertian_face.eps} \\
    (a) & (b) \\
  \end{tabular}
  \caption[Lambertian reflectance.]{ \it (a) A conceptual drawing of the
            Lambertian reflectance model. Light is reflected
            isotropically, the reflected intensity being proportional to
            the cosine between the incident light direction $\mathbf{l}$ and the surface
            normal $\mathbf{n}$. (b) The appearance of a face rendered
            as a texture-free Lambertian surface.  }
  \label{Fig: Lambertian}
  \vspace{6pt}\hrule
\end{figure*}

\begin{figure*}[!t]
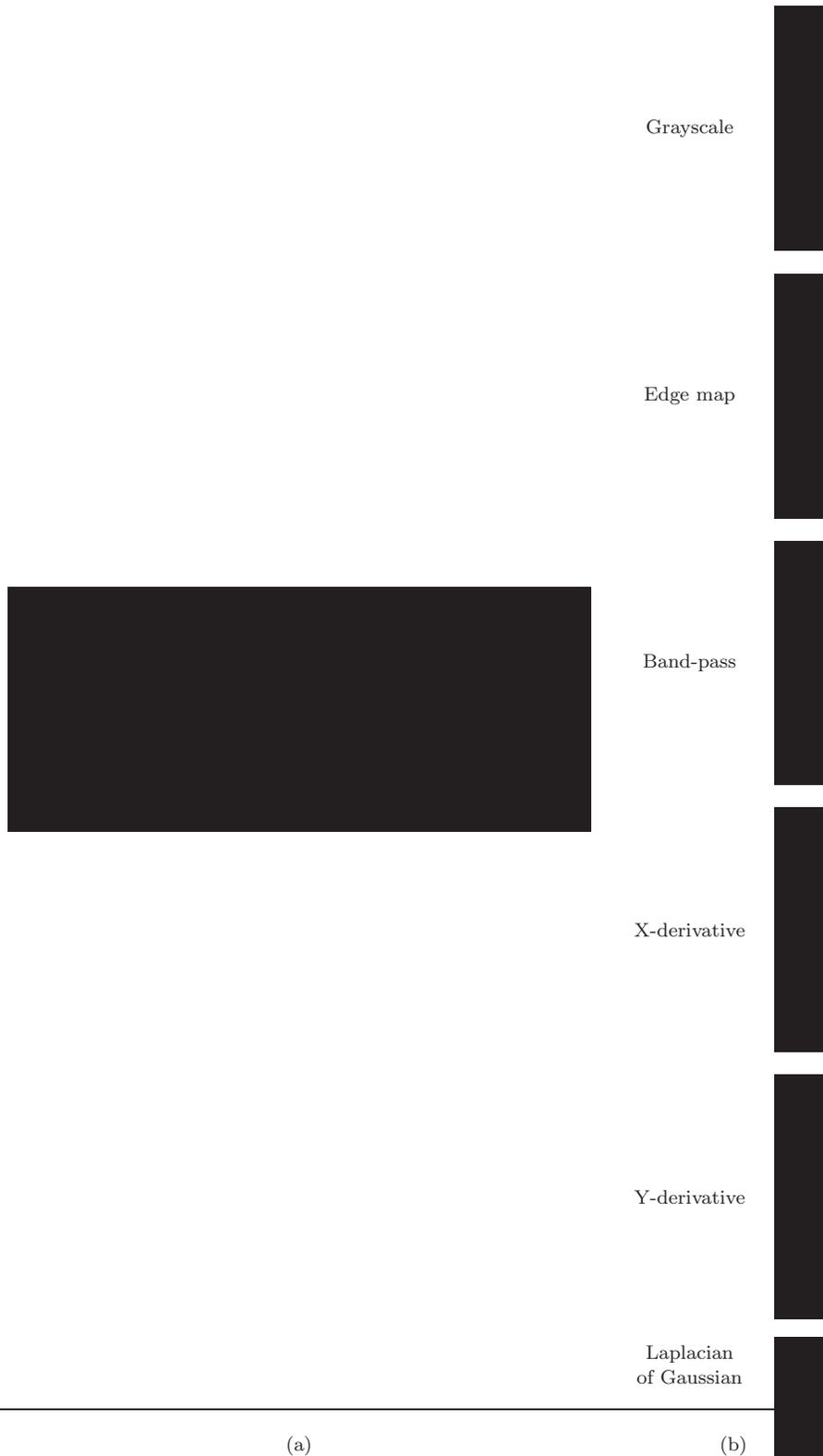

  \centering
  \footnotesize
  \begin{tabular}{VV}
    \includegraphics[width=0.6\textwidth]{freq_model.eps}
    &
    \begin{tabular}{VV}
      Grayscale & \includegraphics[width=0.06\textwidth]{img.eps}    \\\\
      Edge map      & \includegraphics[width=0.06\textwidth]{img_e.eps}  \\\\
      Band-pass     & \includegraphics[width=0.06\textwidth]{img_bp.eps} \\\\
      X-derivative  & \includegraphics[width=0.06\textwidth]{img_dx.eps} \\\\
      Y-derivative  & \includegraphics[width=0.06\textwidth]{img_dy.eps} \\\\
      Laplacian     & \multirow{2}{*}{\includegraphics[width=0.06\textwidth]{img_log.eps}} \\
      of Gaussian   &  \\\\
    \end{tabular}\\\\
    (a) & (b) \\
  \end{tabular}
  \caption[Spatial frequency-based generative model.]{ \it (a) The simplest generative
            model used for face recognition: images are assumed to consist of the
            low-frequency band that mainly corresponds to illumination changes,
            mid-frequency band which contains most of the discriminative, personal
            information and white noise. (b) The results of several most popular
            image filters operating under the assumption of the frequency model. }
  \label{Fig: Freq}
  \vspace{6pt}\hrule
\end{figure*}

\paragraph{2D illumination models.}
The simplest generative models in face recognition are used for
illumination normalization of raw images, as a preprocessing step
cascaded with, typically, appearance-based classification that
follows it. Considering the previously-discussed observation that
the face surface, as well as albedo, are largely smooth, and
assuming a Lambertian reflectance model, illumination effects on
appearance are for the most part slowly spatially varying, see
Figure~\ref{Fig: Lambertian}. On the other hand, discriminative
person-specific information is mostly located around facial features
such as the eye, the mouth and the nose, which contain
discontinuities and give rise to appearance changes of high
frequency, as illustrated in Figure~\ref{Fig: Freq}~(a).

It has been applied in the forms of high-pass and band-pass filters
\cite{AranZiss2005, AranCipo2006c, BuhmLadeEeck1994, FitzZiss2002},
Laplacian-of-Gaussian filters \cite{AdinMoseUllm1997,
AranCipo2006a}, edge maps \cite{AdinMoseUllm1997, AranCipo2006a,
GaoLeun2002, Taka1998} and intensity derivatives
\cite{AdinMoseUllm1997, AranCipo2006a}, to name the few most popular
approaches, see Figure~\ref{Fig: Freq}~(b) (also see
Chapter~\ref{Chp: Filters}). Although widely used due to its
simplicity, numerical efficiency and the lack of assumptions on head
pose, the described spatial frequency model is universally regarded
as insufficiently sophisticated in all but mild illumination
conditions, struggling with cast shadows and specularities, for
example. Various modifications have thus been proposed. In the
\textit{Self-Quotient Image} (SQI) method \cite{WangLiWang2004}, the
mid-frequency, discriminative band is also scaled with local image
intensity, thus normalizing edge strengths in weakly and strongly
illuminated regions of the face.

A principled treatment of illumination invariant recognition for
Lambertian faces, the \textit{Illumination Cones} method, was
proposed in \cite{GeorKrieBelh1998}. In \cite{BelhKrie1996} it was
shown that the set of images of a convex, Lambertian object,
illuminated by an arbitrary number of point light sources at
infinity, forms a polyhedral cone in the image space with dimension
equal to the number of distinct surface normals. Georghiades
\textit{et al.} successfully used this result by reilluminating
images of frontal faces. The key limitations of their method are (i)
the requirement of at least 3 images for each novel face,
illuminated from linearly independent directions and in the same
pose, (ii) lack of accounting non-Lambertian effects. These two
limitations are separately addressed in \cite{NishYama2006} and
\cite{WolfShas2003}. In \cite{NishYama2006}, a simple model of
diffuse reflections of a generic face is used to iteratively
classify face regions as `Lambertian, no cast shadows', `Lambertian,
cast shadow' and `specular', applying SQI-based normalization to
each separately. Is important to observe that the success of this
method which uses a model of only a \emph{single} (generic) face
demonstrates that shape and reflectance similarities across
individuals can also be exploited so as to improve recognition. The
\textit{Quotient Image} (QI) method \cite{WolfShas2003} makes use of
this by making the assumption that all human faces have the same
shape. Using a small ($\sim 10$) bootstrap set of individuals, each
in 3 different illuminations, it is shown how pure albedo and
illumination effects can approximately be separated from a single
image of a novel face. However, unlike the method of Nishiyama and
Yamaguchi \cite{NishYama2006}, QI does not deal well with non-Lambertian effects or cast
shadows.

\paragraph{2D combined shape and appearance models.}
The \textit{Active Appearance Model} (AAM) \cite{CootEdwaTayl1998}
was proposed for modelling objects that vary in shape and
appearance. It has a lot of similarity to the older \textit{Active
Contour Model} \cite{KassWitkTerz1987} and the \textit{Active Shape
Model} \cite{CootTaylCoopGrah1995, HamaAbuGust1998,
HamaAbuGust1998a} (also see \cite{SclaIsid1998}) that model shape
only.

\begin{figure}
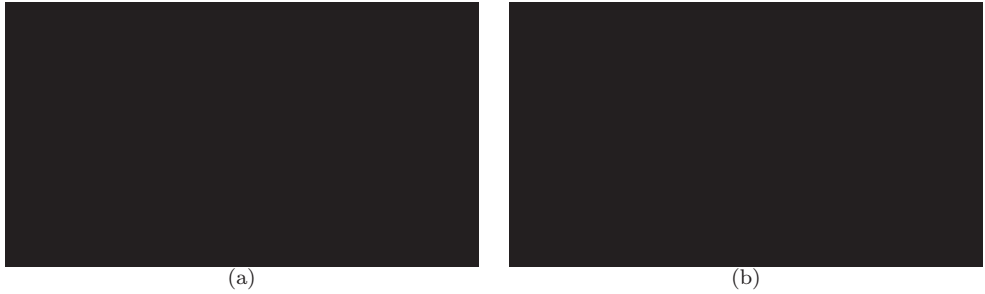

  \centering
  \footnotesize
  \begin{tabular}{VV}
    \includegraphics[width=0.45\textwidth]{aam_1.eps} &
    \includegraphics[width=0.45\textwidth]{aam_2.eps} \\
    (a) & (b) \\
  \end{tabular}
  \caption[AAM mesh fitting.]
      { \it (a) Two input images with correct adaptation (top) and the
                corresponding geometrically normalized images (bottom). \cite{DornAhlb2003}.
            (b) First three modes of the AAM appearance model in
                $\pm 3$ standard deviations \cite{KangCootTayl2002} }
  \label{Fig: AAM}
  \vspace{6pt}\hrule
\end{figure}

In AAM a deformable triangular (c.f.\ EBGM) mesh is fitted to an
image of a face, see Figure~\ref{Fig: AAM}~(a). This is guided by
combined statistical models of shape and shape-free appearance, so
as to best explain the observed image. In \cite{CootEdwaTayl1998}
linear, PCA models are used, see Figure~\ref{Fig: AAM}~(b). Still,
AAM parameter estimation is a difficult optimization problem.
However, given that faces do not vary a lot in either shape or
appearance, the structure of the problem is similar whenever the
minimization is performed. In \cite{CootEdwaTayl1998} and
\cite{CootEdwaTayl1999}, this is exploited by learning a linear
relationship between the current reconstruction error and the model
parameter perturbation required to correct it (for variations on the
basic algorithm also see \cite{CootKitt2002, ScotCootTayl2003}).

AAMs have been successfully used for face recognition
\cite{EdwaCootTayl1998, FaggPaplChin2006}, tracking
\cite{DornAhlb2003} and expression recognition \cite{SaatTown2006}.
The main limitations of the original method are: (i) the sensitivity
to illumination changes in the recognition stage, and (ii) occlusion
(including self-occlusion, due to 3D rotation for example). The
latter problem was recently addressed by Gross \textit{et al.}
\cite{GrosMattBake2006}, a modification to the original algorithm
demonstrating good fitting results with extreme pose changes and
occlusion.

\paragraph{3D combined shape and illumination models.}
The most recent and complex generative models used for face
recognition are largely based on the \textit{3D Morphable Model}
introduced in \cite{BlanVett1999} which builds up on the previous
work on 3D modelling and animation of faces \cite{deCaMetaSton1998,
DPaol1991, MagnMinhAnge+1989, Park1975, Park1982, Park1996}. The
model consists of albedo values at the nodes of a 3-dimensional
triangular mesh describing face geometry. Model fitting is performed
by combining a Gaussian prior on the shape and texture of human
faces with photometric information from an image
\cite{WallBlanVett1999}. The priors are estimated from training data
acquired with a 3D scanner and densely registered using a
regularized 3D optical flow algorithm \cite{BlanVett1999}. 3D model
recovery from an input image is performed using a gradient descent
search in an analysis-by-synthesis loop. Linear
\cite{RomdBlanVett2002} or quadratic \cite{RomdVett2003} error
functions have been successfully used.

\begin{figure}
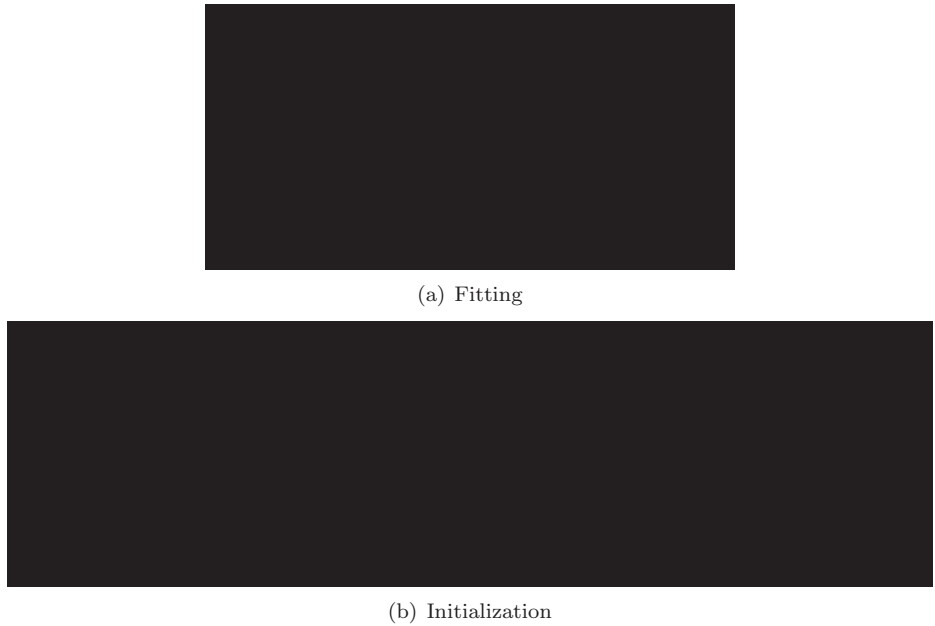

  \centering
  \subfigure[Fitting]{\includegraphics[width=200pt]{3DMM_2.eps}}
  \subfigure[Initialization]{\includegraphics[width=350pt]{3DMM_1.eps}}
  \caption[3D Morphable Model]
   { \it (a) Simultaneous reconstruction of 3D shape and texture of a
            new face from two images taken under different conditions. In the
            centre row, the 3D face is rendered on top of the input
            images~\cite{BlanVett1999}.
         (b) 3D Morphable Model Initialization: seven landmarks for front
            and side views and eight for the profile view are manually
            labelled for each input image \cite{LiJain2004}. }
  \label{Fig: 3DMM}
\end{figure}

An attractive feature of the 3D Morphable Model is that it
explicitly models both intrinsic and extrinsic variables,
respectively: face shape and albedo, and pose and illumination
parameters, see Figure~\ref{Fig: 3DMM}~(a). On the other hand, it
suffers from convergence problems in the presence of background
clutter or facial occlusions (glasses or facial hair). Furthermore
and importantly for the work presented in this thesis, the 3D
Morphable Model requires high quality image input
\cite{EverZiss2004} and struggles with non-Lambertian effects or
multiple light sources. Finally, nontrivial user intervention is
required (localization of up to seven facial landmarks and the
dominant light direction, see Figure~\ref{Fig: 3DMM}~(b)), the
fitting procedure is slow \cite{VettRomd2004} and can get stuck in
local minima \cite{LeeMoghPfisMach2004}.

\subsection{Image set and video-based methods.} Both appearance and
model-based methods have been applied to face recognition using
image set or video sequence matching. In principle, evidence from
multiple shots can be used to optimize model parameter recovery in
model-based methods and reduce the problem of local minima
\cite{EdwaTaylCoot1999}. However, an important limitation lies in
the computational demands of model fitting. Specifically, it is
usually too time consuming to optimize model parameters over an
entire sequence. If, on the other hand, parameter constraints are
merely propagated from the first frame, the fitting can experience
steady deterioration over time, the so-called \textit{drift}.

More variability in the manner in which still-based algorithms are
extended to deal with multi-image input is seen amongst
appearance-based methods. We have stated that the main limitation of
purely appearance-based recognition is that of limited
generalization ability, especially in the presence of greatly
varying illumination conditions. On the other hand, a promising fact
is that data collected from video sequences often contains some
variability in these parameters.

Eigenfaces, for example, have been used on a per-image basis, with
recognition decision cast using majority vote \cite{AranCipo2006}. A
similar voting approach was also successfully used with local
features in \cite{CampFeriCesa2000}, which were extracted by
tracking a face using a Gabor Wavelet Network
\cite{CampFeriCesa2000, KrugHappSomm2000, KrugSomm2002}. In
\cite{TorrLoreVila2000} video information is used only in the
training stage to construct person-specific PCA spaces,
\textit{self-eigenfaces}, while verification was performed from a
single image using the Distance from Feature Space criterion.
Classifiers using different eigenfeature spaces were used in
\cite{PricGree2001} and combined using the sum rule
\cite{KittHateDuinMata1998}. Better use of training data is made
with various discriminative methods such as Fisherfaces, which can
be used to estimate \emph{database-specific} optimal projection
\cite{EdwaTaylCoot1997}.

An interesting extension of appearance correlation-based recognition
to matching sets of faces was proposed by Yamaguchi \textit{et al.}
\cite{YamaFukuMaed1998}. The so-called Mutual Subspace Method (MSM)
has since gained considerable attention in the literature. In MSM,
linear subspaces describing appearance variations within sets or
sequences are matched using canonical correlations \cite{Gitt1985,
Hote1936, Kail1974, Oja1983}. It can be shown that this corresponds
to finding the most similar modes of variation between subspaces
\cite{KimAranCipo2007} (see Chapters~\ref{Chp: Filters}
and~\ref{Chp: BoMPA}, and Appendix~\ref{App: MPMM} for more detail
and criticism of MSM). A discriminative heuristic extension was
proposed in \cite{FukuYama2003} and a more rigourous framework in
\cite{KimKittCipo2006}. This group of methods typically performs
well when some appearance variation between training and novel input
is shared \cite{AranShakFish+2005, AranCipo2006}, but fail to
generalize in the presence of large illumination changes, for
example \cite{AranCipo2006b}. The same can be said of
the methods that use the temporal component to enforce prior
knowledge on likely appearance changes between consecutive frames.
In the algorithm of Zhou \textit{et al.} \cite{ZhouKrueChel2003} the
joint probability distribution of identity and motion is modelled
using sequential importance sampling, yielding the recognition
decision by marginalization. Lee \textit{et al.}
\cite{LeeHoYangKrie2003} approximate face manifolds by a finite
number of infinite extent subspaces and use temporal information to
robustly estimate the operating part of the manifold.

\subsection{Summary.}
\begin{figure}[!t]
  \centering
  \includegraphics[width=1.00\textwidth]{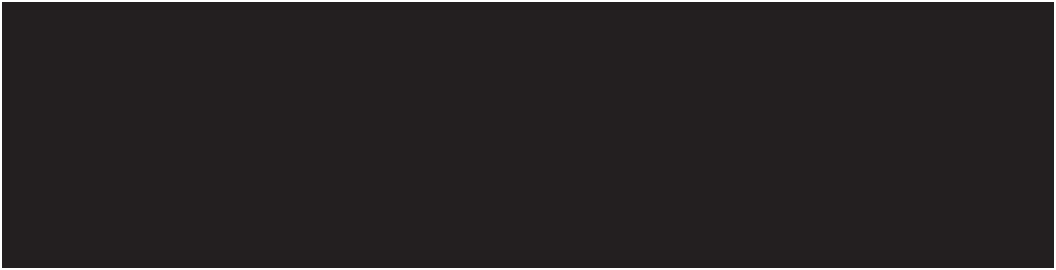}
  \caption[Summary of face recognition approaches.]
  { \it A summary of reviewed face recognition trends in the literature.
        Also see Table~\ref{Tab: Comp} for a comparative summary. }
  \label{Fig: Summ}
            \vspace{6pt}\hrule
\end{figure}

Amongst a great number of developed face recognition algorithms,
we've seen that two drastically different groups of approaches can
be identified: appearance-based and model-based, see
Figure~\ref{Fig: Summ}. The preceding section described the rich
variety of methods within each group and highlighted their
advantages and disadvantages. In closing, we summarize these in
Table~\ref{Tab: Comp}.

\begin{table}[!t]
  \Large
  \caption[Appearance and model-based approach comparison.]
     { \it A qualitative comparison of advantages and disadvantages of the
           two main groups of face recognition methods in the literature.\vspace{10pt} }
  \begin{tabularx}{1.00\textwidth}{c|X|X}
    \Hline
       & \normalsize \bf Appearance-based & \normalsize \bf Model-based\\
    \hline
      \rotatebox{90}{\normalsize \bf Advantages  \hspace{10pt}}
       & \vspace{-55pt} \small \begin{itemize}
           \item Well-understood statistical methods can be applied.
           \item Can be used for poor quality and low resolution input.
         \end{itemize}

       & \vspace{-55pt} \small \begin{itemize}
           \item Explicit modelling and recovery of personal and extrinsic variables.
           \item Prior, domain-specific knowledge is used.
         \end{itemize}
       \\
    \hline
      \rotatebox{90}{\normalsize \bf Disadvantages  \hspace{40pt}}
       & \vspace{-95pt} \small \begin{itemize}
           \item Lacking generalization to unseen pose, illumination etc.
           \item No (or little) use of domain-specific knowledge.
         \end{itemize}

       & \vspace{-95pt} \small \begin{itemize}
           \item High quality input is required.
           \item Model parameter recovery is time-consuming.
           \item Fitting optimization can get stuck in a local minimum.
           \item User intervention is often required for initialization.
           \item Difficult to model complex illumination effects --
           fitting becomes as ill-conditioned problem.
         \end{itemize}
       \\
    \Hline
  \end{tabularx}
  \label{Tab: Comp}
\end{table}

\section{Performance evaluation}\label{CH2: Perf}
To motivate different performance measures used across the
literature, it is useful to first consider the most common paradigms
in which face matching is used. These are
\begin{itemize}
  \item recognition -- 1-to-N matching,
  \item verification -- 1-to-1 matching, and
  \item retrieval.
\end{itemize}
In this context by the term ``matching'' we mean that the result of
a comparison of two face representations yields a scalar, numerical
score $d$ that measures their dissimilarity.

\begin{figure*}[!t]
  \centering
  \subfigure[1-to-N]{\includegraphics[width=0.4\textwidth]{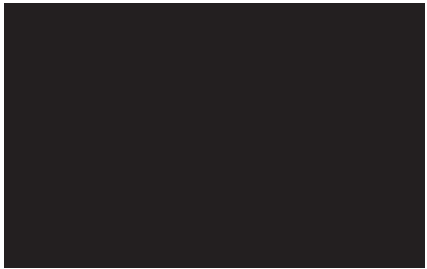}}
  \hspace{10pt}
  \subfigure[1-to-N]{\includegraphics[width=0.5\textwidth]{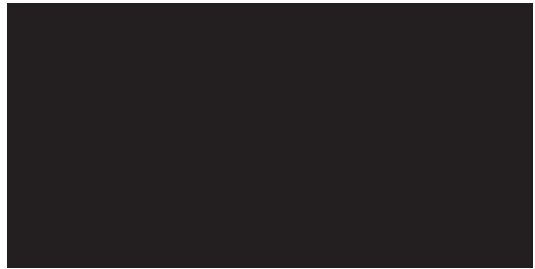}}
  \subfigure[Retrieval]{\includegraphics[width=0.8\textwidth]{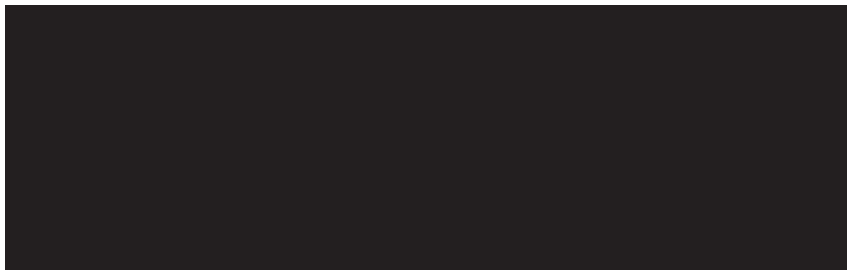}}
  \caption[Matching paradigms.]
    { \it Three matching paradigms give rise to different
          performance measures for face recognition algorithms. }
  \label{Fig: Matching}
  \vspace{6pt}\hrule
\end{figure*}

\paragraph{Paradigm 1: 1-to-N matching.}
In this setup novel input is matched to each of the individuals in a
database of known persons and classified to -- \emph{recognized} as
-- the closest, most similar one. One and only one correct
correspondence is assumed to exist. This is illustrated in
Figure~\ref{Fig: Matching}~(a).

When 1-to-N matching is considered, the most natural and often used
performance measure is the \emph{recognition rate}. We define it as
the ratio of the number of correctly assigned to the total number of
test persons.

\paragraph{Paradigm 2: 1-to-1 matching.}
In 1-to-1 matching, only a single comparison is considered at a time
and the question asked is if two people are the same. This is
equivalent to thresholding the dissimilarity measure $d$ used for
matching, see Figure~\ref{Fig: Matching}~(b).

Given a particular distance threshold $d^*$, the \emph{true positive
rate} (TPR) $p_t(d^*)$ is the proportion of intra-personal
comparisons that yields distances within the threshold. Similarly,
the \emph{false positive rate} (FPR) $p_f(d^*)$ is the proportion of
inter-personal comparisons that yields distances within the
threshold. As $d^*$ is varied, the changes in $p_t(d^*)$ and
$p_f(d^*)$ are often visualized using the so-called
\emph{Receiver-Operator Characteristic} (ROC) curve, see
Figure~\ref{Fig: ROC}.

\begin{figure*}[!t]
  \centering
  \includegraphics[width=0.5\textwidth]{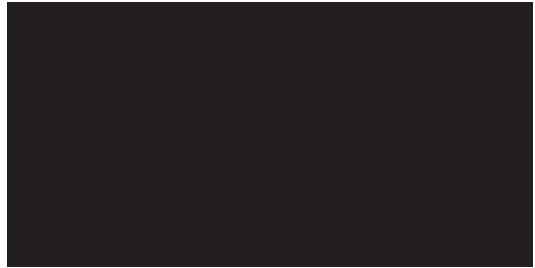}
  \caption[Receiver-Operator Characteristic curve.]
    { \it The variations in the true positive and false positive rates,
          as functions of the distance threshold, are often visualized
          using the Receiver-Operator Characteristic (ROC) curve, by
          plotting them against each other. Shown is a family of ROC
          curves and the Equal Error Rate (EER) line. Better
          performing algorithms have ROC curves closer to the 100\%
          true positive and 0\% false positive rate. }
  \label{Fig: ROC}
  \vspace{6pt}\hrule
\end{figure*}

The \emph{Equal Error Rate} (EER) point of the ROC curve is
sometimes used for brevity:
\begin{align}
  EER = p_f(d_{EER}) & \text{, where: } p_f(d_{EER}) = 1 - p_t(d_{EER}),
\end{align}
see Figure~\ref{Fig: ROC}.

\paragraph{Paradigm 3: retrieval.}
In the retrieval paradigm the novel person is now a query to the
database, which may contain several instances of any individual. The
result of a query is an \emph{ordering} of the entire database using
the dissimilarity measure $d$, see Figure~\ref{Fig: Matching}~(c).
More successful orderings have instances of the query individual
first (i.e.\ with a lower recall index).

From the above, it can seen that the normalized sum of indexes
corresponding to in-class faces is a meaningful measure of the
recall accuracy. We call this the rank ordering score and compute it
as follows:
\begin{align}\label{Eqn: OrdScore}
    \rho = 1 - \frac {S-m} {M},
\end{align}
where $S$ is the sum of indexes of retrieved in-class faces, and $m$
and $M$, respectively, the minimal and maximal values $S$ and
$(S-m)$ can take.

The score of $\rho=1.0$ corresponds to orderings which correctly
cluster all the data (all the in-class faces are recalled first),
0.0 to those that invert the classes (the in-class faces are
recalled last), while 0.5 is the expected score of a random
ordering. The {\em average normalized rank} \cite{SaltMcGi1983} is
equivalent to $1 - \rho$.

\subsection{Data}
Most algorithms in this thesis were evaluated on three large data sets of
video sequences -- the \textit{CamFace}, \textit{ToshFace} and
\textit{Face Video Database}. These are briefly desribed next. Other
data, used only in a few specific chapters, is explained in the
corresponding evaluation sections. The algorithm we used to autmatically
extract faces from video is described in
Appendix~\ref{AppC:AutomaticExtraction}.

\begin{figure*}[!t]
  \centering
  \subfigure[\textit{CamFace}]
  { \includegraphics[width=0.97\textwidth]{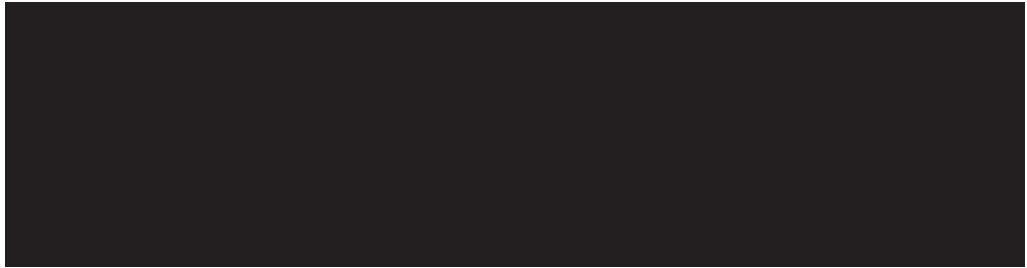} }
  \subfigure[\textit{ToshFace}]
  {
  \includegraphics[width=0.98\textwidth]{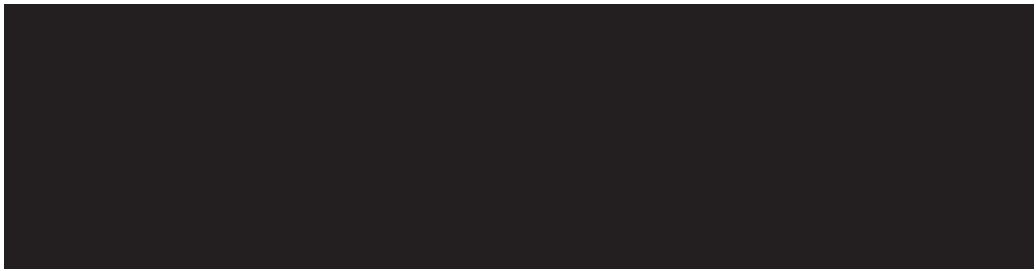} }
  \subfigure[\textit{Face Video DB}]
  {
  \includegraphics[width=1.0\textwidth]{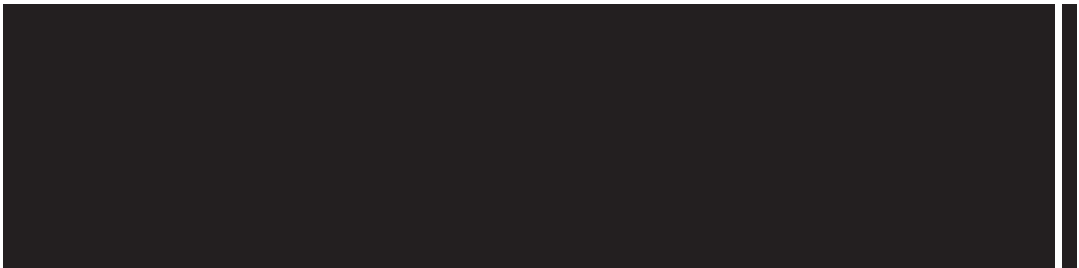} }

  \caption[Frames from our video databases.]{ \it
            Frames from typical video sequences from the 3 databases used
            for evaluation of most recognition algorithms in this thesis. }
  \label{Fig: Faces Input}
  \vspace{6pt}\hrule
\end{figure*}

\begin{figure}[!t]
  \centering
  \subfigure[FaceDB100]{
  \includegraphics[width=0.76\textwidth]{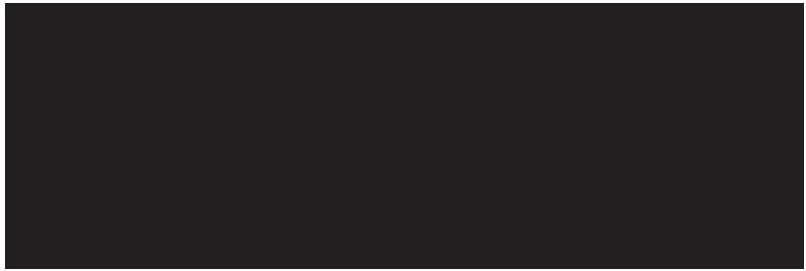}}
  \subfigure[FaceDB60]{
  \includegraphics[width=0.95\textwidth]{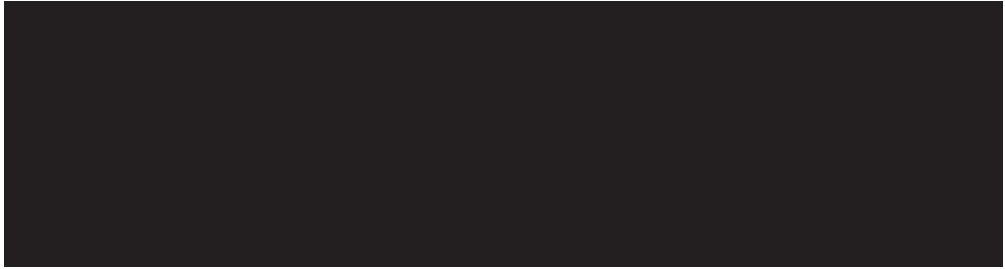}}
  \caption[Illuminations in \textit{CamFace} and \textit{ToshFace} data sets]
  { \it (a) Illuminations 1--7 from CamFace data set and
        (b) illuminations 1--10 from ToshFace data set. }
  \label{Fig: Illuminations}
            \vspace{6pt}\hrule
\end{figure}

\paragraph{The \textit{CamFace} dataset.}
This database contains 100 individuals of varying age and ethnicity,
and equally represented genders. For each person in the database
there are 7 video sequences of the person in arbitrary motion
(significant translation, yaw and pitch, negligible roll), each in a
different illumination setting, see Fig.~\ref{Fig: Faces Input}~(a)
and~\ref{Fig: Illuminations}, for 10s at 10fps and $320 \times 240$ pixel
resolution (face size $\approx 60$ pixels). For more information see
Appendix~\ref{App: CamFace} in which this database is thoroughly
described.

\paragraph{The \textit{ToshFace} dataset.}
This database was kindly provided to us by Toshiba Corporation. It
contains 60 individuals of varying age, mostly male Japanese, and 10
sequences per person. Each 10s sequence corresponds to a different
illumination setting, acquired at 10fps and $320 \times 240$ pixel resolution
(face size $\approx 60$ pixels), see Fig.~\ref{Fig: Faces
Input}~(b).

\paragraph{The \textit{Face Video Database.}}
This database is freely available and described in \cite{Goro2005a}.
Briefly, it contains 11 individuals and 2 sequences per person,
little variation in illumination, but extreme and uncontrolled
variations in pose, acquired for 10-20s at 25fps and $160 \times 120$ pixel
resolution (face size $\approx 45$ pixels), see Fig.~\ref{Fig: Faces
Input}~(c).

\section{Summary and conclusions} \label{Conclusions}
This chapter finished laying out the foundations for understanding
the novelty of this thesis. The challenges of face recognition were
explored by presenting a detailed account of previous research
attempts at solving the problem at hand. It was established that
both major methodologies, discriminative model-based and generative
model-based, suffer from serious limitations when dealing with data
acquired in realistic, practical conditions.

The second part of the chapter addressed the issue of evaluating and
comparing face recognition algorithms. We described a number of
useful performance measures and three large data sets that will be
used extensively throughout this work.

\part{Access Control}

\graphicspath{{./03mDiv/}}
\chapter{Manifold Density Divergence}
\label{Chp: MDD}
\newcommand{\pface}[1]{\ensuremath{p^{(#1)}}}

\begin{center}
  \footnotesize
  \vspace{-20pt}
  \framebox{\includegraphics[width=0.55\textwidth]{title_img.eps}}\\
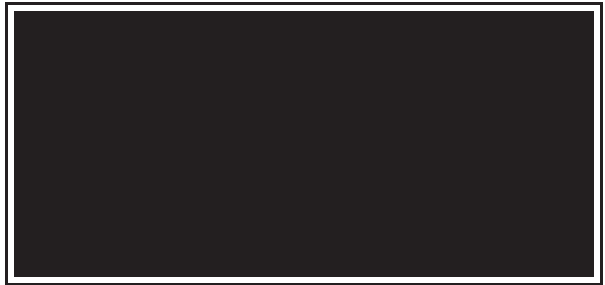
  Albrecht D\"{u}rer. \textit{Melancholia}\\
  1514, Engraving, 24.1 x 18.8 cm\\
  Albright-Knox Art Gallery, Buffalo
\end{center}

\cleardoublepage

The preceding two chapters introduced the problem of face
recognition from video, placed it into the context of
biometrics-based identification methods and current practical
demands on them, in broad strokes describing relevant research with
its limitations. In this chapter we adopt the
\emph{appearance-based} recognition approach and set up the grounds
for most of the material in the chapters that follow by formalizing
the face manifold model. The first contribution of this thesis is
also introduced -- the \emph{Manifold Density Divergence (MDD)}
algorithm.

Specifically, we address the problem of matching a novel face video
sequence to a set of faces containing typical, or expected,
appearance variations. We propose a flexible, semi-parametric model
for learning probability densities confined to highly non-linear but
intrinsically low-dimensional manifolds. The model leads to a
statistical formulation of the recognition problem in terms of
minimizing the divergence between densities estimated on these
manifolds. The proposed method is evaluated on the \textit{CamFace}
data set and is shown to match the best and outperform other
state-of-the-art algorithms in the literature, achieving 94\%
recognition rate on average.

\section{Introduction}
Training a system in certain imaging conditions (single
illumination, pose and motion pattern) and being able to recognize
under arbitrary  changes in these conditions can be considered to be
the hardest problem formulation for automatic face recognition.
However, in many practical applications this is too strong of a
requirement. For example, it is often possible to ask a subject to
perform random head motion under varying illumination conditions. It
is often not reasonable, however, to request that the user perform a
strictly defined motion, assume strictly defined poses or illuminate
the face with lights in a specific setup. We therefore assume that
the training data available to an AFR system is organized in a
database where a \emph{set} of images for each individual represents
significant (typical) variability in illumination and pose, but does
not exhibit temporal coherence and is not obtained in scripted
conditions.

The test data -- that is, the input to an AFR system -- also often
consist of a \emph{set} of images, rather than a single image. For
instance, this is the case when the data is extracted from
surveillance videos. In such cases the recognition problem can be
formulated as taking a set of face images from an unknown individual
and finding the best matching set in the database of labelled sets.
This is the recognition paradigm we are concerned with in this
chapter.

We approach the task of recognition with image sets from a
statistical perspective, as an instance of the more general task of measuring similarity
between two probability density functions that generated two sets
of observations. Specifically, we model these densities as  Gaussian Mixture
Models (GMMs) defined on low-dimensional nonlinear manifolds
embedded in the image space, and evaluate the similarity between
the estimated densities via the Kullback-Leibler divergence. The
divergence, which for GMMs cannot be computed in closed form, is
efficiently evaluated by a Monte Carlo algorithm.

In the next section, we introduce our model and discuss the proposed
method for learning and comparing face appearance manifolds.
Extensive experimental evaluation of the proposed model and its
comparison to state-of-the-art methods are reported in
Section~\ref{sec:experiments}, followed by discussion of the results
and a conclusion.

\section{Modelling face manifold densities}\label{sec:manifold}
Under the standard representation of an image as a raster-ordered
pixel array, images of a given size can be viewed as points in a
Euclidean \emph{image space}. The dimensionality, $D$, of this space
is equal to the number of pixels. Usually $D$ is high enough to
cause problems associated with the \emph{curse of dimensionality} in
learning and estimation algorithms. However, surfaces of faces are
mostly smooth and have regular texture, making their appearance
quite constrained. As a result, it can be expected that face images
are confined to a \emph{face space}, a manifold of lower dimension
$d\ll D$ embedded in the image space \cite{BichPent1994}. We next
formalize this notion and propose an algorithm for comparing
estimated densities on manifolds.

\subsection{Manifold density model} 
\begin{figure*}[htbp]
\centering 

\subfigure[First three PCs]{
\includegraphics[width=.65\textwidth]{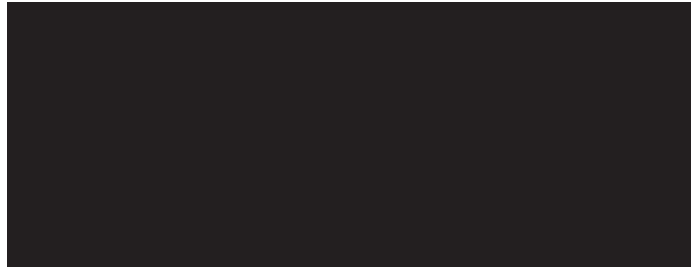}}
\subfigure[Second three PCs]{
\includegraphics[width=.65\textwidth]{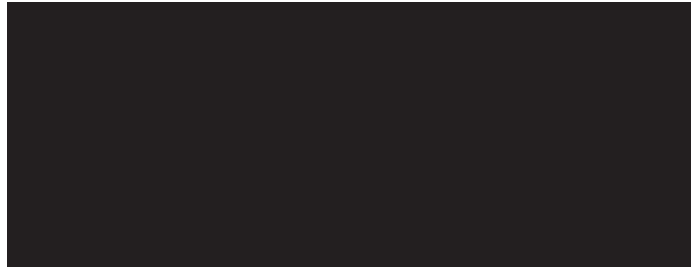}}

\caption[Appearance manifolds.]
         { \it   A typical manifold of face images in a training
                (small blue dots) and a test (large red dots) set. Data used come
                from the same person and  shown projected to the first three
                (a) and second three (b) principal components. The
                nonlinearity and smoothness of the manifolds are apparent.
                Although globally quite dissimilar, the training
                and test manifolds have locally similar structures.
            }
            \label{fig:manifolds}
\end{figure*}

The assumption of an
underlying manifold subject to additive sensor noise leads to the
following statistical model: An image $\mathbf{x}$ of subject
$i$'s face is drawn from the probability density function
(\emph{pdf}) $p_F^{(i)}(\mathbf{x})$ within the face space, and
embedded in the image space by means of a mapping function
$f^{(i)}:\mathbb{R}^d\to\mathbb{R}^D$. The resulting point in the
$D$-dimensional space is further perturbed by noise drawn from a
noise distribution $p_n$ (note that the noise operates in the
image space) to form the observed image $\mathbf{X}$. Therefore
the distribution of the observed face images of the subject $i$ is
given by:
\begin{equation}\label{eq:manifold_pdf}
\pface{i}(\mathbf{X}) = \int p_F^{(i)}(\mathbf{x})
p_n\left(f^{(i)}(\mathbf{x}) - \mathbf{X}\right) d\mathbf{x}
\end{equation}
Note that both the manifold embedding function $f$ and the density
$p_F$ on the manifold are subject-specific, as denoted by the
superscripts, while the noise distribution $p_n$ is assumed to be
common for all subjects. Following accepted practice, we model
$p_n$ by an isotropic, zero-mean Gaussian.
Figure~\ref{fig:manifolds} shows an example of a face image set
projected onto a few principal components estimated from the data,
and illustrates the validity of the manifold notion.

Let the training database consist of sets $S_1,\ldots,S_K$,
corresponding to $K$ individuals. $S_i$ is assumed to be a set of
independent and identically distributed (i.i.d.) observations drawn
from $\pface{i}$~\eqref{eq:manifold_pdf}. Similarly, the input set
$S_0$ is assumed to be i.i.d.~drawn from the test subject's face
image density $\pface{0}$. The recognition task can then be
formulated as selecting one among $K$ hypotheses, the $k$-th
hypothesis postulating that $\pface{0}=\pface{k}$. The
Neyman-Pearson lemma \cite{DudaHartStor2001} states that the optimal
solution for this task consists of choosing the model under which
$S_0$ has the highest likelihood. Since the underlying densities are
unknown, and the number of samples is limited, relying on direct
likelihood estimation is problematic. Instead, we use
Kullback-Leibler divergence as a ``proxy'' for the likelihood
statistic needed in this $K$-ary hypothesis test
\cite{ShakFishDarr2002}.



\subsection{Kullback-Leibler divergence}\label{sec:kld} 
\begin{figure}
  \centering
  \includegraphics[width=.6\textwidth]{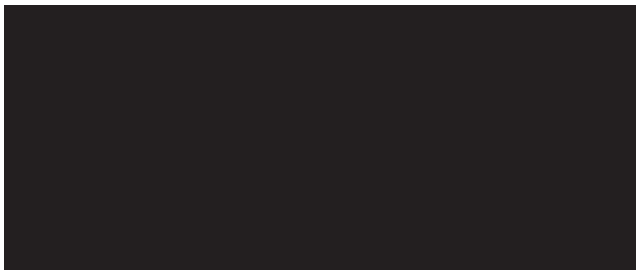}
  \caption[GMM description lengths.]
      { \it Description lengths for varying numbers of GMM
            components for training (solid) and test (dashed)
            sets. The lines show the average plus/minus one standard
            deviation across sets.}
            \label{Figure: DLs}
  \vspace{6pt}\hrule
\end{figure}

The
Kullback-Leibler (KL) divergence \cite{CoveThom1991} quantifies how
well a particular pdf $q(\mathbf{x})$ describes samples from anther
pdf $p(\mathbf{x})$:

\begin{equation}\label{Equation: KLD Definition}
    D_{KL}(p||q) = \int p(\mathbf{x})\log\left(\frac{p(\mathbf{x})}{q(\mathbf{x})}\right)d\mathbf{x}
\end{equation}
It is nonnegative and equal to zero iff $p \equiv q$.
Consider the integrand in~\eqref{Equation: KLD Definition}. It can
be seen that the regions of the image space
with a large contribution to the divergence are those in which
$p(\mathbf{x})$ is significant and $p(\mathbf{x}) \gg
q(\mathbf{x})$. On the other hand, regions in which $p(\mathbf{x})$
is small contribute comparatively little. We expect the sets in the
training data to be significantly more extensive than the input set,
and as a result $\pface{i}$ to have broader support than
$\pface{0}$. We therefore use $D_{KL}(\pface{0}||\pface{i})$ as a
``distance measure'' between training and test sets. This
expectation is confirmed empirically, see Figure~\ref{Figure: DLs}.
The novel patterns not represented in the training set are heavily
penalized, but there is no requirement that all variation seen
during training should be present in the novel distribution.

We have formulated recognition in terms of minimizing the divergence
between densities on face manifolds. Two problems still remain to
be solved. First, since the analytical form for
neither the densities nor the embedding functions is known,
these must be estimated from the data. Second, the KL divergence
between the estimated densities must be evaluated. In the remainder of
this section,
we describe our solution for these two problems.

\subsection{Gaussian mixture models}\label{sec:gmms} Our goal is
to estimate the density defined on a complex nonlinear manifold
embedded in a high-dimensional image space. Global parametric models
typically fail to adequately capture such manifolds. We therefore
opt for a more flexible mixture model for $\pface{i}$: the Gaussian
Mixture Model (GMM). This choice has a number of advantages:
\begin{itemize}
  \item It is a flexible, semi-parametric
  model, yet simple enough to allow efficient estimation.
    \item The model is generative and offers interpolation and
    extrapolation of face pattern variation based on local
    manifold structure.
    \item Principled model order selection is possible.
\end{itemize}

The multivariate Gaussian components of a GMM in our method need not
be semantic (corresponding to a specific view or illumination) and
can be estimated using the Expectation Maximization (EM) algorithm
\cite{DudaHartStor2001}. The EM is initialized by K-means
clustering, and constrained to diagonal covariance matrices. As with
any mixture model, it is important to select an appropriate number
of components in order to allow sufficient flexibility while
avoiding overfitting. This can be done in a principled way with the
Minimal Description Length (MDL) criterion \cite{BarrRissYu1998}.
Briefly, MDL assigns to a model a cost related to the amount of
information necessary to encode the model and the data \emph{given}
the model. This cost, known as the description length, is
proportional to the likelihood of the training data under that model
penalized by the model complexity, measured as the number of free
parameters in the model.

\begin{figure}
  \footnotesize
  \centering
  \includegraphics[width=0.90\textwidth]{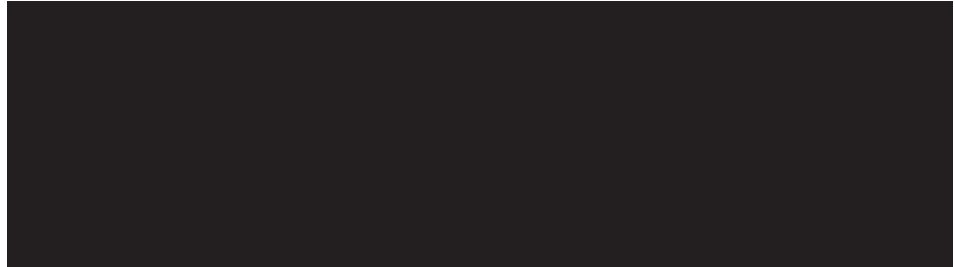}\\
  (a)\\ \vspace{10pt}
  \includegraphics[width=0.52\textwidth]{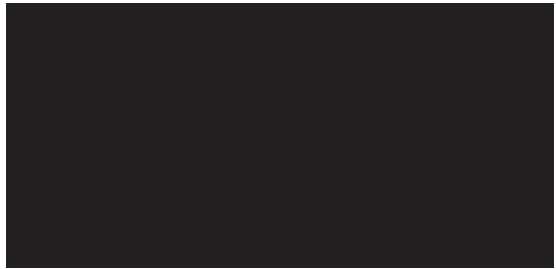} \\ (b)
  \caption[Training and test manifold models.]
      { \it Centres of the MDL GMM approximation to a typical
            training face manifold, displayed as images (a) (also
            see Figure~\ref{Figure: Manifold GMM}). These
            appear to correspond to different pose/illumination
            combinations. Similarly, centres for a typical face
            manifold used for recognition are shown in (b). As this
            manifold corresponds to a video in fixed illumination,
            the number of Gaussian clusters is much smaller.
            In this case clusters correspond to
            different poses only: frontal, looking down, up, left and right.}
            \label{Figure: Training Centres}
\end{figure}

\begin{figure}
  \centering
  \includegraphics[width=0.95\textwidth]{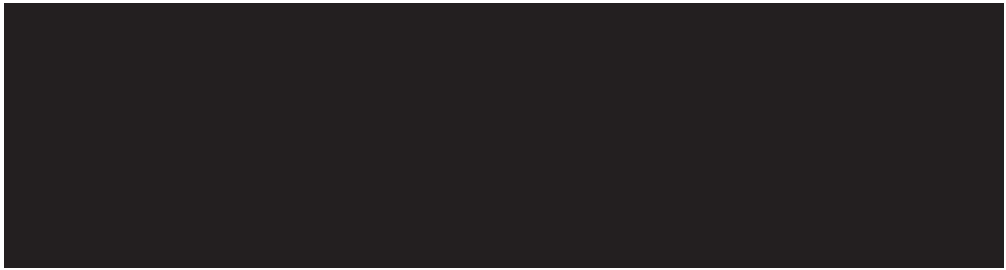}
  \caption[Appearance variations captured by the manifold model.]
      { \it Synthetically generated images from a single Gaussian
            component in a GMM of a training image set. It can be seen
            that local manifold structure, corresponding to varying head
            pose in fixed illumination, is well captured. }
            \label{fig:synthetic}
  \vspace{6pt}\hrule
\end{figure}

\begin{figure}
  \centering
  \includegraphics[width=.6\textwidth]{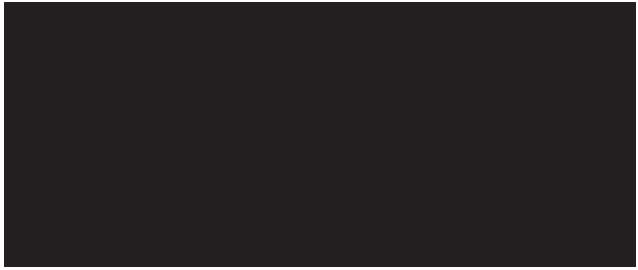}\\
  \caption[Training data and the manifold model in 3PC space.]
      { \it A training face manifold (blue dots) and the centres
            of Gaussian clusters of the corresponding MDL GMM
            model of the data (circles), projected on the first
            three principal components. }
            \label{Figure: Manifold GMM}
  \vspace{6pt}\hrule
\end{figure}

Average description lengths for different numbers of components for
the data sets used in this chapter are shown in Figure~\ref{Figure:
DLs}. Typically, the optimal (in the MDL sense) number of components
for a training manifold was found to be 18, while 5 was typical for
the manifolds used for recognition. This is illustrated in
Figures~\ref{Figure: Training Centres},~\ref{fig:synthetic}
and~\ref{Figure: Manifold GMM}.

\subsection{Estimating KL divergence}\label{sec:kld_compute}
Unlike in the case of Gaussian distributions, the KL divergence
cannot be computed in a closed form when $\hat{p}(\mathbf{x})$ and
$\hat{q}(\mathbf{x})$ are GMMs. However, it is straightforward to
sample from a GMM. The KL divergence in~\eqref{Equation: KLD
Definition} is the expectation of the log-ratio of the two densities
w.r.t. the density $p$. According to the law of large numbers
\cite{GrimStir1992}, this expectation can be evaluated by a
Monte-Carlo simulation. Specifically, we can draw a sample
$\mathbf{x}_i$ from the estimated density $\hat{p}$, compute the
log-ratio of $\hat{p}$ and $\hat{q}$, and average this over $M$
samples:
\begin{equation}\label{Equation: Estimating KL Divergence 1}
    D_{KL}(\hat{p} || \hat{q}) \approx \frac {1}{M}
    \sum_{i=1}^M log \left( \frac {\hat{p}(\mathbf{x}_i)} {\hat{q}(\mathbf{x}_i)} \right)
\end{equation}

Drawing from $\hat{p}$ involves selecting a GMM component and then
drawing a sample from the corresponding multi-variate
Gaussian. Figure~\ref{fig:synthetic} shows a few examples of samples
drawn in this manner.
In summary, we
use the following approximation for the KL divergence between the test
set and the $k$-th subject's training set:

\begin{equation}\label{Equation: Estimating KL Divergence 2}
    D_{KL}\left(\hat{p}^{(0)} || \hat{p}^{(k)}\right) \approx \frac {1} {M}
    \sum_{i=1}^M log \left( \frac {\hat{p}^{(0)}(\mathbf{x}_i)} {\hat{p}^{(k)}(\mathbf{x}_i)} \right)
\end{equation}
In our experiments we used $M=1000$ samples.

\section{Empirical evaluation}\label{sec:experiments}
We compared the performance of our recognition algorithm on the
\textit{CamFace} data set to that of:
\begin{itemize}
    \item KL divergence-based algorithm of Shakhnarovich et al. (Simple KLD) \cite{ShakFishDarr2002},

    \item Mutual Subspace Method (MSM) \cite{YamaFukuMaed1998},

    \item Constrained MSM (CMSM) \cite{FukuYama2003}  which projects
    the data onto a linear subspace before applying MSM,

    \item Nearest Neighbour (NN) in the set distance sense; that
    is, achieving\\$\min_{\mathbf{x}\in S_0}\min_{\mathbf{y}\in S_i} d(\mathbf{x},\mathbf{y})$.
\end{itemize}

In Simple KLD, we used a principal subspace that captured 90\% of
the data variance. In MSM, the dimensionality of PCA subspaces was
set to 9 \cite{FukuYama2003}, with the first three principal angles
used for recognition. The constraint subspace dimensionality in CMSM
(see \cite{FukuYama2003}) was chosen to be 70. All algorithms were
preceded with PCA performed on the entire dataset, which resulted in
dimensionality reduction to 150 (while retaining 95\% of the
variance).

In each experiment we used all of the sets from one illumination
setup as test inputs and the remaining sets as training data, see
Appendix~\ref{App: CamFace}.

\subsection{Results}\label{sec:results}
A summary of the experimental results is shown in Table~\ref{Figure:
Results}. Notice the relatively good performance of the simple NN
classifier. This supports our intuition that for training, even
random illumination variation coupled with head motion is sufficient
for gathering a representative set of samples from the
illumination-pose face manifold.

\begin{table}
  \centering
  \Large
  \caption[Recognition results.]{\it Recognition accuracy (\%) of the various methods using
           different training/testing illumination combinations.\vspace{10pt}}\label{Figure: Results}
  \begin{tabular*}{1.0\textwidth}{@{\extracolsep{\fill}}ll|ccccc}
    \Hline
    \bf \normalsize ~Method                             &                & \normalsize MDD  & \normalsize  Simple KLD & \normalsize MSM  & \normalsize CMSM  & \normalsize  Set NN \\
    \hline
    \multirow{2}{*}{\bf  \normalsize ~Recognition rate}  & \normalsize mean & \small           94     & \small     69    & \small 83         & \small   92  & \small 89\\
                                                         & \normalsize std  & \small            8     & \small      5    & \small 10         & \small    7  & \small 9\\
    \Hline
  \end{tabular*}
\end{table}

Both MSM-based methods scored relatively well, with CMSM
achieving the best performance of all of the algorithms besides the proposed
method. That is an interesting result, given that this algorithm
has not received significant attention in the AFR community; to
the best of our knowledge, this is the first report of CMSM's
performance on a data set of this size, with such illumination and
pose variability. On the other hand, the lack of a probabilistic
model underlying CMSM may make it somewhat less appealing.

Finally, the performance of the two statistical methods evaluated,
the Simple KLD method and the proposed algorithm, are
very interesting. The former performed worst, while the latter
produced the highest recognition rates out of the methods
compared. This suggests several conclusions. Firstly, that the
approach to statistical modelling of manifolds of faces is a
promising research direction. Secondly, it is confirmed that our
flexible GMM-based model captures the modes of the data variation
well, producing good generalization results even when the test
illumination is not present in the training data set. And lastly,
our argument in Section~\ref{sec:manifold} for the choice of the
direction of KL
divergence is empirically confirmed, as our method performs well
even when the subject's pose is only very loosely controlled.

\section{Summary and conclusions} \label{sec:summary} In this
chapter we introduced a new statistical approach to face recognition
with image sets. Our main contribution is the formulation of a
flexible mixture model that is able to accurately capture the modes
of face appearance under broad variation in imaging conditions. The
basis of our approach is the semi-parametric estimate of probability
densities confined to intrinsically low-dimensional, but highly
nonlinear face manifolds embedded in the high-dimensional image
space. The proposed recognition algorithm is based on a stochastic
approximation of Kullback-Leibler divergence between the estimated
densities. Empirical evaluation on a database with 100 subjects has
shown that the proposed method, integrated into a practical
automatic face recognition system, is successful in recognition
across illumination and pose. Its performance was shown to match the
best performing state-of-the-art method in the literature and exceed
others.

\section*{Related publications}

The following publications resulted from the work presented in this
chapter:

\begin{itemize}
  \item O. Arandjelovi{\'c}, G. Shakhnarovich, J. Fisher, R. Cipolla, and T. Darrell.
                  Face recognition with image sets using manifold density divergence. In
                  \textit{Proc. IEEE Conference on Computer Vision and Pattern Recognition (CVPR)},
                  \textbf{1}:pages 581--588, June 2005.
                  \cite{AranShakFish+2005}
\end{itemize}

\graphicspath{{./04krad/}}
\chapter{Unfolding Face Manifolds}
\label{Chp: KRAD}
\begin{center}
  \vspace{-20pt}
  \footnotesize
  \framebox{\includegraphics[width=0.50\textwidth]{title_img.eps}}\\\vspace{20pt}
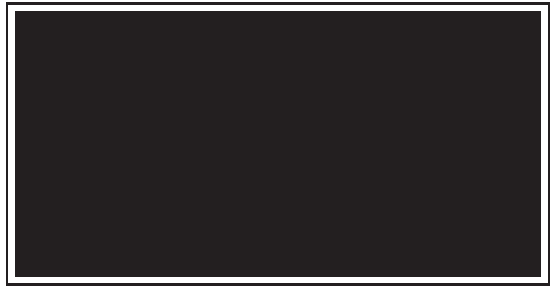
  Athanadoros, Hagesandros, and Polydoros of Rhodes. \textit{Laoco\"{o}n and His Sons}\\
  Early 1st century, Marble\\
  Museo Pio Clementino, Vatican\\
  \vspace{50pt}
\end{center}

\cleardoublepage

In the previous chapter we addressed the problem of matching a novel
face video sequence to a set of faces containing typical, or
expected, appearance variations. In this chapter we move away from
the assumption of having available such a large training corpus and
instead match a novel sequence against a database which too contains
only a single sequence per known individual. To solve the problem we
propose the \emph{Robust Kernel RAD} algorithm.

Following the adopted appearance-based approach, we motivate the use
of the Resistor-Average Distance (RAD) as a dissimilarity measure
between densities corresponding to appearances of faces in a single
sequence. We then introduce a kernel-based algorithm that makes use
of the simplicity of the closed-form expression for RAD between two
Gaussian densities, while allowing for modelling of complex but
intrinsically low-dimensional face manifolds. Additionally, it is
shown how geodesically local appearance manifold structure can be
modelled, naturally leading to a stochastic algorithm for
generalizing to unseen modes of data variation. On the
\textit{CamFace} data set our method is demonstrated to exceed the
performance of state-of-the-art algorithms achieving the correct
recognition rate of 98\% in the presence of mild illumination
variations.

\section{Dissimilarity between manifolds}
\begin{figure*}[!h]
  \centering
  \begin{tabular}{c}
    \includegraphics[width=240pt]{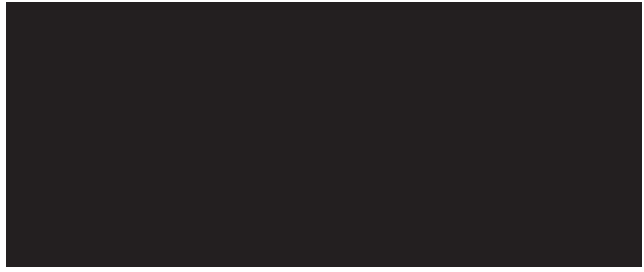}\vspace{5pt}\\
    \footnotesize (a)\vspace{30pt}\\
    \includegraphics[width=240pt]{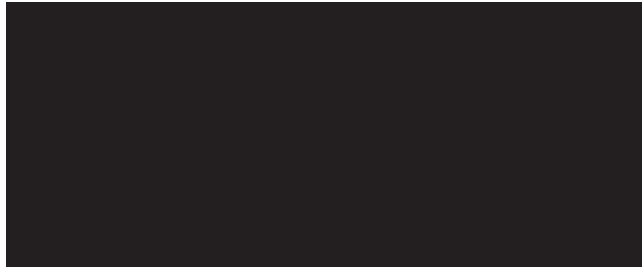}\vspace{10pt}\\
    \footnotesize (b)\\
  \end{tabular}

  \caption[KL divergence asymmetry and RAD.]
      { \it A 1D illustration of the asymmetry of the KL divergence (a). $D_{KL}(q||p)$ is an order of
            magnitude greater than $D_{KL}(p||q)$ -- the ``wider'' distribution $q(\mathbf{x})$
            explains the ``narrower'' $p(\mathbf{x})$ better than the other way round.
            In (b), $D_{RAD}(p,q)$ is plotted as a function of $D_{KL}(p||q)$ and $D_{KL}(q||p)$.}
            \label{Fig: KLD}
\end{figure*}

Consider the Kullback-Leibler $D_{KL}(p||q)$ divergence employed in
Chapter~\ref{Chp: MDD}. As previously discussed, the regions of the
observation space that produce a large contribution to
$D_{KL}(p||q)$ are those that are well explained by $p(\mathbf{x})$,
but not by $q(\mathbf{x})$. The asymmetry of the KL divergence makes
it suitable in cases when it is known a priori that one of the
densities $p(\mathbf{x})$ or $q(\mathbf{x})$ describes a wider range
of data variation than the other. This is conceptually illustrated
in Figure~\ref{Fig: KLD}~(a).

\begin{figure*}[!t]
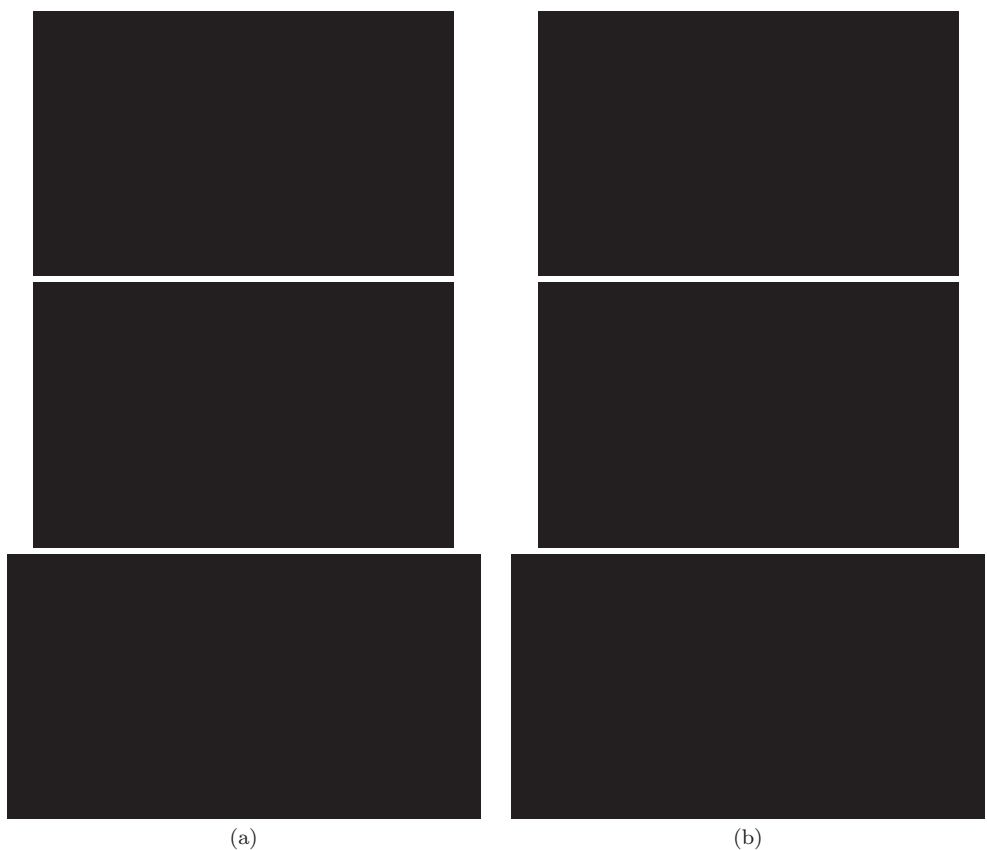

  \footnotesize
  \centering
  \begin{tabular}{cc}
    \includegraphics[width=0.4\textwidth]{BMVC2005_data30_1.ps} &
    \includegraphics[width=0.4\textwidth]{BMVC2005_data83_1.ps} \\
    \includegraphics[width=0.4\textwidth]{BMVC2005_data30_2.ps} &
    \includegraphics[width=0.4\textwidth]{BMVC2005_data83_2.ps} \\
    \includegraphics[width=0.45\textwidth]{BMVC2005_manifold_30.eps} &
    \includegraphics[width=0.45\textwidth]{BMVC2005_manifold_83.eps}\\
    (a) & (b) \\
  \end{tabular}
  \caption[Example appearance manifolds.]
      { \it A subset of 10 samples from two typical face sets used to illustrate
            concepts addressee in this chapter (top) and the corresponding
            patterns in the 3D principal component subspaces (bottom),
            estimated from data. The sets capture appearance changes of
            faces of two different individuals as they performed
            unconstrained head motion in front of a fixed camera. The
            corresponding pattern variations (blue circles) are highly
            nonlinear, with a number of outliers present (red stars).
            }
            \label{Fig: Input Data Sets}
  \vspace{6pt}\hrule
\end{figure*}

However, in the proposed recognition framework, this is not the case
-- pitch and yaw changes of a face are expected to be the dominant
modes of variation in both training and novel data, see
Figure~\ref{Fig: Input Data Sets}. Additionally, exact head poses
assumed by the user are expected to somewhat vary from sequence to
sequence and the robustness to variations not seen in either is
desired. This motivates the use of a symmetric ``distance'' measure.

\subsection{Resistor-Average distance.} We propose to use the
Resistor-Average distance (RAD) as a measure of dissimilarity
between two probability densities. It is defined as:
\begin{equation}\label{RAD Definition}
    D_{RAD}(p,q) \doteq \left[D_{KL}(p||q)^{-1} + D_{KL}(q||p)^{-1}\right]^{-1}
\end{equation}

Much like the KL divergence from which it is derived, it is
nonnegative and equal to zero iff $p(\mathbf{x}) \equiv
q(\mathbf{x})$, but unlike it, it is symmetric. Another important
property of the Resistor-Average distance is that when two classes
of patterns $\mathcal{C}_p$ and $\mathcal{C}_q$ are distributed
according to, respectively, $p(\mathbf{x})$ and $q(\mathbf{x})$,
$D_{RAD}(p,q)$ reflects the error rate of the Bayes-optimal
classifier between $\mathcal{C}_p$ and $\mathcal{C}_q$
\cite{JohnSina2001}.

To see in what manner RAD differs from the KL divergence, it is
instructive to consider two special cases: when divergences in
both directions between two pdfs are approximately equal and when
one of them is much greater than the other:

\begin{itemize}
    \item $D_{KL}(p||q) \approx D_{KL}(q||p) \equiv D$\\
            \begin{equation}\label{Eq: RAD case 1}
                D_{RAD}(p, q) \approx D/2
            \end{equation}

    \item $D_{KL}(p||q) \gg D_{KL}(q||p)$ or\\ $D_{KL}(p||q) \ll D_{KL}(q||p)$\\
            \begin{equation}\label{Eq: RAD case 2}
                D_{RAD}(p, q) \approx
                \min \left(D_{KL}(p||q),D_{KL}(q||p)\right)
            \end{equation}
\end{itemize}
It can be seen that RAD very much behaves like a smooth min of
$D_{KL}(p||q)$ and $D_{KL}(q||p)$ (up to a multiplicative constant),
also illustrated in Figure~\ref{Fig: KLD}~(b).

\section{Estimating RAD for nonlinear densities} Following the
choice of the Resistor-Average distance as a means of quantifying
the similarity of manifolds, we turn to the question of estimating
this distance for two arbitrary, nonlinear face manifolds. For a
general case there is no closed-form expression for RAD. However,
when $p(\mathbf{x})$ and $q(\mathbf{x})$ are two normal
distributions \cite{YoshTana1999}:
\begin{align}\label{Normal KLD}
        D_{KL} (p||q) =& \frac{1}{2} \log_2\left(\frac{|\mathbf{\Sigma}_q|}{|\mathbf{\Sigma}_p|}\right)
    + \frac{1}{2} \text{Tr}\Big[\mathbf{\Sigma}_p \mathbf{\Sigma}_q^{-1} +
    \mathbf{\Sigma}_q^{-1}
    (\bar{\mathbf{x}}_q-\bar{\mathbf{x}}_p)(\bar{\mathbf{x}}_q-\bar{\mathbf{x}}_p)^T\Big] -
    \frac{D}{2}
\end{align}
where $D$ is the dimensionality of data, $\bar{\mathbf{x}}_p$ and
$\bar{\mathbf{x}}_q$ data means, and $\mathbf{\Sigma}_p$ and
$\mathbf{\Sigma}_q$ the corresponding covariance matrices.

To achieve both expressive modelling of nonlinear manifolds as well
as an efficient procedure for comparing them, in the proposed method
a nonlinear projection of data using Kernel Principal Component
Analysis (Kernel PCA) is performed first. We show that with an
appropriate choice of the kernel type and bandwidth, the assumption
of normally distributed face patterns in the projection space
produces good KL divergence estimates. With a reference to our
generative model in \eqref{eq:manifold_pdf}, an appearance manifold
is effectively unfolded from the embedding image space.

\section{Kernel PCA} PCA is a technique in
which an orthogonal basis transformation is applied such that the
data covariance matrix $\mathbf{C} = \langle (\mathbf{x}_i - \langle
\mathbf{x}_j \rangle)(\mathbf{x}_i - \langle \mathbf{x}_j \rangle)^T
\rangle $ is diagonalized. When data $\{\mathbf{x}_i\}$ lies on a
linear manifold, the corresponding linear subspace is spanned by the
dominant (in the eigenvalue sense) eigenvectors of $\mathbf{C}$.
However, in the case of nonlinearly distributed data, PCA does not
capture the true modes of variation well.

The idea behind KPCA is to map data into a high-dimensional space in
which it \emph{is} approximately linear -- then the true modes of
data variation can be found using standard PCA. Performing this
mapping explicitly is prohibitive for computational reasons and
inherently problematic due to the ``curse of dimensionality''. This
is why a technique widely known as the ``kernel trick'' is used to
implicitly realize the mapping. Let function $\mathbf{\Phi}$ map the
original data from input space to a high-dimensional feature space
in which it is (approximately) linear,
$\mathbf{\Phi}:\mathbb{R}^D\to\mathbb{R}^\Delta$, $\Delta \gg D$. In
KPCA the choice of mappings $\Phi$ is restricted to the set such
that there is a function $k$ (the kernel) for which:
\begin{equation}\label{Eqn: Kernel Def}
    \mathbf{\Phi}(\mathbf{x}_i)^T \mathbf{\Phi}(\mathbf{x}_j) =
\textit{k}(\mathbf{x}_i, \mathbf{x}_j)
\end{equation}
In this case, the principal components of the data in
$\mathbb{R}^\Delta$ space can be found by performing computations
in the input, $\mathbb{R}^D$ space only.

Assuming zero-centred data in the feature space (for information on
centring data in the feature space as well as a more detailed
treatment of KPCA see \cite{SchoSmolMull1999}), the problem of
finding principal components in this space is equivalent to solving
the eigenvalue problem:
\begin{equation}\label{Equation: KPCA_1}
    \mathbf{Ku}_i = \lambda_i \mathbf{u}_i
\end{equation}
where $\mathbf{K}$ is the kernel matrix:
\begin{equation}\label{Equation: KPCA_2}
    \mathbf{K}_{j, k} = \textit{k}(\mathbf{x}_j, \mathbf{x}_k) =
    \mathbf{\Phi}(\mathbf{x}_j)^T \mathbf{\Phi}(\mathbf{x}_k)
\end{equation}

The projection of a data point $\mathbf{x}$ to the $i$-th kernel
principal component is computed using the following expression
\cite{SchoSmolMull1999}:
\begin{equation}\label{Equation: KPCA_Projection}
    a_i = \sum_{m=1}^N u_i^{(m)} k(\mathbf{x}_m, \mathbf{x})
\end{equation}

\section{Combining RAD and kernel PCA} 
\begin{figure*}[!h]
  \centering
  \begin{tabular}{c}
    \includegraphics[width=230pt]{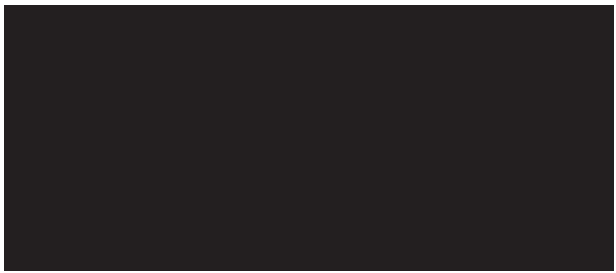} \\
    (a) \\
    \includegraphics[width=230pt]{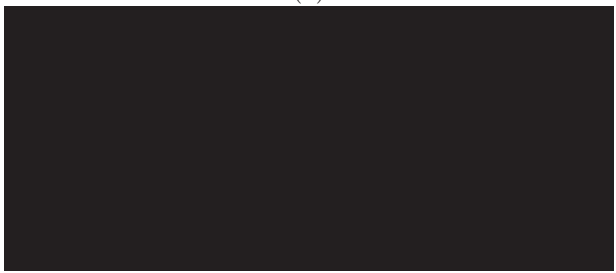}\\
    (b) \\
  \end{tabular}
  \caption[Nonlinear manifold unfolding.]
      { \it A typical face motion manifold in the input, image space
            exhibits high nonlinearity (a). The ``unfolded'' manifold
            is shown in (b). It can be seen that Kernel PCA captures
            the modes of data variation well, producing a Gaussian-looking
            distribution of patterns, confined to a roughly 2-dimensional space
            (corresponding to the intrinsic dimensionality of the manifold).
            In both (a) and (b) shown are projections to the first three
            principal components. }
            \label{Fig: Input and KPCA space manifolds}
  \vspace{6pt}\hrule
\end{figure*}

\label{Combining RAD and
KPCA} The variation of face patterns is highly nonlinear (see
Figure~\ref{Fig: Input and KPCA space manifolds}~(a)), making the
task of estimating RAD between two sparsely sampled face manifolds
in the image space difficult. The approach taken in this work is
that of mapping the data from the input, image space into a space in
which it lies on a nearly linear manifold. As before, we would not
like to compute this mapping explicitly. Also, note that the
inversions of data covariance matrices and the computation of their
determinants in the expression for the KL divergence between two
normal distributions \eqref{Normal KLD} limit the maximal practical
dimensionality of the feature space.

In our method both of these problems are solved using Kernel PCA.
The key observation is that regardless of how high the feature space
dimensionality is, the data has covariance in at most $N$
directions, where $N$ is the number of data points. Therefore, given
two data sets of faces, each describing a smooth manifold, we first
find the kernel principal components of their union. After
dimensionality reduction is performed by projecting the data onto
the first $M$ kernel principal components, the RAD between the two
densities, each now assumed Gaussian, is computed. Note that the
implicit nonlinear map is different for each data set pair. The
importance of this can be seen by noticing that the intrinsic
dimensionality of the manifold that \emph{both} sets lie on is lower
than of the manifold that all data in a database lies on, resulting
in its more accurate ``unfolding'', see Figure~\ref{Fig: Input and
KPCA space manifolds}~(b).

We estimate covariance matrices in the Kernel PCA space using
probabilistic PCA (PPCA) \cite{TippBish1999}. In short,
probabilistic PCA is an extension of the traditional PCA that
recovers parameters of a linear generative model of data (i.e.\ the
full corresponding covariance matrix), with the assumption of
isotropic Gaussian noise: $\mathbf{C} =
\mathbf{V}\mathbf{\Lambda}\mathbf{V}^T + \sigma\mathbf{I}$. Note the
model of noise density in~\eqref{eq:manifold_pdf} that this
assumption implies: $g^{(i)}(p_n(\mathbf{x})) \sim
\mathcal{N}(\mathbf{0},\sigma\mathbf{I})$, where $g^{(i)}\left(
f^{(i)}(\mathbf{x}) \right) = \mathbf{x}$.

\section{Synthetically repopulating manifolds} \label{Sec: Populating
FMMs} In most applications, due to the practical limitations in the
data acquisition process, AFR algorithms have to work with sparsely
populated face manifolds. Furthermore, some modes of data variation
may not be present in full. Specifically, in the AFR for
authentication setup considered in this work, the practical limits
on how long the user can be expected to wait for verification, as
well as how controlled his motion can be required to be, limit the
possible variations that are seen in both training and novel video
sequences. Finally, noise in the face localization process increases
the dimensionality of the manifolds faces lie on, effectively
resulting in even less densely populated manifolds. For a
quantitative insight, it is useful to mention that the face
appearance variations present in a typical video sequence used in
evaluation in this chapter typically lie on a manifold of intrinsic
dimensionality of 3-7, with 85 samples on average.

In this work, appearance manifolds are synthetically repopulated in
a manner that achieves both higher manifold sample density, as well
as some generalization to unseen modes of variation (see work by
Martinez \cite{Mart2002}, and Sung and Poggio \cite{SungPogg1998}
for related approaches). To this end, we use domain-specific
knowledge to learn face transformations in a more sophisticated way
than could be realized by simple interpolation and extrapolation.

Given an image of a face, $\mathbf{x}$, we stochastically repopulate
its geodesic neighbourhood by a set of novel images
$\{\mathbf{x}_j^S\}$. Under the assumption that the embedding
function $f^{(i)}$ in~\eqref{eq:manifold_pdf} is smooth,
geodesically close images correspond to small changes in the imaging
parameters (e.g.\ yaw or pitch). Therefore, using the first-order
Taylor approximation of the effects of a projective camera, the face
motion manifold is locally similar to the \emph{affine warp}
manifold of $\mathbf{x}$. The proposed algorithm then consists of
random draws of a face image $\mathbf{x}$ from the data, stochastic
perturbation of $\mathbf{x}$ by a set of affine warps
$\{\mathbf{A}_j\}$ and finally, the augmentation of data by the
warped images -- see Figure~\ref{Algorithm: RANSAC KPCA}. Writing
the affine warp matrix decomposed to rotation and translation, skew
and scaling:
\begin{align}\label{Eqn: Affine Warp}
    \mathbf{A} =& \left(%
                    \begin{array}{ccc}
                      \cos{\theta} & -\sin{\theta} & t_x \\
                      \sin{\theta} &  \cos{\theta} & t_y \\
                      0            & 0             & 1 \\
                    \end{array}%
                 \right)
                 \left(%
                    \begin{array}{ccc}
                      1 & k & 0 \\
                      0 & 1 & 0 \\
                      0 & 0 & 1 \\
                    \end{array}%
                 \right)
                 \left(%
                    \begin{array}{ccc}
                      1+s_x & 0     & 0 \\
                      0     & 1+s_y & 0 \\
                      0     & 0     & 1 \\
                    \end{array}%
                 \right)
\end{align}
in the proposed method, affine transformation parameters $\theta$,
$t_x$ and $t_y$, $k$, and $s_x$ and $s_y$ are drawn from zero-mean
Gaussian densities.

\begin{figure}[t]
  \centering
  \includegraphics[width=0.6\textwidth]{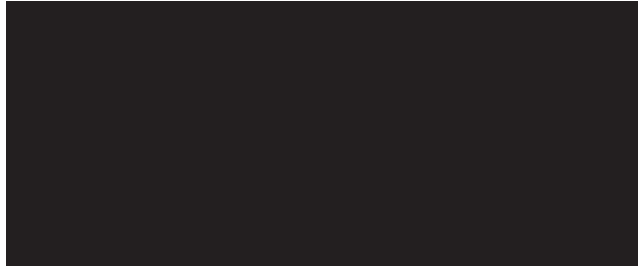} \\
  \caption[Stochastic manifold repopulation.]
      { \it The original, input data (dots) and the result of
            stochastically repopulating the corresponding manifold (circles).
            A few samples from the dense result are shown as images,
            demonstrating that the proposed method successfully captures
            and extrapolates the most significant modes of data variation. }
  \label{Figure: Synthetic Data}
  \vspace{6pt}\hrule
\end{figure}

\subsection{Outlier rejection} In most cases, automatic face
detection in cluttered scenes will result in a considerable number
of incorrect localizations -- \emph{outliers}. Typical outliers
produced by the Viola-Jones face detector employed in this chapter
are reproduced from Appendix~\ref{App: CamFace} in Figure~\ref{4Fig:
Falses}.

\begin{figure}
  \centering
  \includegraphics[width=0.8\textwidth]{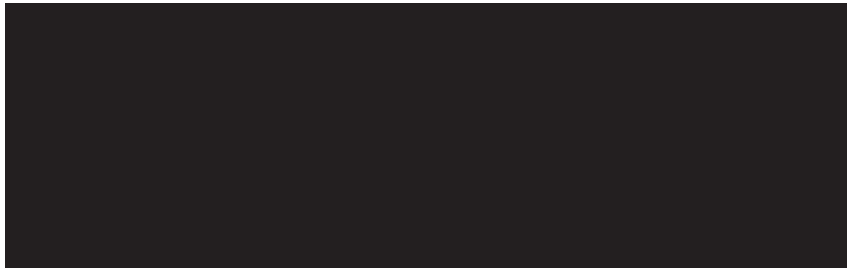}\\
  \caption[False positive face detections.]
    { \it Typical false face detections identified by our algorithm.}
  \label{4Fig: Falses}
  \vspace{6pt}\hrule
\end{figure}

Note that due to the complexity of face manifolds, outliers cannot
be easily removed in the input space. On the other hand, outlier
rejection after Kernel PCA-based manifold ``unfolding'' is trivial.
However, a way of computing the kernel matrix robust to the presence
of outliers is needed. To this end, our algorithm uses RANSAC
\cite{FiscBoll1981} with an underlying Kernel PCA model. The
application of RANSAC in the proposed framework is summarized in
Figure~\ref{Algorithm: RANSAC KPCA}. Finally, the Robust Kernel RAD
algorithm proposed in this chapter is in its entirety shown in
Figure~\ref{Algorithm: Robust Kernel RAD}.

\begin{figure}[!t]
    \centering
    \begin{tabular}{l}
          \begin{tabular}{ll}
              \textbf{Input}:  & set of observations $\{\mathbf{x}_i\}$,\\
                               & KPCA space dimensionality $D$.\\
              \textbf{Output}: & kernel principal components $\{\mathbf{u}_i\}$.\\
          \end{tabular}\vspace{5pt}
          \\ \hline \\
          \begin{tabular}{l}
            \textbf{1: Initialize best minimal sample}\\
            \hspace{10pt}$\mathcal{B} = \emptyset$\\\\

            \textbf{2: RANSAC iteration}\\
            \hspace{10pt}\textbf{for} $it=0$ to $LIMIT$\\\\

             \hspace{15pt}\textbf{3: Random sample draw}\\
             \hspace{25pt}$\{y_i\} \xleftarrow{D} \{x_i\}$\\\\

             \hspace{15pt}\textbf{4: Kernel PCA}\\
             \hspace{25pt}$\{u_i\}$ = KPCA$\left(\{y_i\}\right)$\\\\

             \hspace{15pt}\textbf{5: Nonlinear projection}\\
             \hspace{25pt}$\{x_i^P\} \xleftarrow{\{\mathbf{u}_i\}} \{x_i\}$\\\\

             \hspace{15pt}\textbf{6: Consistent data}\\
             \hspace{25pt}$\mathcal{B}_{it} = |filter(D_{MAH}(x_i,0) < T)|$\\\\

             \hspace{15pt}\textbf{7: Update best minimal sample}\\
             \hspace{25pt}$|\mathcal{B}_{it}| > |\mathcal{B}|~?~\mathcal{B} = \mathcal{B}_{it} $\\\\

             \textbf{8: Kernel PCA using best minimal sample}\\
             \hspace{10pt}$\{u_i\}$ = KPCA$(\mathcal{B})$\\\\
          \end{tabular}\\\hline
    \end{tabular}
    \caption[RANSAC KPCA algorithm.]{\it RANSAC Kernel PCA algorithm for unfolding face appearance manifolds in the presence of outliers. }
    \vspace{6pt}\hrule
    \label{Algorithm: RANSAC KPCA}
\end{figure}

\begin{figure}
    \centering
    \begin{tabular}{l}
          \begin{tabular}{ll}
              \textbf{Input}:  & sets of observations $\{\mathbf{a}_i\}$, $\{\mathbf{b}_i\}$.\\
                               & KPCA space dimensionality $D$.\\
              \textbf{Output}: & inter-manifold distance $D_{RAD}(\{\mathbf{a}_i\}, \{\mathbf{b}_i\})$.\\
          \end{tabular}\vspace{5pt}
          \\ \hline \\
          \begin{tabular}{l}
            \textbf{1: Inliers with RANSAC}\\
            \hspace{10pt}$\mathcal{V} = \{\mathbf{a}_i^V\},\{\mathbf{b}_i^V\} =$
            RANSAC$\left(\{\mathbf{a}_i\},\{\mathbf{b}_i\} \right)$\\\\

            \textbf{2: Synthetic data}\\
            \hspace{10pt}$\mathcal{S} = \{\mathbf{a_i^S}\},\{\mathbf{b_i^S}\}$ = perturb$\left(
                \langle\mathbf{a^V}\rangle,\langle\mathbf{b^V}\rangle \right) $\\\\

             \textbf{3: RANSAC Kernel PCA}\\
             \hspace{10pt}$\{\mathbf{u}_i\}$ =  KPCA$\left( \mathcal{V} \cup \mathcal{S} \right)$\\\\

             \textbf{4: Nonlinear projection}\\
             \hspace{10pt}$\{\mathbf{a}^P_i\},\{\mathbf{b}^P_i\} \xleftarrow{\{\mathbf{u}_i\}} \left( \mathcal{V},\mathcal{S} \right)$\\\\

             \textbf{7: Closed-form RAD}\\
             \hspace{10pt}$D_{RAD}(\{\mathbf{a}^P_i\},\{\mathbf{b}^P_i\})$\\\\

          \end{tabular}\\\hline
    \end{tabular}
    \caption[Robust Kernel RAD algorithm.]{\it Robust Kernel RAD algorithm summary. }
    \vspace{6pt}\hrule
    \label{Algorithm: Robust Kernel RAD}
\end{figure}

\section{Empirical evaluation} \label{Evaluation} We compared the
recognition performance of the following methods\footnote{Methods
were reimplemented through consultation with authors.} on the
\textit{CamFace} data set:
\begin{itemize}
    \item KL divergence-based algorithm of Shakhnarovich et al. (Simple
    KLD) \cite{ShakFishDarr2002},

    \item Simple RAD (based on Simple KLD),

    \item Kernelized Simple KLD algorithm (Kernel KLD),

    \item Kernel RAD,

    \item Robust Kernel RAD,

    \item Mutual Subspace Method (MSM) \cite{YamaFukuMaed1998},

    \item Majority vote using Eigenfaces, and

    \item Nearest Neighbour (NN) in the set distance sense; that
    is, achieving\\$\min_{\mathbf{x}\in S_0}\min_{\mathbf{y}\in S_i}
    \|\mathbf{x}-\mathbf{y}\|_2$.
\end{itemize}

\begin{figure}[t]
  \centering
  \includegraphics[width=0.6\textwidth]{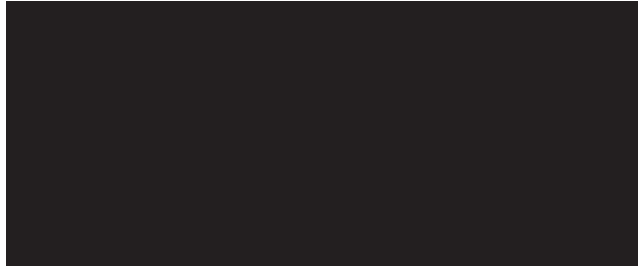}\\
  \caption[Dimensionality reduction by nonlinear unfolding.]{ \it Histograms of the dimensionality of the principal subspace
            in kernelized (dotted line) and non-kernelized (solid line)
            KL divergence-based methods, across the evaluation data set.
            The corresponding average dimensionalities were found to be
            $\sim 4$ and $\sim 16$. The large difference illustrates
            the extent of nonlinearity of face motion manifolds. }
            \label{Fig: PC dimensionality}
  \vspace{6pt}\hrule
\end{figure}

In all KLD and RAD-based methods, 85\% of data energy was explained
by the principal subspaces. In non-kernelized algorithms this
typically resulted in the principal subspace dimensionality of 16,
see Figure~\ref{Fig: PC dimensionality}. In MSM, first 3 principal
angles were used for recognition, while the dimensionality of PCA
subspaces describing the data was set to 9 \cite{YamaFukuMaed1998}.
In the Eigenfaces method, the 150-dimensional principal subspace
used explained $\sim 95\%$ of data energy. A 20-dimensional
nonlinear projection space was used in all kernel-based methods with
the RBF kernel $k(\mathbf{x}_i, \mathbf{x}_j)=\exp{-\gamma
(\mathbf{x}_i-\mathbf{x}_j)^T(\mathbf{x}_i-\mathbf{x}_j)}$. The
optimal value of the parameter $\gamma$ was learnt by optimizing the
recognition performance on a 20 person training data set. Note that
people from this set were not included in the evaluation reported in
Section~\ref{Results}. We used $\gamma = 0.380$ for greyscale images
normalized to have pixel values in the range $[0.0, 1.0]$.

In each experiment we used sets in a single illumination setup, with
test and training sets corresponding to sequences acquired in two
different sessions, see Appendix~\ref{App: CamFace}.

\subsection{Results} \label{Results}

The performance of the evaluated recognition algorithms is
summarized in Table~\ref{Table: Comparison 1}. The results suggest a
number of conclusions.

Firstly, note the relatively poor performance of the two nearest
neighbour-type methods -- the Set NN and the Majority vote using
Eigenfaces. These can be considered as a proxy for gauging the
difficulty of the recognition task, seeing that both can be expected
to perform relatively well if the imaging conditions are not greatly
different between training and test data sets. An inspection of the
incorrect recognitions of these methods offered an interesting
insight in one of their particular weaknesses, see Figure~\ref{Fig:
ROCs}~(a). This reaffirms the conclusion of \cite{SimZhang2004},
showing that it is not only \emph{changes} in the data acquisition
conditions that are challenging but also that there are certain
intrinsically difficult imaging configurations.

The Simple KLD method consistently achieved the poorest results. We
believe that the likely reason for this is the high nonlinearity of
face manifolds corresponding to the training sets used, caused by
near, office lighting used to vary the illumination conditions. This
is supported by the dramatic and consistent increase in the
recognition performance with kernelization. This result confirms the
first premise of this work, showing that sophisticated face manifold
modelling is indeed needed to accurately describe variations that
are expected in realistic imaging conditions. Furthermore, the
improvement observed with the use of Resistor-Average distance
suggests its greater robustness with respect to unseen variations in
face appearance, compared to the KL divergence. The performance of
Kernel RAD was comparable to that of MSM, which ranked second-best
in our experiments. The best performing algorithm was found to be
Robust Kernel RAD. Synthetic manifold repopulation produced a
significant improvement in the recognition rate (of about $10\%$),
the proposed method correctly recognizing 98\% of individuals. ROC
curves corresponding to the methods that best illustrate the
contributions of this chapter are shown in Figure~\ref{Fig:
ROCs}~(b), with Robust Kernel RAD achieving an Equal Error Rate of
2\%.

\begin{table}
  \centering
  \Large
  \caption[Recognition results.]{\it  Results of the comparison of our novel algorithm with existing methods in
           the literature. Shown is the identification rate in \%.\vspace{10pt}}

  \begin{tabular*}{1.00\textwidth}{@{\extracolsep{\fill}}l|ccccccc}
    \Hline
    \bf \normalsize ~Method &
    \rotatebox{90}{\normalsize Robust} \rotatebox{90}{\normalsize Kernel} \rotatebox{90}{\normalsize RAD} &
    \rotatebox{90}{\normalsize MSM} & \rotatebox{90}{\normalsize Kernel} \rotatebox{90}{\normalsize RAD} &
    \rotatebox{90}{\normalsize Kernel} \rotatebox{90}{\normalsize KLD} & \rotatebox{90}{\normalsize Set}
    \rotatebox{90}{\normalsize Nearest} \rotatebox{90}{\normalsize Neighbour } &
    \rotatebox{90}{\normalsize \normalsize Majority} \rotatebox{90}{\normalsize Vote w/}
    \rotatebox{90}{\normalsize Eigenfaces } & \rotatebox{90}{\normalsize Simple} \rotatebox{90}{\normalsize KLD}\\
    \hline
    \bf \normalsize ~Recognition rate & \small 98 & \small 89 & \small 88 & \small 79 & \small 72 &
    \small 71 & \small 52\\
  \Hline
  \end{tabular*}
  \label{Table: Comparison 1}
\end{table}

\begin{figure}[t]
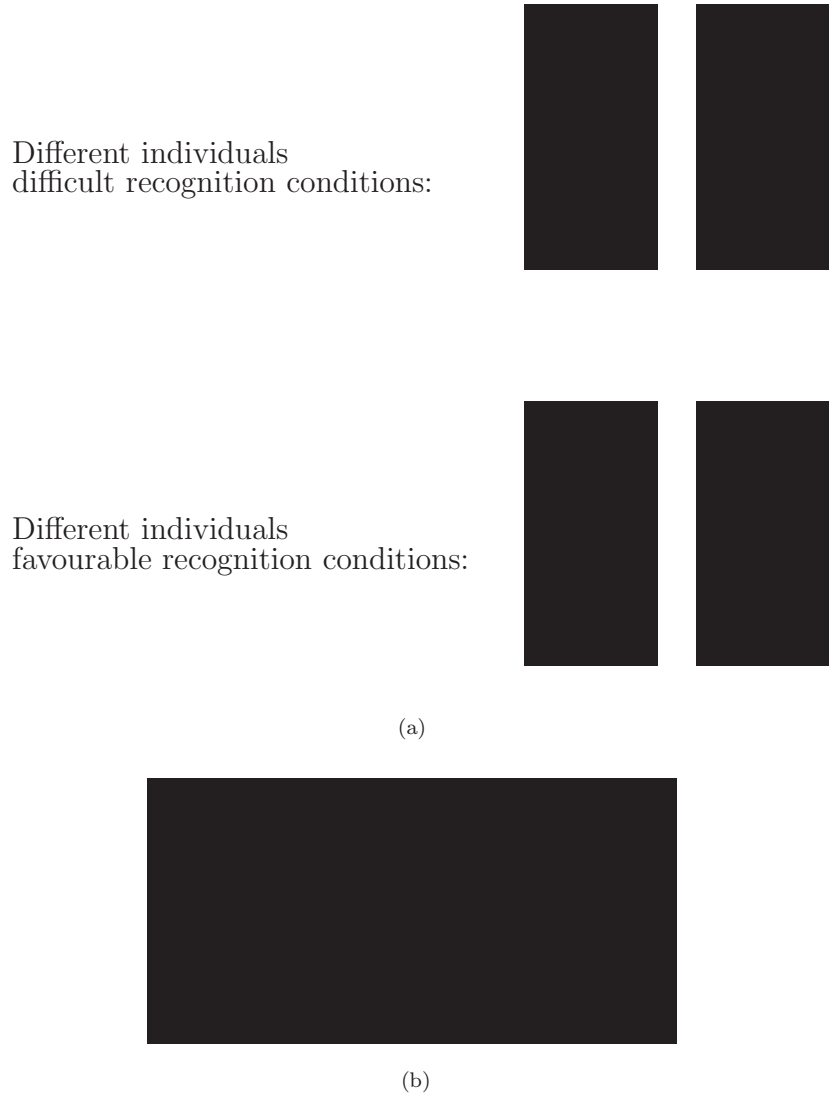

  \centering
  \footnotesize

  \begin{tabular}{V}
    \begin{tabular}{lVV}
       \begin{tabular}{l}{\large Different individuals}\\{\large difficult recognition conditions:}\end{tabular} \vspace{25pt} &
       \includegraphics[width=50pt]{IVC2004_nn_face1.ps} \vspace{25pt} &
       \includegraphics[width=50pt]{IVC2004_nn_face2.ps} \vspace{25pt}\\

       \begin{tabular}{l}{\large Different individuals}\\{\large favourable recognition conditions:}\end{tabular} \vspace{10pt} &
       \includegraphics[width=50pt]{IVC2004_nn_face3.ps} \vspace{10pt} &
       \includegraphics[width=50pt]{IVC2004_nn_face4.ps} \vspace{10pt}\\
    \end{tabular}\\
    (a) \vspace{15pt} \\
    \includegraphics[width=200pt]{IVC2004_rocs.ps} \vspace{10pt} \\
    (b) \\
  \end{tabular}

  \caption[ROC curves and recognition difficulties.]
    { \it (a) The most common failure mode of NN-type recognition algorithms
                        is caused by ``hard'' illumination conditions and head
                        poses. The two top images show faces that due to severe illumination
                        conditions and semi-profile head orientation look very similar
                        in spite of different identities (see \cite{SimZhang2004}) --
                        the Set NN algorithm incorrectly classified these fames as
                        belonging to the same person. Information from other
                        frames (e.g.\ the two bottom images) is not used
                        to achieve increased robustness. (b) Receiver Operator Characteristic (ROC) curves of the Simple
                        KLD, MSM, Kernel KLD and the proposed Robust Kernel RAD methods.
                        The latter exhibits superior performance, achieving
                        an Equal Error Rate of 2\%.
                         }
  \label{Fig: ROCs}
\end{figure}


\section{Summary and conclusions} \label{Conclusions}

In this chapter we introduced a novel method for face recognition
from face appearance manifolds due to head motion. In the proposed
algorithm the Resistor-Average distance computed on nonlinearly
mapped data using Kernel PCA is used as a dissimilarity measure
between distributions of face appearance, derived from video. A
data-driven approach to generalization to unseen modes of variation
was described, resulting in stochastic manifold repopulation.
Finally, the proposed concepts were empirically evaluated on a
database with 100 individuals and mild illumination variation. Our
method consistently achieved a high recognition rate, on average
correctly recognizing in 98\% of the cases and outperforming
state-of-the-art algorithms in the literature.

\section*{Related publications}

The following publications resulted from the work presented in this
chapter:
	
\begin{itemize}
  \item O. Arandjelovi{\'c} and R. Cipolla. Face recognition from face motion manifolds using
                  robust kernel resistor-average distance. In \textit{Proc. IEEE Workshop on
                  Face Processing in Video}, \textbf{5}:page 88, June
                  2004. \cite{AranCipo2004}

  \item O. Arandjelovi{\'c} and R. Cipolla. An information-theoretic approach to face recognition
                  from face motion manifolds. \textit{Image and Vision Computing (special issue
                  on Face Processing in Video Sequences)}, \textbf{24}(6):pages 639--647, June 2006.
                  \cite{AranCipo2006}
\end{itemize}

\graphicspath{{./05therm/}}
\chapter{Fusing Visual and Thermal Face Biometrics}
\label{Chp: Thermal}
\begin{center}
  \footnotesize
  \vspace{-20pt}
  \framebox{\includegraphics[width=0.75\textwidth]{title_img.eps}}\\
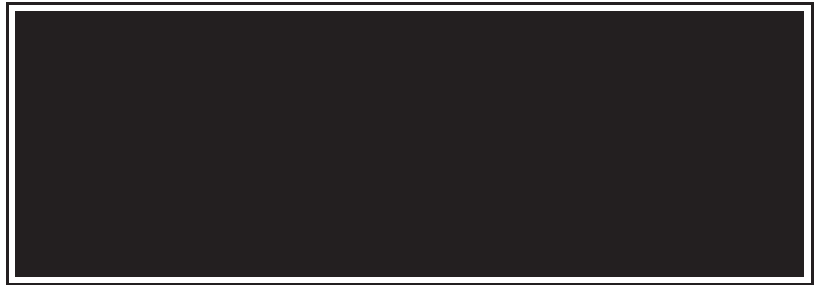
  Claude Monet, Haystack\\
  Oil on canvas, 73.3 x 92.6 cm\\
  Museum of Fine Arts, Boston\\
  \vspace{50pt}
\end{center}

\cleardoublepage

In the preceding chapters we dealt with increasingly difficult
formulations of the face recognition problem. As restrictions on
both training and novel data were relaxed, more generalization was
required. So far we addressed robustness to pose changes of the
user, noise contamination and low spatiotemporal resolution of
video. In this chapter we start exploring the important but
difficult problem of recognition in the presence of changing
illumination conditions in which faces are imaged.

In practice, the effects of changing pose are usually least
problematic and can often be overcome by acquiring data over a time
period e.g.\ by tracking a face in a surveillance video. As before,
we assume that the training image set for each individual contains
some variability in pose, but is not obtained in scripted conditions
or in controlled illumination.

In contrast, illumination is much more difficult to deal with: the
illumination setup is in most cases not practical to control and its
physics is difficult to accurately model. Biometric imagery acquired
in the thermal, or near-infrared electromagnetic spectrum, is useful
in this regard as it is virtually insensitive to illumination
changes. On the other hand, it lacks much of the individual,
discriminating facial detail contained in visual images. In this
sense, the two modalities can be seen as complementing each other.
The key idea behind the system presented in this chapter is that
robustness to extreme illumination changes can be achieved by fusing
the two. This paradigm will further prove useful when we consider
the difficulty of recognition in the presence of occlusion caused by
prescription glasses.


\section{Face recognition in the thermal spectrum}

    \index{face!appearance}
    \index{spectrum!visual}
    \index{spectrum!thermal}
    \index{biometric!fusion}
    \index{spectrum!infrared|see{thermal spectrum}}
    \index{glasses}
    \index{occlusion!glasses|see{glasses}}
    \index{image!set matching}
    \index{set!matching|see{image set matching}}
    \index{data!preprocessing|see{band pass filter}}
    \index{filter!band pass}

A number of recent studies suggest that face recognition in the
thermal spectrum offers a few distinct advantages over the visible
spectrum, including invariance to ambient illumination changes
\cite{WolfSocoEvel2001, SocoSeliNeuh2003, Prok2000, SocoSeli2004}.
This is due to the fact that a thermal infrared sensor measures the
heat energy radiation emitted by the face rather than the light
reflectance.  In outdoor environments, and particularly in direct
sunlight, illumination invariance only holds true to good
approximation for the Long-Wave Infrared (LWIR: $8$--$14\mu m$)
spectrum, which is fortunately measured by the less expensive
uncooled thermal infrared camera technology.  Human skin has high
emissivity in the Long-Wave Infrared (MWIR: $3$--$5\mu m$) spectrum
and even higher emissivity in the LWIR spectrum making face imagery
by and large invariant to illumination variations in these spectra.

Appearance-based face recognition algorithms applied to thermal
infrared imaging consistently performed better than when applied to
visible imagery, under various lighting conditions and facial
expressions \cite{KongHeoAbidPaik+2005, SocoSeli2002, SocoSeliNeuh2003,
equinoxreport02}. Further performance improvements were achieved
using decision-based fusion \cite{SocoSeliNeuh2003}. In contrast to
other techniques, Srivastana and Liu \cite{SrivLiu2003}, performed
face recognition in the space of Bessel function parameters. First,
they decompose each infrared face image using Gabor filters. Then,
they represent the face by modelling the marginal density of the
Gabor filter coefficients using Bessel functions. This approach has
further been improved by Buddharaju \textit{et al.}
\cite{BuddPavlKaka2004}. Recently, Friedrich and Yeshurun
\cite{FrieYesh2003} showed that IR-based recognition is less
sensitive to changes in 3D head pose and facial expression.

A thermal sensor generates imaging features that uncover thermal
characteristics of the face pattern. Another advantage of thermal
infrared imaging in face recognition is the existence of a direct
relationship to underlying physical anatomy such as vasculature.
Indeed, thermal face recognition algorithms attempt to take
advantage of such anatomical information of the human face as unique
signatures.  The use of vessel structure for human identification
has been studied during recent years using traits such as hand
vessel patterns \cite{LinFan2004,ImChoiKim2003}, finger vessel
patterns \cite{ShimShim2004,MiurNagaMiya2003} and vascular networks
from thermal facial images \cite{ProkRied1998}. In
\cite{BuddPavlTsia2005} a novel methodology that consists of a
statistical face segmentation and a physiological feature extraction
algorithm, and a matching procedure of the vascular network from
thermal facial imagery has been proposed.

The downside of employing near infrared and thermal infrared sensors
is that glare reflections and opaque regions appear in presence of
subjects wearing prescription glasses, plastic and sun glasses. For
a large proportion of individuals the regions around the eyes --
that is an area of high interest to face recognition systems --
become occluded and therefore less discriminant
\cite{AranHammCipo2006,LiChuLiao+2007}.

\subsection{Multi-sensor based techniques}
In the biometric literature several classifiers have been used to
concatenate and consolidate the match scores of multiple independent
matchers of biometric traits \cite{ChatBorsPita1999}
\cite{BenAbdeMayo1998,BiguBiguDuc+1997,VerlChol1999,WangTanJain2003}.
In \cite{BrunFala1995} a HyperBF network is used to combine matchers
based on voice and face features.  Ross and Jain \cite{RossJain2003}
use decision tree and linear discriminant classifiers for
classifying the match scores pertaining to the face, fingerprint and
hand geometry modalities.  In \cite{RossGovi2005} three different
colour channels of a face image are independently subjected to LDA
and then combined.

Recently, several successful attempts have been made to fuse the
visual and thermal infrared modalities to increase the performance
of face recognition
\cite{HeoKongAbidAbid2004,GyaoBebiPavl2004,SocoSeli2004,WangSungVenk2004,
ChenFlynBowy2005,KongHeoAbidPaik+2005, BrunFala1995, RossJain2003,
ChenFlynBowy2003,HeoAbiKong+2003}. Visible and thermal sensors are
well-matched candidates for image fusion as limitations of imaging
in one spectrum seem to be precisely the strengths of imaging in the
other. Indeed, as the surface of the face and its temperature have
nothing in common, it would be beneficial to extract and fuse cues
from both sensors that are not redundant and yet complementary.

In \cite{HeoKongAbidAbid2004} two types of visible and thermal fusion
techniques have been proposed. The first fuses low-level data while
the second fuses matching distance scores. Data fusion was
implemented by applying pixel-based weighted averaging of
co-registered visual and thermal images. Decision fusion was
implemented by combining the matching scores of individual
recognition modules.

The fusion at the score level is the most commonly considered
approach in the biometric literature \cite{RossNandJain2006}.
Cappelli et al. \cite{CappMaioMalt2000} use a \emph{double sigmoid
function} for score normalization in a multi-biometric system that
combines different fingerprint matchers. Once the match scores
output by multiple matchers are transformed into a common domain
they can be combined using simple fusion operators such as the sum
of scores, product of scores or order statistics (e.g.,
maximum/minimum of scores or median score). Our proposed method
falls into this category of multi-sensor fusion at the score level.
To deal with occlusions caused by eyeglasses in thermal imagery, Heo
et al. \cite{HeoKongAbidAbid2004} used a simple ellipse fitting
technique to detect the circle-like eyeglass regions in the IR image
and replaced them with an average eye template. Using a commercial
face recognition system, FaceIt \cite{Iden}, they demonstrated
improvements in face recognition accuracy. Our method differs both
in the glasses detection stage, which uses a principled statistical
model of appearance variation, and in the manner it handles detected
occlusions. Instead of using the average eye template, which carries
no discriminative information, we segment out the eye region from
the infrared data, effectively placing more weight on the
discriminative power of the same region extracted from the filtered,
visual imagery.

\section{Method details}
\begin{figure*}[t]

    \index{biometric!fusion}
    \index{features!holistic}
    \index{features!local}
    \index{filter!band pass}
    \index{glasses}

  \centering
  \includegraphics[width=0.9\textwidth]{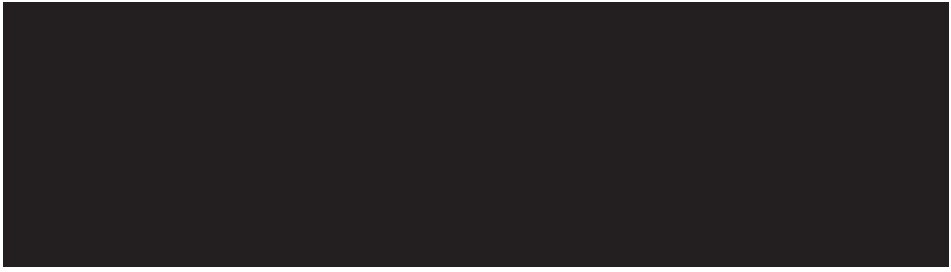}
  \caption[System overview.]
                 {\it Our system consists of three
                      main modules performing (i) data preprocessing and
                      registration, (ii) glasses detection and (iii) fusion of
                      holistic and local face representations using visual and
                      thermal modalities. }
            \label{Fig: Overview}
  \vspace{6pt}\hrule
\end{figure*}

In the sections that follow we explain our system in detail, the
main components of which are conceptually depicted in
Figure~\ref{Fig: Overview}.

\subsection{Matching image sets}\label{SS: Matching image sets}

    \index{image!set matching}

As before, in this chapter too we deal with face recognition from
sets of images, both in the visual and thermal spectrum. We will
show how to achieve illumination invariance using a combination of
simple data preprocessing (Section~\ref{SS: Data preprocessing}), a
combination of holistic and local features (Section~\ref{SS: Single
modality recognition}) and the fusion of two modalities (see
Section~\ref{SS: Fusing modalities}). These stages normalize for the
bulk of appearance changes caused by extrinsic (non person-specific)
factors. Hence, the requirements for our basic set-matching
algorithm are those of (i) some pose generalization and (ii)
robustness to noise. We compare two image sets by modelling the
variations within a set using a linear subspace and comparing two
subspaces by finding the most similar modes of variation within
them.

The face appearance modelling step is a simple application of
Principal Component Analysis (PCA) without mean subtraction. In
other words, given a data matrix $\mathbf{d}$ (each column
representing a rasterized image), the corresponding subspace is
spanned by the eigenvectors of the matrix $\mathbf{C} = \mathbf{d}
\mathbf{d}^T$ corresponding to the largest eigenvalues; we used 5D
subspaces, as sufficiently expressive to on average explain over
90\% of data variation within intrinsically low-dimensional face
appearance changes in a set.

We next formally introduce the concept of principal angles and
motivate their application for face image set comparison. We show
that they can be used to efficiently extract the most similar
appearance variation modes within two sets.

    \index{image!set matching}

\subsubsection{Principal angles}\label{SubSec: Principal Angles}
Principal, or canonical, angles $0 \leq \theta_1 \leq \ldots \leq
\theta_D \leq (\pi/2)$ between two $D$-dimensional linear subspaces
$U_1$ and $U_2$ are recursively uniquely defined as the minimal
angles between any two vectors of the subspaces \cite{Hote1936}:
\begin{equation}\label{Eqn: Principal Angles Definition 1}
    \rho_i = \cos \theta_i=\max_{\mathbf{u}_i \in U_1} \max_{\mathbf{v}_i \in U_2}
                \mathbf{u}_i^T \mathbf{v}_i
\end{equation}
subject to the orthonormality condition:
\begin{equation}\label{Eqn: Principal Angles Definition 2}
   \mathbf{u}_i^T \mathbf{u}_i = \mathbf{v}_i^T \mathbf{v}_i = 1,~
   \mathbf{u}_i^T \mathbf{u}_j = \mathbf{v}_i^T \mathbf{v}_j = 0,~
   j = 1, ..., i-1
\end{equation}
We will refer to $\mathbf{u}_i$ and $\mathbf{v}_i$ as the $i$-th
pair of \emph{principal vectors}, see Figure~\ref{Fig: Principal
Vectors MSM}~(a). The quantity $\rho_i$ is also known as the $i$-th
canonical correlation \cite{Hote1936}. Intuitively, the first pair
of principal vectors corresponds to the most similar modes of
variation within two linear subspaces; every next pair to the most
similar modes orthogonal to all previous ones. We quantify the
similarity of subspaces $U_1$ and $U_2$, corresponding to two face
sets, by the cosine of the smallest angle between two vectors
confined to them i.e.\ $\rho_1$.

\begin{figure}[t]
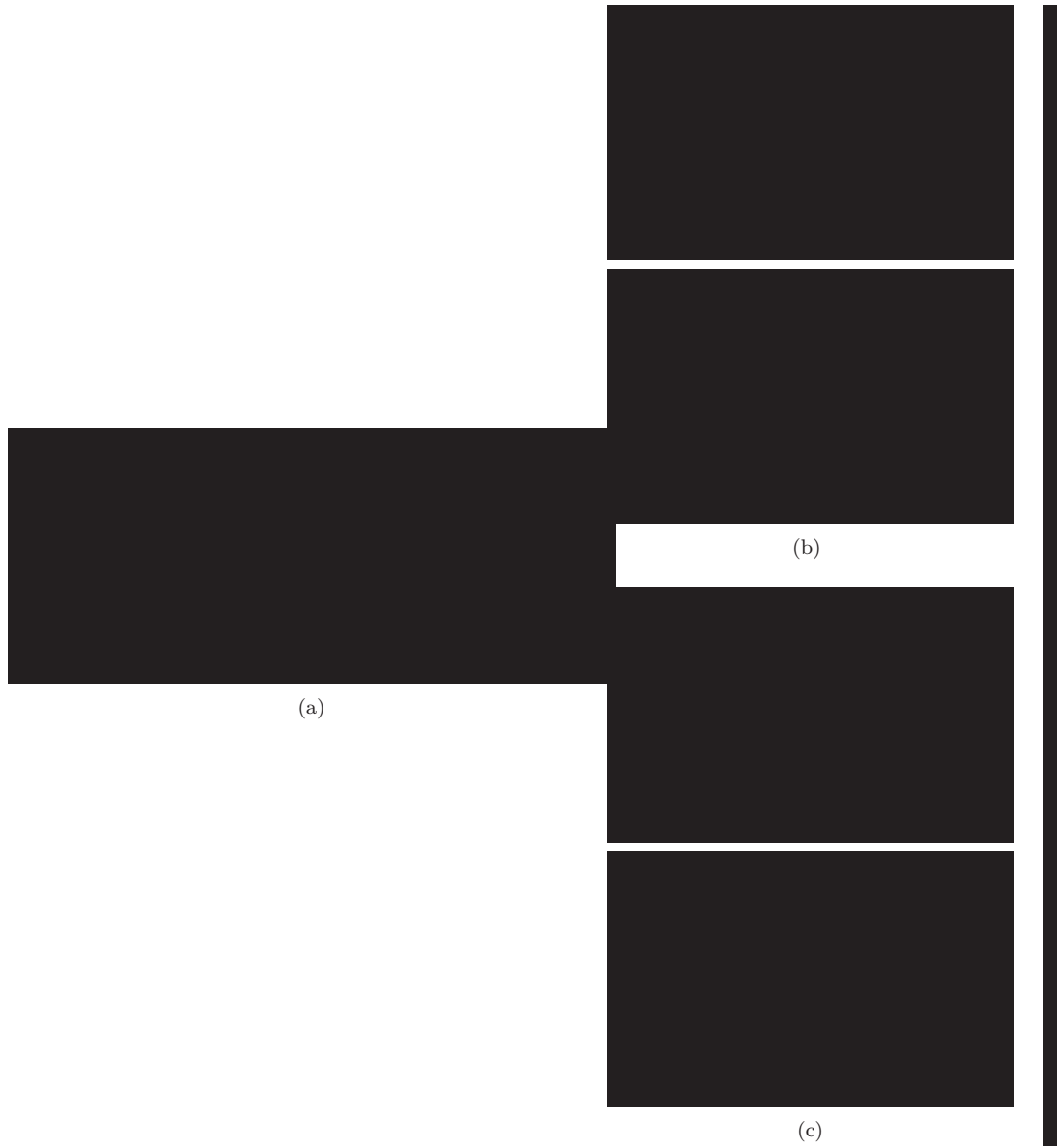

  \centering

  \begin{tabular}{VV}
  \centering
  \begin{tabular}{c}
  \includegraphics[width=0.6\textwidth]{pvs_spaces.eps}  \\
  \footnotesize (a)
  \end{tabular} \hspace{-30pt} &
  \begin{tabular}{c}
    \includegraphics[width=0.4\textwidth]{BMVC2005_pvs_msm_ok1.eps} \\
    \includegraphics[width=0.4\textwidth]{BMVC2005_pvs_msm_ok3.eps} \\
    \footnotesize (b) \vspace{10pt} \\
    \includegraphics[width=0.4\textwidth]{BMVC2005_pvs_msm_ok2.eps} \\
    \includegraphics[width=0.4\textwidth]{BMVC2005_pvs_msm_ok4.eps} \\
    \footnotesize (c) \\
  \end{tabular}
  \end{tabular}

  \caption[Principal vectors between linear subspaces.]
           { \it
        An illustration of the concept of principal angles and principal
                vectors in the case of two 2D subspaces embedded in a 3D
                space. As two such subspaces necessarily intersect,
                the first pair of principal vectors is the same
                (i.e.\ $\mathbf{u}_1 = \mathbf{v}_1$). However, the
                second pair is not, and in this case forms the
                second principal angle of
                $\cos^{-1}\rho_2 = \cos^{-1}(0.8084) \approx 36^{\circ}$.
                The top three pairs of principal vectors, displayed as images, when the
                subspaces correspond to image sets of the same and different
                individuals are displayed in (b) and (c)
                (top rows corresponds to $\mathbf{u}_i$, bottom rows to $\mathbf{v}_i$).
                In (b), the most
                similar modes of pattern variation, represented
                by principal vectors, are very much alike \emph{in spite
                of different illumination conditions} used in
                data acquisition.
            }
            \label{Fig: Principal Vectors MSM}
  \vspace{6pt}\hrule
\end{figure}

This interpretation of principal vectors motivates the suitability
of canonical correlations as a similarity measure when subspaces
$U_1$ and $U_2$ correspond to face images. First, the empirical
observation that face appearance varies smoothly as a function of
camera viewpoint \cite{AranCipo2006b,BichPent1994} is implicitly
exploited: since the computation of the most similar modes of
appearance variation between sets can be seen as an
efficient``search'' over entire subspaces, \emph{generalization} by
means of linear pose interpolation and extrapolation is inherently
achieved. This concept is further illustrated in Figure~\ref{Fig:
Principal Vectors MSM}~(b,c). Furthermore, by being dependent on
only a single (linear) direction within a subspace, by employing the
proposed similarity measure the bulk of data in each set, deemed not
useful in a specific set-to-set comparison, is thrown away. In this
manner \emph{robustness to missing data} is achieved.

An additional appealing feature of comparing two subspaces in this
manner is contained in its computational efficiency. If
$\mathbf{B}_1$ and $\mathbf{B}_2$ are orthonormal basis matrices
corresponding to $U_1$ and $U_2$, then writing the Singular Value
Decomposition (SVD) of the matrix $\mathbf{B}_1^T \mathbf{B}_2$:
\begin{align}
 \mathbf{M} = \mathbf{B}_1^T \mathbf{B}_2 = \mathbf{U} \mathbf{\Sigma} \mathbf{V}^T.
\end{align}
The $i$-th canonical correlation $\rho_i$ is then given by the
$i$-th singular value of $\mathbf{M}$ i.e.\ $\mathbf{\Sigma}_{i,i}$,
and the $i$-th pair of principal vectors $\mathbf{u}_i$ and
$\mathbf{v}_i$ by, respectively, $\mathbf{B}_1 \mathbf{U}$ and
$\mathbf{B}_2 \mathbf{V}$ \cite{BjorGolu1973}. Seeing that in our
case $\mathbf{M}$ is a $5 \times 5$ matrix and that we only use the
largest canonical correlation, $\rho_1$ can be rapidly computed as
the largest eigenvalue of $\mathbf{M}\mathbf{M}^T$
\cite{PresTeukVettFlan1992}.

\subsection{Data preprocessing \& feature extraction}\label{SS: Data preprocessing}

    \index{filter!band pass}

The first stage of our system involves coarse normalization of pose
and illumination. Pose changes are accounted for by in-plane
registration of images, which are then passed through quasi
illumination-invariant image filters.

\begin{figure}

    \index{features!local}

  \centering
  \includegraphics[width=0.95\textwidth]{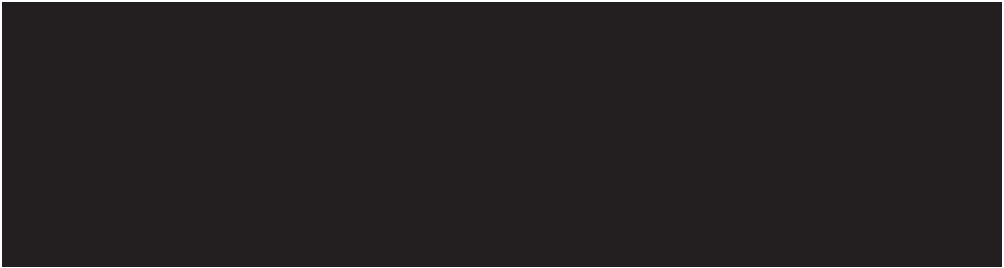}
  \caption[Facial feature localization and registration.]
       {\it Shown is the original
            image in the visual spectrum with detected facial
            features marked by yellow circles (left), the result
            of affine warping the image to the canonical frame
            (centre) and the final registered and cropped facial
            image. }
            \label{Fig: Registration}
  \vspace{6pt}\hrule
\end{figure}

We register all faces, both in the visual and thermal domain, to
have the salient facial features aligned. Specifically, we align the
eyes and the mouth due to the ease of detection of these features
(e.g.\ see \cite{AranZiss2005, BergBergEdwa+2004, CrisCootSco2004,
FelzHutt2005, TruOlagHamm+2005}). The 3 point correspondences,
between the detected and the canonical features' locations, uniquely
define an affine transformation which is applied to the original
image. Faces are then cropped to $80 \times 80$ pixels, as shown in
Figure~\ref{Fig: Registration}.

Coarse brightness normalization is performed by band-pass filtering
the images \cite{AranZiss2005, FitzZiss2002}. The aim is to reduce
the amount of high-frequency noise as well as extrinsic appearance
variations confined to a low-frequency band containing little
discriminating information. Most obviously, in visual imagery, the
latter are caused by illumination changes, owing to the smoothness
of the surface and albedo of faces \cite{AdinMoseUllm1997}.

\begin{figure*}
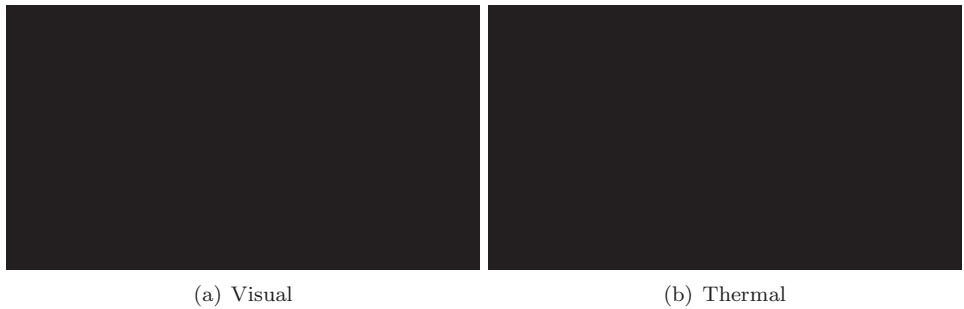


    \index{filter!band pass}

  \centering
  \subfigure[Visual]
    {\includegraphics[width=0.45\textwidth]{bp_visual.eps}}
  \subfigure[Thermal]
    {\includegraphics[width=0.45\textwidth]{bp_thermal.eps}}
  \caption[Performance changes with filtering in visual and thermal spectra.]
       {\it The optimal combination of
            the lower and upper band-pass filter thresholds is estimated
            from a small training corpus. The plots show the recognition
            rate using a single modality, (a) visual and (b) thermal, as
            a function of the widths $W_1$ and $W_2$ of the two Gaussian
            kernels in \eqref{Eqn: Band-Pass}. It is interesting to note
            that the optimal band-pass filter for the visual spectrum
            passes a rather narrow, mid-frequency band, whereas the optimal
            filter for the thermal spectrum is in fact a \emph{low-pass}
            filter. }
            \label{Fig: BP}
\end{figure*}

We consider the following type of a band-pass filter:
\begin{equation}
    \label{Eqn: Band-Pass}
    \mathbf{I}_F = \mathbf{I} \ast \mathbf{G}_{\sigma = W_1} - \mathbf{I} \ast \mathbf{G}_{\sigma =
    W_2},
\end{equation}
which has two parameters - the widths $W_1$ and $W_2$ of isotropic
Gaussian kernels. These are estimated from a small training corpus
of individuals in different illuminations. Figure~\ref{Fig: BP}
shows the recognition rate across the corpus as the values of the
two parameters are varied. The optimal values were found to be $2.3$
and $6.2$ for visual data; the optimal filter for thermal data was
found to be a \emph{low-pass} filter with $W_2 = 2.8$ (i.e.\ $W_1$
was found to be very large). Examples are shown in Figure~\ref{Fig:
Preprocessing}. It is important to note from Figure~\ref{Fig: BP}
that the recognition rate varied smoothly with changes in kernel
widths, showing that the method is not very sensitive to their exact
values, which is suggestive of good generalization to unseen data.

    \index{filter!band pass}

The result of filtering visual data is further scaled by a smooth
version of the original image:
\begin{equation}
    \label{Eqn: SQI}
    \hat{\mathbf{I}}_F(x,y) = \mathbf{I}_F(x,y) ./ (\mathbf{I} \ast \mathbf{G}_{\sigma =
    W_2}),
\end{equation}
where $./$ represents element-wise division. The purpose of local
scaling is to equalize edge strengths in dark (weak edges) and
bright (strong edges) regions of the face; this is similar to the
Self Quotient Image of Wang \textit{et al.} \cite{WangLiWang2004}.
This step further improves the robustness of the representation to
illumination changes, see Section~\ref{S: Empirical Evaluation}.

\begin{figure}
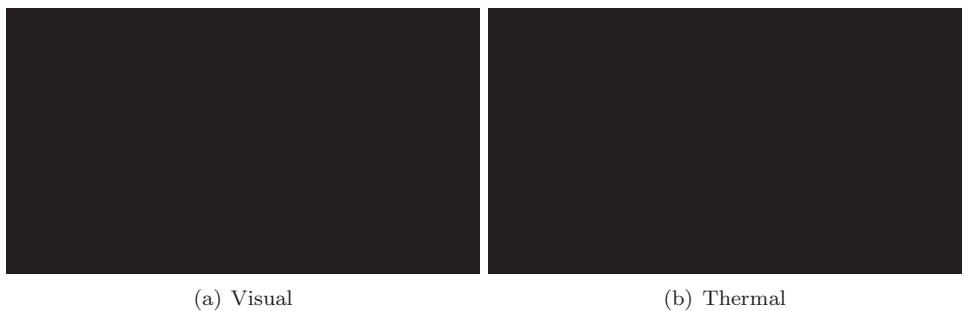


    \index{filter!band pass}

  \centering
  \subfigure[Visual]{\includegraphics[width=0.45\textwidth]{preprocess_V.eps}}
  \subfigure[Thermal]{\includegraphics[width=0.45\textwidth]{preprocess_T.eps}}
  \caption[Filtered examples in visual and thermal spectra.]
      {\it The effects of the optimal band-pass
            filters on registered and cropped faces in (a) visual and (b) thermal
            spectra. }
            \label{Fig: Preprocessing}
\end{figure}

\subsection{Single modality-based recognition}\label{SS: Single modality recognition}

    \index{features!holistic}
    \index{features!local}
    \index{biometric!fusion}
    \index{face!appearance}

We compute the similarity of two individuals using only a single
modality (visual or thermal) by combining the holistic face
representation described in Section~\ref{SS: Data preprocessing} and
a representation based on local image patches. These have been shown
to benefit recognition in the presence of large pose changes
\cite{SiviEverZiss2005}.

\begin{figure}
  \centering
  \includegraphics[width=0.9\textwidth]{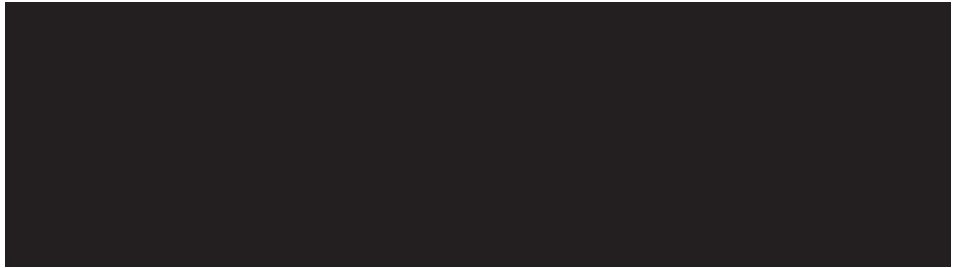}
  \caption[Local features.]
          {\it  In both the visual and the thermal spectrum our algorithm combines the similarities
                obtained by matching the holistic face appearance and the appearance of three salient
                local features -- the eyes and the mouth. }
            \label{Fig: Regions}
  \vspace{6pt}\hrule
\end{figure}

As before, we use the eyes and the mouth as the most discriminative
regions, by extracting rectangular patches centred at the
detections, see Figure~\ref{Fig: Regions}. The overall similarity
score is obtained by weighted summation:
\begin{equation}\label{Eqn: Fusion Regions}
    \rho_{v/t} = \underbrace{\omega_h \cdot \rho_h}_{\text{Holistic contribution}} +
                \underbrace{\omega_m \cdot \rho_m + (1 - \omega_h - \omega_m) \cdot \rho_e}_{\text{Local features contribution}},
\end{equation}
where $\rho_m$, $\rho_e$  and $\rho_h$ are the scores of separately
matching, respectively, the mouth, the eyes and the entire face
regions, and $\omega_h$ and $\omega_m$ the weighting constants.

The optimal values of the weights were estimated from the offline
training corpus. As expected, eyes were shown to carry a significant
amount of discriminative information, as for the visual spectrum we
obtained $\omega_e = 0.3$. On the other hand, the mouth region,
highly variable in appearance in the presence of facial expression
changes, was found not to improve recognition (i.e.\ $\omega_m
\approx 0.0$).

The relative magnitudes of the weights were found to be different in
the thermal spectrum, both the eye and the mouth region contributing
equally to the overall score: $\omega_m = 0.1,~\omega_h = 0.8$.
Notice the rather insignificant contribution of individual facial
features. This is most likely due to inherently spatially slowly
varying nature of heat radiated by the human body.

    \index{features!local}

\subsection{Fusing modalities}\label{SS: Fusing modalities}

    \index{biometric!fusion}

Until now we have focused on deriving a similarity score between two
individuals given sets of images in either thermal or visual
spectrum. A combination of holistic and local features was employed
in the computation of both. However, the greatest power of our
system comes from the fusion of the two modalities.

Given $\rho_v$ and $\rho_t$, the similarity scores corresponding to
visual and thermal data, we compute the joint similarity as:
\begin{equation}
    \rho_f = \underbrace{\omega_v(\rho_v) \cdot \rho_v}_{\text{Optical contribution}} +
        \underbrace{\big[1 - \omega_v(\rho_v)\big] \cdot \rho_t}_{\text{Thermal contribution}}.
\end{equation}
Notice that the weighting factors are no longer constants, but
\emph{functions}. The key idea is that if the visual spectrum match
is very good (i.e.\ $\rho_v$ is close to $1.0$), we can be confident
that illumination difference between the two images sets compared is
mild and well compensated for by the visual spectrum preprocessing
of Section~\ref{SS: Data preprocessing}. In this case, visual
spectrum should be given relatively more weight than when the match
is bad and the illumination change is likely more drastic. The value
of $\omega_v(\rho_v)$ can then be interpreted as statistically the
optimal choice of the mixing coefficient $\omega_v$ given the visual
domain similarity $\rho_v$. Formalizing this we can write
\begin{align}
    \omega_v(\rho_v) = \arg \max_{\omega} p(\omega | \rho_v),
    \label{Eqn: Omega Def}
\end{align}
or, equivalently
\begin{align}
    \omega_v(\rho_v) = \arg \max_{\omega} \frac {p(\omega, \rho_v)}
    {p(\rho_v)}.
\end{align}
Under the assumption of a uniform prior on the degree of visual
similarity, $p(\rho_v)$
\begin{align}
    p(\alpha | \rho_v) \propto p(\alpha, \rho_v),
\end{align}
and
\begin{align}
    \omega_v(\rho_v) = \arg \max_{\omega} p(\omega, \rho_v).
    \label{Eqn: Omega Est}
\end{align}

\subsubsection{Learning the weighting function}
The function $\omega_v \equiv \omega_v(\rho_v)$ is estimated in
three stages: first (i) we estimate $p(\omega_v, \rho_v)$, then (ii)
compute $\omega(\rho_v)$ using \eqref{Eqn: Omega Est} and finally
(iii) make an analytic fit to the obtained marginal distribution.
Step (i) is challenging and we describe it next.

\paragraph{Iterative density estimate}
The principal difficulty of estimating $p(\omega_v, \rho_v)$ is of
practical nature: in order to obtain an accurate estimate (i.e.\ a
well-sampled distribution), a prohibitively large training database
is needed. Instead, we employ a heuristic alternative. Much like
before, the estimation is performed using the offline training
corpus.

\begin{figure}
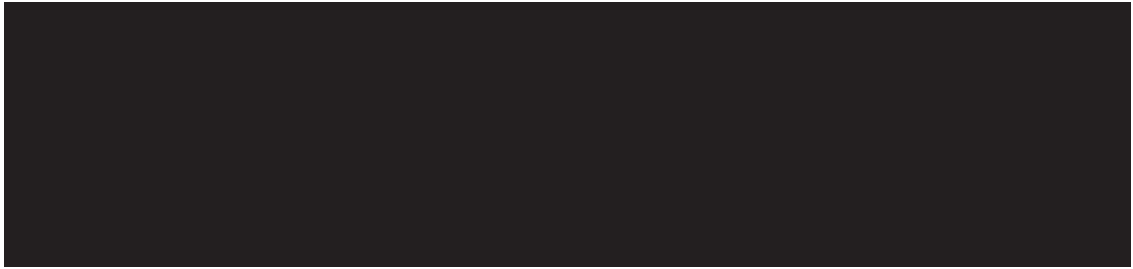


  \centering
  \begin{tabular}{VV}
    \includegraphics[width=0.55\textwidth]{sigmoid_f.eps} \hspace{-30pt} \vspace{10pt} &
    \includegraphics[width=0.55\textwidth]{similarity.eps} \vspace{10pt} \\
    \footnotesize (a) $y$-axis shifted sigmoid function &
    \footnotesize (b) Weighting function\\
  \end{tabular}
  \caption[Learnt weighting function for modality fusion.]{\it The contribution of visual matching,
               as a function of the similarity of visual imagery. A low similarity
               score between image sets in the visual domain is indicative of large
               illumination changes and consequently our algorithm leant that
               more weight should be placed on the illumination-invariant thermal
               spectrum. }
            \label{Fig: Coeff}
  \vspace{6pt}\hrule
\end{figure}

Our algorithm is based on an iterative incremental update of the
density, initialized as uniform over the domain $\omega_v, \rho_v
\in [0,1]$. We iteratively simulate matching of an unknown person
against a set gallery individuals. In each iteration of the
algorithm, these are randomly drawn from the offline training
database. Since the ground truth identities of all persons in the
offline database are known, for each $\omega_v = k \Delta \omega_v$
we can compute (i) the initial visual spectrum similarity
$\rho_v^{p,p}$ of the novel and the corresponding gallery sequences,
and (ii) the resulting separation $\delta(k \Delta \omega_v)$ i.e.\
the difference between the similarities of the test set and the set
corresponding to it in identity, and that between the test set and
the most similar set that does \emph{not} correspond to it in
identity. This gives us information about the usefulness of a
particular value of $\omega_v$ for observed $\rho_v^{p,p}$. Hence,
the density estimate $\hat{p}(\omega_v, \rho_v)$ is then updated at
$(k \Delta \omega_v, \rho^{p,p}),\text{}k=1\ldots$. We increment it
proportionally to $\delta(k \Delta \omega_v)$ after passing through
a $y$-axis shifted sigmoid function:
\begin{align}
    \hat{p}(k \Delta \omega_v, \rho_v^{p,p})_{[n+1]} = \hat{p}(k \Delta \omega_v, \rho_v^{p,p})_{[n]} +
        \Big[\text{sig}(C \cdot \delta(k \Delta \omega_v)) - 0.5\Big],
\end{align}
where subscript $[n]$ signifies the $n$-th iteration step and
\begin{align}
    \text{sig}(x) = \frac {1} {1 + e^{-x}},
\end{align}
as shown in Figure~\ref{Fig: Coeff}~(a). The sigmoid function has
the effect of reducing the overly confident weight updates for the
values of $\omega_v$ that result in extremely good or bad
separations $\delta(k \Delta \omega_v)$. The purpose of this can be
seen by noting that we are using separation as a proxy for the
statistical goodness of $\omega_v$, while in fact attempting to
maximize the average recognition rate (i.e.\ the average number of
cases for which $\delta(k \Delta \omega_v) > 0$).

\begin{figure}[!ht]
    \centering

    \begin{tabular}{l}
          \begin{tabular}{ll}
              \textbf{Input}:  & visual data $d_v(person, illumination)$,\\
                               & thermal data $d_t(person, illumination)$.\\
              \textbf{Output}: & density estimate $\hat{p}(\omega_v, \rho_v)$.\\
          \end{tabular}\vspace{5pt}
          \\ \hline \\
          \begin{tabular}{l}
            \textbf{1: Initialization}\\
            \hspace{10pt}$\hat{p}(\omega_v, \rho_v) = 0$,\\\\

            \textbf{2: Iteration}\\
            \hspace{10pt}\textbf{for}$\text{ all illuminations } i,~j$ \textbf{ and} $\text{ persons }
            p$\\\\

             \hspace{15pt}\textbf{3: Iteration}\\
             \hspace{25pt}\textbf{for}$\text{ all } k = 0, \dots, 1 / \Delta \omega_v,~\omega_v = k \Delta \omega_v$\\\\

             \hspace{30pt}\textbf{5: Separation given $\omega$}\\
             \hspace{40pt}$\delta(k \Delta \omega_v) = \min_{q \neq p} \large[ \omega_v \rho_v^{p,p} +
                                                                 (1 - \omega_v)\rho_t^{p,p}$\\
                                                     \hspace{107pt}$- \omega_v \rho_v^{p,q} +
                                                                 (1 - \omega_v)\rho_t^{p,q} \Large]$\\\\

             \hspace{30pt}\textbf{6: Update density estimate}\\
             \hspace{40pt}$\hat{p}(k \Delta \omega_v, \rho_v^{p,p}) = \hat{p}(k \Delta \omega_v, \rho_v^{p,p})$\\
              \hspace{87pt}$+ \Big[\text{sig}(C \cdot \delta(k \Delta \omega_v)) - 0.5\Big]$\\\\

             \textbf{7: Smooth the output}\\
             \hspace{10pt}$\hat{p}(\omega_v, \rho_v) =  \hat{p}(\omega_v, \rho_v) \ast \mathbf{G}_{\sigma=0.05} $\\\\

             \textbf{8: Normalize to unit integral}\\
             \hspace{10pt}$\hat{p}(\omega_v, \rho_v) =  \hat{p}(\omega_v, \rho_v) / \int_{\omega_v} \int_{\rho_v} \hat{p}(\omega_v, \rho_v) d\rho_v
             d\omega_v$\\\\
             \hline
          \end{tabular}\\
    \end{tabular}
    \caption[Optimal fusion algorithm.]{\it The proposed fusion learning algorithm, used offline. }
    \vspace{15pt}\hrule
    \label{Fig: Algorithm}
\end{figure}

Figure~\ref{Fig: Algorithm} summarizes the proposed offline learning
algorithm. An analytic fit to $\omega_v(\rho_v)$ in the form
$(1+e^{a})/(1 + e^{a / \rho_v})$ is shown in Figure~\ref{Fig:
Coeff}~(b).

\subsection{Prescription glasses}\label{SS: Dealing with glasses}

    \index{glasses}

The appeal of using the thermal spectrum for face recognition stems
mainly from its invariance to illumination changes, in sharp
contrast to visual spectrum data. The exact opposite is true in the
case of prescription glasses, which appear as dark patches in
thermal imagery, see Figure~\ref{Fig: Preprocessing}. The practical
importance of this can be seen by noting that in the US in 2000
roughly 96 million people, or 34\% of the total population, wore
prescription glasses \cite{WalkMill2001}.

In our system, the otherwise undesired, gross appearance distortion
that glasses cause in thermal imagery is used to help recognition by
detecting their presence. If the subject is not wearing glasses,
then both holistic and all local patches-based face representations
can be used in recognition; otherwise the eye regions in thermal
images are ignored as they contain no useful recognition
(discriminative) information.

\paragraph{Glasses detection.}

    \index{glasses}
    \index{image!set matching}

\begin{figure*}[t]
  \centering

  \index{glasses}

  \includegraphics[width=1\textwidth]{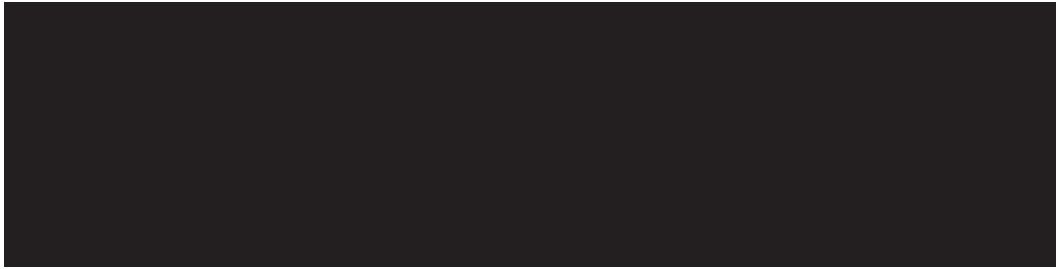}
  \caption[Glasses detection: training data.]{\it Shown are examples of glasses-on (top) and glasses-off (bottom)
                thermal data used to construct the corresponding appearance
                models for our glasses detector. }
            \label{Fig: Glasses Data}
  \vspace{6pt}\hrule
\end{figure*}

\begin{figure}
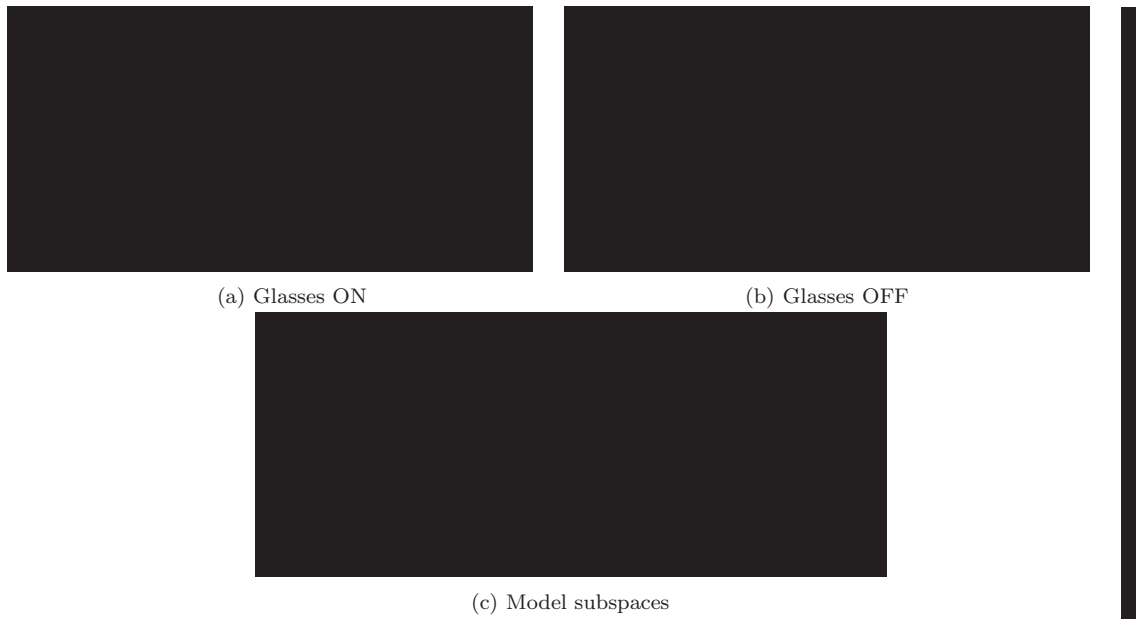

  \centering
  \begin{tabular}{c}
      \begin{tabular}{cc}
        \hspace{-20pt} \includegraphics[width=0.5\textwidth]{sim_glasses_on.eps} &
        \includegraphics[width=0.5\textwidth]{sim_glasses_off.eps}\\
        \footnotesize (a) Glasses ON
        &
        \footnotesize (b) Glasses OFF\\
      \end{tabular} \\
      \begin{tabular}{c}
        \includegraphics[width=0.6\textwidth]{glasses_spaces1.eps}\\
        \footnotesize (c) Model subspaces
      \end{tabular}
  \end{tabular}
  \caption[Prescription glasses detector response.]
        {\it (a) Inter- and (b) intra- class (glasses on and off) similarities
               across our data set. (c) Good discrimination by principal
                angles is also motivated qualitatively as the subspaces modelling appearance
                variations of the eye region with and without glasses on show very different orientations
                even when projected to the 3D principal subspace. As expected, the ``glasses off''
                subspace describes more appearance variation, as illustrated by the larger
                extent of the linear patch representing it in the plot.
                }
            \label{Fig: Glasses Similarity}
\end{figure}

We detect the presence of glasses by building representations for
the left eye region (due to the symmetry of faces, a detector for
only one side is needed) with and without glasses, in the thermal
spectrum. The foundations of our classifier are laid out in
\S\ref{SS: Matching image sets}. Appearance variations of the eye
region with out without glasses are represented by two 6D linear
subspaces estimated from the training data corpus, see
Fig.~\ref{Fig: Glasses Data} for examples of training data used for
subspace estimations. The linear subspace corresponding to eye
region patches extracted from a set of thermal imagery of a novel
person is then compared with ``glasses on'' and ``glasses off''
subspaces using principal angles. The presence of glasses is deduced
when the corresponding subspace results in a higher similarity
score. We obtain close to flawless performance on our data set (also
see \S\ref{S: Empirical Evaluation} for description), as shown in
Fig.~\ref{Fig: Glasses Similarity}~(a,b). Good discriminative
ability of principal angles in this case is also supported by visual
inspection of the ``glasses on'' and ``glasses off'' subspaces; this
is illustrated in Fig.~\ref{Fig: Glasses Similarity}~(c) which shows
the first two dominant modes of each, embedded in the 3D principal
subspace.

The presence of glasses severely limits what can be achieved with
thermal imagery, the occlusion heavily affecting both the holistic
face appearance as well as that of the eye regions. This is the
point at which our method heavily relies on decision fusion with
visual data, limiting the contribution of the thermal spectrum to
matching using mouth appearance only i.e.\ setting $\omega_m = 1.0$
in \eqref{Eqn: Fusion Regions}.

\section{Empirical evaluation}\label{S: Empirical Evaluation}

    \index{face!database}
    \index{spectrum!visual}
    \index{spectrum!thermal}

\begin{figure}
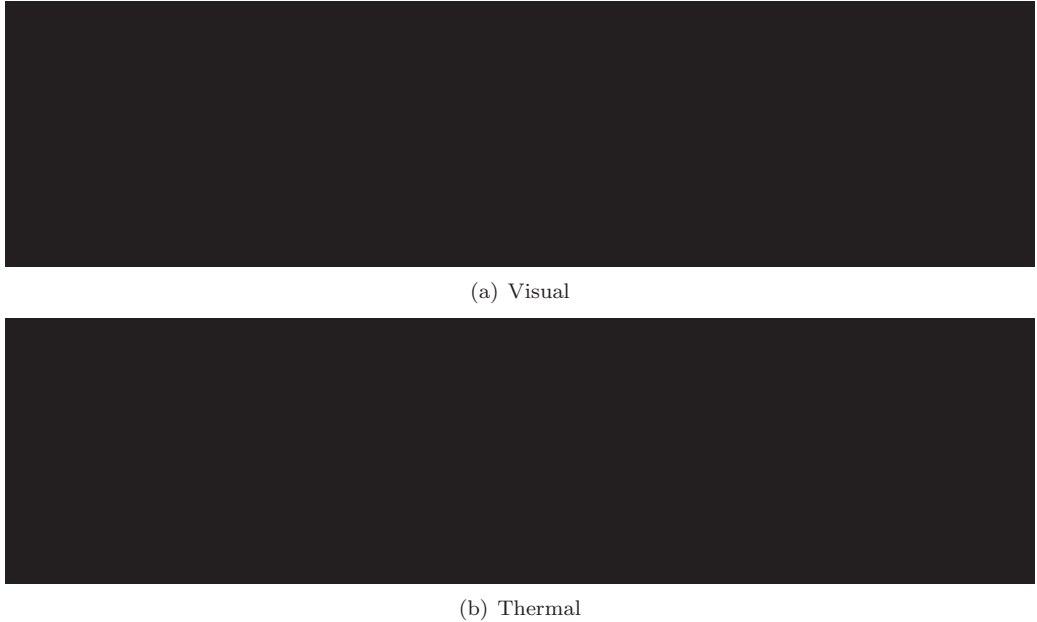


    \index{face!database}
    \index{spectrum!visual}
    \index{spectrum!thermal}
    \index{glasses}

  \centering
  \subfigure[Visual]
    {\includegraphics[width=0.98\textwidth]{data_V.eps}}
  \subfigure[Thermal]
    {\includegraphics[width=0.98\textwidth]{data_T.eps}}
  \caption[Data set examples.]{\it Each row corresponds to an example of a
                single training (or test) set of images used for our algorithm in
                the (a) visual and (b) thermal spectrum. Note the extreme
                changes in illumination, as well as that in some sets the user is
                wearing glasses and in some not. }
            \label{Fig: Example Data}
\end{figure}

We evaluated the described system on the \emph{``Dataset 02: IRIS
Thermal/Visible Face Database''} subset of the \emph{Object Tracking
and Classification Beyond the Visible Spectrum (OTCBVS)}
database\footnote{IEEE OTCBVS WS Series Bench; DOE University
Research Program in Robotics under grant DOE-DE-FG02-86NE37968;
DOD/TACOM/NAC/ARC Program under grant R01-1344-18; FAA/NSSA grant
R01-1344-48/49; Office of Naval Research under grant
\#N000143010022.}, freely available for download at {\small
\url{http://www.cse.ohio-state.edu/OTCBVS-BENCH/}}. Briefly, this
database contains 29 individuals, 11 roughly matching poses in
visual and thermal spectra and large illumination variations (some
of these are exemplified in Figure~\ref{Fig: Example Data}). Images
were acquired using the Raytheon Palm-IR-Pro camera in the thermal
and Panasonic WV-CP234 camera in the visual spectrum, in the
resolution of $240 \times 320$ pixels.

\begin{figure*}[!t]
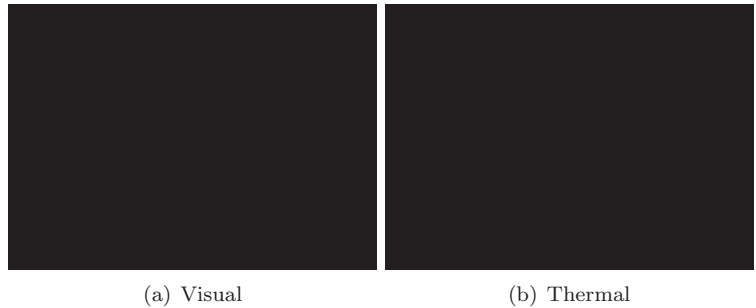


    \index{image!set matching}

  \centering
  \subfigure[Visual]
    {\includegraphics[width=0.35\textwidth]{visual_nfaces.eps}}
  \subfigure[Thermal]
    {\includegraphics[width=0.35\textwidth]{thermal_nfaces.eps}}
  \caption[Number of training images across the data set.]
                   {\it Shown are histograms of the number of
                        images per person used to train our algorithm. Depending on the
                        exact head poses assumed by the user we typically obtained
                        7-8 visual spectrum images and typically a slightly lower number
                        for the thermal spectrum. The range of yaw angles covered is roughly
                        $\pm 45^{\circ}$ measured from the frontal face orientation.  }
            \label{Fig: N Faces}
  \vspace{6pt}\hrule
\end{figure*}

Our algorithm was trained using all images in a single illumination
in which all 3 salient facial features could be detected. This
typically resulted in 7-8 images in the visual and 6-7 in the
thermal spectrum, see Figure~\ref{Fig: N Faces}, and roughly $\pm
45^{\circ}$ yaw range, as measured from the frontal face
orientation.

The performance of the algorithm was evaluated both in 1-to-N and
1-to-1 matching scenarios. In the former case, we assumed that test
data corresponded to one of people in the training set and
recognition was performed by associating it with the closest match.
Verification (or 1-to-1 matching, \textit{``is this the same
person?''}) performance was quantified by looking at the true
positive admittance rate for a threshold that corresponds to 1
admitted intruder in 100.

\subsection{Results}

    \index{biometric!fusion}
    \index{features!local}
    \index{features!holistic}
    \index{filter!band pass}
    \index{verification}

A summary of 1-to-N matching results is shown in Table~\ref{Fig:
Results}.

Firstly, note the poor performance achieved using both raw visual as
well as raw thermal data. The former is suggestive of challenging
illumination changes present in the OTCBVS data set. This is further
confirmed by significant improvements gained with both band-pass
filtering and the Self-Quotient Image which increased the average
recognition rate for, respectively, 35\% and 47\%. The same is
corroborated by the Receiver-Operator Characteristic curves in
Figure~\ref{Fig: Holistic ROCs} and 1-to-1 matching results in
Table~\ref{Fig: Holistic}.

On the other hand, the reason for low recognition rate of raw
thermal imagery is twofold: it was previously argued that the two
main limitations of this modality are the inherently lower
discriminative power and occlusions caused by prescription glasses.
The addition of the glasses detection module is of little help at
this point - some benefit is gained by steering away from
misleadingly good matches between any two people wearing glasses,
but it is limited in extent as a very discriminative region of the
face is lost. Furthermore, the improvement achieved by optimal
band-pass filtering in thermal imagery is much more modest than with
visual data, increasing performance respectively by 35\% and 8\%.
Similar increase was obtained in true admittance rate (42\% vs.
8\%), see Table~\ref{Fig: Holistic ROCs}.

\begin{figure}
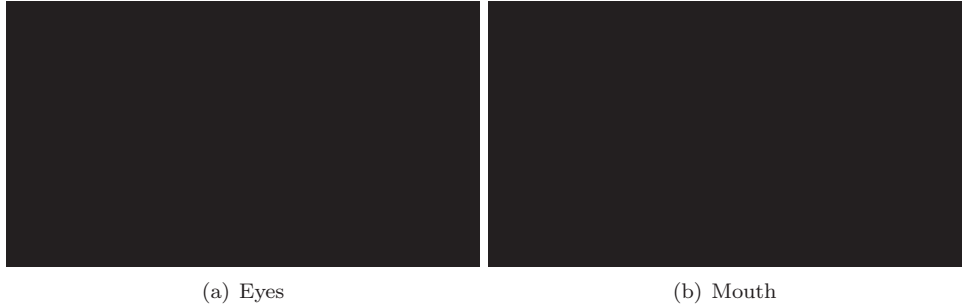


    \index{face!appearance}
    \index{spectrum!visual}
    \index{spectrum!thermal}
    \index{verification}

  \centering
  \subfigure[Eyes]
    {\includegraphics[width=0.45\textwidth]{roc_eyes.eps}}
  \subfigure[Mouth]
    {\includegraphics[width=0.45\textwidth]{roc_mouth.eps}}
  \caption[ROC curves: local features]{\it Isolated local features Receiver-Operator Characteristics (ROC):
               for visual (blue) and thermal (red) spectra. }
            \label{Fig: Local ROCs}
\end{figure}

\begin{figure}
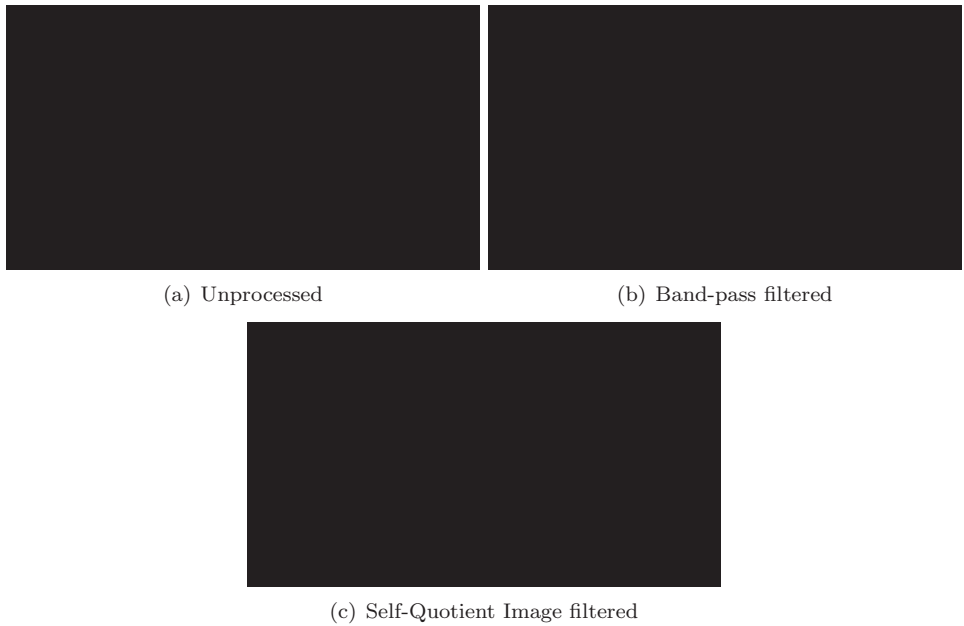


    \index{face!appearance}
    \index{spectrum!visual}
    \index{spectrum!thermal}
    \index{filter!band pass}
    \index{verification}

  \centering
  \subfigure[Unprocessed]
    {\includegraphics[width=0.45\textwidth]{roc_raw.eps}}
  \subfigure[Band-pass filtered]
    {\includegraphics[width=0.45\textwidth]{roc_bp.eps}}
    \subfigure[Self-Quotient Image filtered]
    {\includegraphics[width=0.45\textwidth]{roc_sqi.eps}}
  \caption[ROC curves: holistic representations]
                   {\it Holistic representations Receiver-Operator Characteristics (ROC) for
                        visual (blue) and thermal (red) spectra. }
            \label{Fig: Holistic ROCs}
\end{figure}

Neither the eyes or the mouth regions, in either the visual or
thermal spectrum, proved very discriminative when used in isolation,
see Figure~\ref{Fig: Local ROCs}. Only 10-12\% true positive
admittance was achieved, as shown in Table~\ref{Fig: Features}.
However, the proposed fusion of holistic and local appearance
offered a consistent and statistically significant improvement. In
1-to-1 matching the true positive admittance rated increased for
4-6\%, while the average correct 1-to-N matching improved for
roughly 2-3\%.

The greatest power of the method becomes apparent when the two
modalities, visual and thermal, are fused. In this case the role of
the glasses detection module is much more prominent, drastically
decreasing the average error rate from 10\% down to 3\%, see
Table~\ref{Fig: Results}. Similarly, the true admission rate
increases to 74\% when data is fused without special handling of
glasses, and to 80\% when glasses are taken into account.

\begin{table}

    \index{face!appearance}
    \index{spectrum!visual}
    \index{spectrum!thermal}
    \index{biometric!fusion}
    \index{glasses}
    \index{filter!band pass}

  \centering
  \caption[Recognition results summary.]{\it Shown is the average rank-1
                        recognition rate using different representations across all combinations
                        of illuminations. Note the performance increase with each of the main features
                        of our system: image filtering, combination of holistic and local features,
                        modality fusion and prescription glasses detection. \vspace{10pt}}
  \Large
  \begin{tabular*}{1.00\textwidth}{@{\extracolsep{\fill}}l|l|c}
    \Hline
      \multicolumn{2}{c|}{\textbf{\normalsize Representation}}   & \textbf{\normalsize Recognition}\\
    \hline
      \multirow{4}{*}{\normalsize ~Visual} & \small Holistic raw data   & \small 0.58 \\
                              & \small Holistic, band-pass              & \small 0.78 \\
                              & \small Holistic, SQI filtered           & \small 0.85 \\
                              & \small Mouth+eyes+holistic & \multirow{2}{*}{\small 0.87} \\
                              & \small data fusion, SQI filtered   & \\
    \hline
      \multirow{5}{*}{\normalsize ~Thermal} & \small Holistic raw data  & \small 0.74 \\
                               & \small Holistic raw w/   & \multirow{2}{*}{\small 0.77} \\
                               & \small glasses detection & \\
                               & \small Holistic, low-pass filtered     & \small 0.80 \\
                               & \small Mouth+eyes+holistic             & \multirow{2}{*}{\small 0.82 } \\
                               & \small data fusion, low-pass filtered~~~  &  \\
    \hline
      \multirow{2}{*}{\normalsize ~Proposed thermal + visual fusion~~}     & \small w/o glasses detection & \small 0.90 \\
                                                & \small w/ glasses detection  & \textit{\small 0.97} \\
    \Hline
  \end{tabular*}
  \label{Fig: Results}
\end{table}

\begin{table}

    \index{face!appearance}
    \index{spectrum!visual}
    \index{spectrum!thermal}
    \index{filter!band pass}
    \index{verification}

  \centering
  \caption[Verification results summary.]{\it A summary of the comparison of different image processing filters for 1
                        in 100 intruder acceptance rate. Both the simple band-pass filter, and even
                        further its locally-scaled variant, greatly improve performance. This is
                        most significant in the visual spectrum, in which image intensity in the
                        low spatial frequency is most affected by illumination changes.\vspace{10pt} }
  \Large
  \begin{tabular*}{1.00\textwidth}{@{\extracolsep{\fill}}l|cc}
    \Hline
      \textbf{\normalsize ~Representation} & \textbf{\normalsize Visual} & \textbf{\normalsize Thermal}\\
    \hline
      \multicolumn{3}{c}{\normalsize 1\% intruder acceptance}\\
    \hline
      {\normalsize ~Unprocessed/raw}           & \small 0.2850 & \small 0.5803 \\
      {\normalsize ~Band-pass filtered (BP)}   & \small 0.4933 & \small 0.6287 \\
      {\normalsize ~Self-quotient image (SQI)} & \small 0.6410 & \small 0.6301 \\
    \Hline
  \end{tabular*}
  \label{Fig: Holistic}
\end{table}

\begin{table}

    \index{face!appearance}
    \index{spectrum!visual}
    \index{spectrum!thermal}

  \centering
  \caption[Verification results: local features.]{\it A summary of the results for 1 in 100 intruder acceptance rate.
               Local features in isolation perform very poorly.\vspace{10pt} }
  \Large
  \begin{tabular*}{1.00\textwidth}{@{\extracolsep{\fill}}l|cc}
    \Hline
      \textbf{\normalsize ~Representation} & \textbf{\normalsize Visual (SQI)} & \textbf{\normalsize Thermal (BP)}\\
    \hline
      \multicolumn{3}{c}{\normalsize 1\% intruder acceptance}\\
    \hline
      {\normalsize ~Eyes}       & \small 0.1016 & \small 0.2984 \\
      {\normalsize ~Mouth}      & \small 0.1223 & \small 0.3037 \\
    \Hline
  \end{tabular*}
  \label{Fig: Features}
\end{table}

\begin{table}

    \index{face!appearance}
    \index{spectrum!visual}
    \index{spectrum!thermal}
    \index{biometric!fusion}
    \index{verification}

  \centering
  \caption[Verification results: hybrid representations.]
  {\it Holistic \& local features -- a summary of 1-to-1 matching (verification) results.\vspace{10pt} }
  \Large
  \begin{tabular*}{1.00\textwidth}{@{\extracolsep{\fill}}l|cc}
    \Hline
      \textbf{\normalsize ~Representation} & \textbf{\normalsize Visual (SQI)} & \textbf{\normalsize Thermal (BP)}\\
    \hline
      \multicolumn{3}{c}{\normalsize 1\% intruder acceptance}\\
    \hline
      {\normalsize ~Holistic + Eyes}         & \small 0.6782 & \small 0.6499 \\
      {\normalsize ~Holistic + Mouth}        & \small 0.6410 & \small 0.6501 \\
      {\normalsize ~Holistic + Eyes + Mouth} & \small 0.6782 & \small 0.6558 \\
    \Hline
  \end{tabular*}
  \label{Fig: Comb}
\end{table}

\begin{table}

    \index{face!appearance}
    \index{spectrum!visual}
    \index{spectrum!thermal}
    \index{biometric!fusion}
    \index{glasses}
    \index{verification}

  \centering
  \caption[Verification results: glasses detection.]{\it Feature and modality fusion -- a summary of the 1-to-1 matching (verification) results. \vspace{10pt}}
  \Large
  \begin{tabular*}{1.00\textwidth}{@{\extracolsep{\fill}}l|cc}
    \Hline
      \textbf{\normalsize ~Representation} & \textbf{\normalsize True admission rate}\\
    \hline
      \multicolumn{2}{c}{\normalsize 1\% intruder acceptance}\\
    \hline
      {\normalsize ~Without glasses detection} & \small 0.7435 \\
      {\normalsize ~With glasses detection} & \small 0.8014 \\
    \Hline
  \end{tabular*}
  \label{Fig: Fusion}
\end{table}

\section{Summary and conclusions}

    \index{spectrum!visual}
    \index{spectrum!thermal}
    \index{filter!band pass}
    \index{biometric!fusion}
    \index{glasses}

In this chapter we described a system for personal identification
based on a face biometric that uses cues from visual and thermal
imagery. The two modalities are shown to complement each other,
their fusion providing good illumination invariance and
discriminative power between individuals. Prescription glasses, a
major difficulty in the thermal spectrum, are reliably detected by
our method, restricting the matching to non-affected face regions.
Finally, we examined how different preprocessing methods affect
recognition in the two spectra, as well as holistic and local
feature-based face representations. The proposed method was shown to
achieve a high recognition rate (97\%) using only a small number of
training images (5-7) in the presence of large illumination changes.

\section*{Related publications}

The following publications resulted from the work presented in this
chapter:

\begin{itemize}
  \item O. Arandjelovi\'c, R. Hammoud and R. Cipolla. Multi-sensory face biometric fusion (for
                  personal identification). In \textit{Proc. IEEE Workshop on Object Tracking and
                  Classification Beyond the Visible Spectrum (OTCBVS)}, page 52, June
                  2006. \cite{AranHammCipo2006}

  \item O. Arandjelovi\'c, R. Hammoud and R. Cipolla. \textit{Face
                  Biometrics for Personal Identification}, chapter Towards person authentication
          by fusing visual and thermal face biometrics. Springer-Verlag, 2007.
          ISBN 978-3-540-49344-0. \cite{AranHammCipo2007}

  \item O. Arandjelovi\'c, R. Hammoud and R. Cipolla. On face recognition by fusing visual and
                  thermal face biometrics. In \textit{Proc. IEEE International Conference on
                  Advanced Video and Signal Based Surveillance (AVSS)}, pages 50--56, November 2006.
                  \cite{AranHammCipo2006a}

  \item O. Arandjelovi\'c, R. Hammoud and R. Cipolla. Thermal and reflectance based personal identification methodology
                  in challenging variable illuminations, \textit{Pattern Recognition}, \textbf{43}(5):pages 1801--1813, May 2010.
                  \cite{AranHammCipo2010}

\end{itemize}

\graphicspath{{./06filters/}}
\chapter{Illumination Invariance using Image Filters}
\label{Chp: Filters}
\begin{center}
  \footnotesize
  \vspace{-20pt}
  \framebox{\includegraphics[trim=0cm 0.3cm 0cm 0cm, clip=true,width=0.80\textwidth]{title_img.eps}}\\
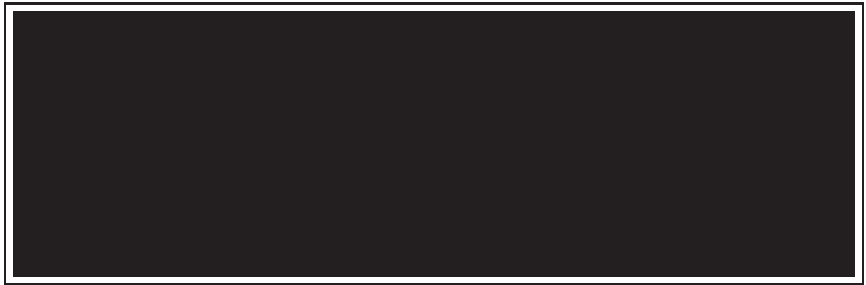
  El Greco. \textit{The Purification of the Temple}\\
  1571-76, Oil on canvas, 117 x 150 cm\\
  Institute of Arts, Minneapolis
\end{center}

\cleardoublepage
In the previous chapter recognition the invariance to illumination
condition was achieved by fusing face biometrics acquired in the
visual and thermal spectrum. While successful, in practice this
approach suffers from the limited availability and high cost of
thermal imagers. We wish to achieve the same using visual data only,
acquired with an inexpensive and readily available optical camera.

In this chapter we show that image processed visual data can be used
to much the same effect as we used thermal data, fusing it with raw
visual data. The framework is based on simple image processing
filters that compete with unprocessed greyscale input to yield a
single matching score between individuals. It is shown how the
discrepancy between illumination conditions between novel input and
the training data set can be estimated and used to weigh the
contribution of two competing representations. Evaluated on \textit{CamFace},
\textit{ToshFace} and \textit{Face Video} databases, our algorithm consistently
demonstrated a dramatic performance improvement over traditional
filtering approaches. We demonstrate a reduction of 50--75\% in
recognition error rates, the best performing method-filter
combination correctly recognizing 96\% of the individuals.

\section{Adapting to data acquisition conditions}
The framework proposed in this chapter is most closely motivated by
the findings first reported in \cite{AranCipo2006b}. In that paper
several face recognition algorithms were evaluated on a large database using (i)
raw greyscale input, (ii) a high-pass (HP) filter and (iii) the
Self-Quotient Image (QI) \cite{WangLiWang2004}. Both the high-pass
and even further Self Quotient Image representations produced an
improvement in recognition for all methods over raw grayscale, which
is consistent with previous findings in the literature
\cite{AdinMoseUllm1997, AranZiss2005, FitzZiss2002, WangLiWang2004}. Of
importance to this work is that it was also examined in which cases
these filters help and how much depending on the data acquisition
conditions. It was found, consistently over different algorithms,
that recognition rates using greyscale and either the HP or the QI
filter negatively correlated (with $\rho \approx -0.7$), as illustrated in Figure~\ref{Fig:
Correlation}.

\begin{figure*}
  \centering
  \includegraphics[width=0.65\textwidth]{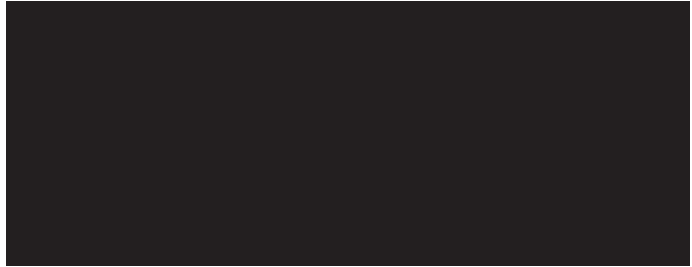}
  \caption[Image filtering paradox.]
    { \it A plot of the performance improvement with HP and QI filters against
          the performance of unprocessed, raw imagery across different
          illumination combinations used in training and test. The tests
          are shown in the order of increasing raw data performance for
          easier visualization. }
  \label{Fig: Correlation}
  \vspace{6pt}\hrule\vspace{15pt}
\end{figure*}

This is an interesting result: it means that while on average both
representations increase the recognition rate, they actually
\emph{worsen} it in ``easy'' recognition conditions when no
normalization is needed. The observed phenomenon is well understood
in the context of energy of intrinsic and extrinsic image
differences and noise (see \cite{WangTang2003} for a thorough
discussion). Higher than average recognition rates for raw input
correspond to small changes in imaging conditions between training
and test, and hence lower energy of extrinsic variation. In this
case, the two filters decrease the signal-to-noise ratio, worsening the performance.
On the other hand, when the imaging conditions between training and
test are very different, normalization of extrinsic variation is the
dominant factor and performance is improved, see Figure~\ref{Fig: Signal
E}~(b).

\begin{figure*}[!ht]
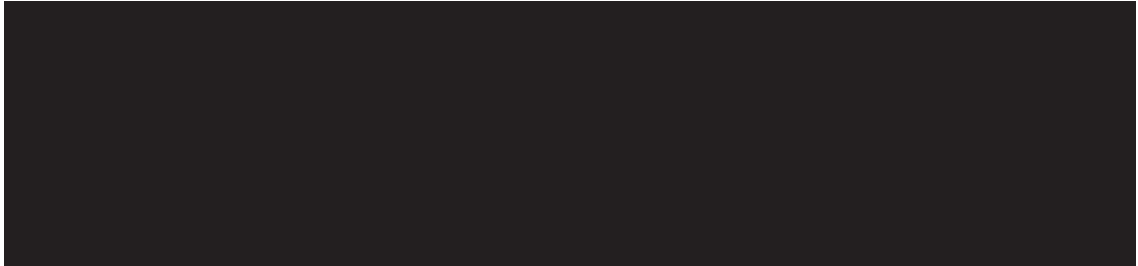
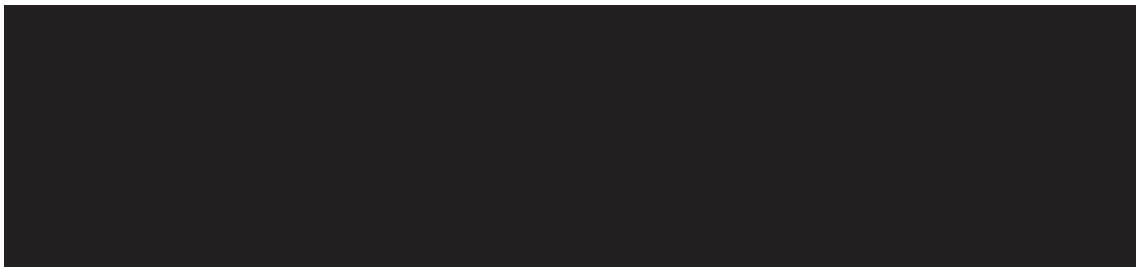

  \centering
  \subfigure[Similar acquisition conditions between sequences]{
    \begin{tabular}{VVV}
      \hspace{-25pt} \includegraphics[width=0.5\textwidth]{easy_1.eps} & \hspace{-20pt} \includegraphics[width=0.1\textwidth]{filter_arrow.eps} & \hspace{-20pt} \includegraphics[width=0.5\textwidth]{easy_2.eps}
    \end{tabular}
  }
  \subfigure[Different acquisition conditions between sequences]{
    \begin{tabular}{VVV}
      \hspace{-25pt}\includegraphics[width=0.5\textwidth]{hard_1.eps} & \hspace{-20pt} \includegraphics[width=0.1\textwidth]{filter_arrow.eps} & \hspace{-20pt} \includegraphics[width=0.5\textwidth]{hard_2.eps}
    \end{tabular}
  }
  \caption[Filtering effects on signal energy distribution.]
    { \it A conceptual illustration of the distributions of intrinsic, extrinsic and noise
          signal energies across frequencies in the cases when training and test data
          acquisition conditions are (a) similar and (b) different, before (left) and
          after (right) band-pass filtering.  }
  \label{Fig: Signal E}
  \vspace{6pt}\hrule\vspace{25pt}
\end{figure*}

This is an important observation: it suggests that the performance
of a method that uses either of the representations can be increased
further by detecting the difficulty of recognition conditions. In
this chapter we propose a novel learning framework to do exactly this.

\subsection{Adaptive framework} Our goal is to implicitly learn how
similar the novel and training (or \emph{gallery}) illumination
conditions are, to appropriately emphasize either the raw input
guided face comparisons or of its filtered output. Figure~\ref{Fig: No
Separation} shows the difficulty of this task: different classes
(i.e.\ persons) are not well separated in the space of 2D feature
vectors obtained by stacking raw and filtered similarity scores.

\begin{figure}
  \centering
  \includegraphics[width=0.65\textwidth]{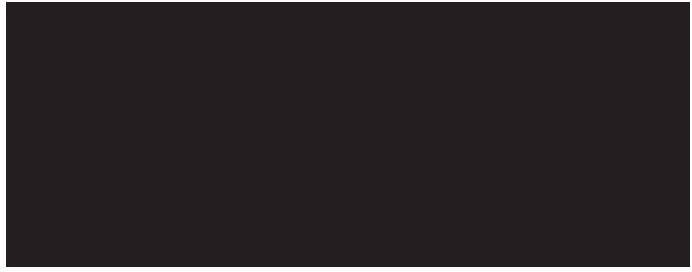}
  \caption[Class separation with simple decision-level fusion.]
                 { \it  Distances ($0-1$) between sets of faces -- interpersonal and
                        intrapersonal comparisons are shown respectively as large
                        red and small blue dots. Individuals are poorly separated. }
  \label{Fig: No Separation}
  \vspace{6pt}\hrule
\end{figure}

Let $\{ \mathcal{X}_1, \dots, \mathcal{X}_N \} $ be a database of
known individuals, $\mathcal{X}$ novel input corresponding to one of
the gallery classes and $\rho()$ and $F()$, respectively, a given
similarity function and a quasi illumination-invariant filter. We
then express the degree of belief $\eta$ that two face sets
$\mathcal{X}$ and $\mathcal{X}_i$ belong to the same person as a
weighted combination of similarities between the corresponding
unprocessed and filtered image sets:
\begin{equation}
    \eta = (1 - \alpha^*) \rho(\mathcal{X}, \mathcal{X}_i) + \alpha^* \rho(F(\mathcal{X}),
    F(\mathcal{X}_i))
    \label{Eqn: Similarity}
\end{equation}

In the light of the previous discussion, we want $\alpha^*$ to be
small (closer to $0.0$) when novel and the corresponding gallery
data have been acquired in similar illuminations, and large (closer
to $1.0$) when in very different ones. We show that $\alpha^*$ can
be learnt as a function:
\begin{equation}
    \alpha^* = \alpha^*(\mu),
    \label{Eqn: Alpha Fn}
\end{equation}
where $\mu$ is the \emph{confusion margin} -- the difference between
the similarities of the two  $\mathcal{X}_i$ most similar to
$\mathcal{X}$. As in Chapter~\ref{Chp: Thermal}, we compute an estimate
of $\alpha^*(\mu)$ in a maximum a posteriori sense:
\begin{equation}
    \alpha^*(\mu) = \arg \max_{\alpha} p(\alpha | \mu),
\end{equation}
which, under the assumption of a uniform prior on the confusion margin $\mu$, reduces to:
\begin{equation}
    \alpha^*(\mu) = \arg \max_{\alpha} p(\alpha, \mu),
    \label{Eqn: Alpha Def}
\end{equation}
where $p(\alpha, x)$ is the probability that $\alpha$ is the optimal
value of the mixing coefficient. The proposed offline learning algorithm
entirely analogous to the algorithm described in
Section~\ref{SS: Fusing modalities}, so here we just summarize it in
Figure~\ref{Fig: Algorithm} with a typical evolution of $p(\alpha,
\mu)$ shown in Figure~\ref{Fig: Iterations}. The final stage of the offline learning in our method involves
imposing the monotonicity constraint on $\alpha^*(\mu)$ and
smoothing of the result, see Figure~\ref{Fig: Alpha Dists}.

\begin{figure}[!h]
    \centering
    \begin{tabular}{l}
     \hline\\
          \begin{tabular}{ll}
              \textbf{Input}:  & training data $D(person, illumination)$,\\
                               & filtered data $F(person, illumination)$,\\
                               & similarity function $\rho$,\\
                               & filter $F$.\\
              \textbf{Output}: & estimate $\hat{p}(\alpha, \mu)$.\\
          \end{tabular}\vspace{5pt}
          \\ \hline \\
          \begin{tabular}{l}
            \textbf{1: Init}\\
            \hspace{10pt}$\hat{p}(\alpha, \mu) = 0$,\\\\

            \textbf{2: Iteration}\\
            \hspace{10pt}\textbf{for}$\text{ all illuminations } i,~j$ \textbf{ and} $\text{ persons }
            p$\\\\

            \hspace{15pt}\textbf{3: Initial separation}\\
            \hspace{25pt}$\delta_0 = \min_{q \neq p} \left[ \rho(D(p, i), D(q, j)) - \rho(D(p, i), D(p, j))
            \right]$\\\\

             \hspace{15pt}\textbf{4: Iteration}\\
             \hspace{25pt}\textbf{for}$\text{ all } k = 0, \dots, 1 / \Delta \alpha,~\alpha = k \Delta \alpha$\\\\

             \hspace{30pt}\textbf{5: Separation given $\alpha$}\\
             \hspace{40pt}$\delta(k \Delta \alpha) = \min_{q \neq p} \large[ \alpha \rho(F(p, i), F(q, j))$\\
                                                     \hspace{112pt}$- \alpha \rho(F(p, i), F(p, j))$\\
                                                     \hspace{112pt}$+ (1 - \alpha)\rho(D(p, i), D(q, j))$\\
                                                     \hspace{112pt}$- (1 - \alpha)\rho(D(p, i), D(p, j))\Large]$\\\\

             \hspace{30pt}\textbf{6: Update density estimate}\\
             \hspace{40pt}$\hat{p}(k \Delta \alpha, \delta_0) = \hat{p}(k \Delta \alpha, \delta_0) + \delta(k \Delta
             \alpha)$\\\\

             \textbf{7: Smooth the output}\\
             \hspace{10pt}$\hat{p}(\alpha, \mu) =  \hat{p}(\alpha, \mu) \ast \mathbf{G}_{\sigma=0.05} $\\\\

             \textbf{8: Normalize to unit integral}\\
             \hspace{10pt}$\hat{p}(\alpha, \mu) =  \hat{p}(\alpha, \mu) / \int_{\alpha} \int_{x} \hat{p}(\alpha, x) dx
             d\alpha$\\\\\hline
          \end{tabular}\\
    \end{tabular}
    \caption[Fusion algorithm.]{ \it Offline training algorithm. }
    \label{Fig: Algorithm}
\end{figure}

\clearpage

\begin{figure*}
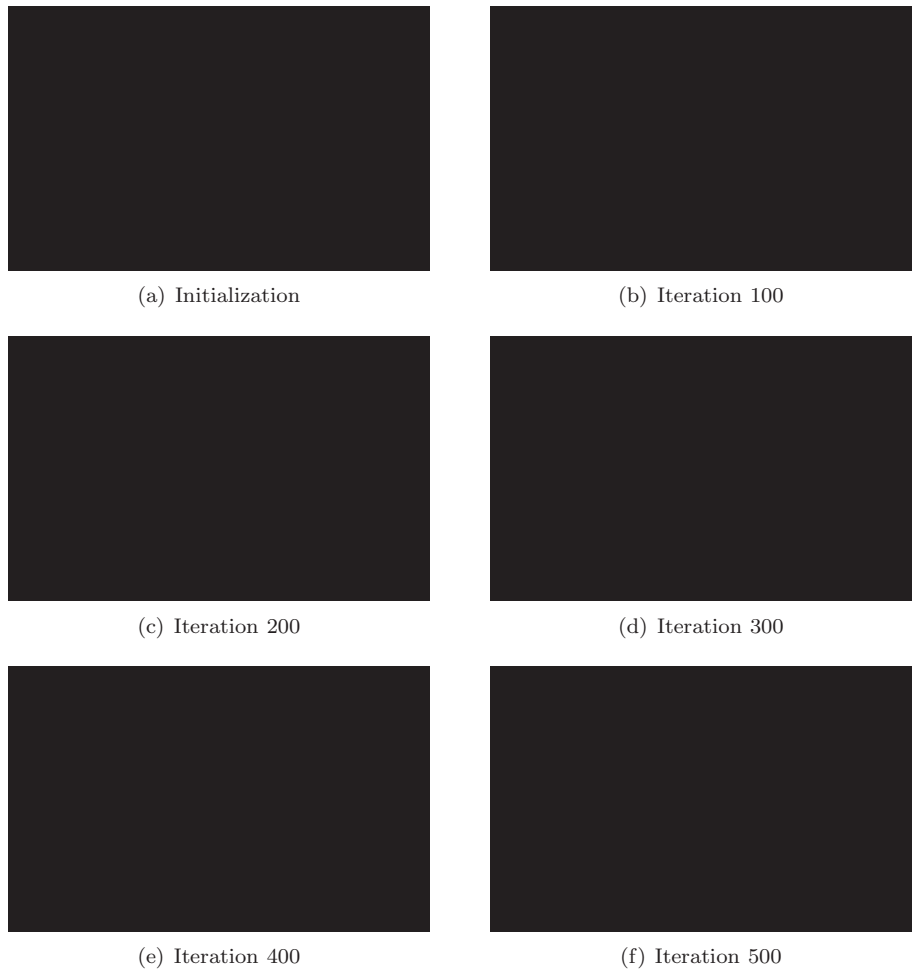

  \centering
  \subfigure[Initialization]{\includegraphics[width=0.40\textwidth]{map_0001.eps}}~~~~~~
  \subfigure[Iteration 100]{\includegraphics[width=0.40\textwidth]{map_0003.eps}}\\
  \subfigure[Iteration 200]{\includegraphics[width=0.40\textwidth]{map_0005.eps}}~~~~~~
  \subfigure[Iteration 300]{\includegraphics[width=0.40\textwidth]{map_0007.eps}}\\
  \subfigure[Iteration 400]{\includegraphics[width=0.40\textwidth]{map_0009.eps}}~~~~~~
  \subfigure[Iteration 500]{\includegraphics[width=0.40\textwidth]{map_0011.eps}}
  \caption[Iterative density estimate.]{ \it The estimate of the joint density $p(\alpha, \mu)$ through 500 iterations
            for a band-pass filter used for the evaluation of the proposed framework in
            Section~\ref{SS: Results}. }
            \label{Fig: Iterations}
\end{figure*}

\clearpage

\begin{figure*}
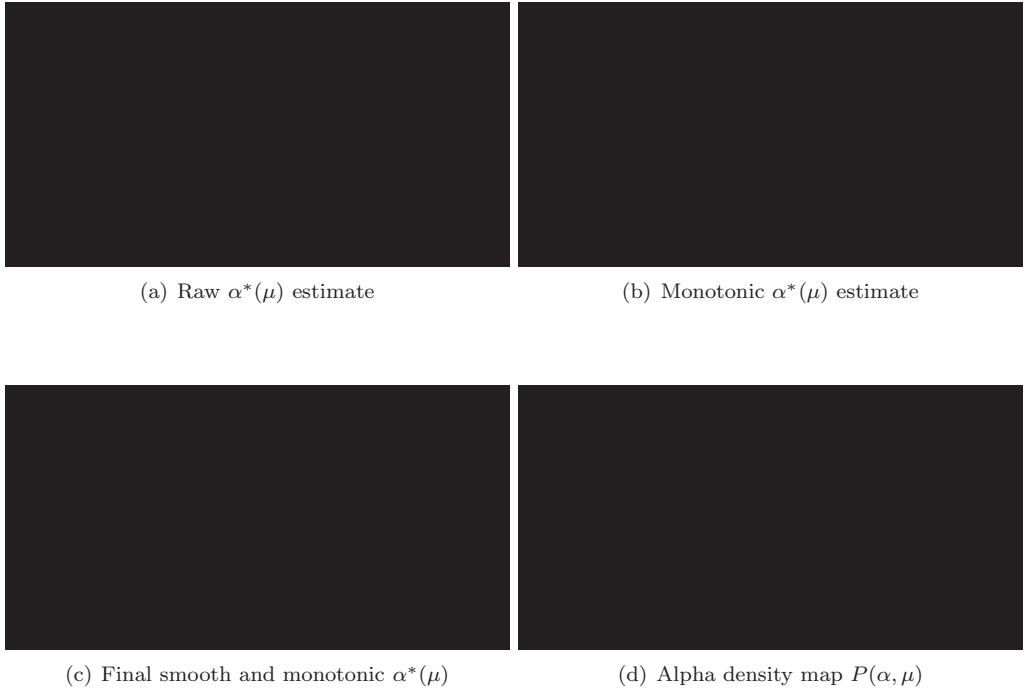

  \centering
  \subfigure[Raw $\alpha^*(\mu)$ estimate]{\includegraphics[width=0.48\textwidth]{alpha1.eps}}
  \subfigure[Monotonic $\alpha^*(\mu)$ estimate]{\includegraphics[width=0.48\textwidth]{alpha2.eps}}
  \\\vspace{20pt}
  \subfigure[Final smooth and monotonic $\alpha^*(\mu)$]{\includegraphics[width=0.48\textwidth]{alpha3.eps}}
  \subfigure[Alpha density map $P(\alpha, \mu)$]{\includegraphics[width=0.48\textwidth]{alpha_map_dx.eps}}
  \caption[Learning the $alpha$-function.]
    { \it Typical estimates of the $\alpha$-function plotted against confusion margin $\mu$ (a--c).
                    The estimate shown was computed using 40 individuals in 5 illumination conditions for
                    a Gaussian high-pass filter. As expected, $\alpha^*$ assumes low values for small
                    confusion margins and high values for large confusion margins (see \eqref{Eqn: Similarity}).
                    Learnt probability density $p(\alpha, \mu)$ (greyscale surface) and
                    a superimposed raw estimate of the $\alpha$-function (solid red line) for
                    a high-pass filter are shown in (d).
                     }
  \label{Fig: Alpha Dists}
\end{figure*}

\section{Empirical evaluation}\label{Sec: Empirical Evaluation}
\begin{figure}
  \centering
  \includegraphics[width=0.9\textwidth]{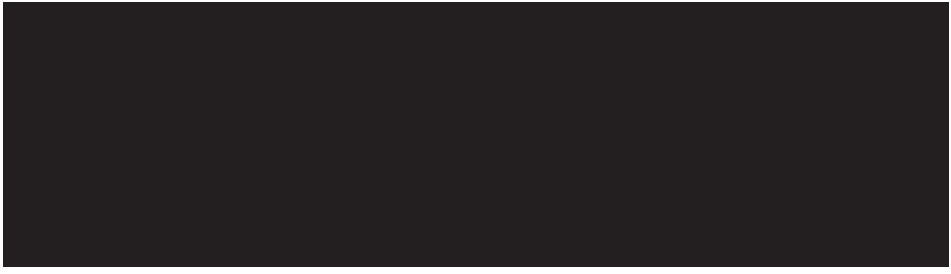}
  \caption[Evaluated filters.]{ \it An example of the original image and
            the 6 corresponding filtered representations we evaluated.}
            \label{Fig: Representations}
\end{figure}

The proposed framework was evaluated using the following filters (illustrated in
Figure~\ref{Fig: Representations}):
\begin{itemize}
    \item Gaussian high-pass filtered images \cite{AranZiss2005,FitzZiss2002} (HP):
        \begin{align}
          \mathbf{X}_H = \mathbf{X} - (\mathbf{X} \ast \mathbf{G}_{\sigma =
          1.5}),
      \end{align}

    \item local intensity-normalized high-pass filtered images -- similar to the Self-Quotient Image \cite{WangLiWang2004} (QI):
      \begin{align}
        \mathbf{X}_Q = \mathbf{X}_H~./~\mathbf{X}_L \equiv \mathbf{X}_H~./~(\mathbf{X} - \mathbf{X}_H),
      \end{align}
      the division being element-wise,

    \item distance-transformed edge map \cite{AranCipo2006b,Cann1986} (ED):
      \begin{align}
        \mathbf{X}_{ED} = &\text{DistanceTransform}\big[ \mathbf{X}_E \big] \\
                          &\equiv \text{DistanceTransform}\big[\text{Canny}(\mathbf{X})\big],
      \end{align}

    \item Laplacian-of-Gaussian \cite{AdinMoseUllm1997} (LG):
      \begin{align}
        \mathbf{X}_L = \mathbf{X} \ast \nabla \mathbf{G}_{\sigma=3},
      \end{align}
      where $\ast$ denotes convolution, and

    \item directional grey-scale derivatives \cite{AdinMoseUllm1997,EverZiss2004} (DX, DY):
      {\begin{align}
        &\mathbf{X}_x = \mathbf{X} \ast \frac {\partial} {\partial x} \mathbf{G}_{\sigma_x =
        6}\\\\
        &\mathbf{X}_y = \mathbf{X} \ast \frac {\partial} {\partial y} \mathbf{G}_{\sigma_y = 6}.
      \end{align}}
\end{itemize}

To demonstrate the contribution of the proposed framework, we evaluated it
with two well-established methods in the literature:
\begin{itemize}
    \item Constrained MSM (CMSM) \cite{FukuYama2003} used in a state-of-the-art
    commercial system FacePass$^{\circledR}$ \cite{Tosh}, and

    \item Mutual Subspace Method (MSM) \cite{FukuYama2003}.
\end{itemize}
In all tests, both training data for each person in the gallery, as
well as test data, consisted of only a single sequence. Offline
training of the proposed algorithm was performed using 40
individuals in 5 illuminations from the \textit{CamFace} data set. We
emphasize that these were not used as test input for the evaluations
reported in this section.

\subsection{Results}\label{SS: Results}
We evaluated the performance of CMSM and MSM using each of the
7 face image representations (raw input and 6 filter outputs).
Recognition results for the 3 databases are shown in blue in
Figure~\ref{Fig: Results CFDB} (the results on \textit{Face Video} data set are
tabulated in Figure~\ref{Fig: Results CFDB}~(c), for the ease of
visualization). Confirming the first premise of this work as well as
previous research findings, all of the filters produced an
improvement in average recognition rates. Little interaction between
method/filter combinations was found, Laplacian-of-Gaussian and the
horizontal intensity derivative producing the best results and
bringing the best and average recognition errors down to 12\% and
9\% respectively.

\begin{figure}
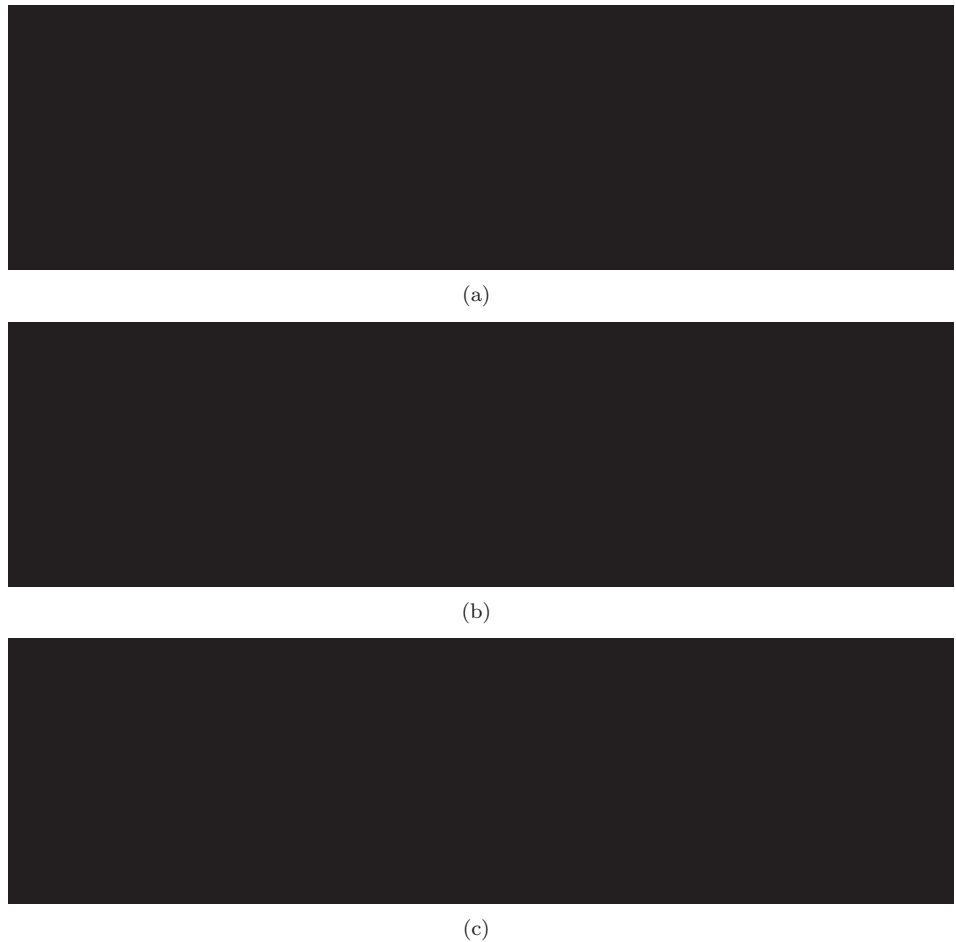

  \centering
  \subfigure[]{\includegraphics[width=0.9\textwidth]{representations_pvs.eps}}
  \subfigure[]{\includegraphics[width=0.9\textwidth]{seq1.eps}}
  \subfigure[]{\includegraphics[width=0.9\textwidth]{seq2.eps}}
  \caption[Principal vectors using different filters.]
      { \it (a) The first pair of principal vectors (top and
                bottom) corresponding to the sequences (b) and (c)
                (every 4$^{\text{th}}$ detection is shown for
                compactness),
                for each of the 7 representations used in
                the empirical evaluation described in this chapter. A
                higher degree of similarity between the two vectors
                indicates a greater degree of illumination invariance
                of the corresponding filter. }
            \label{Fig: PVs}
\end{figure}

In the last set of experiments, we employed each of the 6 filters in
the proposed data-adaptive framework. Recognition results for the 3
databases are shown in red in Figure~\ref{Fig: Results CFDB}. The
proposed method produced a dramatic performance improvement in the
case of all filters, reducing the average recognition error rate to
only 4\% in the case of CMSM/Laplacian-of-Gaussian combination.
An improvement in the robustness to illumination changes can also be seen in the
significantly reduced standard deviation of the recognition.
Finally, it should be emphasized that the demonstrated improvement
is obtained with a negligible increase in the computational cost as
all time-demanding learning is performed offline.

\begin{figure*}[t]
  \centering
  \small
  \subfigure[CamFace]{
    \includegraphics[width=0.5\textwidth]{res_cfdb.eps}
    \includegraphics[width=0.5\textwidth]{res_cfdb_std.eps}
  }

  \subfigure[ToshFace]{
    \includegraphics[width=0.5\textwidth]{res_tfdb.eps}
    \includegraphics[width=0.5\textwidth]{res_tfdb_std.eps}
  }

  \subfigure[Face Video Database, mean error (\%)]{
  \Large
  \begin{tabular}{c|ccccccc}
    \Hline
            & \normalsize RW   & \normalsize HP   & \normalsize QI   & \normalsize ED   & \normalsize LG   & \normalsize DX   & \normalsize DY   \\
    \hline
    \normalsize MSM     & \small 0.00 & \small 0.00 & \small 0.00 & \small 0.00 & \small 9.09 & \small 0.00 & \small 0.00 \\
    \normalsize MSM-AD  & \small 0.00 & \small 0.00 & \small 0.00 & \small 0.00 & \small 0.00 & \small 0.00 & \small 0.00 \\
    \normalsize CMSM    & \small 0.00 & \small 9.09 & \small 0.00 & \small 0.00 & \small 0.00 & \small 0.00 & \small 0.00 \\
    \normalsize CMSM-AD & \small 0.00 & \small 0.00 & \small 0.00 & \small 0.00 & \small 0.00 & \small 0.00 & \small 0.00 \\
    \Hline
  \end{tabular}
  }

  \caption[Recognition results.]{ \it Error rate statistics. The proposed framework
            (-AD suffix) dramatically improved recognition
            performance on all method/filter combinations, as
            witnessed by the reduction in both error rate averages
            and their standard deviations. }
  \label{Fig: Results CFDB}
\end{figure*}

\subsection{Failure modes}
In the discussion of failure modes of the described framework, it is
necessary to distinguish between errors introduced by a
\emph{particular} image processing filter used, and the fusion
algorithm itself. As generally recognized across literature (e.g.\
see \cite{AdinMoseUllm1997}), qualitative inspection of incorrect
recognitions using filtered representations indicates that the main
difficulties are posed by those illumination effects which most
significantly deviate from the underlying frequency model (see
Section~\ref{CH2: Models}) such as: cast shadows,
specularities (especially commonly observed for users with glasses)
and photo-sensor saturation.

On the other hand, any failure modes of our fusion framework were
difficult to clearly identify, due to such a low frequency of
erroneous recognition decisions. Even these were in virtually all of
the cases due to overly confident decisions in the filtered
pipeline. Overall, this makes the methodology proposed in this chapter
extremely promising as a robust and efficient way of matching face
appearance image sets, and suggests that future work should
concentrate on developing appropriately robust image filters that
can deal with more complex illumination effects.

\section{Summary and conclusions}
In this chapter we described a novel framework for
increasing the robustness of simple image filters for automatic face recognition
in the presence of varying illumination. The proposed framework is general and is
applicable to matching face sets or sequences, as well as single shots. It is
based on simple image processing filters
that compete with unprocessed greyscale input to yield a single
matching score between individuals. By performing all numerically
consuming computation offline, our method both (i) retains the
matching efficiency of simple image filters, but (ii) with a greatly
increased robustness, as all online processing is performed in
the closed-form. Evaluated on a large, real-world data corpus, the proposed method was shown
to dramatically improve video-based recognition across a wide range of illumination,
pose and face motion pattern changes.

\section*{Related publications}

The following publications resulted from the work presented in this
chapter:

\begin{itemize}
  \item O. Arandjelovi\'c and R. Cipolla. A new look at filtering techniques for illumination
                  invariance in automatic face recognition. In \textit{Proc. IEEE Conference on Automatic
                  Face and Gesture Recognition (FGR)}, pages 449--454, April 2006. \cite{AranCipo2006a}

  \item G. Brostow, M. Johnson, J. Shotton, O. Arandjelovi\'c, V. Kwatra and R. Cipolla. Semantic photo
                  synthesis. In \textit{Proc. Eurographics}, September 2006. \cite{BrosJohnShot+2006}
\end{itemize}

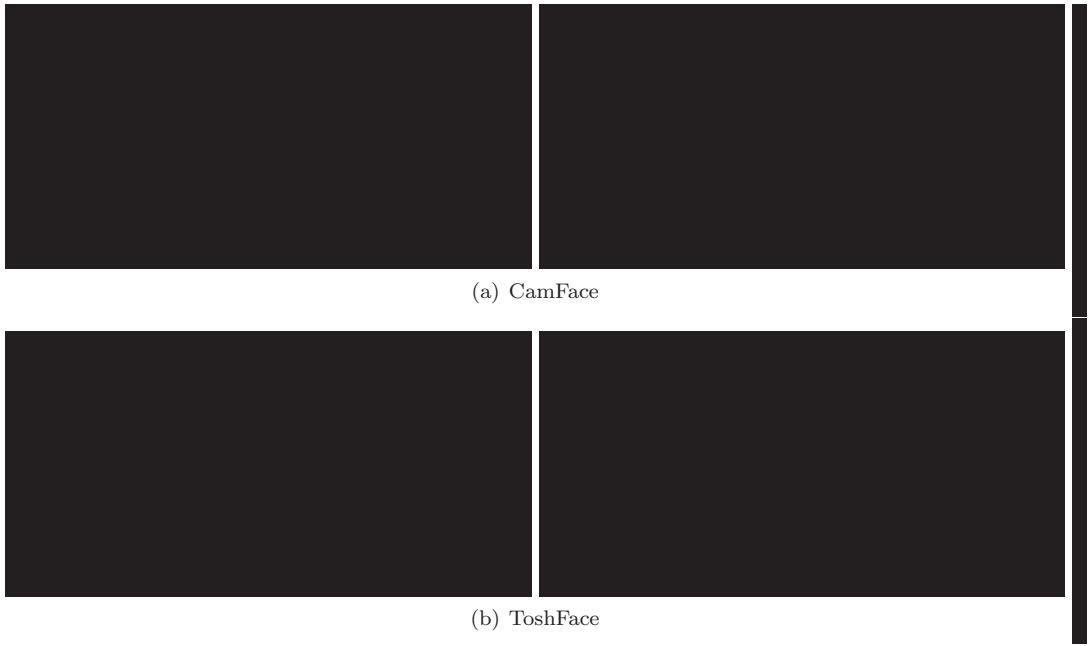
\graphicspath{{./07bompa/}}
\chapter{Boosted Manifold Principal Angles}
\label{Chp: BoMPA}
\begin{center}
  \vspace{-20pt}
  \footnotesize
  \framebox{\includegraphics[width=0.8\textwidth]{title_img.eps}}\\
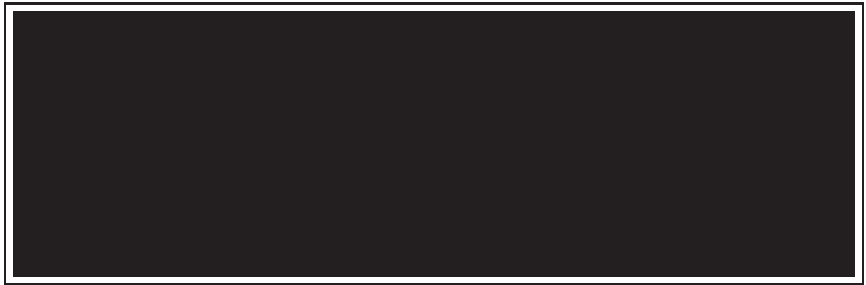
  Joseph M. W. Turner. \textit{Snow Storm: Steamboat off a Harbour's
  Mouth}\\
  1842, Oil on Canvas, 91.4 x 121.9 cm\\
  Tate Gallery, London
\end{center}

\cleardoublepage

The method introduced in the previous chapter suffers from two major
drawbacks. Firstly, the image formation model implicit in the
derivation of the employed quasi-illumination invariant image
filters is too simplistic. Secondly, illumination normalization is
performed on a frame-by-frame basis, not exploiting in fullness all
the available data from a head motion sequence.

In this chapter we focus on the latter problem. We return to
considering face appearance manifolds and identify a manifold
illumination invariant. We show that under the assumption of a
commonly used illumination model by which illumination effects on
the appearance are slowly spatially varying, tangent planes of the
manifold retain their orientation under the set of transformations
caused by face illumination changes. To exploit the invariant, we
propose a novel method based on comparisons between linear subspaces
corresponding to linear patches, piece-wise approximating appearance
manifolds. In particular, there are two main areas of novelty: (i)
we extend the concept of principal angles between linear subspaces
to manifolds with arbitrary nonlinearities; (ii) it is demonstrated
how boosting can be used for application-optimal principal angle
fusion.

\section{Manifold illumination invariants} \label{Sec: Manifold
Illumination Invariants}

Let us start by formalizing our recognition framework. Let
$\mathbf{x}$ be an image of a face and $\mathbf{x} \in
\mathbb{R}^D$, where $D$ is the number of pixels in the image and
$\mathbb{R}^D$ the corresponding image space. Then
$\mathbf{f}(\mathbf{x},\mathbf{\Theta})$ is an image of the same
face after the rotation with parameter $\mathbf{\Theta} \in
\mathbb{R}^3$ (yaw, pitch and roll). Function $\mathbf{f}$ is a
generative function of the corresponding face motion manifold,
obtained by varying $\Theta$\footnote{As a slight digression, note
that strictly speaking, $\mathbf{f}$ should be
\emph{person-specific}. Due to self-occlusion of parts of the face,
$\mathbf{f}$ cannot produce plausible images of rotated faces simply
from a single image $\mathbf{x}$. However, in our work, the range of
head rotations is sufficiently restricted that under the standard
assumption of face symmetry \cite{AranZiss2005}, $\mathbf{f}$ can be
considered generic.}.

\begin{figure}[!t]
  \footnotesize
  \centering
  \includegraphics[width=0.7\textwidth]{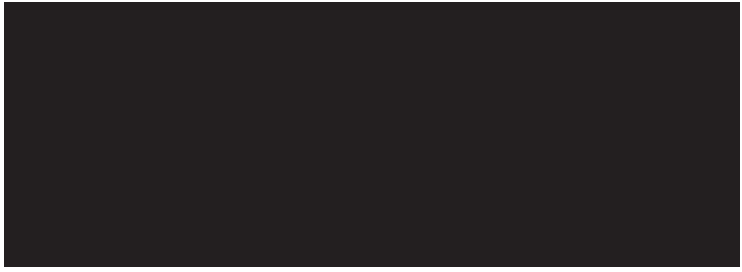}
  \caption[Manifold illumination invariant.]
          { \it Under the assumption that illumination effects on the
                appearance faces are spatially slowly varying,
                appearance manifold tangent planes retrain their
                orientation in the image space with changes in
                lighting conditions. }
            \label{Fig: Tangents}
  \vspace{6pt}\hrule
\end{figure}

\paragraph{Rotation affected appearance changes.}
Now, consider the appearance change of a face due to small rotation
$\Delta \mathbf{\Theta}$: {\begin{align}
  \Delta \mathbf{x} = \mathbf{f}(\mathbf{x}, \Delta \mathbf{\Theta} ) -
  \mathbf{x}.
\end{align}}
For small rotations, geodesic neighbourhood of $\mathbf{x}$ is
linear and using Taylor's theorem we get: {\begin{align}
  \mathbf{f}(\mathbf{x}, \Delta \mathbf{\Theta}) - \mathbf{x} \approx
  \mathbf{f}(\mathbf{x}, \mathbf{0}) + \nabla \mathbf{f}|_{(\mathbf{x},
  \mathbf{0})} \cdot \Delta \mathbf{\Theta}
 - \mathbf{x}.
\end{align}}
where $\nabla \mathbf{f}|_{(\mathbf{x}, \mathbf{\Theta})}$ is the
Jacobian matrix evaluated at $(\mathbf{x}, \mathbf{\Theta})$. Noting
that $\mathbf{f}(\mathbf{x}, \mathbf{0}) = \mathbf{x}$ and writing
$\mathbf{x}$ as a sum of its low and high frequency components
$\mathbf{x} = \mathbf{x}_L + \mathbf{x}_H$: {
\begin{align}
  \Delta \mathbf{x} \approx & \nabla \mathbf{f}|_{\mathbf{x}, \mathbf{0}} \cdot \Delta \mathbf{\Theta} =
   \nabla \mathbf{f}|_{\mathbf{x}_L, \mathbf{0}} \cdot \Delta \mathbf{\Theta} +
  \nabla \mathbf{f}|_{\mathbf{x}_H, \mathbf{0}} \cdot \Delta \mathbf{\Theta}
\end{align}}
But $\mathbf{x}_L$ is by definition slowly spatially varying and
therefore: {\begin{align}
  \bigl\|\nabla \mathbf{f}|_{\mathbf{x}_L, \mathbf{0}} \cdot \Delta \mathbf{\Theta} \bigr\| \ll
  \bigl\|\nabla \mathbf{f}|_{\mathbf{x}_H, \mathbf{0}} \cdot \Delta \mathbf{\Theta} \bigr\|,
\end{align}}
and {\begin{align}
  \Delta \mathbf{x} \approx \nabla \mathbf{f}|_{\mathbf{x}_H, \mathbf{0}} \cdot \Delta \mathbf{\Theta}.
\end{align}}
It can be seen that $\Delta \mathbf{x}$ is a function of the
person-specific $\mathbf{x}_H$ but not the illumination affected
$\mathbf{x}_L$. Hence, the directions (in $\mathbb{R}^D$) of face
appearance changes due to small head rotations form a \emph{local
manifold invariant} with respect to illumination variation, see
Figure~\ref{Fig: Tangents}.

The manifold illumination invariant we identified explicitly
motivates the use of principal angles between tangent planes as a
similarity measure between manifolds. We now address two questions
that remain:
\begin{itemize}
  \item given principal angles between two tangent planes, what
        contribution should each principal angle have, and
  \item given similarities between different tangent planes of two
        manifolds, how to obtain a similarity measure between the
        manifolds themselves.
\end{itemize}
We now turn to the first of these problems.

\section{Boosted principal angles}
In general, each principal angle $\theta_i$ carries some information
for discrimination between the corresponding two subspaces. We use
this to build simple weak classifiers $\mathcal{M}(\theta_i) = \sign
\left[ \cos(\theta_i) - C\right]$. In the proposed method, these are
combined using the now acclaimed AdaBoost
algorithm~\cite{FreuScha1995}. In summary, AdaBoost learns a
weighting $\{w_i\}$ of decisions cast by weak learners to form a
classifier $\mathcal{M}(\Theta)$:
\begin{equation}\label{Eqn: AdaBoost}
    \mathcal{M}(\Theta) = \sign
        \left[ \sum_{i=1}^N w_i \mathcal{M}(\theta_i) - \frac{1}{2}\sum_{i=1}^N w_i \right]
\end{equation}
In an iterative update scheme classifier performance is optimized on
training data which consists of in-class and out-of-class features
(i.e.\ principal angles). Let the training database consist of sets
$S_1,\ldots,S_K \equiv \{S_i\}$, corresponding to $K$ classes. In
the framework described, the $K(K-1)/2$ out-of-class principal
angles are computed between pairs of linear subspaces corresponding
to training data sets $\{S_i\}$, estimated using Principal Component
Analysis (PCA). On the other hand, the $K$ in-class principal angles
are computed between a pair of randomly drawn subsets for each
$S_i$.

We use the learnt weights $\{w_i\}$ for computing the following
similarity measure between two linear subspaces:
\begin{equation}\label{Eqn: Linear Subspace Similarity Function}
    f(\Theta) = \frac{1}{N} \frac {\sum_{i=1}^N w_i \cos(\theta_i)} {\sum_{i=1}^N
    w_i}
\end{equation}

\begin{figure}
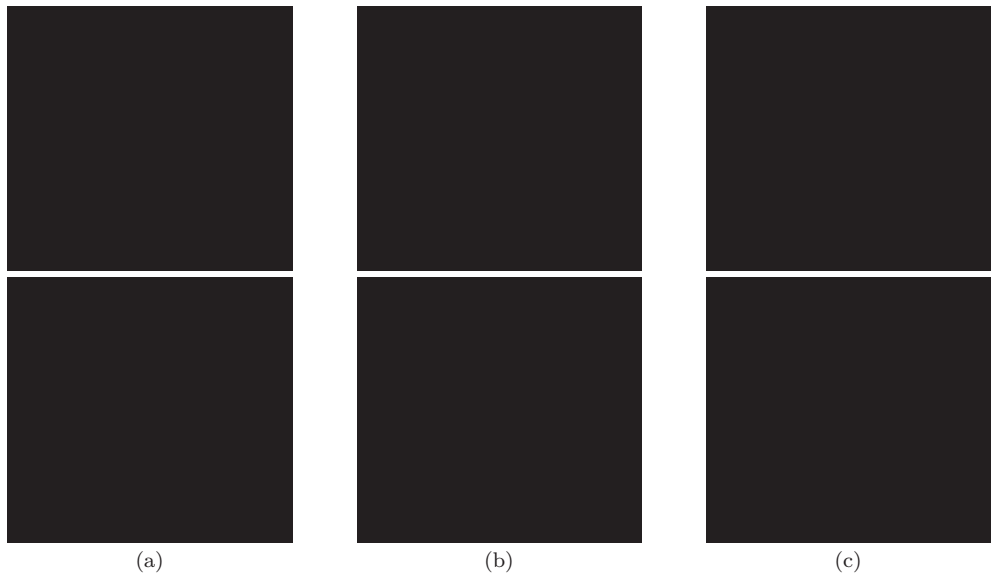

  \footnotesize
  \centering
  \begin{tabular*}{0.95\textwidth}{@{\extracolsep{\fill}}ccc}
    \includegraphics[width=0.27\textwidth]{BMVC2005_pvs_msm_error1_1.eps} &
    \includegraphics[width=0.27\textwidth]{BMVC2005_pvs_bpa_ok1_1.eps} &
    \includegraphics[width=0.27\textwidth]{BMVC2005_pvs_bompa_ok1_1.eps} \\
    \includegraphics[width=0.27\textwidth]{BMVC2005_pvs_msm_error2_1.eps} &
    \includegraphics[width=0.27\textwidth]{BMVC2005_pvs_bpa_ok2_1.eps} &
    \includegraphics[width=0.27\textwidth]{BMVC2005_pvs_bompa_ok2_1.eps} \\
    (a) & (b) & (c)\\
  \end{tabular*}

  \caption[MSM, Boosted Principal Angles and BoMPA.]
          { \it (a) The first 3 principal
                vectors between two linear subspaces which MSM
                incorrectly classifies as corresponding to the
                same person. In spite of
                different identities, the most similar modes of variation are very much
                alike and can be seen to correspond to especially
                difficult illuminations. (b) Boosted Principal
                Angles (BPA), on the other hand, chooses different
                principal vectors as the most discriminating --
                these modes of variation are now less similar
                between the two sets. (c) Modelling of nonlinear
                manifolds corresponding to the two image sets
                produces a further improvement. Shown are the
                most similar modes of variation amongst all
                pairs of linear manifold patches. Local
                information is well captured and even these
                principal vectors are now very dissimilar.
            }
            \label{Fig: Principal Vectors Prob}
  \vspace{6pt}\hrule
\end{figure}

\begin{figure}
  \footnotesize
  \centering
  \subfigure[Optimal angle weighting]{\includegraphics[width=0.7\textwidth]{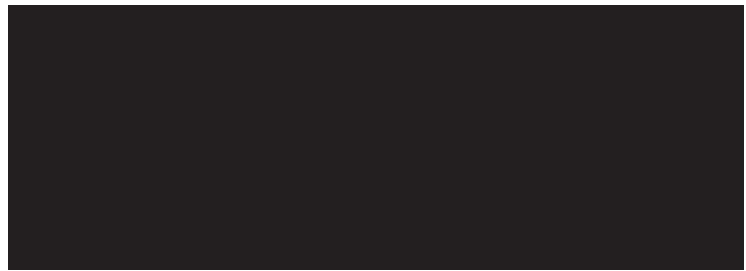}}
  \subfigure[Performance improvement]{\includegraphics[width=0.7\textwidth]{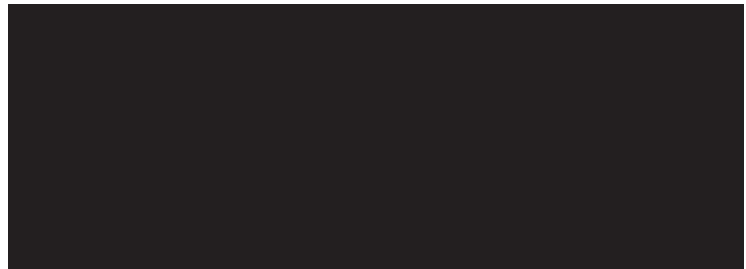}}

  \caption[Weak classier weights and performance improvement with boosting.]
    {\it (a) A typical set of weights corresponding to weak principal
            angle-based classifiers, obtained using AdaBoost. This
            figure confirms our criticism of MSM-based methods for (i)
            their simplistic fusion of information from different
            principal angles and (ii) the use of only the first few angles.
            (b) The average performance of a simple MSM classifier and
            our boosted variant. }
            \label{Fig: Weights}
\end{figure}

A typical set of weights $\{w_i\}$ we obtained is shown graphically
in Figure~\ref{Fig: Weights}~(a). The plot shows an interesting
result: the weight corresponding to the first principal angle is not
the greatest. Rather it is the second principal angle that is most
discriminating, followed by the third one. This shows that the most
similar mode of variation across two subspaces can indeed be due an
extrinsic factor. Figure~\ref{Fig: Principal Vectors Prob}~(b) shows
the 3 most discriminating principal vector pairs selected by our
algorithm for data incorrectly classified by MSM -- the most
weighted principal vectors are now much less similar. The gain
achieved with boosting is also apparent from Figure~\ref{Fig:
Weights}~(b). A significant improvement can be seen both for a small
and a large number of principal angles. In the former case this is
because our algorithm chooses not the first but the most
discriminating set of angles. The latter case is practically more
important -- as more principal angles are added to MSM, its
performance first improves, but after a certain point it starts
\emph{worsening}. This highly undesirable behaviour is caused by
effectively equal weighting of base classifiers in MSM. In contrast,
the performance of our algorithm never decreases as more information
is added. As a consequence, no special provision for choosing the
optimal number of principal angles is needed.

At this point it is worthwhile mentioning the work of Maeda
\textit{et al.}~\cite{MaedYamaFuku2004} in which the third principal
angle was found to be useful for discriminating between sets of
images of a face and its photograph. Much like in MSM and CMSM, the
use of a single principal angle was motivated only empirically --
the framework described in this chapter can be used for a more
principled feature selection in this setting as well.

\section{Nonlinear subspaces}\label{Subsec: Nonlinear Subspaces}
Our aim is to extend the described framework of boosted principal
angles to being able to effectively capture nonlinear data
behaviour. We propose a method that combines \emph{global} manifold
variations with more subtle, \emph{local} ones.

Without the loss of generality, let $S_1$ and $S_2$ be two sets of
face appearance images and $\Theta$ the set of principal angles
between two linear subspaces. We derive a measure of similarity
$\rho$ between $S_1$ and $S_2$ by comparing the corresponding linear
subspaces $U_{1,2}$ and locally linear patches $L_{1,2}^{(i)}$
corresponding to piece-wise linear approximations of manifolds of
$S_1$ and $S_2$:
\begin{equation}\label{Eqn: Similarity}
    \rho\left( S_1, S_2 \right) = \underbrace{(1 - \alpha) f_G \left[\Theta \left( U_1, U_2
    \right) \right]}_{\text{Global manifold similarity contribution}} + \underbrace{\alpha \max_{i, j} f_L \left[\Theta (L_1^{(i)},
    L_2^{(j)}) \right]}_{\text{Local manifold similarity contribution}}
\end{equation}
where $f_G$ and $f_L$ have the same functional form as $f$
in~\eqref{Eqn: Linear Subspace Similarity Function}, but separately
learnt base classifier weights $\{w_i\}$. Put in words, the
proximity between two manifolds is computed as a weighted average of
the similarity between global modes of data variation and the best
matching local behaviour. The two terms complement each other: the
former provides (i) robustness to noise, whereas the latter ensures
(ii) graceful performance degradation with missing data (e.g.\
unseen poses) and (iii) illumination invariance, see
Figure~\ref{Fig: Principal Vectors Prob}~(c).

\subsubsection{Finding stable locally linear patches} In the proposed
framework, stable locally linear manifold patches are found using
Mixtures of Probabilistic PCA (PPCA)~\cite{TippBish1999a}. The main
difficulty in fitting of a PPCA mixture is the requirement for the
local principal subspace dimensionality to be set \textit{a priori}.
We solve this problem by performing the fitting in two stages. In
the first stage, a Gaussian Mixture Model (GMM) constrained to
diagonal covariance matrices is fitted first. This model is crude as
it is insufficiently expressive to model local variable
correlations, yet too complex (in terms of free parameters) as it
does not encapsulate the notion of intrinsic manifold dimensionality
and additive noise. However, what it is useful for is the
\emph{estimation} of the intrinsic manifold dimensionality $d$, from
the eigenspectra of its covariance matrices, see Figure~\ref{Fig:
GMM Fitting}~(a). Once $d$ is estimated (typically $d \ll D$), the
fitting is repeated using a Mixture of PPCA.

\begin{figure}
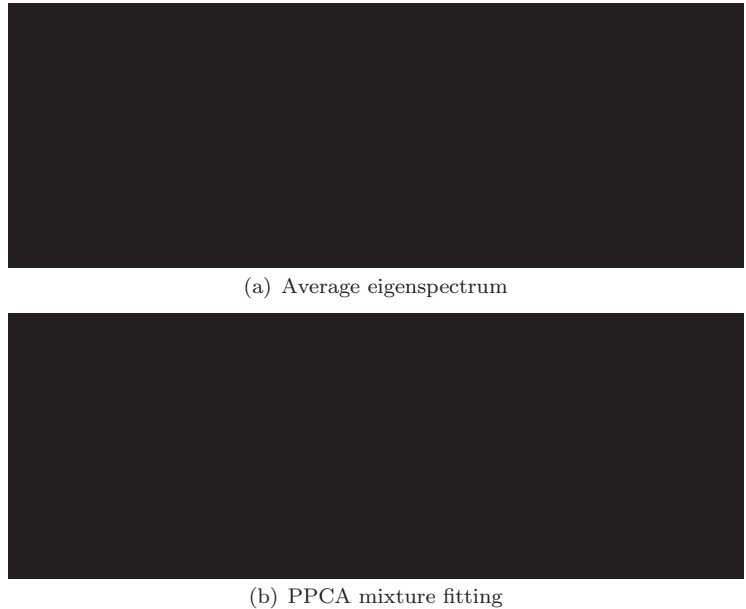

  \centering
  \footnotesize
  \subfigure[Average eigenspectrum]{\includegraphics[width=0.7\textwidth]{BMVC2005_eigenspectrum_L.eps}}
  \subfigure[PPCA mixture fitting]{\includegraphics[width=0.7\textwidth]{BMVC2005_dl_L.eps}}

  \caption[Finding stable tangent planes.]
           {\it
            (a) Average eigenspectrum of diagonal covariance matrices
            in a typical intermediate GMM fit. The approximate
            intrinsic manifold dimensionality can be seen to be around 10.
            (b) Description length as a function of the number of
            Gaussian components in the intermediate and final,
            PPCA-based GMM fitting on a typical data set. The
            latter results in fewer components and a significantly
            lower MDL. }
            \label{Fig: GMM Fitting}
\end{figure}

Both the intermediate diagonal and the final PPCA mixtures are
estimated using the Expectation Maximization (EM)
algorithm~\cite{DudaHartStor2001} which is initialized by K-means
clustering. Automatic model order selection is performed using the
well-known Minimum Description Length (MDL)
criterion~\cite{DudaHartStor2001}, see Figure~\ref{Fig: GMM
Fitting}~(b). Typically, the optimal (in the MDL sense) number of
components for face data sets used in Section~\ref{Sec: Evaluation}
was 3.

\section{Empirical evaluation}\label{Sec: Evaluation} Methods in
this chapter were evaluated on the \textit{CamFace} data set, see
Appendix~\ref{App: CamFace}. We compared the performance of our
algorithm, without and with boosted feature selection (respectively
MPA and BoMPA), to that of:
\begin{itemize}
    \item KL divergence algorithm (KLD) of Shakhnarovich \textit{et al.} \cite{ShakFishDarr2002}\footnote{The algorithm was reimplemented through consultation with the authors.},
    \item Mutual Subspace Method (MSM) of Yamaguchi \textit{et al.} \cite{YamaFukuMaed1998}\footnotemark[\value{footnote}],
    \item Kernel Principal Angles (KPA) of Wolf and Shashua \cite{WolfShas2003}\footnote{We used the original authors' implementation.}, and
    \item Nearest Neighbour (NN) in the Hausdorff distance sense
    in (i) LDA \cite{BelhHespKrie1997} and (ii) PCA \cite{TurkPent1991} subspaces,
    estimated from data.
\end{itemize}
In KLD 90\% of data energy was explained by the principal subspace
used. In MSM, the dimensionality of PCA subspaces was set to
9~\cite{FukuYama2003}. A sixth degree monomial expansion kernel was
used for KPA~\cite{WolfShas2003}. In BoMPA, we set the value of
parameter $\alpha$ in \eqref{Eqn: Similarity} to 0.5. All algorithms
were preceded with PCA estimated from the entire training dataset
which, depending on the illumination setting used for training,
resulted in dimensionality reduction to around 150 (while retaining
95\% of data energy).

In each experiment we used performed training using sequences in a
single illumination setup and tested recognition with sequences in
each different illumination setup in turn.

\subsection{BoMPA implementation} From a practical stand, there are
two key points in the implementation of the proposed method: (i) the
computation of principal angles between linear subspaces and (ii)
time efficiency. These are now briefly summarized for the
implementation used in the evaluation reported in this chapter. We
compute the cosines of principal angles using the method of
Bj{\"o}rck and Golub~\cite{BjorGolu1973}, as singular values of the
matrix $B_1^T B_2$ where $B_{1,2}$ are orthonormal basis of two
linear subspaces. This method is numerically more stable than the
eigenvalue decomposition-based method used
in~\cite{YamaFukuMaed1998} and with roughly the same computational
demands, see~\cite{BjorGolu1973} for a thorough discussion on
numerical issues pertaining to the computation of principal angles.
A computationally far more demanding stage of the proposed method is
the PPCA mixture estimation. In our implementation, a significant
improvement was achieved by dimensionality reduction using the
incremental PCA algorithm of Hall \emph{et al.}
\cite{HallMarsMart2000}. Finally, we note that the proposed model of
pattern variation within a set inherently places low demands on
storage space.

\subsection{Results}
The performance of evaluated recognition algorithms is summarized in
Table~\ref{Tab:BOMPAResults}. Firstly, note the relatively poor
performance of the two nearest neighbour-type methods -- the
Hausdorff NN in PCA and LDA subspaces. These can be considered as
proxies for gauging the difficulty of the recognition task, seeing
that both can be expected to perform relatively well if the imaging
conditions do not greatly differ between training and test data
sets. Specifically, LDA-based methods have long been established in
the single-shot face recognition literature, e.g.\ see
\cite{BelhHespKrie1997,ZhaoChelKris1998,SadeKitt2004,WangTang2004,KimKitt2005}.
The KL-divergence based method achieved by far the worst recognition
rate. Seeing that the illumination conditions varied across data and
that the face motion was largely unconstrained, the distribution of
intra-class face patterns was significant making this result
unsurprising. This is consistent with results reported in the
literature~\cite{AranShakFish+2005}.

\begin{table}
    \centering
    \Large
    \caption[Recognition results.]{\it The mean recognition rate and its standard
                 deviation across different training/test
                 illuminations (in \%). The last row
                 shows the average time in seconds for 100
                 set comparisons.\vspace{10pt} }

\begin{tabular*}{1.00\textwidth}{@{\extracolsep{\fill}}ll|ccccccc}
  \Hline
  \bf \normalsize ~Method  &         & \normalsize KLD  & \normalsize NN-LDA & \normalsize NN-PCA & \normalsize MSM  & \normalsize KPA  & \normalsize MPA  & \normalsize BoMPA \\
  \hline
  \multirow{3}{*}{\bf \normalsize ~Recognition } & \normalsize mean  & \small 19.8 & \small 40.7   & \small 44.6   & \small 84.9 & \small 89.1 & \small 89.7 & \small 92.6 \\
                                                 & \normalsize std   & \small 9.7  & \small 6.6    & \small 7.9    & \small 6.8  & \small 10.1 & \small 5.5  & \small 4.3 \\
                                                 & \normalsize time  & \small 7.8  & \small 11.8   & \small 11.8   & \small 0.8  & \small 45   & \small 7.0  & \small 7.0  \\
  \Hline
\end{tabular*}
\label{Tab:BOMPAResults}
\end{table}

The performance of the four principal angle-based methods confirms
the premises of our work. Basic MSM performed well, but worst of the
four. The inclusion of nonlinear manifold modelling, either by using
the ``kernel trick'' or a mixture of linear subspaces, achieved an
increase in the recognition rate of about 5\%. While the difference
in the average performance of MPA and the KPA methods is probably
statistically insignificant, it is worth noting the greater
robustness to specific imaging conditions of our MPA, as witnessed
by a much lower standard deviation of the recognition rate. Further
performance increase of 3\% is seen with the use of boosted angles,
the proposed BoMPA algorithm correctly recognizing 92.6\% of the
individuals with the lowest standard deviation of all methods
compared. An illustration of the improvement provided by each novel
step in the proposed algorithm is shown in Figure~\ref{Fig: Rank
Recognition}. Finally, its computational superiority to the best
performing method in the literature, Wolf and Shashua's KPA, is
clear from a 7-fold difference in the average recognition time.

\begin{figure}
   \centering
   \subfigure[Per-case rank-$N$ performance]{\includegraphics[width=0.9\textwidth]{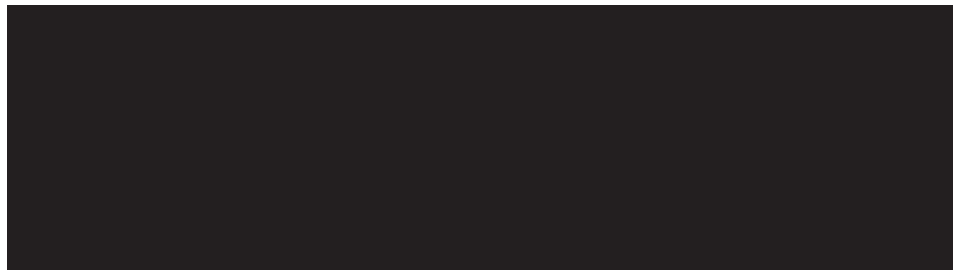}}
   \subfigure[Average rank-$N$ performance]{\includegraphics[width=0.6\textwidth]{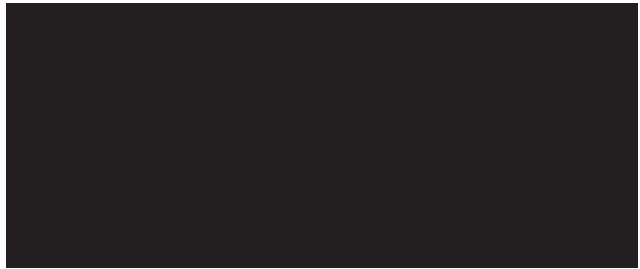}}

   \caption[ROC curves.]{\it
            Shown is the improvement in rank-$N$ recognition accuracy
            of the basic MSM, MPA and BoMPA algorithms for (a) each
            training/test combination and (b) on average. A consistent
            and significant improvement is seen with nonlinear
            manifold modelling, which is further increased using
            boosted principal angles. }
            \label{Fig: Rank Recognition}

\end{figure}

\section{Summary and conclusions}
In this chapter we showed how appearance manifolds can be used to
integrate information across a face motion video sequence to achieve
illumination invariance. This was done by combining (i) an
illumination model, and (ii) observed appearance changes, to derive
a manifold illumination invariant. A novel method, the
\textit{Boosted Manifold Principal Angles} (BoMPA), was proposed to
exploit the invariant. We used a boosting framework by which focus
is put on the most discriminative regions of invariant tangent
planes and introduced a method for fusing their similarities to
obtain the overall manifold similarity. The method was shown to be
successful in recognition across large changes in illumination.

\section*{Related publications}

The following publications resulted from the work presented in this
chapter:

\begin{itemize}
  \item T-K. Kim, O. Arandjelovi{\'c} and R. Cipolla. Learning over Sets using Boosted Manifold
                  Principal Angles (BoMPA). In \textit{Proc. IAPR British Machine Vision Conference (BMVC)},
                  \textbf{2}:pages 779--788, September 2005.
                  \cite{KimAranCipo2005}

  \item O. Arandjelovi\'c and R. Cipolla. Face set classification using maximally probable mutual
                  modes.  In \textit{Proc. IEEE International Conference on Pattern Recognition (ICPR)},
                  pages 511-514, August 2006. \cite{AranCipo2006e}

  \item T-K. Kim, O. Arandjelovi\'c and R. Cipolla. Boosted manifold principal angles for image
                  set-based recognition. \textit{Pattern Recognition}, \textbf{40}(9):2475--2484,
          September 2007. \cite{KimAranCipo2007}
\end{itemize}

\graphicspath{{./08acc/}}
\chapter{Pose-Wise Linear Illumination Manifold Model}
\label{Chp: Auth}
\begin{center}
  \vspace{-10pt}
  \footnotesize
  \framebox{\includegraphics[width=0.75\textwidth]{title_img.eps}}\\
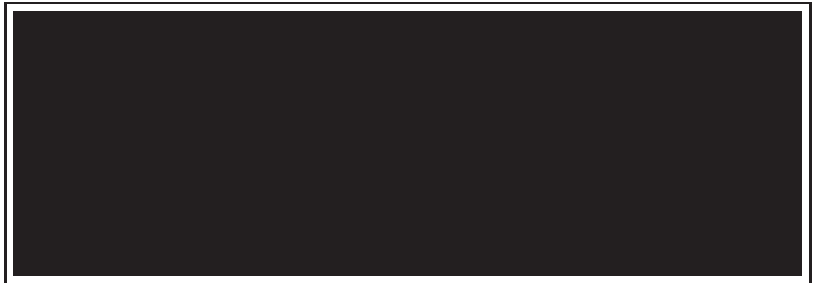
  Pablo Picasso. \textit{Bull, 11th State}\\
  1946, Lithograph, 29 x 37.5 cm\\
  Mus\'{e}e Picasso, Paris
\end{center}

\cleardoublepage

In the proceeding chapters, illumination invariance was achieved
largely by employing \textit{a priori} domain knowledge, such as the
smoothness of faces and their largely Lambertian reflectance
properties. Subtle, yet important effects of the underlying complex
photometric process were not captured, cast shadows and
specularities both causing incorrect recognition decisions. In this
chapter we take a further step towards the goal of combining models
stemming from our understanding of image formation and learning from
available data.

In particular there are two major areas of novelty: (i) illumination
generalization is achieved using a two-stage method, combining
coarse region-based gamma intensity correction with normalization
based on a pose-specific illumination subspace, learnt offline; (ii)
pose robustness is achieved by decomposing each appearance manifold
into semantic Gaussian pose clusters, comparing the corresponding
clusters and fusing the results using an RBF network. On the
\textit{ToshFace} data set, the proposed algorithm consistently
demonstrated a very high recognition rate (95\% on average),
significantly outperforming state-of-the-art methods from the
literature.


\section{Overview}
A video sequence of a moving face carries information about its 3D
shape and texture. In terms of recognition, this information can be
used either explicitly, by recovering parameters of a generative
model of the face (e.g.\ as in \cite{BlanVett2003}), or implicitly
by modelling face appearance and trying to achieve invariance to
extrinsic causes of its variation (e.g.\ as in \cite{AranZiss2005}).
In this chapter we employ the latter approach, as more suited for
low-resolution input data (see Section~\ref{Sec: Evaluation} for
typical data quality) \cite{EverZiss2004}.

In the proposed method, manifolds
\cite{AranShakFish+2005,BichPent1994} of face appearance are
modelled using at most three Gaussian pose clusters describing small
face motion around different head poses. Given two such manifolds,
first (i) the pose clusters are determined, then (ii) those
corresponding in pose are compared and finally, (iii) the results of
pairwise cluster comparisons are combined to give a unified measure
of similarity of the manifolds themselves. Each of the steps, aimed
at achieving robustness to a specific set of nuisance parameters, is
described in detail next.

\section{Face registration}\label{SubSec: Registration} 
\begin{figure}[!t]
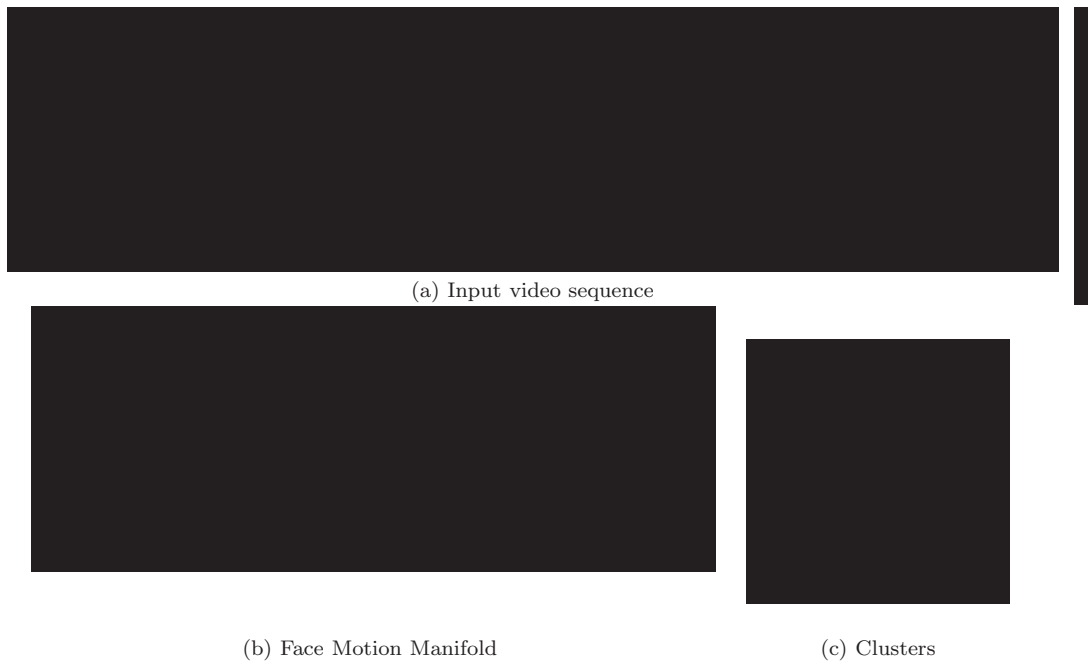

  \centering
  \footnotesize
  \begin{tabular}{c}
      \includegraphics[width=1.00\textwidth]{PR2006_sequence1.eps} \\
      (a) Input video sequence
  \end{tabular}\\

  \begin{tabular}{VV}
  \vspace{25pt}
  \includegraphics[width=0.65\textwidth]{PR2006_manifold.eps}
  &
  \includegraphics[width=0.25\textwidth]{PR2006_clusters.eps}\\
  (b) Face Motion Manifold & (c) Clusters\\
  \end{tabular}
  \caption[Affine-registered faces -- manifold and poses.]{\it
           A typical input video sequence of random head motion performed by the user
           (a) and the corresponding face appearance manifold (b). Shown is the projection
           of affine-registered data (see Section~\ref{SubSec: Registration}) to the first
           three linear principal components. Note that while highly nonlinear, the
           manifold is continuous and smooth. Different poses are marked in different
           styles (red stars, blue dots and green squares). Examples of faces from the
           three clusters can be seen in (b) (also affine-registered and cropped). }
            \label{Fig: Face Manifolds}
  \vspace{6pt}\hrule
\end{figure}

Using the standard
appearance representation of a face as a raster-ordered pixel array,
it can be observed that the corresponding variations due to head
motion, i.e.\ pose changes, are highly nonlinear, see
Figure~\ref{Fig: Face Manifolds}~(a,b). A part of the difficulty of
recognition from appearance manifolds is then contained in the
problem of what is an appropriate way of representing them, in a way
suitable for the analysis of the effects of varying illumination or
pose.

In the proposed method, face appearance manifolds are represented in
piece-wise linear manner, by a set of semantic Gaussian \emph{pose
clusters}, see Figure~\ref{Fig: Face Manifolds}~(b,c). Seeing that each
cluster describes a locally linear mode of variation, this approach to
modelling manifolds becomes increasingly difficult as their intrinsic
dimensionality is increased. Therefore, it is advantageous to normalize
the raw, input frames as much as possible so as to minimize this
dimensionality. In this first step of our method, this is done by
\emph{registering} faces i.e.\ by warping them to have a set of salient
facial features aligned. For related approaches see
\cite{AranZiss2005,BergBergEdwa+2004}.

We compute warps that align each face with a canonical frame using
four point correspondences: the locations of pupils (2) and nostrils
(2). These are detected using a two-stage feature detector of Fukui
and Yamaguchi \cite{FukuYama1998}\footnote{We thank the authors for
kindly providing us with the original code of their algorithm.}.
Briefly, in the first stage, shape matching is used to rapidly
remove a large number of locations in the input image that do not
contain features of interest. Out of the remaining, `promising'
features, true locations are chosen using the appearance-based,
distance from feature space criterion. We found that the
described method reliably detected pupils and nostrils across a wide
variation in illumination conditions and pose.

\begin{figure*}
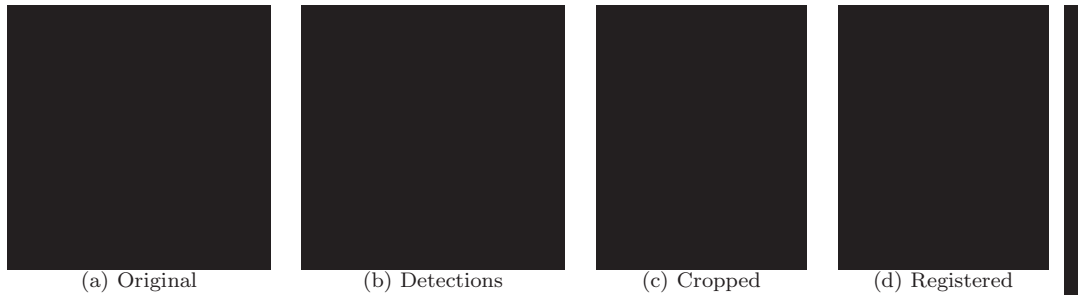

  \centering
  \footnotesize
  \begin{tabular}{VVVV}
    \includegraphics[width=0.25\textwidth]{PR2006_face_original.ps}
    &
    \includegraphics[width=0.25\textwidth]{PR2006_face_features.ps}
    &
    \includegraphics[width=0.2\textwidth]{PR2006_face_localized.eps}
    &
    \includegraphics[width=0.2\textwidth]{PR2006_face_registered.eps}\\
    (a) Original & (b) Detections & (c) Cropped & (d) Registered \\
  \end{tabular}
  \caption[Face registration and cropping.]
          {\it          (a) Original input frame (resolution $320 \times 240$ pixels),
                        (b) superimposed detections of the two pupils and nostrils
                        (as white circles),
                        (c) cropped face regions with background clutter removed, and
                        (d) the final affine registered and cropped image of the face
                        (resolution $30 \times 30$ pixels).}
  \label{Fig: Registration Pipeline}
  \vspace{6pt}\hrule
\end{figure*}

From the four point correspondences between the locations of the
facial features and their canonical locations (we chose canonical
locations to be the mean values of true feature locations) we
compute optimal affine warps on a per-frame basis. Since four
correspondences over-determine the affine transformation parameters (8
equations with 6 unknown parameters), we estimate them in the
minimum $L_2$ error sense. Finally, the resulting images are
cropped, so as to remove background clutter, and resized to the
uniform scale of $30 \times 30$ pixels. An example of a face registered
and cropped in the described manner is shown in Figure~\ref{Fig: Registration Pipeline}
(also see Figure~\ref{Fig: Face Manifolds}~(c)).

\section{Pose-invariant recognition}\label{SubSec: Pose Invariance}
Achieving invariance to varying pose is one of the most challenging
aspects of face recognition and yet a prerequisite condition for
most practical applications. This problem is complicated further by
variations in illumination conditions, which inevitably occur due to
movement of the user relative to the light sources.

We propose to handle changing pose in two, complementary stages: (i)
in the first stage an appearance manifold is \emph{decomposed} to
Gaussian pose clusters, effectively reducing the problem to
recognition under a small variation in pose parameters; (ii) in the
second stage, fixed-pose recognition results are \emph{fused} using
a neural network, trained offline. The former stage is addressed
next, while the latter is the topic of Section~\ref{SubSubSec:
Intermanifold Distance}.

\subsection{Defining pose clusters}\label{SubSubSec: Defining Pose
Clusters} Inspection of manifolds of registered faces in random
motion around the fronto-parallel face shows that they are dominated
by the first nonlinear principal component. This principal component
corresponds to lateral head rotation, i.e.\ changes in the face yaw,
see Figure~\ref{Fig: Face Manifolds}~(a,b). The reason for this lies
in the greater smoothness of the face surface in the vertical than
in the horizontal direction -- pitch changes (``nodding'') are
largely compensated for by using the affine registration described
in Section~\ref{SubSec: Registration}. This is not the case with
significant changes, when self-occlusion occurs.

Therefore, the centres of Gaussian clusters used to linearize an
appearance manifold correspond to different yaw angle values. In
this work we describe the manifolds using three Gaussian clusters,
corresponding to the frontal face orientation, face left and face
right, see Figure~\ref{Fig: Face Manifolds}~(a,b).

\subsection{Finding pose clusters}\label{SubSubSec: Finding Pose
Clusters} As the extent of lateral rotation, as well as the number
of frames corresponding to each cluster, can vary between video
sequences, a generic clustering algorithm, such as the k-means
algorithm, is unsuitable for finding the three Gaussians.

With the prior knowledge of the semantics of clusters, we decide on
a single face image membership on a frame-by-frame basis. We show
that this can be done in a very simple and rapid manner from already
detected locations of the four characteristic facial features: the
pupils and nostrils, see Section~\ref{SubSec: Registration}.

\begin{figure*}
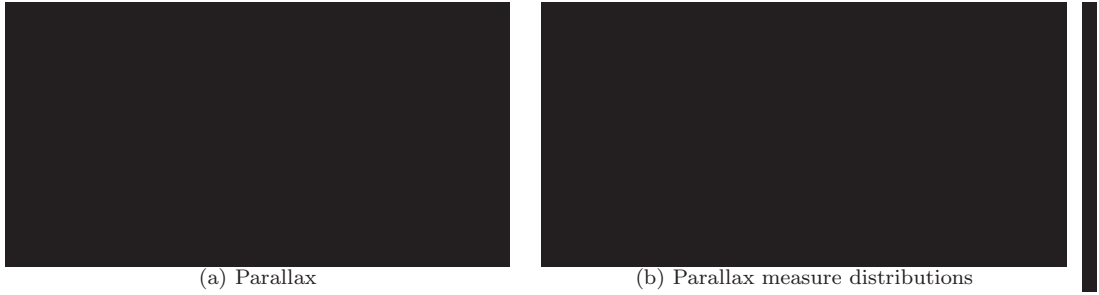

  \centering
  \footnotesize
  \begin{tabular}{VV}
    \includegraphics[width=0.48\textwidth]{PR2006_parallax1.eps} &
    \includegraphics[width=0.5\textwidth]{PR2006_parallax_dists.eps}\\
    (a) Parallax & (b) Parallax measure distributions\\
  \end{tabular}

  \caption[Motion parallax.]
      {\it (a) A schematic illustration of the motion parallax
               used for coarse pose clustering of input faces (the diagram is based on
               a figure taken from \cite{Gray1918}). (b) The distributions of the scale-normalized
               parallax measure $\hat{\eta}$ defined in \eqref{Eqn: Eta} for the
               three pose clusters on the offline training data set. Good separation is demonstrated. }
  \label{Fig: Parallax}
  \vspace{6pt}\hrule
\end{figure*}

The proposed method relies on motion parallax based on
inherent properties of the shape of faces. Consider the anatomy of a
human head shown in profile view in Figure~\ref{Fig: Parallax}~(a). It
can be seen that the pupils are further away than the nostrils from
the vertical axis defined by the neck. Hence, assuming no head roll takes place,
as the head rotates laterally, nostrils travel a longer projected path in the image.
Using this observation, we define the quantity $\eta$ as follows:
\begin{align}
    \eta = x_e^c - x_n^c
\end{align}
where $x_e^c$ and $x_n^c$ are the mid-points between, respectively,
the eyes and the nostrils:
\begin{align}
    x_e^c  = \frac {x_{e1} + x_{e2}} {2} &&
    x_n^c  = \frac {x_{n1} + x_{n2}} {2}.
\end{align}
It can now be understood that $\eta$ approximates the discrepancy
between distances travelled by the mid-points between the eyes and
nostrils, measured from the frontal face orientation. Finally, we
normalize $\eta$ by dividing it with the distance between the eyes,
to obtain $\hat{\eta}$, the scale-invariant parallax measure:
\begin{align}\label{Eqn: Eta}
    \hat{\eta} = \frac {\eta} { \|x_{e1} - x_{e2}\|} = \frac {x_e^c - x_n^c} { \|x_{e1} - x_{e2}\|}
\end{align}

\paragraph*{Learning the parallax model.} In our method, discrete poses used for
linearizing appearance manifolds are automatically learnt from a
small training corpus of video sequences of faces in random motion.
To learn the model, we took 20 sequences of 100 frames each,
acquired at 10fps, and computed the value of $\hat{\eta}$ for each
registered face. We then applied the k-means clustering algorithm
\cite{DudaHartStor2001} on the obtained set of parallax measure
values and fitted a 1D Gaussian to each, see Figure~\ref{Fig:
Parallax}~(b).

\begin{figure}
  \centering
  \includegraphics[width=0.7\textwidth]{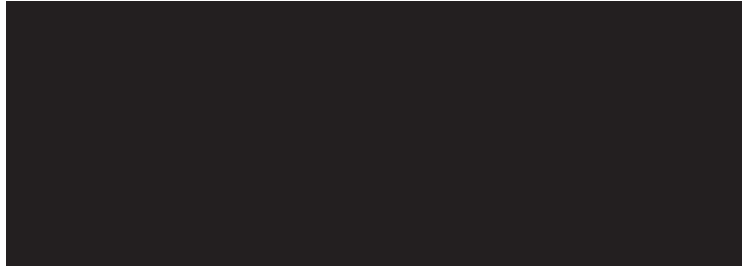}
  \caption[Distributions of faces across pose clusters.]
          {\it Histograms of the number of correctly registered faces using four
               point correspondences between detected facial features (pupils and
               nostrils) for each of the three discrete poses and in total for each
               sequence.
               }
  \label{Fig: Registration Hists}
  \vspace{6pt}\hrule
\end{figure}

To apply the learnt model, a frame in our method is classified to
the maximal likelihood pose. In other words, when a novel face is to
be classified to one of the three pose clusters (i.e.\ head poses),
we evaluate pose likelihood given each of the learnt distributions
and classify it to the one giving the highest probability of the
observation. Figure~\ref{Fig: Registration Hists} shows the
proportions of faces belonging to each pose cluster.

\section{Illumination-invariant recognition}\label{SubSec:
Illumination Invariance} Illumination variation of face patterns is
extremely complex due to varying surface reflectance properties,
face shape, and type and distance of lighting sources. Hence, in
such a general setup, this is a difficult problem to approach in a
purely discriminative fashion.

Our method for compensating for illumination changes is based on the
observation that on the coarse level most of the variation can be
described by the \emph{dominant light} direction e.g.\ `strong light
from the left'. Such variations are addressed much easier.
We will also demonstrate that it is the case that \emph{once normalized}
at this, coarse level, the learning of residual illumination changes is
significantly simplified as well. This motivates the two-stage, per-pose
illumination normalization employed in the proposed method:
\begin{enumerate}
    \item \textbf{Coarse level:} Region-based gamma intensity correction (GIC), followed by
    \item \textbf{Fine level:} Illumination subspace normalization.
\end{enumerate}
The algorithm is summarized in Figure~\ref{Fig: Illumination
Overview} while its details are explained in the sections that
follow.

\begin{figure}[!t]
    \centering
    \begin{tabular}{l}
          \begin{tabular}{ll}
              \textbf{Input}:  & pose clusters $\mathcal{C}_1 = \{ \mathbf{x}^{(1)}_i \}$, $\mathcal{C}_2 = \{ \mathbf{x}^{(2)}_i \}$,\\
                               & face regions mask $\mathbf{r}$,\\
                               & mean face (for pose) $\mathbf{m}$,\\
                               & pose illumination subspace basis matrix $\mathbf{B}_I$.\\
              \textbf{Output}: & pose cluster $\hat{\mathcal{C}}_1$ normalized to $\mathcal{C}_2$.\\
          \end{tabular}\vspace{5pt}
          \\ \hline \\
          \begin{tabular}{l}
            \textbf{1: Per-frame region-based GIC, sequence 1}\\
            \hspace{10pt}$\forall i.~\mathbf{x}^{(1)}_i  = \text{region\_GIC}(\mathbf{r},
                   \mathbf{m}, \mathbf{x}^{(1)}_i)$\\\\

            \textbf{2: Per-frame region-based GIC, sequence 2}\\
            \hspace{10pt}$\forall i.~\mathbf{x}^{(2)}_i  = \text{region\_GIC}(\mathbf{r},
                   \mathbf{m}, \mathbf{x}^{(2)}_i)$\\\\

            \textbf{3: Per-frame illumination subspace compensation}\\
            \hspace{10pt}$\forall i.~\hat{\mathbf{x}}^{(1)}_i = \mathbf{B}_I \mathbf{a}^*_i + \mathbf{x}^{(1)}_i$\\
            $~~~~~~~~$where $\mathbf{a}^*_i = \arg \min_{\mathbf{a}_i} D_{MAH}\left[\mathbf{B}_I \mathbf{a}_i +
                   \mathbf{x}^{(1)}_i - \langle \mathcal{C}_2
                   \rangle;~\mathcal{C}_2\right]$\\\\

             \textbf{4: The result is the normalized cluster $\hat{\mathcal{C}}_1$}\\
             \hspace{10pt}$\hat{\mathcal{C}}_1 = \{ \hat{\mathbf{x}}^{(1)}_i\}$\\\\
          \end{tabular}\\\hline
    \end{tabular}
    \caption[Two-stage illumination normalization algorithm.]
                  { \it Proposed illumination normalization algorithm. Coarse appearance changes due to illumination
                        variation are normalized using
                        region-based gamma intensity correction, while the residual variation is modelled
                        using a linear, pose-specific illumination subspace, learnt offline. Local manifold
                        shape is employed as a constraint in the second, `fine' stage of normalization in the form
                        of Mahalanobis distance for the computation of the optimal additive illumination subspace
                        component. }
    \vspace{6pt}\hrule
    \label{Fig: Illumination Overview}
\end{figure}

\subsection{Gamma intensity correction}\label{SubSubSec: Gamma
intensity correction} Gamma Intensity Correction (GIC) is a
well-known image intensity histogram transformation technique that
is used to compensate for global brightness changes
\cite{GonzWood1992}. It transforms pixel values (normalized to lie
in the range $[0.0, 1.0]$) by exponentiation so as to best match a
\emph{canonically illuminated image}. This form of the operator is
motivated by non-linear exposure-image intensity response of the
photographic film that it approximates well over a wide range of
exposure. Formally, given an image $I$ and a canonically illuminated
image $I_C$, the gamma intensity corrected image $I^*$ is defined as
follows:
\begin{align}
  I^*(x, y) = I(x, y)^{\gamma^*},
\end{align}
where $\gamma^*$ is the optimal gamma value and is computed using
\begin{align}
  \gamma^*  =&\arg \min_{\gamma} \| \mathbf{I}^{\gamma} - \mathbf{I}_C \| =\\
             &\arg \min_{\gamma} \sum_{x, y} \left[I(x, y)^{\gamma} - I_C(x, y)\right]^2.
\end{align}

This is a nonlinear optimization problem in 1D. In our
implementation of the proposed method it is solved using the Golden
Section search with parabolic interpolation, see
\cite{PresTeukVettFlan1992} for details.

\paragraph*{Region-based gamma intensity correction.}
Gamma intensity correction can be used across a wide range of types
of input to correct for \emph{global} brightness changes. However,
in the case of objects with a highly variable surface normal, such
as faces, it is unable to correct for the effects of side lighting.
This is recognized as one of the most difficult problems in face
recognition \cite{AdinMoseUllm1997}.

\begin{figure*}[!t]
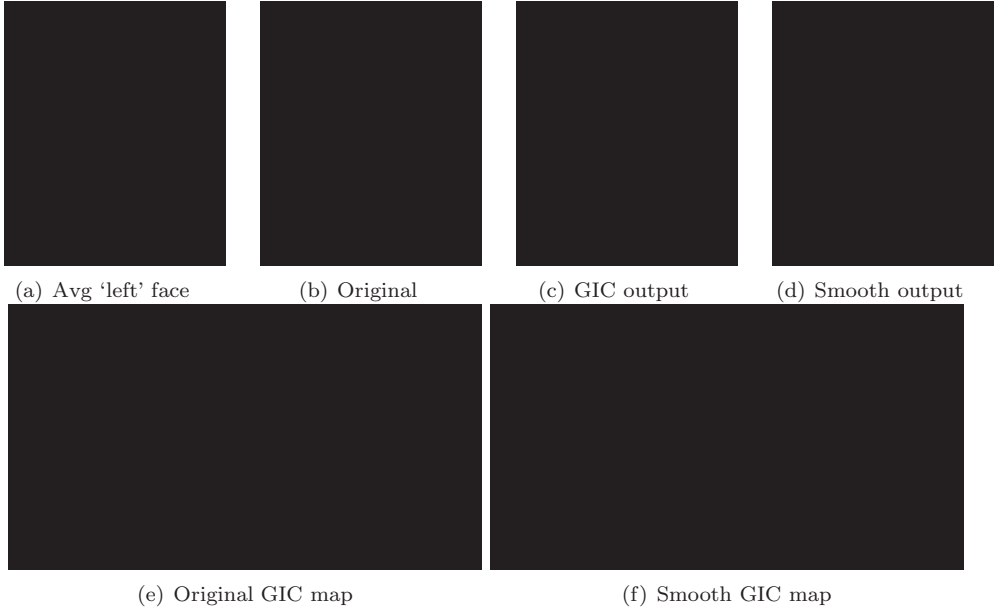

  \centering

  \subfigure[Avg `left' face]{\hspace{10pt}\includegraphics[width=0.21\textwidth]{PR2006_rgic_2.eps}}
  \subfigure[Original]{\hspace{10pt}\includegraphics[width=0.21\textwidth]{PR2006_rgic_originalimg_2.eps}}
  \subfigure[GIC output]{\hspace{10pt}\includegraphics[width=0.21\textwidth]{PR2006_rgic_edgy_2.eps}}
  \subfigure[Smooth output]{\hspace{10pt}\includegraphics[width=0.21\textwidth]{PR2006_rgic_smooth_2.eps}} \\
  \vspace{-10pt}
  \subfigure[Original GIC map]
     {\includegraphics[width=0.45\textwidth]{BMVC2004_gamma_edgy_1.eps}}
  \subfigure[Smooth GIC map]{\includegraphics[width=0.45\textwidth]{BMVC2004_gamma_smooth_1.eps}}\\

  \caption[Region-based gamma intensity correction.]
    {\it Canonical illumination
  image and the regions used in region-based GIC (a), original unprocessed face image (b),
  region-based GIC corrected image without smoothing (c), and region-based GIC corrected
  image with smoothing (d), gamma value map (e), smoothed gamma value map (f). Notice
  artefacts at region boundaries in the gamma corrected image (c). The output of the
  proposed smooth region-based GIC in (d) does not have the same problem. Finally, note
  that the coarse effects of the strong side lighting in (b) have been greatly removed.
  Gamma value maps corresponding to the original and the proposed methods is shown
  under, respectively, (e) and (f). }
  \label{Fig: Gamma Correction}
 \vspace{6pt}\hrule
\end{figure*}

Region-based GIC proposes to overcome this problem by dividing the
image (and hence implicitly the imaged object/face as well) into
regions corresponding to surfaces with near-constant surface normal.
Regular gamma intensity correction is then applied to each region
separately, see Figure~\ref{Fig: Gamma Correction}.

An undesirable result of this method is that it tends to produce
artificial intensity discontinuities at region boundaries
\cite{ShanGaoCaoZhao2003}. This occurs due to discontinuities in the
computed gamma values between neighbouring regions. We propose to
first Gaussian-blur the obtained gamma value map image
$\mathbf{\Gamma}^*$:
\begin{align}
  \mathbf{\Gamma}^*_S  = \mathbf{\Gamma}^* \ast
  \mathbf{G}_{\sigma=2},
\end{align}
before applying it to an input image to give the final, region-based
gamma corrected output $\mathbf{I}^*_S$:
\begin{align}
   \forall x,y.~I^*_S(x, y) = I(x, y)^{\Gamma^*_S(x, y)}
\end{align}
This method almost entirely remedies the problem with  boundary
artefacts, as illustrated in Figure~\ref{Fig: Gamma Correction}.
Note that because smoothing is performed on the gamma map, not the
processed image, the artefacts are removed without any loss of
discriminative, high frequency detail, see Figure~\ref{Fig: Gamma
Correction 1}.

\begin{figure}
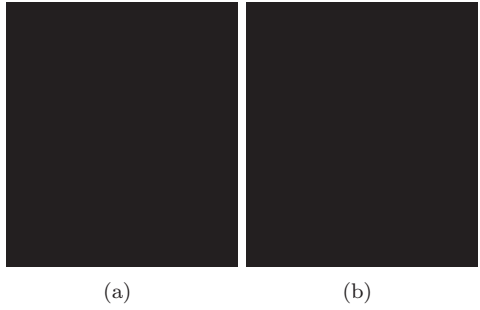

  \centering

  \subfigure[]{\includegraphics[width=0.22\textwidth]{PR2006_rgic_smooth_2.eps}}
  \subfigure[]{\includegraphics[width=0.22\textwidth]{PR2006_rgic_edgy_sm_2.eps}}

  \caption[Seamless vs. smoothed region-based GIC.]
      { \it
        (a) Seamless output of the proposed smooth region-based GIC. Boundary artefacts are
        removed without blurring of the image. Contrast this with output of a
        original region-based GIC, after Gaussian smoothing (b). Image quality is
        significantly reduced, with boundary edges still clearly visible.  }
  \label{Fig: Gamma Correction 1}
  \vspace{6pt}\hrule
\end{figure}

\subsection{Pose-specific illumination subspace normalization}
After region-based GIC is applied to all images, for each of the
pose clusters, it is assumed that the lighting variation can be
modelled using a linear, \emph{pose illumination subspace}. Given a
reference and a novel cluster corresponding to the same pose, each
frame of the novel cluster is normalized for the illumination
change. This is done by adding a vector from the pose illumination
subspace to the frame so that its distance from the reference
cluster's centre is minimal.

\paragraph*{Learning the model.} We define a
pose-specific illumination subspace to be a linear manifold that
explains \emph{intra-personal} appearance variations due to
illumination changes across a narrow range of poses. In other words,
this is the principal subspace of the within-class scatter.

Formalizing the definition above, given that $\mathbf{x}_{i,j}^k$ is
the $k$-th of $N_f(i, j)$ frames of person $i$ under the illumination
$j$ (out of $N_l(i)$), the within-class scatter matrix is:
\begin{align}
  \label{Eqn: WClass Scatter}
  \mathbf{S}_B = \sum_{i=1}^{N_p} \sum_{j=1}^{N_l(i)} \sum_{k=1}^{N_f(i,j)} (\mathbf{x}_{i, j}^k -
  \bar{\mathbf{x}}_i)(\mathbf{x}_{i, j}^k -
  \bar{\mathbf{x}}_i)^T,
\end{align}
where $N_p$ is the total number of training individuals and
$\bar{\mathbf{x}}_i$ is the mean face of the person in the
range of considered poses:
\begin{align}
  \bar{\mathbf{x}}_i = \frac {\sum_{j=1}^{N_l(i)} \sum_{k=1}^{N_f(i,j)} \mathbf{x}_{i, j}^k} { \sum_j N_f(i,j)}.
\end{align}

The pose-specific illumination subspace basis $\mathbf{B}_I$ is then
computed by eigendecomposition of $\mathbf{S}_B$ as the principal
subspace explaining 90\% of data energy variation.

For offline learning of illumination subspaces we used 10s video
sequences of 20 individuals, each in 5 illumination conditions,
acquired at 10fps. The first few basis vectors learnt in the
described manner are shown as images in Figure~\ref{Fig:
Illumination Subspaces}.

\begin{figure}[!t]
  \centering

  \subfigure[Frontal]{\includegraphics[width=0.65\textwidth]{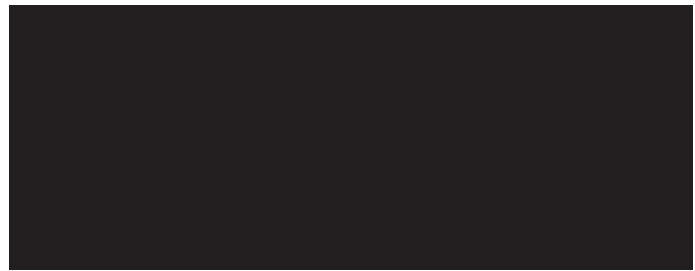}}
  \subfigure[Left]   {\includegraphics[width=0.65\textwidth]{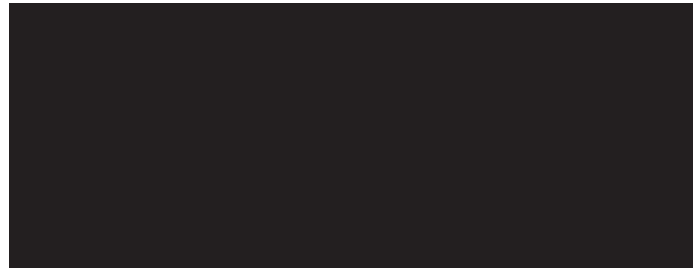}}\\
  \subfigure[Eigenvalues]   {\includegraphics[width=0.65\textwidth]{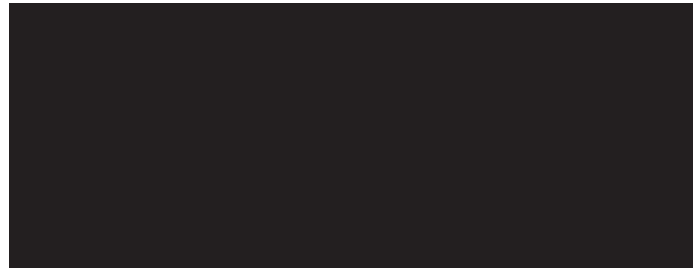}}\\

  \caption[Pose-specific illumination subspaces.]{\it Shown as images are the first 5 bases of pose-specific illumination subspaces for the (a) frontal
                        and (b) left head orientations. The distribution of energy for pose-specific
                        illumination variation across principal directions is shown in (c). }
  \label{Fig: Illumination Subspaces}
  \vspace{6pt}\hrule
\end{figure}

\paragraph*{Employing the model.} Let $\mathcal{C}_1 = \{ \mathbf{x}^{(1)}_1, \dots, \mathbf{x}^{(1)}_{N_1} \}$ and
$\mathcal{C}_2 = \{ \mathbf{x}^{(2)}_1, \dots, \mathbf{x}^{(2)}_{N_2} \}$ be two corresponding pose
clusters of different appearance manifolds, previously preprocessed
using the region-based gamma correction algorithm described in
Section~\ref{SubSubSec: Gamma intensity correction}. Cluster
$\mathcal{C}_1$ is then illumination-normalized with respect to
$\mathcal{C}_2$ (we will therefore refer to $\mathcal{C}_2$ as the
\emph{reference cluster}), under the null assumption that the
identities of the two people they represent are the same. The
normalization is performed on a frame-by-frame basis, by adding a
vector $\mathbf{B}_I \mathbf{a}^*_i$ from the estimated
pose-specific illumination subspace:
\begin{align}
    \forall i.~\hat{\mathbf{x}}^{(1)}_i = \mathbf{B}_I \mathbf{a}^*_i + \mathbf{x}^{(1)}_i
    \label{Eqn: Basic Normalization}
\end{align}
where we define $\mathbf{a}^*_i$ as:
\begin{align}
  \label{Eqn: IL sspace component}
  \mathbf{a}^*_i = \arg \min_{\mathbf{a}_i} \|\mathbf{B}_I \mathbf{a}_i +
                   \mathbf{x}^{(1)}_i - \langle \mathcal{C}_2
                   \rangle\|,
\end{align}
and $\| \dots \|$ is a vector norm and $\langle \mathcal{C}_2
\rangle$ the mean face of cluster $\mathcal{C}_2$. We then define
cluster $\mathcal{C}_1$ normalized to $\mathcal{C}_2$ to be
$\hat{\mathcal{C}}_1 = \{ \hat{\mathbf{x}}^{(1)}_i  \}$. This form
is directly motivated by the definition of a pose-specific subspace.

To understand the next step, which is the choice of the vector norm
in \eqref{Eqn: IL sspace component}, it is important to
notice in the definition of the pose-specific illumination subspace,
that the basis $\mathbf{B}_I$ explains not only appearance
variations caused by illumination: reflectance properties of faces used
in training (e.g.\ their albedos), as well as subjects' pose changes
also affect it. This is especially the case as we do not make the
common assumption that surfaces of faces are Lambertian, or that
light sources are point lights at infinity.

The significance of this observation is that the subspace of a
dimensionality sufficiently high to explain the modelled phenomenon
(illumination changes) will, undesirably, also be able to explain
`distracting' phenomena, such as differing identity. The problem is
therefore that of \emph{constraining} the region of interest of the
subspace to that which is most likely to be due to illumination
changes for a particular individual. For this purpose we propose to
exploit the local structure of appearance manifolds, which are
smooth. We do this by employing the Mahalanobis distance (using the
probability density corresponding to the reference cluster) when
computing the illumination subspace correction for each novel frame
using \eqref{Eqn: IL sspace component}. Formally:
\begin{align}
  \mathbf{a}^*_i = &\arg \min_{\mathbf{a}_i} \left(\mathbf{B}_I \mathbf{a}_i +
                     \mathbf{x}^{(1)}_i - \langle \mathcal{C}_2
                     \rangle \right)^T \cdot
                   \mathbf{B}_2 \mathbf{\Lambda}_2^{-1}
                   \mathbf{B}_2^T
                   \left(\mathbf{B}_I \mathbf{a}_i +
                   \mathbf{x}^{(1)}_i - \langle \mathcal{C}_2
                   \rangle \right),
\end{align}
where $\mathbf{B}_2$ and $\mathbf{\Lambda}_2$ are, respectively,
reference cluster's orthonormal basis and the diagonalized
covariance matrix. We found that the use of Mahalanobis distance, as
opposed to the usual Euclidean distance, achieved better explanation
of novel images when the person's identity was the same, and worse
when it was different, achieving better inter-to-intra class
separation.

This quadratic minimization problem is solved by differentiation and
the minimum is achieved for:
\begin{align}\label{Eqn: Illumination Normalization 3}
  \mathbf{a}_i^* = &\left(\mathbf{B}_I^T \mathbf{B}_2 \mathbf{\Lambda}_2^{-1} \mathbf{B}_2^T
  \mathbf{B}_I\right)^{-1} \cdot
  \mathbf{B}_I^T \mathbf{B}_2 \mathbf{\Lambda}_2^{-1} \mathbf{B}_2^T \left(\langle
\mathcal{C}_2 \rangle - \mathbf{x}\right)
\end{align}

Examples of registered and cropped face images before and after
illumination normalization can be seen in Figure~\ref{Fig:
Illumination Corrected Frames}~(a).

\paragraph*{Practical considerations.} The computation of the optimal value
$\mathbf{a}^*$ using \eqref{Eqn: Illumination Normalization
3} involves inversion and Principal Component Analysis (PCA) on
matrices of size $D \times D$, where $D$ is the number of pixels in
a face image (in our case equal to 900, see Section~\ref{SubSec:
Registration}). Both of these operations put high demands on
computer resources. To reduce the computational overhead, we exploit
the assumption that data modelled is of much lower dimensionality
than $D$.

Formalizing the model of low-dimensional face manifolds, we assume
that an image $\mathbf{y}$ of subject $i$'s face is drawn from the
probability density $p^{(i)}_F(\mathbf{y})$ within the face space,
and embedded in the image space by means of a mapping function
$f^{(i)}:\mathbb{R}^d\to\mathbb{R}^D$. The resulting point in the
$D$-dimensional space is further perturbed by noise drawn from a
noise distribution $p_n$ (note that the noise operates in the image
space) to form the observed image $\mathbf{x}$. Therefore the
distribution of the observed face images of the subject $i$ is given
by the integral:
\begin{align}\label{Eqn: Manifold_pdf}
    p^{(i)}(\mathbf{x}) = \int p_F^{(i)}(\mathbf{y}) p_n(f_i(\mathbf{y}) - \mathbf{x}) d\mathbf{y}
\end{align}

This model is then used in two stages:
\begin{enumerate}
  \item Pose-specific PCA dimensionality reduction,
  \item Exact computation of the linear principal and rapid estimation of
  the complementary subspace of a pose cluster.
\end{enumerate}

Specifically, we first perform a linear projection of all images in
a specific pose cluster to a \emph{pose-specific face subspace} that
explains 95\% of data variation in a specific pose. This achieves
data dimensionality reduction from 900 to 250.

Referring back to \eqref{Eqn: Manifold_pdf}, to
additionally speed up the process, we estimate the intrinsic
dimensionality of face manifolds (defined as explaining 95\% of
within-cluster data variability) and assume that all other variation
is due to isotropic Gaussian noise $p_n$. Hence, we can write the
basis of the PCA subspace corresponding to the reference cluster as
consisting of a principal and complementary subspaces
\cite{TippBish1999} represented by orthonormal basis matrices,
respectively $\mathbf{V}_P$ and $\mathbf{V}_C$:
\begin{align}
  \mathbf{B}_2 = [\mathbf{V}_P \mathbf{V}_C]
  \label{Eqn: Cluster Density}
\end{align}
where $\mathbf{V}_P \in \mathbb{R}^{250 \times 6}$ and $\mathbf{V}_C
\in \mathbb{R}^{250 \times 244}$. The principal subspace and the
associated eigenvectors $\mathbf{v}_1, \dots, \mathbf{v}_6$ are
rapidly computed, e.g.\ using \cite{BaglCalvReic1996}. The isotropic
noise covariance and the complementary subspace basis are then
estimated in the following manner:
\begin{align}
  \lambda_n = \omega \sum_{i=1}^6 \lambda_i && \mathbf{V}_C = \text{null}\left( \mathbf{V}_P \right)
\end{align}
where the nullspace of the principal subspace is computed using the
QR-decomposition \cite{PresTeukVettFlan1992} and the value of
$\omega$ estimated from a small training corpus; we obtained $\omega
\approx 2.2 e{-4}$. The diagonalized covariance matrix is then
simply:
\begin{align}
  \mathbf{\Lambda}_2 = \diag (
  \lambda_1, \ldots, \lambda_6,\overbrace{\lambda_n, \ldots,
  \lambda_n}^{244})
\end{align}

\begin{figure}[!t]
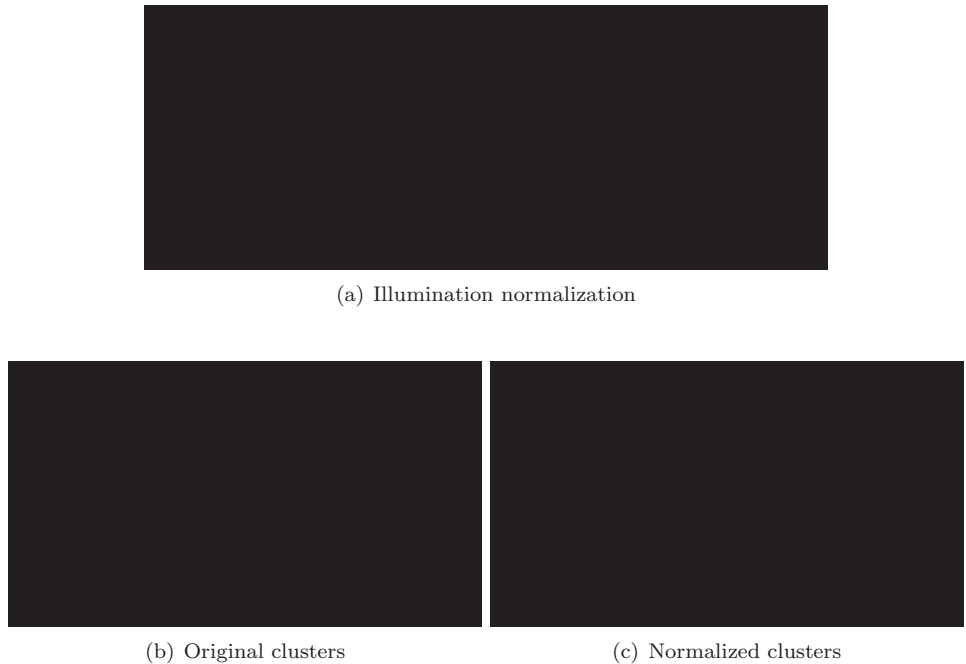

  \centering

  \subfigure[Illumination normalization]{\includegraphics[width=0.65\textwidth]{PR2006_faces_normalization_2.eps}}
    \\ \vspace{10pt}

  \subfigure[Original clusters]{\includegraphics[width=0.45\textwidth]{PR2006_clusters_before_2.eps}}
  \subfigure[Normalized clusters]{\includegraphics[width=0.45\textwidth]{PR2006_clusters_after_2.eps}}

  \caption[Illumination normalization effects in the image space.]
         {\it
           In (a) are respectively, top to bottom, shown the
           original registered and cropped face images from
           an input video sequence, the same faces after the
           proposed illumination normalization and a sample
           from the \emph{reference} video sequence. The
           effects of strong side lighting have been greatly
           removed, while at the same time a high level of
           detail is retained. The corresponding data from
           the two sequences, before and after illumination
           compensation are shown under (b) and (c). Shown
           are their projections to the first two principal
           components. Notice that initially the clusters
           were completely non-overlapping. Illumination
           normalization has adjusted the location of centre
           of the blue cluster, but has also affected its
           spread. after normalization, while overlapping,
           the two sets of patterns are still distributed
           quite differently. }
  \label{Fig: Illumination Corrected Frames}
  \vspace{6pt}\hrule
\end{figure}

\section{Comparing normalized pose clusters}\label{SubSec:
Comparing Clusters} Having illumination normalized one face cluster
to match another, we want to compute a similarity measure between
them, a \emph{distance}, expressing our degree of belief that they
belong to the same person.

At this point it is instructive to examine the effects of the
described method for illumination normalization on the face
patterns. Two clusters before, and after one has been normalized,
are shown in Figure~\ref{Fig: Illumination Corrected Frames}~(b,c).
An interesting artefact can be observed: the spread of the
normalized cluster is significantly reduced. This is easily
understood by referring back to \eqref{Eqn: Basic
Normalization}-\eqref{Eqn: IL sspace component} and noticing that
the normalization is performed frame-by-frame, trying to make each
normalized face as close as possible to the reference cluster's
mean, i.e.\ a \emph{single point}. For this reason, dissimilarity
measures between probability densities common in the literature,
such as such as the Bhattacharyya distance, the Kullback-Leibler
divergence \cite{AranShakFish+2005,ShakFishDarr2002} or the
Resistor-Average distance \cite{AranCipo2006,JohnSina2001}, are not
suitable choices. Instead, we propose to use the simple Euclidean
distance between normalized cluster centres:
\begin{align}
D(\mathcal{C}_1, \mathcal{C}_2) = \frac {\sum_{i=1}^{N_1}
\hat{\mathbf{x}}_i^{(1)}} { N_1 } - \frac {\sum_{j=1}^{N_2}
\mathbf{x}_j^{(2)}} { N_2 }.
\end{align}

\subsubsection{Inter-manifold distance}\label{SubSubSec: Intermanifold Distance}
The last stage in the proposed method is the computation of an
inter-manifold distance, or an inter-manifold dissimilarity measure,
based on the distances between corresponding pose clusters. There
are two main challenges in this problem: (i) depending on the poses
assumed by the subjects, one or more clusters, and hence the
corresponding distances, may be void; (ii) different poses are not
equally important, or discriminative, in terms of face recognition
\cite{SimZhang2004}.

Writing $\mathbf{d}$ for the vector containing the three pose
cluster distances, we want to classify a novel appearance manifold
to the gallery class giving the highest probability of corresponding
to it in identity, $P(s|\mathbf{d})$. Then, using the Bayes'
theorem:
\begin{align}
    P(&s|\mathbf{d}) =\frac { p(\mathbf{d}|s)P(s) } { p(\mathbf{d}) }\\
     &= \frac { p(\mathbf{d}|s)P(s) } { p(\mathbf{d}|s)P(s) + p(\mathbf{d}|\neg s)P(\neg s)
     }\\
     &=\frac { 1 } { 1 + p(\mathbf{d}|\neg s)P(\neg s) / p(\mathbf{d}|s)P(s) }
\end{align}

Assuming that the ratio of same-identity to differing-identities
priors $P(\neg s) / P(s)$ is a constant across individuals, it is
clear than classifying to the class with the highest
$P(s|\mathbf{d})$ is equivalent to classifying to the class
with the highest \emph{likelihood ratio}:
\begin{align}\label{Eqn: Final Decision}
    \mu(\mathbf{d}) = \frac { p(\mathbf{d} | s) }{ p(\mathbf{d} | \neg
    s) }
\end{align}

\paragraph*{Learning pose likelihood ratios.}
Understanding that $\mathbf{d} = [D_1, D_2, D_3]^T$ we assume
statistical independence between pose cluster distances:
\begin{align}
  &p(\mathbf{d} | s)     = \prod_{i=1}^3 p(D_i | s) \\
  &p(\mathbf{d} | \neg s) = \prod_{i=1}^3 p(D_i | \neg s)
\end{align}

We propose to learn likelihood ratios $\mu(D_i) = p(\mathbf{d} | s)
/ p(\mathbf{d} | \neg s)$ offline, from a small data corpus,
labelled by the identity, in two stages. First, (i) we obtain a
Parzen window estimate of intra- and inter- personal pose distances
by comparing all pairs of training appearance manifolds; then (ii)
we refine the estimates using a Radial Basis Functions (RBF)
artificial neural network trained for each pose.

\begin{figure}
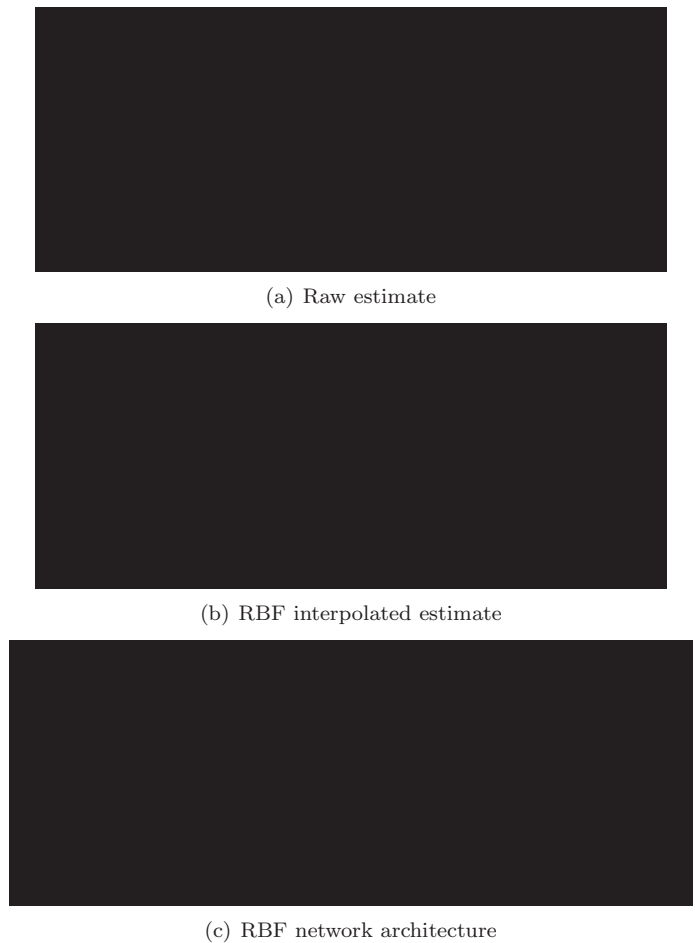

  \centering
  \subfigure[Raw estimate]{\includegraphics[width=0.6\textwidth]{PR2006_lr_front_raw.eps}}
  \subfigure[RBF interpolated estimate]{\includegraphics[width=0.6\textwidth]{PR2006_lr_front.eps}}
  \subfigure[RBF network architecture]{\includegraphics[width=0.65\textwidth]{PR2006_RBF_net1.eps}}

  \caption[Learnt pose-specific likelihood ratios.]{\it Likelihood ratio corresponding
                  to frontal head pose obtained from the training corpus using Parzen
                  windows (a) and the RBF network-based likelihood ratio (b). The corresponding
                  RBF network architecture is shown in (c). Note that the initial estimate (a)
                  is not monotonically decreasing, while (b) is. }
                  \label{Fig: Posteriors}
\end{figure}

A Parzen window-based \cite{DudaHartStor2001} estimate of $\mu(D)$
for the frontal head orientation, obtained by directly comparing
appearance manifolds as described in Sections~\ref{SubSec:
Registration}-\ref{SubSec: Comparing Clusters} is shown in
Figure~\ref{Fig: Posteriors}~(a). In the proposed method, this, and
the similar likelihood ratio estimates for the other two head poses
are not used directly for recognition as they suffer from an
important limitation: the estimates are ill-defined in domain
regions sparsely populated with training data. Specifically, an
artefact caused by this problem can be observed by noting that the
likelihood ratios are not monotonically decreasing. What this means
is that \emph{more distant} pose clusters can result in higher
chance of classifying two sequences as originating from the
\emph{same individual}.

To overcome the problem of insufficient training data, we train a
two-layer RBF-based neural network for each of the discrete poses
used in approximating face appearance manifolds, see
Figure~\ref{Fig: Posteriors}~(c). In its basic form, this means that
the estimate $\hat{\mu}(D_i)$ is given by the following expression:
\begin{align}
  \hat{\mu}(D_i) = \sum_j \alpha_j \mathcal{G}(D_i; \mathbf{\mu}_j, \sigma_j
  ),
\end{align}
where:
\begin{align}
  \mathcal{G}(D_i; &\mathbf{\mu}_j, \sigma_j) = \notag \\
  &\frac {1} { \sigma \sqrt{2\pi }}
    \exp{- \frac{ (D_i - \mathbf{\mu}_j)^2 }{2\sigma^2}  }.
\end{align}

In the proposed method, this is modified so as to enforce prior
knowledge on the functional form of $\mu(D_i)$ in the form of its
monotonicity:
\begin{align}
  \hat{\mu}^*(&D_i) = \notag \\
   &\max_{\delta>D_i} \left\{\sum_j \alpha_j \mathcal{G}(D_i; \mathbf{\mu}_j,
  \sigma_j),~\hat{\mu}(\delta)\right\}
  \label{Eqn: RBF Estimate}
\end{align}

\begin{figure}
  \centering
  \includegraphics[width=0.65\textwidth]{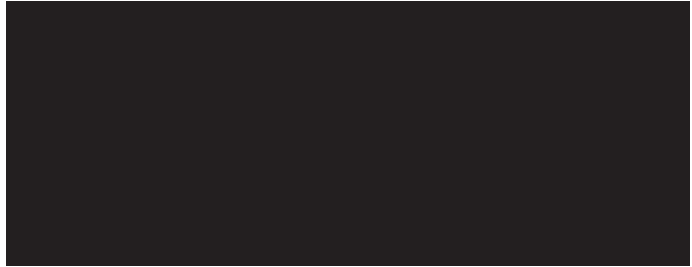}

  \caption[Joint likelihood ratio.]{\it Joint RBF network-based likelihood
                      ratio for the frontal and left head orientations. }
  \vspace{6pt}\hrule
  \label{Fig: Joint LR}
\end{figure}

Finally, to ensure that the networks are trained using reliable data
(in the context of training sample density in the training domain),
we use only local peaks of Parzen window-based estimates. Results
using six second-layer neurons, each with the spread of $\sigma_j =
60$, see \eqref{Eqn: RBF Estimate}, are summarized in
Figures~\ref{Fig: Posteriors} and \ref{Fig: Joint LR}.

\section{Empirical evaluation} \label{Sec: Evaluation} Methods in
this chapter were evaluated on the \textit{ToshFace} data set. To
establish baseline performance, we compared our recognition
algorithm to:
\begin{itemize}
    \item Mutual Subspace Method (MSM) of Fukui and Yamaguchi \cite{FukuYama2003},

    \item KL divergence-based algorithm of Shakhnarovich \textit{et al.} (KLD)
        \cite{ShakFishDarr2002},

    \item Majority vote across all pairs of frames using Eigenfaces of Turk and Pentland \cite{TurkPent1991}.
\end{itemize}
In the KL divergence-based method we used principal subspaces that
explain 85\% of data variation energy. In MSM we set the
dimensionality of linear subspaces to 9 and used the first 3
principal angles for recognition, as suggested by the authors in
\cite{FukuYama2003}. For the Eigenfaces method, the 22-dimensional
eigenspace used explained 90\% of total training data energy.

Offline training, i.e.\ learning of the pose-specific illumination
subspaces and likelihood ratios, was performed using 20 randomly
chosen individuals in 5 illumination settings, for a total of 100
sequences. These were not used for neither gallery data nor test
input for the evaluation reported in this section.

Recognition performance of the proposed system was assessed by
training it with the remaining 40 individuals in a single
illumination setting, and using the rest of the data as test input.
In all tests, both training data for each person in the gallery, as
well as test data, consisted of only a single sequence.

\subsection{Results} \label{SubSec: Results}
The performance of the proposed method is summarized in
Table~\ref{Tab:AccResults}. We tabulated the recognition rates
achieved across different combinations of illuminations used for
training and test input, so as to illustrate its degree of
sensitivity to the particular choice of data acquisition conditions.
An average rate of 95\% was achieved, with a mean standard deviation
of only 4.7\%. Therefore, we conclude that the proposed method is
successful in recognition across illumination, pose and motion
pattern variation, with high robustness to the exact imaging setup
used to provide a set of gallery videos.

\begin{table*}
   \centering
   \caption[Recognition results.]{\it Recognition performance (\%) of the proposed method
             using different illuminations for training and test
             input. Excellent results are demonstrated with little
             dependence of the recognition rate on the data
             acquisition conditions. \vspace{10pt}}
           \label{Tab:AccResults}
   \Large

   \begin{tabular*}{1.00\textwidth}{@{\extracolsep{\fill}}l|ccccc|cc}
    \Hline
         & \normalsize IL.\ 1 &  \normalsize IL.\ 2 & \normalsize IL.\ 3  & \normalsize IL.\ 4  &  \normalsize IL.\ 5  & \normalsize mean & \normalsize std\\
    \hline
    \normalsize ~IL.\ 1  & \small 100 &  \small 90  & \small 95  &  \small 95  &  \small 90  &  \small 94     & \small 4.2 \\
    \normalsize ~IL.\ 2  & \small 95  &  \small 95  & \small 95  &  \small 95  &  \small 90  &  \small 94     & \small 2.2 \\
    \normalsize ~IL.\ 3  & \small 95  &  \small 95  & \small 100 &  \small 95  &  \small 100 &  \small 97     & \small 2.7 \\
    \normalsize ~IL.\ 4  & \small 95  &  \small 90  & \small 100 &  \small 100 &  \small 95  &  \small 96     & \small 4.2 \\
    \normalsize ~IL.\ 5  & \small 100 &  \small 80  & \small 100 &  \small 95  &  \small 100 &  \small 95     & \small 8.7 \\
    \hline
    \normalsize ~mean      &  \small 97 &  \small 90  &  \small 98  &  \small 96  &   \small 95  &  \small 95.2   & \small 4.5 \\
    \Hline
  \end{tabular*}

\end{table*}

\begin{figure}[!t]
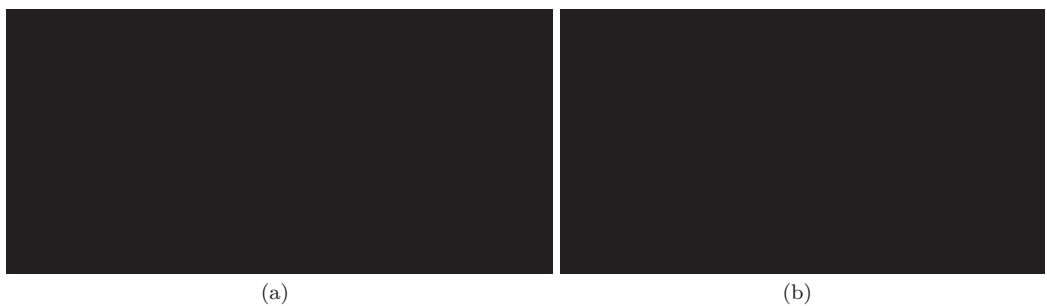

  \centering
  \footnotesize
  \subfigure[]{\includegraphics[width=0.52\textwidth]{PR2006_separability.eps}}
  \subfigure[]{\includegraphics[width=0.46\textwidth]{PR2006_ROC.eps}}

  \caption[Separation of intra- and inter-personal manifold distances.]{ \it Cumulative distributions
            of intra-personal (dashed line) and inter-personal (solid line)
            distances (a). Good separability is demonstrated. The corresponding
            ROC curve can be seen in (b) -- less than 0.5\% of false positive
            rate is attained for 91\% true positive rate. The corresponding
            distance threshold choice is numerically well-conditioned, as
            witnessed by close-to-zero derivatives of the plots in (a) at
            the corresponding point. }
  \label{Fig: Separability}
  \vspace{6pt}\hrule
\end{figure}

This conclusion is further corroborated by Figure~\ref{Fig:
Separability}~(a), which shows cumulative distributions of inter-
and intra-personal manifold distances (see Section~\ref{SubSubSec:
Intermanifold Distance}) and Figure~\ref{Fig: Separability}~(b)
which plots the Receiver-Operator Characteristic of the proposed
algorithm. Good class separation can be seen in both, illustrating
the suitability of our method for verification (one-against-one
matching) applications: less than 0.5\% false positive rate is
attained for 91.2\% true positive rate. Additionally, it is
important to note that good separation is maintained across a wide
range of distances, as can be seen in Figure~\ref{Fig:
Separability}~(a) from low gradients of inter- and intra- class
distributions e.g.\ on the interval between $1.0$ and $15.0$. This is
significant as it implies that the interclass threshold choice is
not very numerically sensitive: by choosing a threshold in the
middle of this range, we can expect the recognition performance to
generalize well to different data sets.

\subsubsection*{Pose clusters}
One of the main premises that this work rests on is the idea that
illumination and pose robustness in recognition can be achieved by
decomposing an appearance manifold into a set of pose ranges (see
Section~\ref{SubSubSec: Defining Pose Clusters}) which are, after
being processed independently, probabilistically combined (see
Section~\ref{SubSubSec: Intermanifold Distance}). We investigated
the discriminating power of each of the three pose clusters used in
the proposed context by performing recognition using the
inter-cluster distance defined in Section~\ref{SubSec: Comparing
Clusters}. Table~\ref{Tab: Distances} show a summary of the results.
High recognition rates were achieved even using only a single pose
cluster. Furthermore, the proposed method for integrating cluster
distance into a single inter-manifold distance can be seen to
improve the average performance of the most discriminative pose. In
the described recognition framework, side poses contributed more
discriminative information to the distance than the frontal pose (in
spite of a lower average number of side faces per sequence, see
Figure~\ref{Fig: Registration Hists} in Section~\ref{SubSec:
Registration}), as witnessed by both a higher average recognition
accuracy and lower standard deviation of recognition. It is
interesting to observe that this is in agreement with the finding
that appearance in a roughly semi-profile head pose is inherently
most discriminative for recognition~\cite{SimZhang2004}.

\begin{table*}
  \centering
  \caption[Discriminative power of different poses.]{ \it A comparison of
            identification statistics for recognition using each of the pose-specific cluster
            distances separately and the proposed method for combining them using an RBF-based
            neural network. In addition to the expected performance improvement when using all
            over only some poses, it is interesting to note different contributions of side
            and frontal pose clusters, the latter being more discriminative in the context of
            the proposed method.\vspace{10pt}
          }
  \Large
  \begin{tabular*}{1.00\textwidth}{@{\extracolsep{\fill}}l|ccc}
    \Hline

    \bf \normalsize ~Measure   & {\normalsize ~Manifold distance} & {\normalsize Front clusters distance} & {\normalsize Side clusters distance} \\
    \hline
    \bf \normalsize ~mean & {\small 95}    & {\small 90}             & {\small 93} \\
    \bf \normalsize ~std  & {\small 4.7}   & {\small 5.7}            & {\small 3.6} \\
  \Hline
  \end{tabular*} \\
  \label{Tab: Distances}
\end{table*}

\subsubsection*{Other algorithms}
The result of the comparison with the other evaluated methods is
shown in Table~\ref{Tab: Comparison}. The proposed algorithm
outperformed others by a significant margin. Majority vote using
Eigenfaces and the KL divergence algorithm performed with
statistically insignificant difference, while MSM showed least
robustness to the extreme changes in illumination conditions.
It is interesting to note that all three algorithms achieved
perfect recognition when training and test sequences were acquired in
the same illumination conditions. Considering the simplicity and
computational efficiency of these methods, investigation of their
behaviour when used on preprocessed data (e.g.\ high-pass filtered images
\cite{AranZiss2005,FitzZiss2002} or self-quotient images
\cite{WangLiWang2004}) appears to be a promising research direction.

\begin{table*}
  \centering
  \caption[Comparison with algorithms in the literature.]
      { \it Average recognition
            rates (\%) of the compared methods across different illumination
            conditions used for training and test. The performance of the
            proposed method is by far the best, both in terms of the average
            recognition rate and its variance.\vspace{10pt} }
  \Large
  \begin{tabular*}{1.00\textwidth}{@{\extracolsep{\fill}}l|cccc}
    \Hline

    \bf \normalsize ~Method  & \normalsize Proposed method & \normalsize Majority vote, Eigenfaces & \normalsize KLD  & \normalsize MSM  \\
    \hline
    \bf \normalsize ~mean    & \small 95  & \small 43    & \small 39   & \small 24   \\
    \bf \normalsize ~std     & \small 4.7 & \small 31.9  & \small 32.5 & \small 38.9 \\
  \Hline
  \end{tabular*} \\
  \label{Tab: Comparison}
\end{table*}

\subsubsection*{Failure modes} Finally, we investigated the main
failure models of our algorithm. An inspection of failed
recognitions suggests that the largest difficulty was caused by
significant user motion to and from the camera. During the data
acquisition, for some of the illumination conditions the dominant
light sources were relatively close to the user (from $\approx
0.5m$). This invalidated the implicit assumption that illumination
conditions were unchanging within a single video sequence i.e.\ that
the main cause of appearance changes in images was head rotation.

Another limitation of the method was observed in cases when only few
faces were clustered to a particular pose, either because of facial
feature detection failure or because the user did not spend enough
time in a certain range of head poses. The noisy estimate of the
corresponding cluster density in \eqref{Eqn: Cluster Density}
propagated the estimation error to illumination normalized images and
finally to the overall manifold distance, reducing the separation
between classes.

\section{Summary and conclusions} \label{Sec: Conclusions}
In this chapter we introduced a novel algorithm for face recognition
from video, robust to changes in illumination, pose and the motion
pattern of the user. This was achieved by combining person-specific
face motion appearance manifolds with generic pose-specific
illumination manifolds, which were assumed to be linear. Integrated
into a fully automatic practical system, the method has demonstrated
a high recognition rate in realistic, uncontrolled data acquisition
conditions.

\section*{Related publications}

The following publications resulted from the work presented in this
chapter:

\begin{itemize}
  \item O. Arandjelovi{\'c} and R. Cipolla.  An illumination invariant face recognition system for
                  access control using video. In \textit{Proc. IAPR British Machine Vision Conference (BMVC)},
                  pages 537--546, September 2004.
                  \cite{AranCipo2004a}
\end{itemize}

\graphicspath{{./09gSIM/}}
\chapter{Generic Shape-Illumination Manifold}
\label{Chp: gSIM}
\begin{center}
  \footnotesize
  \vspace{-20pt}
  \framebox{\includegraphics[width=0.65\textwidth]{title_img.eps}}\\
  Maurits C. Escher. \textit{Prentententoonstelling}\\
  1956, Lithograph
\end{center}

\cleardoublepage

In the previous chapter it was shown how \textit{a priori}
domain-specific knowledge can be combined with data-driven learning
to reliably recognize in the presence of illumination, pose and
motion pattern variations. The main limitations of the proposed
method are: (i) the assumption of linearity of pose-specific
illumination subspaces, (ii) the coarse pose-based fusion of
discriminative information from different frames, and (iii) the
appearance distribution artifacts introduced during pose-specific
illumination normalization.

This chapter finalizes the part of the thesis that deals with
robustly comparing two face motion sequences. We describe the
\textit{Generic Shape-Illumination Manifold} recognition algorithm
that in a principled manner handles all of the aforementioned
limitations.

In particular there are three areas of novelty: (i) we show how a
photometric model of image formation can be combined with a
statistical model of generic face appearance variation to generalize
in the presence of extreme illumination changes; (ii) we use the
smoothness of geodesically local appearance manifold structure and a
robust same-identity likelihood to achieve robustness to unseen head
poses; and (iii) we introduce a precise video sequence
``reillumination'' algorithm to achieve robustness to face motion
patterns in video.

The proposed algorithm consistently demonstrated a nearly perfect
recognition rate (over 99.5\% on \textit{CamFace}, \textit{ToshFace}
and \textit{Face Video} data sets), significantly outperforming
state-of-the-art commercial software and methods from the
literature.

\section{Synthetic reillumination of face motion
manifolds}
\begin{figure}
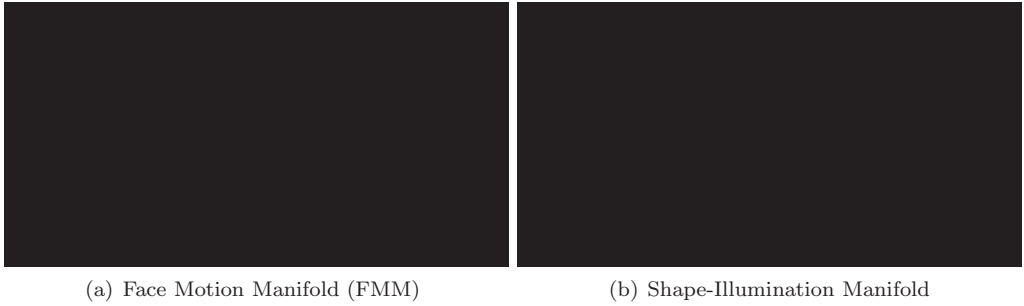

  \centering
  \small
  \subfigure[Face Motion Manifold (FMM)]{
  \includegraphics[width=0.48\textwidth]{ICCV2005_fmm2.ps}}
  \subfigure[Shape-Illumination Manifold]{\includegraphics[width=0.48\textwidth]{ICCV2005_sim2.ps}}

    \caption[Appearance and albedo-free appearance manifolds.]
            { \it   Manifolds of (a) face appearance and (b) albedo-free appearance
                    i.e.\ the effects of illumination and pose changes, in a single
                    motion sequence. Shown are projections to the first
                    3 linear principal components, with a typical manifold
                    sample on the top-right. }
            \label{Fig: Manifolds}
  \vspace{6pt}\hrule
\end{figure}

\label{SubSec: Synthetic Reillumination of Face Motion
Sequences} One of the key ideas of this chapter is the algorithm for
\emph{reillumination} of video sequences. Our goal is to take two
input sequences of faces and produces a third, synthetic one, that
contains the same poses as the first in the illumination of the
second one. For the proposed method, the crucial properties are the
(i) continuity and (ii) smoothness of face motion manifolds, see
Figure~\ref{Fig: Manifolds}

The proposed method consists of two stages. First, each face from
the first sequence is matched with the face from the second that
corresponds to it best in terms of pose. Then, a number of faces
close to the matched one are used to finely reconstruct the
reilluminated version of the original face. Our algorithm is
therefore global, unlike most of the previous methods which use a
sparse set of detected salient points for registration, e.g.\
\cite{AranZiss2005,BergBergEdwa+2004,FukuYama2003}. We have found
these to fail on our data set due to the severity of illumination
conditions (see Section~\ref{Sec: Empirical Evaluation}). The two
stages of the proposed algorithm are next described in detail.

\subsection{Stage 1: pose matching}\label{SubSec: Stage 1} Let $\{ \mathbf{X}_i \}^{(1)}$
and $\{ \mathbf{X}_i \}^{(2)}$ be two motion sequences of a person's
face in two different illuminations. Then, for each
$\mathbf{X}_i^{(1)}$ we are interested in finding
$\mathbf{X}_{c(i)}^{(2)}$ that corresponds to it best in terms of
head pose. Finding the unknown mapping $c$ on a frame-by-frame basis
is difficult in the presence of extreme illumination changes and
when face images are of low resolution. Instead, we exploit the face
manifold smoothness by formulating the problem as a minimization
task with the fitness function taking on the form:
\begin{align}
  f(c) &= f_{match}(c) + \omega f_{reg}(c) \label{Eqn: Fitness Basic} \\
  &= \underbrace{\sum_j d_E\left(\textbf{X}_j^{(1)}, \textbf{X}_{c(j)}^{(2)} \right)^2}_{\text{Matching term}}
  +
  \omega \underbrace{\sum_j \sum_k \frac
  {d_G^{(2)}\left(\textbf{X}_{c(j)}^{(2)}, \textbf{X}_{c(n(j, k))}^{(2)};
   \{ \textbf{X}_j \}^{(2)} \right)}
   {d_G^{(1)}\left(\textbf{X}_j^{(1)}, \textbf{X}_{n(j, k)}^{(1)}; \{ \textbf{X}_j
   \}^{(1)}\right)}}_{\text{Regularization term}} \label{Eqn: Fitness Function}
\end{align}
where $n(i, j)$ is the $j$-th of $K$ nearest neighbours of face $i$,
$d_E$ a pose dissimilarity function and $d_G^{(k)}$ a geodesic
distance estimate along the FMM of sequence $k$. The first term is
easily understood as a penalty for dissimilarity of matched
pose-signatures. The latter is a regularizing term that enforces a
\emph{globally} good matching by favouring mappings that map
geodesically close points from the domain manifold to geodesically
close points on the codomain manifold.

\begin{figure}
  \centering
  \includegraphics[width=0.75\textwidth]{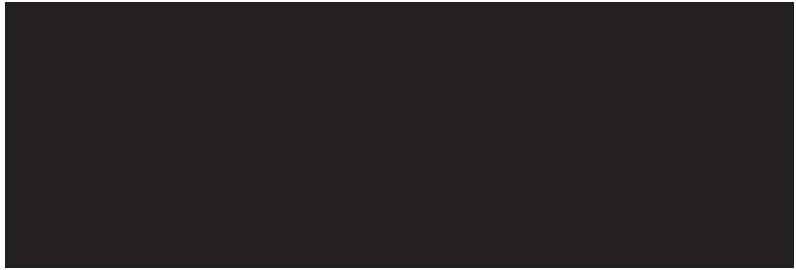}
  \caption[Pose matching.]{ \it Manifold-to-manifold pose matching: geodesic distances
            between neighbouring faces on the domain manifold and the corresponding
            faces on the codomain manifold are used to regularize the solution. }
            \label{fig:matching}
\end{figure}

\begin{figure}
  \centering
  \small
  \subfigure[Original]
    {\includegraphics[width=0.55\textwidth]{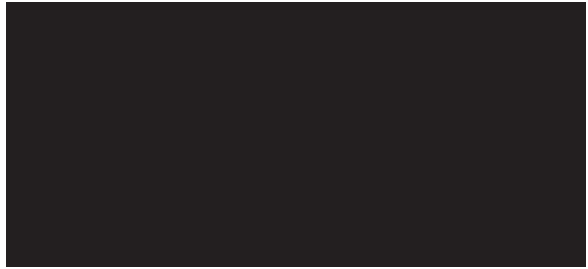}}
  \subfigure[Reilluminated]
    {\includegraphics[width=0.55\textwidth]{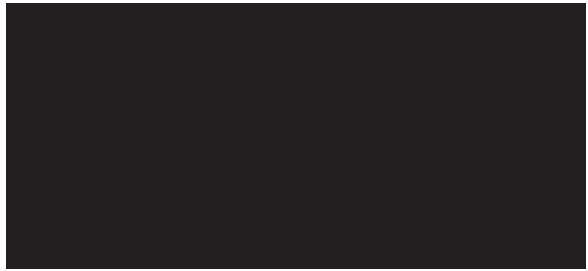}}

  \caption[Sequence reillumination example.]
       {\it (a) Original images from a novel video sequence
            and (b) the result of reillumination using the proposed genetic
            algorithm with nearest neighbour-based reconstruction. }
            \label{Fig: Correspondences}
\end{figure}

\paragraph*{Regularization.} The manifold-oriented nature of the regularizing
function $f_{reg}(c)$ in \eqref{Eqn: Fitness Function} has
significant advantages over alternatives that use some form temporal
smoothing. Firstly, it is unaffected by changes on the motion
pattern of the user (i.e.\ sequential ordering of $\{ \mathbf{X}_i
\}^{(j)}$). On top of the inherent benefit (a person's motion should
not affect recognition), this is important for several practical
reasons, e.g.
\begin{itemize}
  \item face images need not originate from a single sequence - multiple sequences are
        easily combined together by computing the union of their
        frame sets, and

  \item regularization works even if there are bursts of missed or incorrect face
        detections (see Section~\ref{Sec: Empirical Evaluation}).
\end{itemize}

To understand the form of the regularizing function note that the
mapping function $c$ only affects the numerator of each summation
term in $f_{reg}(c)$. Its effect is then to penalize cases in which
neighbouring faces of the domain manifold map to geodesically
distant faces on the codomain manifold. The penalty is further
weighted by the inverse of the original geodesic distance {\small
$d_G^{(1)}(\textbf{X}_j^{(1)}, \textbf{X}_{n(j, k)}^{(1)}; \{
\textbf{X}_j \}^{(1)})$} to place more emphasis on local pose
agreement.

\paragraph{Pose-matching function.} The performance of
function $d_E$ in \eqref{Eqn: Fitness Function} at estimating the
goodness of a frame match is crucial for making the overall
optimization scheme work well. Our approach consists of filtering
the original face image to produce a quasi illumination-invariant
\emph{pose-signature}, which is then compared with other
pose-signatures using the Euclidean distance:
\begin{align}
  d_E\left(\textbf{X}_j^{(1)}, \textbf{X}_{c(j)}^{(2)} \right) = \left\|
  \textbf{X}_j^{(1)} -
  \textbf{X}_{c(j)}^{(2)} \right\|_2
\end{align}
Note that the signatures are \emph{only} used for frame matching and
thus need not retain any power of discrimination between individuals
-- all that is needed is sufficient pose information. We use a
distance-transformed edge map of the face image as a pose-signature,
motivated by the success of this representation in
object-configuration matching across other computer vision
applications, e.g.\ \cite{Gavr2000,StenThayTorrCipo2003}.

\paragraph{Minimizing the fitness function.} Exact minimization of
the fitness function \eqref{Eqn: Fitness Function} over all
functions $c$ is an NP-complete problem. However, since the final
synthesis of novel faces (Stage 2) involves an entire geodesic
neighbouring of the paired faces, it is inherently robust to some
non-optimality of this matching. Therefore, in practice, it is
sufficient to find a good match, not necessarily the optimal one.

\begin{figure*}[t]
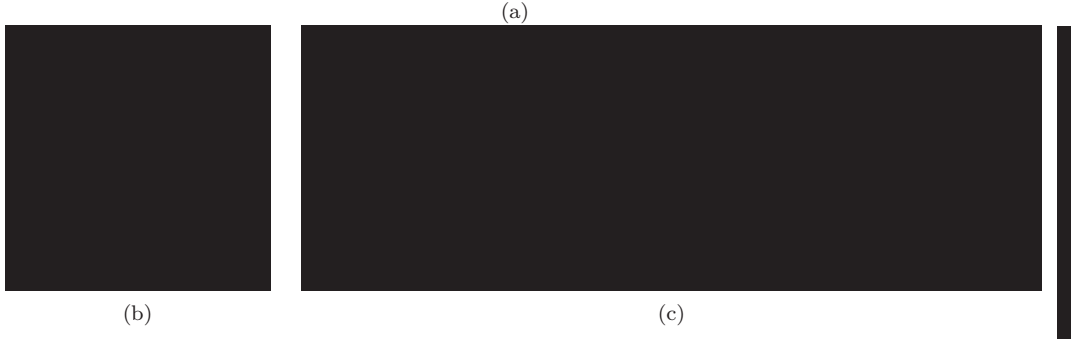

  \centering
  \Large
  \begin{tabular*}{1.00\textwidth}{@{\extracolsep{\fill}}l|cccccc}
    \Hline \vspace{-5pt}
    \normalsize \textbf{~Property} & \normalsize Population & \normalsize Elite        & \normalsize Mutation & \normalsize Migration & \normalsize Crossover & \normalsize Max. \\
                                   & \normalsize size       & \normalsize survival no. & \normalsize (\%)     & \normalsize (\%)      & \normalsize (\%)      & \normalsize generations \\
    \hline
    \normalsize \textbf{~Value}    & \small 20              & \small 2                 & \small 5             & \small 20             & \small 80             & \small 200\\
    \Hline
  \end{tabular*}\\ \vspace{10pt}
  {\footnotesize (a)}\\

  \begin{tabular}{VV}
   \includegraphics[width=100pt]{ICCV2005_chromosome.eps} &
  \includegraphics[width=280pt]{ICCV2005_fitness.eps}\\
  {\footnotesize (b)} & {\footnotesize (c)}\\
  \end{tabular}

  \caption[Pose matching as genetic algorithm-based optimization.]{
            \it
            (a) The parameters of the proposed GA optimization,
            (b) the corresponding chromosome structure and (c)
            the population fitness (see~\eqref{Eqn: Fitness Function})
            in a typical evolution. Maximal generation count of 200 was
            chosen as a trade-off between accuracy
            and matching speed. }
  \label{Fig: GA Parameters}
  \vspace{6pt}\hrule
\end{figure*}

We propose to use a genetic algorithm (GA) \cite{DudaHartStor2001}
as a particularly suitable approach to minimization for our problem.
GAs rely on the property of many optimization problems that
sub-solutions of good solutions are good themselves. Specifically,
this means that if we have a globally good manifold match, then
local matching can be expected to be good too. Hence, combining two
good matches is a reasonable attempt at improving the solution. This
motivates the chromosome structure we use, depicted in
Figure~\ref{Fig: GA Parameters}~(a), with the $i$-th gene in a
chromosome being the value of $c(i)$. GA parameters were determined
experimentally from a small training set and are summarized in
Figure~\ref{Fig: GA Parameters}~(b,c).


\paragraph{Estimating geodesic distances.}
The definition of the fitness function in \eqref{Eqn: Fitness
Function} involves estimates of geodesic distances along manifolds.
Due to the nonlinearity of FMMs
\cite{AranShakFish+2005,LeeHoYangKrie2003} it is not well
approximated by the Euclidean distance. We estimate the geodesic
distance between every two faces from a manifold using the Floyd's
algorithm \cite{CormLeisRive1990} on a constructed undirected graph
whose nodes correspond to face images (also see
\cite{TeneSilvLang2000}). Then, if $\textbf{X}_i$ is one of the $K$
nearest neighbours of $\textbf{X}_j$\footnote{Note that the converse
does not hold as $\textbf{X}_i$ being one of the $K$ nearest
neighbours of $\textbf{X}_j$ does not imply that $\textbf{X}_j$ is
one of the $K$ nearest neighbours of $\textbf{X}_i$. Therefore the
edge relation of this graph is a superset of the ``in $K$-nearest
neighbours'' relation on $\textbf{X}$s.}:
\begin{align}
  \label{Eqn: Floyd 1}
  d_G(\textbf{X}_i, \textbf{X}_j) = \left \| \textbf{X}_i -
  \textbf{X}_j \right \|_2.
\end{align}
Otherwise: \begin{align}
  d_G(\textbf{X}_i, \textbf{X}_j) = \min_k  \left[ d_G(\textbf{X}_i,
  \textbf{X}_k) + d_G(\textbf{X}_k,
  \textbf{X}_j) \right].
\end{align}

\subsection{Stage 2: fine reillumination} 
\begin{figure}
  \centering
  \includegraphics[width=0.9\textwidth]{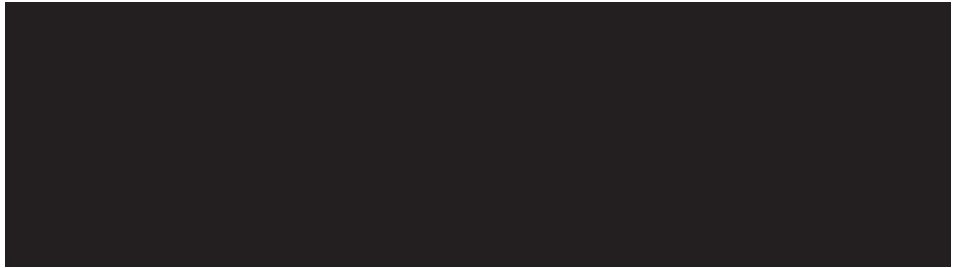}
  \caption[Appearance and pose-signature manifold structures.]{ \it A face motion manifold in the input image space and the corresponding
            pose-signature manifold (both shown in their respective 3D principal subspaces).
            Much like the original appearance manifold, the latter is continuous and smooth,
            as ensured by distance transforming the face edge maps. While not necessarily similar
            globally, the two manifolds retain the same \emph{local} structure, which is
            crucial for the proposed fine illumination algorithm. }
            \label{Fig: Signature Manifolds}
  \vspace{6pt}\hrule
\end{figure}

\begin{figure}
  \centering
  \includegraphics[width=0.7\textwidth]{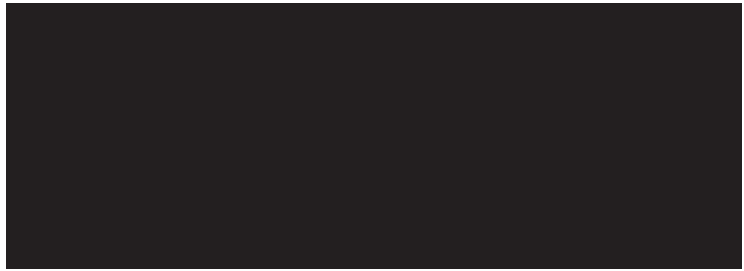}
  \caption[Computing fine reillumination coefficients.]{ \it Face reillumination: the coefficients for
            linearly combining face appearance images (bottom row)
            are computed using the corresponding pose-signatures
            (top row). Also see Figure~\ref{Fig: Signature Manifolds}. }
            \label{Fig: Reillumination Concept}
  \vspace{6pt}\hrule
\end{figure}

Having computed a
pose-matching function $c^*$, we turn to the problem of
reilluminating frames $\textbf{X}_i^{(1)}$. We exploit the
smoothness of pose-signature manifolds (which was ensured by
distance-transforming face edge maps), illustrated in
Figure~\ref{Fig: Signature Manifolds}, by computing
$\textbf{Y}_i^{(1)}$, the reilluminated frame $\textbf{X}_i^{(1)}$,
as a linear combination of $K$ nearest-neighbour frames of
$\textbf{X}_{c^*(i)}^{(2)}$. Linear combining coefficients
$\alpha_1, \dots \alpha_K$ are found from the corresponding
pose-signatures by solving the following constrained minimization
problem:
\begin{align}
    &\{ \alpha_j \} = \arg \min_{\{ \alpha_j \}}
    \left\| \textbf{x}_i^{(1)} - \sum_{k=1}^K \alpha_k
    \textbf{x}_{n(c^*(i), k)}^{(2)} \right\|_2
    \label{Eqn: Reconstruction}
\end{align}
subject to $\sum_{k=1}^K \alpha_k = 1.0$, where $\mathbf{x}_i^{(j)}$
is the pose-signature corresponding to $\mathbf{X}_i^{(j)}$. In
other words, the pose-signature of a novel face is first
reconstructed using the pose-signatures of $K$ training faces (in
the target illumination), which are then combined in the same
fashion to synthesize a reilluminated face, see Figure~\ref{Fig:
Correspondences} and~\ref{Fig: Reillumination Concept}. We restrict
the set of frames used for reillumination to the $K$-nearest
neighbours for two reasons. Firstly, the computational time of using
all faces would make this highly unpractical. Secondly, the
nonlinearity of both face appearance manifolds and pose-signature
manifolds, demands that only the faces in the local, Euclidean-like
neighbourhood are used.

Optimization of \eqref{Eqn: Reconstruction} is readily performed by
differentiation giving:
\begin{align}
    \begin{bmatrix}
      \alpha_2 \\
      \alpha_3 \\
      \vdots \\
      \alpha_K \\
    \end{bmatrix}
     = \mathbf{R}^{-1} \mathbf{t},
    \label{Eqn: Reconstruction 1}
\end{align}
where:
\begin{align}
R(j,k) &= \left(\textbf{x}_{n(c^*(i), 1)}^{(2)} -
\textbf{x}_{n(c^*(i), j)}^{(2)}\right) \cdot
\left(\textbf{x}_{n(c^*(i), 1)}^{(2)} - \textbf{x}_{n(c^*(i),
k)}^{(2)}\right),\\
t(j) &= \left(\textbf{x}_{n(c^*(i), 1)}^{(2)} -
\textbf{x}_{n(c^*(i), j)}^{(2)}\right) \cdot
\left(\textbf{x}_{n(c^*(i), 1)}^{(2)} - \textbf{x}_i^{(1)}\right).
\end{align}

\section{The shape-illumination manifold}\label{Sec: SIM}
In most practical applications, specularities, multiple or non-point
light sources significantly affect the appearance of faces. We
believe that the difficulty of dealing with these effects is one of
the main reasons for poor performance of most face recognition
systems when put to use in a realistic environment. In this work we
make a very weak assumption on the process of image formation: the
only assumption made is that the intensity of each pixel is a linear
function of the albedo $a(j)$ of the corresponding 3D point:
\begin{equation}
  X(j) = a(j) \cdot s(j)
  \label{Eqn: Image Formation Model}
\end{equation}
where $\mathbf{s}$ is a function of illumination, shape and other
parameters not modelled explicitly. This is similar to the
reflectance-lighting model used in Retinex-based algorithms
\cite{KimmEladShak+2003}, the main difference being that we make no
further assumptions on the functional form of $\mathbf{s}$. Note
that the commonly-used (e.g.\ see
\cite{BlanVett2003,GeorKrieBelh1998,RiklShas2001}) Lambertian
reflectance model is a special case of \eqref{Eqn: Image Formation
Model} \cite{BelhKrie1998}:
\begin{equation}
  s(j) = \sum_i \max(\mathbf{n}_j \cdot \mathbf{L}_i, 0)
\end{equation}
where $\mathbf{n}_i$ is the corresponding surface normal and
$\{\mathbf{L}_i\}$ the intensity-scaled illumination directions at
the point.

The image formation model introduced in \eqref{Eqn: Image Formation
Model} leaves the image pixel intensity as an unspecified function
of face shape or illumination parameters. Instead of formulating a
complex model of the geometry and photometry behind this function
(and then needing to recover a large number of model parameters), we
propose to learn it implicitly. Consider two images, $\textbf{X}_1$
and $\textbf{X}_2$ of the same person, in the same pose, but
different illuminations. Then from~\eqref{Eqn: Image Formation
Model}:
\begin{equation}
  \Delta \log X(j) = \log s_2(j) - \log s_1(j) \equiv d_s(j)
  \label{Eqn: SIM Sample}
\end{equation}
In other words, the difference between these logarithm-transformed
images is not a function of face albedo. As before, due to the
smoothness of faces, as the pose of the subject varies the
difference-of-logs vector $\mathbf{d}_s$ describes a manifold in the
corresponding embedding vector space. These is the
Shape-Illumination manifold (SIM) corresponding to a particular pair
of video sequences, refer back to Figure~\ref{Fig: Manifolds}~(b).

\paragraph{The generic SIM.}
A crucial assumption of our work is that the Shape-Illumination
Manifold of all possible illuminations and head poses is
\emph{generic for human faces} (gSIM). This is motivated by a number
of independent results reported in the literature that have shown
face shape to be less discriminating than albedo across different
models \cite{CrawCostKatoAkam1999,GrosMattBake2004a} or have
reported good results in synthetic reillumination of faces using the
constant-shape assumption \cite{RiklShas2001}. In the context of
face manifolds this means that the effects of \emph{illumination and
shape} can be learnt offline from a training corpus containing
typical modes of pose and illumination variation.

It is worth emphasizing the key difference in the proposed offline
learning from previous approaches in the literature which try to
learn the \emph{albedo} of human faces. Since offline training is
performed on persons not in the online gallery, in the case when
albedo is learnt it is necessary to have means of generalization
i.e.\ learning what \emph{possible} albedos human faces can have
from a small subset. In \cite{RiklShas2001}, for example, the
authors demonstrate generalization to albedos in the rational span
of those in the offline training set. This approach is not only
unintuitive, but also without a meaningful theoretical
justification. On the other hand, previous research indicates that
illumination effects can be learnt \emph{directly} without the need
for generalization \cite{AranShakFish+2005}.

\paragraph{Training data organization.} The proposed
method consists of two training stages -- a one-time offline
learning performed using \emph{offline training data} and a stage
when \emph{gallery data} of known individuals with associated
identities is collected. The former (explained next) is used for
learning the generic face shape contribution to face appearance
under varying illumination, while the latter is used for
subject-specific learning.

\subsection{Offline stage: learning the generic SIM (gSIM)}\label{SubSec:
Offline - Learning the Global SIM} Let $\mathbf{X}_i^{(j, k)}$ be
the $i$-th face of the $j$-th person in the $k$-th illumination,
same indexes corresponding in pose, as ensured by the proposed
reillumination algorithm in Section~\ref{SubSec: Synthetic
Reillumination of Face Motion Sequences}. Then from \eqref{Eqn: SIM
Sample}, samples from the generic Shape-Illumination manifold can be
computed by logarithm-transforming all images and subtracting those
corresponding in identity and pose: \begin{align}
  \mathbf{d} = \log \mathbf{X}_i^{(j, p)} - \log \mathbf{X}_i^{(j, q)}
  \label{Eqn: SIM samples}
\end{align}
Provided that training data contains typical variations in pose and
illumination (i.e. that the p.d.f. confined to the generic SIM is
well sampled), this becomes a standard statistical problem of
high-dimensional density estimation. We employ the Gaussian Mixture
Model (GMM). In the proposed framework, this representation is
motivated by: (i) the assumed low-dimensional manifold model
\eqref{eq:manifold_pdf}, (ii) its compactness and (iii) the
existence of incremental model parameter estimation algorithms
(e.g.\ \cite{AranCipo2005a,HallMarsMart2000}).

Briefly, we estimate multivariate Gaussian components using the
Expectation Maximization (EM) algorithm \cite{DudaHartStor2001},
initialized by $k$-means clustering. Automatic model order selection
is performed using the well-known Minimum Description Length
criterion \cite{DudaHartStor2001} while the principal subspace
dimensionality of PPCA components was estimated from eigenspectra of
covariance matrices of a diagonal GMM fit, performed first. Fitting
was then repeated using a PPCA mixture. From $6123$ gSIM samples
computed from 100 video sequences, we obtained 12 mixture
components, each with a 6D principal subspace. Figure~\ref{Fig: SIM
Samples} shows an example of subtle illumination effects learnt with
this model.

\begin{figure}
  \centering
  \small
  \includegraphics[width=0.6\textwidth]{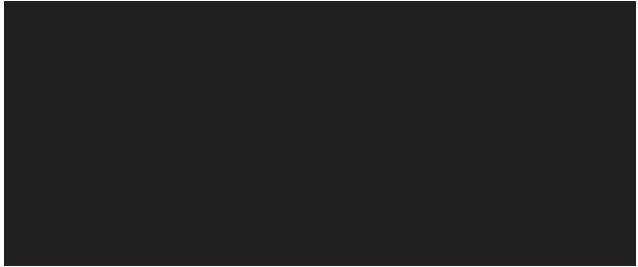}\\
  \caption[Generic SIM: example of learnt illumination effects.]{ \it
            Learning complex illumination effects:
            Shown is the variation along the 1st mode of a
            single PPCA space in our SIM mixture model. Cast
            shadows (e.g.\ from the nose) and the locations of
            specularities (on the nose and above the eyes)
            are learnt as the illumination source moves from
            directly overhead to side-overhead. }
            \label{Fig: SIM Samples}
  \vspace{6pt}\hrule
\end{figure}

\section{Novel sequence classification}\label{SubSec: Robust
Likelihood for Novel Sequence Classification} The discussion so far
has concentrated on offline training and building an illumination
model for faces - the Generic Shape-Illumination manifold. Central
to the proposed algorithm was a method for reilluminating a face
motion sequence of a person with another sequence of the \emph{same}
person (see Section~\ref{SubSec: Synthetic Reillumination of Face
Motion Sequences}). We now show how the same method can be used to
compute a similarity between two unknown individuals, given a single
training sequence for each and the Generic SIM.

Let gallery data consist of sequences $\{ \mathbf{X}_i
\}^{(1)},\ldots,\{ \mathbf{X}_i \}^{(N)}$, corresponding to $N$
individuals, $\{ \mathbf{X}_i \}^{(0)}$ be a novel sequence of one
of these individuals and $\mathcal{G}\left(\mathbf{x};
\mathbf{\Theta}\right)$ a Mixture of Probabilistic PCA corresponding
to the generic SIM. Using the reillumination algorithm of
Section~\ref{SubSec: Synthetic Reillumination of Face Motion
Sequences}, the novel sequence can be reilluminated with each $\{
\mathbf{X}_i \}^{(j)}$ from the gallery, producing samples $\{
\mathbf{d}_i \}^{(j)}$. We assume these to be identically and
independently distributed according to a density corresponding to a
\emph{postulated} subject-specific SIM. We then compute the
probability of these under $\mathcal{G}\left(\mathbf{x};
\mathbf{\Theta}\right)$:
\begin{equation}
    p_i^{(j)} = \mathcal{G}\left(\mathbf{d}_i^{(j)}; \mathbf{\Theta}\right)
    \label{Eqn: SIM Likelihoods}
\end{equation}

\begin{figure}
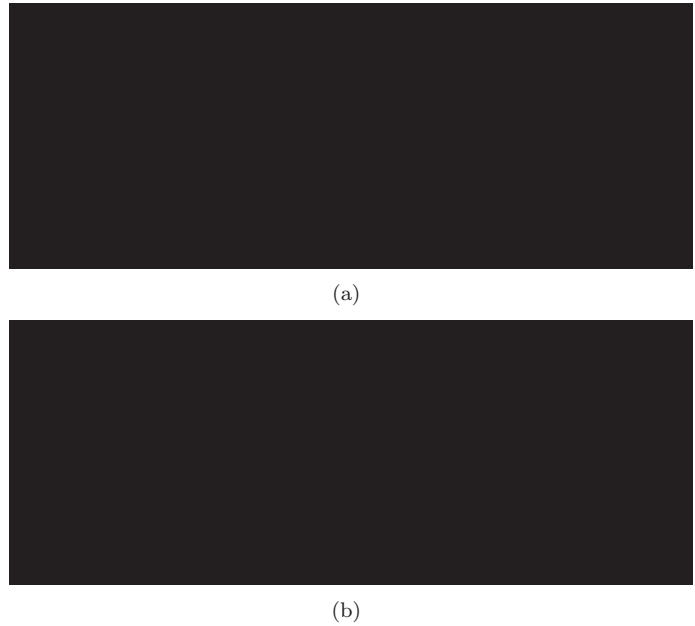

  \centering
  \subfigure[]{\includegraphics[width=0.65\textwidth]{PAMI2006_diff1.eps}}
  \subfigure[]{\includegraphics[width=0.65\textwidth]{PAMI2006_diff2.eps}}
  \caption[Inter-personal ``reillumination'' results.]{ \it An example of ``reillumination'' results when the two compared sequences
            do not correspond to the same individual: the target sequence is shown on the left,
            the output of our algorithm on the right. Most of the frames do not contain faces
            which correspond in pose. }
            \label{Fig: Different Reillumination}
  \vspace{6pt}\hrule
\end{figure}

When $\{ \mathbf{X}_i \}^{(0)}$ and $\{ \mathbf{X}_i \}^{(j)}$
correspond in identity, from the way the Generic SIM is learnt, it
can be seen that the probabilities $p_i^{(j)}$ will be large. The
more interesting question arises when the two compared sequences do
not correspond to the same person. In this case, the reillumination
algorithm will typically fail to produce a meaningful result - the
output frames will not correspond in pose to the target sequence,
see Figure~\ref{Fig: Different Reillumination}. Consequently, the
observed appearance difference will have a low probability under the
hypothesis that it is caused purely by an illumination change. A
similar result is obtained if the two individuals share sufficiently
similar facial lines and poses are correctly matched. In this case
it is the differences in face surface albedo that are not explained
well by the Generic SIM, producing low $p_i^{(j)}$ in \eqref{Eqn:
SIM Likelihoods}.

\begin{figure}
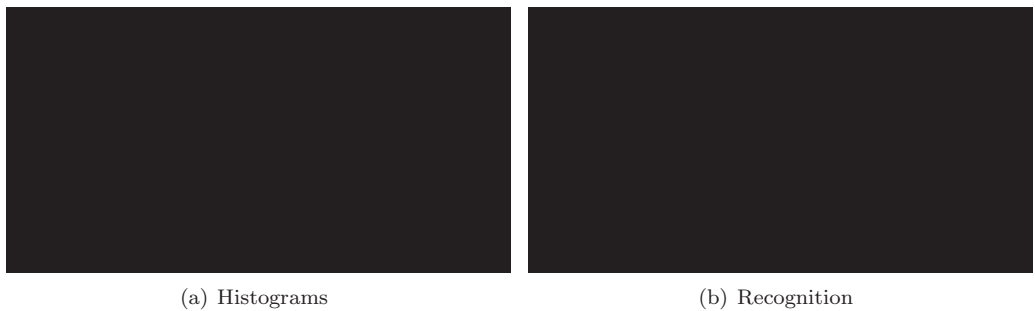

  \centering
  \subfigure[Histograms]{
    \includegraphics[width=0.48\textwidth]{ECCV2005_likelihood_hists.eps}}
  \subfigure[Recognition]{
    \includegraphics[width=0.48\textwidth]{ECCV2006_reliable_rr.eps}}

  \caption[Robust likelihood.]{ \it (a) Histograms of intra-personal likelihoods
            across frames of a sequence when two sequences compared correspond
            to the same (red) and different (blue) people. (b) Recognition rate
            as a function of the number of frames deemed `reliable'. }
            \label{Fig: Likelihood Histograms}
  \vspace{6pt}\hrule
\end{figure}

\paragraph{Varying pose and robust likelihood.}
Instead of basing the classification of $\{ \mathbf{X}_i \}^{(0)}$
on the likelihood of observing the \emph{entire} set $\{
\mathbf{d}_i \}^{(j)}$ in \eqref{Eqn: SIM Likelihoods}, we propose a
more robust measure. To appreciate the need for robustness, consider
the histograms in Figure~\ref{Fig: Likelihood Histograms}~(a). It
can be observed that the likelihood of the most similar faces in an
inter-personal comparison, in terms of \eqref{Eqn: SIM Likelihoods},
approaches that of the most \emph{dissimilar} faces in an
\emph{intra-personal} comparison (sometimes even exceeding it). This
occurs when the correct gallery sequence contains poses that are
very dissimilar to even the most similar ones in the novel sequence,
or vice versa (note that small dissimilarities are extrapolated well
from local manifold structure using \eqref{Eqn: Reconstruction}). In
our method, the robustness to these, unseen modes of pose variation
is achieved by considering the mean log-likelihood of only the most
likely faces. In our experiments we used the top 15\% of the faces,
but we found the algorithm to exhibit little sensitivity to the
exact choice of this number, see Figure~\ref{Fig: Likelihood
Histograms}~(b). A summary of the proposed algorithms is shown in
Figure~\ref{Fig: Alg 1} and~\ref{Fig: Alg 2}.

\begin{figure}
    \centering
    \begin{tabular}{l}
          \begin{tabular}{ll}
              \textbf{Input}:  & database of sequences $\{\mathbf{X}_i \}^{(j)}$.\\
              \textbf{Output}: & model of gSIM $\mathcal{G}\left(\mathbf{d}; \mathbf{\Theta}\right)$.\\
          \end{tabular}\vspace{5pt}
          \\ \hline \\
          \begin{tabular}{l}
            \textbf{1: gSIM iteration}\\
            \hspace{10pt}\textbf{for} all persons $i$ and illuminations $j$, $k$\\\\

            \hspace{15pt}\textbf{2: Reilluminate using $\{ \mathbf{X}_i \}^{(k)}$}\\
            \hspace{25pt}$\{ \mathbf{Y}_i \}^{(j)}$ = reilluminate$\left(\{ \mathbf{X}_i \}^{(j)}; \{ \mathbf{X}_i \}^{(k)}\right)$\\\\

            \hspace{15pt}\textbf{3: Add gSIM samples}\\
            \hspace{25pt}$\mathbb{D} = \mathbb{D}~\bigcup~\left\{ \mathbf{Y}_i^{(j)} - \mathbf{X}_i^{(j)} : i=1\dots \right\}$\\\\

             \textbf{4: Fit GMM $\mathcal{G}$ from gSIM samples}\\
             \hspace{10pt}$\mathcal{G}\left(\mathbf{d}; \mathbf{\Theta}\right) = $EM\_GMM$(\mathbb{D})$\\\\
          \end{tabular}\\\hline
    \end{tabular}
    \caption[Generic SIM model learning algorithm.]{ \it A summary of the proposed offline learning algorithm. Illumination effects on the appearance of faces
                  are learnt as a probability density, in the proposed method approximated with a Gaussian mixture
                  $\mathcal{G}\left(\mathbf{d}; \mathbf{\Theta}\right)$. }
    \label{Fig: Alg 1}
\end{figure}

\begin{figure}
    \centering
    \begin{tabular}{l}
          \begin{tabular}{ll}
              \textbf{Input}:  & sequences $\{ \mathbf{X}_i \}^{(G)}$, $\{ \mathbf{X}_i \}^{(N)}$.\\
              \textbf{Output}: & same-identity likelihood $\rho$.\\
          \end{tabular}\vspace{5pt}
          \\ \hline \\
          \begin{tabular}{l}
            \textbf{1: Reilluminate using $\{ \mathbf{X}_i \}^{(G)}$}\\
            \hspace{10pt}$\{ \mathbf{Y}_i \}^{(N)}$ = reilluminate$\left(\{ \mathbf{X}_i \}^{(N)}; \{ \mathbf{X}_i \}^{(G)}\right)$\\\\

            \textbf{2: Postulate SIM samples}\\
            \hspace{10pt}$\mathbf{d}_i = \log \mathbf{X}_i^{(N)} - \log \mathbf{Y}_i^{(N)}$\\\\

            \textbf{3: Compute likelihoods of $\{ \mathbf{d}_i\}$}\\
            \hspace{10pt}$p_i = \mathcal{G}\left(\mathbf{d}_i; \mathbf{\Theta}\right)$\\\\

            \textbf{4: Order $\{ \mathbf{d}_i\}$ by likelihood}\\
            \hspace{10pt}$p_{s(1)} \geq \dots \geq p_{s(N)} \geq \dots$\\\\

            \textbf{5: Inter-manifold similarity $\rho$}\\
            \hspace{10pt}\hspace{25pt} $\rho = \sum_{i=1}^N \log p_{s(i)} / N$\\\\
          \end{tabular}\\\hline
    \end{tabular}
    \caption[Recognition algorithm based on the learnt Generic SIM model.]{ \it A summary of the proposed online recognition algorithm. }
    \vspace{6pt}\hrule
    \label{Fig: Alg 2}
\end{figure}

\section{Empirical evaluation}\label{Sec: Empirical Evaluation}
We compared the performance of our recognition algorithm with and
without the robust likelihood of Section~\ref{SubSec: Robust
Likelihood for Novel Sequence Classification} (i.e.\ using only the
most reliable vs.\ all detected and reilluminated faces) on
\textit{CamFace}, \textit{ToshFace} and \textit{Face Video} data
sets to that of:
\begin{itemize}
    \item State-of-the-art commercial system FaceIt$^{\circledR}$ by
        Identix \cite{Iden} (the best performing software in the most
        recent Face Recognition Vendor Test
        \cite{PhilGrotMichBlac+2003}),

    \item Constrained MSM (CMSM) \cite{FukuYama2003} used in Toshiba's state-of-the-art
        commercial system FacePass$^{\circledR}$
        \cite{Tosh}\footnote{The algorithm was reimplemented through consultation with the authors.},

    \item Mutual Subspace Method (MSM) \cite{FukuYama2003,MaedYamaFuku2004}\footnotemark[\value{footnote}],

    \item Kernel Principal Angles (KPA) of Wolf and Shashua \cite{WolfShas2003}\footnote{We used the original authors' implementation.}, and

    \item KL divergence-based algorithm of Shakhnarovich \textit{et al.} (KLD)
        \cite{ShakFishDarr2002}\footnotemark[\value{footnote}].\\
\end{itemize}

In all tests, both training data for each person in the gallery, as
well as test data, consisted of only a single sequence. Offline
training of the proposed algorithm was performed using 20
individuals in 5 illuminations from the \textit{CamFace} data set --
we emphasize that these were not used as test input for the
evaluations reported in this section. The methods were evaluated
using 3 face representations:
\begin{itemize}
    \item raw appearance images $\mathbf{X}$,

    \item Gaussian high-pass filtered images -- used for face recognition in
         \cite{AranZiss2005,FitzZiss2002}:
        \begin{equation}
          \mathbf{X}_H = \mathbf{X} - (\mathbf{X} \ast \mathbf{G}_{\sigma =
          1.5}),
      \end{equation}
    \item local intensity-normalized high-pass filtered images -- similar to the Self Quotient Image \cite{WangLiWang2004} (also
    see \cite{AranCipo2006a}):
      \begin{equation}
        \mathbf{X}_Q = \mathbf{X}_H ./ (\mathbf{X} - \mathbf{X}_H),
      \end{equation}
      the division being element-wise.
\end{itemize}

\begin{figure}
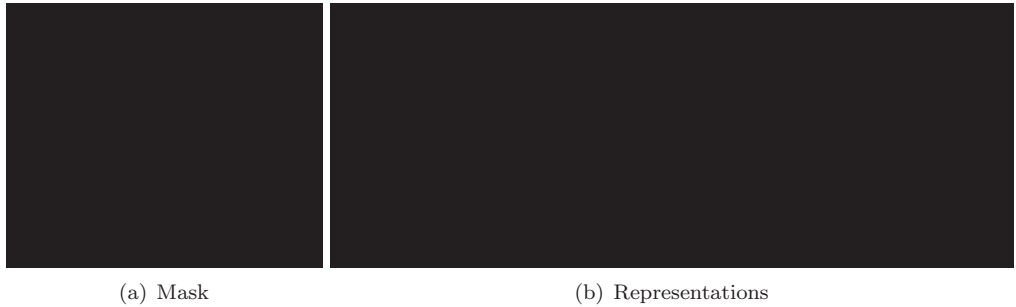

  \centering
  \subfigure[Mask]{\includegraphics[width=.3\textwidth]{feather_mask_bw.eps}}
  \subfigure[Representations]{\includegraphics[width=.65\textwidth]{PAMI2006_representations.eps}}
  \caption[Face representations used in evaluation.]{ \it (a) The weighting mask used to suppress background clutter. (b) The three
            face representations used in evaluation, shown as images, before (top row) and after
            (bottom row) the weighting mask was applied. }
            \label{Fig: Mask}
  \vspace{6pt}\hrule
\end{figure}

Background clutter was suppressed using a weighting mask
$\mathbf{M}_F$, produced by feathering the mean face outline
$\mathbf{M}$ in a manner similar to \cite{AranZiss2005} and as shown
in Figure~\ref{Fig: Mask}:
\begin{align}\label{Eqn: Feathering}
    &\mathbf{M}_F = \mathbf{M} \ast \exp {-\left\{ \frac {r^2(x, y)} {8} \right\}}
\end{align}

\subsection{Results}\label{gSIM: SubSec: Results} A summary of
experimental results is shown in Table~\ref{gSIM: Tab: Results}. The
proposed algorithm greatly outperformed other methods, achieving a
nearly perfect recognition (99.3+\%) on all 3 databases. This is an
extremely high recognition rate for such unconstrained conditions
(see Figure~\ref{Fig: Faces Input}), small amount of training data
per gallery individual and the degree of illumination, pose and
motion pattern variation between different sequences. This is
witnessed by the performance of Simple KLD method which can be
considered a proxy for gauging the difficulty of the task, seeing
that it is expected to perform well if imaging conditions are not
greatly different between training and test \cite{ShakFishDarr2002}.
Additionally, it is important to note the excellent performance of
our algorithm on the Japanese database, even though offline training
was performed using Caucasian individuals only.

\begin{table}
  \centering
  \Large
  \caption[Recognition results.]
   {\it Average recognition rates (\%) and their standard deviations (if applicable).\vspace{10pt} }
           \label{gSIM: Tab: Results}

    \begin{tabular*}{1.00\textwidth}{@{\extracolsep{\fill}}l|lllllll}
      \Hline
                           & \normalsize \textbf{gSIM, rob.} & \normalsize gSIM     & \normalsize FaceIt      & \normalsize CMSM      & \normalsize KPA       & \normalsize MSM       & \normalsize KLD         \\ 
      \Hline
      {\scriptsize\textit{CamFace}}        &&&&&&&\\
      ~~~\normalsize $\mathbf{X}$      & \small \textbf{99.7/0.8}     & \small 97.7/2.3 & \small 64.1/9.2  & \small 73.6/22.5 & \small 63.1/21.2 & \small 58.3/24.3 & \small 17.0/8.8  \\  
      ~~~\normalsize $\mathbf{X}_H$    & \small --                    & \small --       & \small --        & \small 85.0/12.0 & \small 83.1/14.0 & \small 82.8/14.3 & \small 35.4/14.2 \\  
      ~~~\normalsize $\mathbf{X}_Q$    & \small --                    & \small --       & \small --        & \small 87.0/11.4 & \small 87.1/9.0 & \small 83.4/8.4  & \small 42.8/16.8 \\  
      \hline
      {\scriptsize\textit{ToshFace}}         &&&&&&&\\
      ~~~\normalsize $\mathbf{X}$      & \small \textbf{99.9/0.5}     & \small 96.7/5.5 & \small 81.8/9.6  & \small 79.3/18.6 & \small 49.3/25.0 & \small 46.6/28.3 & \small 23.0/15.7 \\  
      ~~~\normalsize $\mathbf{X}_H$    & \small --                    & \small --       & \small --        & \small 83.2/17.1 & \small 61.0/18.9 & \small 56.5/20.2 & \small 30.5/13.3 \\  
      ~~~\normalsize $\mathbf{X}_Q$    & \small --                    & \small --       & \small --        & \small 91.1/8.3  & \small 87.7/11.2 & \small 83.3/10.8 & \small 39.7/15.7 \\  
      \hline
      {\scriptsize\textit{Face Video}}      &&&&&&&\\
      ~~~\normalsize $\mathbf{X}$      & \small \textbf{100.0}        & \small 91.9    & \small 91.9       & \small 91.9      & \small 91.9 & \small 81.8      & \small 59.1      \\  
      ~~~\normalsize $\mathbf{X}_H$    & \small --                    & \small --      & \small --         & \small 100.0     & \small 91.9 & \small 81.8      & \small 63.6      \\  
      ~~~\normalsize $\mathbf{X}_Q$    & \small --                    & \small --      & \small --         & \small 91.9      & \small 91.9 & \small 81.8      & \small 63.6      \\  
      \Hline
    \end{tabular*}
\end{table}

\begin{figure}
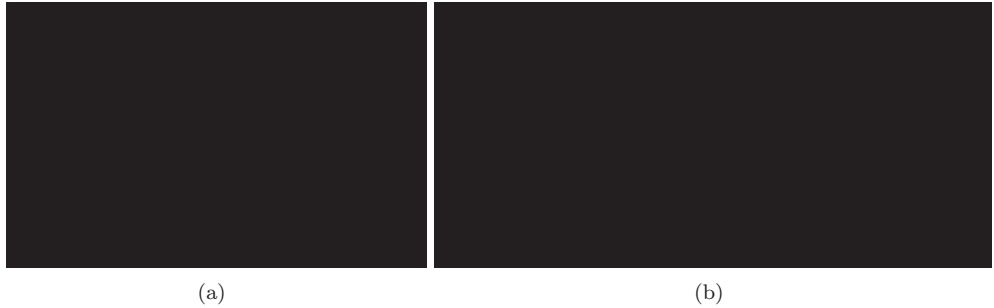

  \centering
  \subfigure[]{\includegraphics[width=.4\textwidth]{PAMI2006_roc.eps}}
  \subfigure[]{\includegraphics[width=.53\textwidth]{PAMI2006_reco_frames.eps}}
  \caption[Generic SIM ROC curves.]{ \it (a) The Receiver-Operator Characteristic (ROC) curves of
            the gSIM method, with and without the robust likelihood proposed in
            Section~\ref{SubSec: Robust Likelihood for Novel Sequence Classification}
            estimated from CamFace and ToshFace data sets. }
            \label{gSIM: Fig: ROC}
  \vspace{6pt}\hrule
\end{figure}

As expected, when plain likelihood was used instead of the robust
version proposed in Section~\ref{SubSec: Robust Likelihood for Novel
Sequence Classification}, the recognition rate was lower, but still
significantly higher than that of other methods. The high
performance of non-robust gSIM is important as an estimate of the
expected recognition rate in the ``still-to-video'' scenario of the
proposed method. We conclude that our algorithm's performance seems
very promising in this setup as well. An inspection of the
Receiver-Operator Characteristics Figure~\ref{gSIM: Fig: ROC}~(a) of the
two methods shows an ever more drastic improvement. This is an
insightful observation: it shows that the use of the proposed robust
likelihood yields less variation in the estimated similarity between
individuals across different sequences.

Finally, note that the standard deviation of our algorithm's
performance across different training and test illuminations is much
lower than that of other methods, showing less dependency on the
exact imaging conditions used for data acquisition.

\paragraph{Representations.} Both the high-pass and
even further Self Quotient Image representations produced an
improvement in recognition for all methods over the raw grayscale.
This is consistent with previous findings in the literature
\cite{AdinMoseUllm1997,AranZiss2005,FitzZiss2002,WangLiWang2004}.

However, unlike in previous reports of performance evaluation of
these filters, we also ask the question of \emph{when} they help and
how much in \emph{each case}. To quantify this, consider
``performance vectors'' $\mathbf{s}_R$ and $\mathbf{s}_F$,
corresponding to respectively raw and filtered input, whose each
component is equal to the recognition rate of a method on a
particular training/test data combination. Then the vector $\Delta
\mathbf{s}_R \equiv \mathbf{s}_R - \overline{\mathbf{s}}_R$ contains
relative recognition rates to its average on raw input, and $\Delta
\mathbf{s} \equiv \mathbf{s}_F - \mathbf{s}_R$ the improvement with
the filtered representation. We then considered the \emph{angle}
$\phi$ between vectors $\Delta \mathbf{s}_R$ and $\Delta
\mathbf{s}$, using both the high-pass and Self Quotient Image
representations. In both cases, we found the angle to be $\phi
\approx 136^\circ$.

This is an interesting result: it means that while on average both
representations increase the recognition rate, they actually
\emph{worsen} it in ``easy'' recognition conditions. The observed
phenomenon is well understood in the context of energy of intrinsic
and extrinsic image differences and noise (see \cite{WangTang2003}
for a thorough discussion). Higher than average recognition rates
for raw input correspond to small changes in imaging conditions
between training and test, and hence lower energy of extrinsic
variation. In this case the training and test data sets are already
normalized to have the same illumination and the two filters can
only decrease the signal-to-noise ratio, thereby worsening the
recognition performance. On the other hand, when the imaging
conditions between training and test are very different,
normalization of extrinsic variation is the dominant factor and the
performance is improved.

This is an important observation, as it suggests that the
performance of a method that uses either of the representations can
be increased further in a very straightforward manner by detecting
the difficulty of recognition conditions. This is exploited in
\cite{AranCipo2006a}.

\paragraph{Imaging conditions.} We were
interested if the evaluation results on our database support the
observation in the literature that some illumination conditions are
intrinsically more difficult for recognition than others
\cite{SimZhang2004}. An inspection of the performance of the
evaluated methods has shown a remarkable correlation in relative
performance across illuminations, despite the very different models
used for recognition. We found that relative recognition rates
across illuminations correlate on average with $\rho = 0.96$.

\paragraph{Faces and individuals.} Finally, in the similar manner as
previously for different illumination conditions, we were interested
to see if certain individuals were more difficult for recognition
than others. In other words, are incorrect recognitions roughly
equally distributed across the database, or does a relatively small
number of people account for most? Our robust algorithm failed in
too few cases to make a statistically significant observation, so we
instead looked at the performance of the non-robust gSIM which
failed at about an order of magnitude greater frequency.

\begin{figure*}[!t]
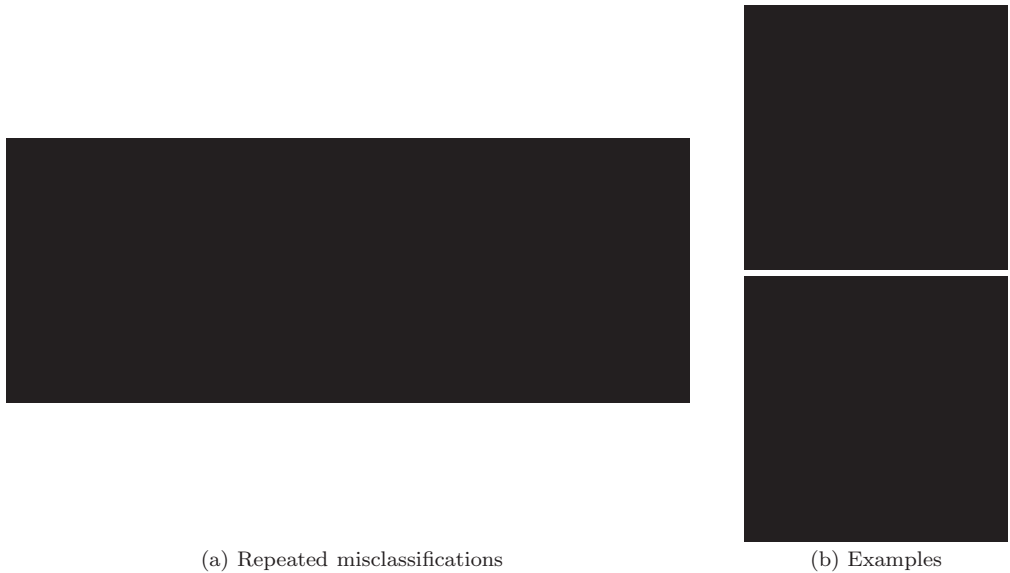

  \centering
  \footnotesize
  \begin{tabular}{VV}
  \includegraphics[width=0.65\textwidth]{PAMI2006_repeat_errs.eps}
  \vspace{5pt}
  &
  \begin{tabular}{c}
    \includegraphics[width=0.25\textwidth]{PAMI2006_errs_5.eps}\\
    \includegraphics[width=0.25\textwidth]{PAMI2006_errs_4.eps}
  \end{tabular}\\
  (a) Repeated misclassifications & (b) Examples \\
  \end{tabular}

  \caption[Failure modes.]{ \it (a) A histogram of non-robust gSIM recognition failures across
            individuals in the ToshFace data set. The majority of errors are
            repeated, of which two of the most common ones are shown in (b). Visual
            inspection of these suggests that these individuals are indeed
            inherently similar in appearance. }
            \label{Fig: Errs}
  \vspace{6pt}\hrule
\end{figure*}

A histogram of recognition errors across individuals in
\textit{ToshFace} data set is shown Figure~\ref{Fig: Errs}~(a),
showing that most errors were indeed repeated. It is difficult to
ascertain if this is a consequence of an inherent similarity between
these individuals or a modelling limitation of our algorithm. A
subjective qualitative inspection of the individuals most commonly
confused, shown in Figure~\ref{Fig: Errs}~(b), tends to suggest that
the former is the dominant cause.

\paragraph{Computational complexity.}
We conclude this section with an analysis of the computational
demands of our algorithm. We focus on the online, novel sequence
recognition (see Section~\ref{SubSec: Robust Likelihood for Novel
Sequence Classification}), as this is of most practical interest. It
consists of the following stages (at this point the reader may find
it useful to refer back to the summary in Figures~\ref{Fig: Alg 1}
and~\ref{Fig: Alg 2}):
\begin{enumerate}
  \item $K$-nearest neighbour computation for each face,
  \item geodesic distance estimation for all pairs of faces,
  \item genetic algorithm optimization,
  \item fine reillumination of all faces, and
  \item robust likelihood computation.
\end{enumerate}
We use the following notation: $N$ is the number of frames in a
sequence, $D$ the number of face pixels, $K$ the number of frames
used in fine reillumination, $N_{gen}$ the number of genetic
algorithm generations, $N_{chr}$ the number of chromosomes in each
generation and $N_{comp}$ the number of Gaussian components in the
Generic SIM GMM.

\begin{figure*}
  \centering
  \Large
  \subfigure[Asymptotic]{
  \begin{tabular*}{1.0\textwidth}{l|c}
    \Hline
    \normalsize \textbf{~Algorithm stage} & \normalsize \textbf{Asymptotic complexity} \\
    \hline
    ~\normalsize $K$-nearest neighbours (see Section~\ref{SubSec: Stage 1}) & \small $N(N D + N \log N)$ \\
    ~\normalsize geodesic distances (see Section~\ref{SubSec: Stage 1})     & \small $N^3 + N K$      \\
    ~\normalsize genetic algorithm (see Section~\ref{SubSec: Stage 1})      & \small $N_{gen} N_{chr} (N D + N K)$ \\
    ~\normalsize fine reillumination (see Section~\ref{SubSec: Stage 1})    & \small $N K^3$  \\
    ~\normalsize robust likelihood (see Section~\ref{SubSec: Robust
Likelihood for Novel Sequence Classification})                              & \small $N N_{comp} D^2 + N\log N$ \\
    \Hline
  \end{tabular*}}
  \subfigure[Measured]{\includegraphics[width=0.8\textwidth]{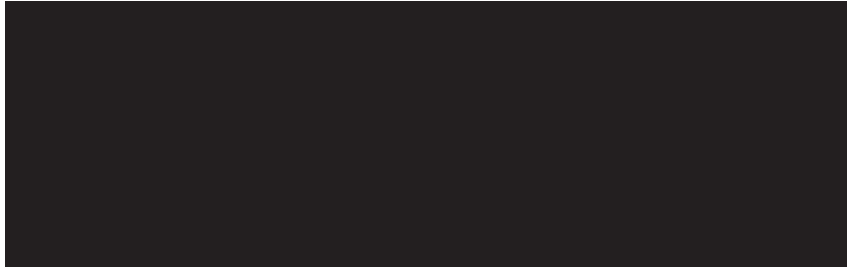}}

  \caption[Theoretical and measured computational demands.]{\it (a) Asymptotic complexity of different stages of the proposed
            online, novel sequence recognition algorithm and (b) the actual measured
            times for our Matlab implementation. }
  \label{Fig: Running Time}
\end{figure*}

For each face, the $K$-nearest neighbour computation consists of
computing its distances to all other faces, $O(D N)$, and ordering
them to find the nearest $K$, $O(N\log N)$. The estimation of
geodesic distances involves initialization, $O(N K)$, and an
application of Floyd's algorithm, $O(N^3)$. In a generation of the
genetic algorithm, for each chromosome we compute the similarity of
all pose-signatures, $O(N D)$, and look-up geodesic distances in all
$K$-neighbourhoods, $O(N K)$. Finally, robust likelihoods are
computed for all faces, $O(N_{comp} D^2)$, which are then ordered,
$O(N \log N)$. Treating everything but $N$ as a constant, the
overall asymptotic complexity of the algorithm is $O(N^3)$. A
summary is presented in Figure~\ref{Fig: Running Time}~(a).

We next profiled our implementation of the algorithm. It should be
stressed that this code was written in Matlab an consequently the
running times reported are not indicative of its actual
practicability. In all experiments only the number of faces per
sequence was varied: we used $N = 25, 50, 100, 200, 400$ and $800$
faces. Mean computation times for different stages of the algorithm
are plotted in Fig~\ref{Fig: Running Time}~(b). In this range of
$N$, the measured asymptote slopes were typically lower than
predicted, which was especially noticeable for the most demanding
computations (e.g.\ of geodesic distances). The most likely reason
for this phenomenon are large constants associated with Matlab's
\textit{for}-loops and data allocation routines.

\section{Summary and conclusions} \label{Sec: Summary}
In this chapter we described a novel algorithm for face recognition
that uses video to achieve invariance to illumination, pose and user
motion pattern variation. We introduced the concept of the Generic
Shape-Illumination manifold as a model of illumination effects on
faces and showed how it can be learnt offline from a small training
corpus. This was made possible by the proposed ``reillumination''
algorithm which is used extensively both in the offline and online
stages of the method.

Our method was demonstrated to achieve a nearly perfect recognition
on 3 databases containing extreme variation in acquisition
conditions. It was compared to and has significantly outperformed
state-of-the-art commercial software and methods in the literature.
Furthermore, an analysis of a large-scale performance evaluation (i)
showed that the method is promising for image-to-sequence matching,
(ii) suggested a direction of research to improve image filtering
for illumination invariance, and (iii) confirmed that certain
illuminations and individuals are inherently particularly
challenging for recognition.

There are several avenues for future work that we would like to
explore. Firstly, we would like to make further use of offline
training data, by constructing the gSIM while taking into account
probabilities of both intra- and inter-personal differences.
Additionally, we would like to improve the computational efficiency
of the method, e.g.\ by representing each FMM by a strategically
chosen set of sparse samples. Finally, we are evaluating the
performance of image-to-sequence matching and looking into
increasing its robustness, in particular to pose.

\section*{Related publications}

The following publications resulted from the work presented in this
chapter:

\begin{itemize}
  \item O. Arandjelovi\'c and R. Cipolla. Face recognition from video using the generic
                  shape-illumination manifold. In \textit{Proc. IEEE European Conference on Computer
                  Vision}, \textbf{4}:pages 27--40, May 2006. \cite{AranCipo2006b}
  \item O. Arandjelovi\'c and R. Cipolla. Achieving robust face recognition from video by combining a weak photometric model and a learnt generic face invariant. \textit{Pattern Recognition}, \textbf{46}(1):9--23, January 2013. \cite{AranCipo2013}
\end{itemize}

\part{Multimedia Organization and Retrieval}

\graphicspath{{./10chreco/}}
\chapter{Film Character Retrieval}
\label{Chp: Char}
\begin{center}
  \footnotesize
  \vspace{-20pt}
  \framebox{\includegraphics[width=0.60\textwidth]{title_img.eps}}\\
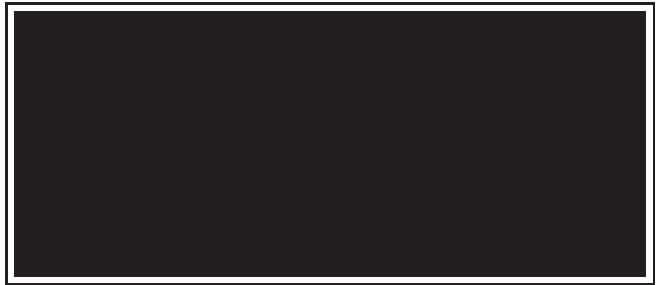
  Charlie Chaplin. \textit{The Gold Rush}\\
  1925, screen capture
\end{center}

\cleardoublepage
    \index{recognition|see{face}}
    \index{features|see{face}}
    \index{registration|see{face}}
    \index{detection|see{face}}
    \index{expression|see{face}}
    \index{pose|see{face}}
    \index{clutter!removal|see{face segmentation}}
    \index{matching!robust|see{distance, robust}}

    \index{face!recognition}
    \index{retrieval!content-based}
    \index{content|see{retrieval}}
    \index{image!signature}
    \index{films!feature length}

The preceding chapters concentrated on the user authentication
paradigm of face recognition. The aim was to reliably compare two
video sequences of random head motion performed by the users. In
contrast, the objective of this work is to recognize all faces of a
character in the closed world of a movie or situation comedy.

This is challenging because faces in a feature-length film are
relatively uncontrolled with a wide variability of scale, pose,
illumination, and expressions, and also may be partially occluded.
Furthermore, unlike in the previous chapters, a continuous video
stream does not contain a face of only a single person, which
increases the difficulty of data extraction. In this chapter
recognition is performed given a small number of query faces, all of
which are specified by the user.

We develop and describe a recognition method based on a cascade of
processing steps that normalize for the effects of the changing
imaging environment. In particular there are three areas of novelty:
(i) we suppress the background surrounding the face, enabling the
maximum area of the face to be retained for recognition rather than
a subset; (ii) we include a pose refinement step to optimize the
registration between the test image and face exemplar; and (iii) we
use robust distance to a sub-space to allow for partial occlusion
and expression change.

The method is applied and evaluated on several feature length films.
It is demonstrated that high recall rates (over 92\%) can be
achieved whilst maintaining good precision (over 93\%).

\index{face!detection}
\section{Introduction}
We consider face recognition for content-based multimedia retrieval:
our aim is to retrieve, and \emph{rank} by confidence, film shots
based on the presence of specific actors. A query to the system
consists of the user choosing the person of interest in one or more
keyframes. Possible applications include:
\begin{enumerate}
    \item \textbf{DVD browsing:} Current DVD technology allows users
            to quickly jump to the chosen part of a film using an
            on-screen index. However, the available locations are
            predefined. Face recognition technology could allow the user to rapidly
            browse scenes by formulating queries based on the
            presence of specific actors.
    \item \textbf{Content-based web search:} Many web search engines
            have very popular image search features (e.g.\ {\small \url{http://www.google.co.uk/imghp}}).
            Currently, the search is performed based on the keywords that
            appear in picture filenames or in the surrounding web page content.
            Face recognition can make the retrieval much more accurate by
            focusing on the content of images.
\end{enumerate}

\begin{figure}
  \centering
  \includegraphics[width=0.9\textwidth]{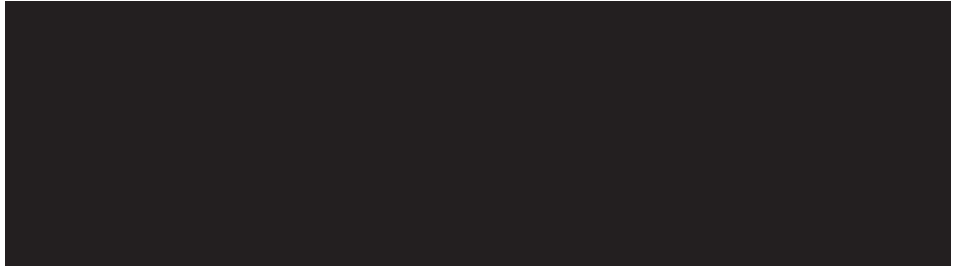}\\
  \caption[Faces in films.]{\it  Automatically detected faces in a typical frame from
                the feature-length film ``Groundhog day''. The
                background is cluttered, pose, expression and
                illumination very variable. }
    \label{Figure: Face Detection}
    \index{face!detection}
  \vspace{6pt}\hrule
\end{figure}

As before, we proceed from the face detection stage, assuming
localized faces. We use a local implementation of the method of
Schneiderman and Kanade \cite{Schneiderman2000} and consider a face
to be correctly detected if both eyes and the mouth are visible, see
Figure~\ref{Figure: Face Detection}. In a typical feature-length
film, using every 10th frame, we obtain 2000-5000 face detections
which result from a cast of 10-20 primary and secondary characters
(see Section~\ref{Section: Evaluation and Results}).

\begin{figure}
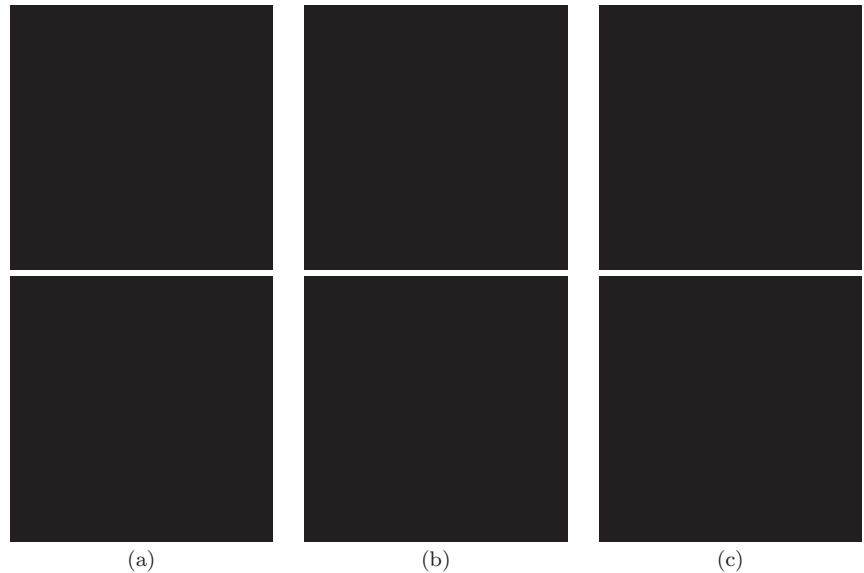

  \centering
  \footnotesize
  \begin{tabular}{ccc}
    \includegraphics[width=0.25\textwidth]{CVPR2005_lighting_01.ps} &
    \includegraphics[width=0.25\textwidth]{CVPR2005_pose_01.ps} &
    \includegraphics[width=0.25\textwidth]{CVPR2005_expression_01.ps}\\
    \includegraphics[width=0.25\textwidth]{CVPR2005_lighting_02.ps} &
    \includegraphics[width=0.25\textwidth]{CVPR2005_pose_02.ps} &
    \includegraphics[width=0.25\textwidth]{CVPR2005_expression_02.ps}\\
    (a) & (b) & (c) \\
  \end{tabular}
  \caption[Appearance variations in films.]{\it The effects of imaging conditions -- illumination (a),
            pose (b) and expression (c) -- on the appearance of a face
            are dramatic and present the main difficulty to automatic face recognition.}
            \label{Figure: Effects of Imaging Conditions}
  \vspace{6pt}\hrule
\end{figure}

\paragraph{Method overview.}
Our approach consists of computing a numerical value, a
distance\index{distance}, expressing the degree of belief that two
face images belong to the same person. Low distance, ideally zero,
signifies that images are of the same person, whilst a large one
signifies that they are of different people.

The method involves computing a series of transformations of the
original image, each aimed at removing the effects of a particular
extrinsic imaging factor. The end result is a \emph{signature image}
of a person, which depends mainly on the person's identity (and
expression, see Section~\ref{SubSection: Multiple Reference Images})
and can be readily classified. The preprocessing stages of our
algorithm are summarized in Figure~\ref{Figure: Face Normalization}
and Figure~\ref{Alg: Overview}.

\subsection{Previous work} Most previous work on face recognition focuses on user
authentication applications, few authors addressing it in a setup
similar to ours. Fitzgibbon and Zisserman~\cite{FitzZiss2002}
investigated face clustering in feature films, though without
explicitly using facial features for registration.  Berg \textit{et
al.}~\cite{BergBergEdwa+2004} consider the problem of clustering
detected frontal faces extracted from web news pages. In a similar
manner to us, affine registration with an underlying SVM-based
facial feature detector is used for face rectification. The
classification is then performed in a Kernel PCA space using
combined image and contextual text-based features.  The problem we
consider is more difficult in two respects: (i) the variation in
imaging conditions in films is typically greater than in newspaper
photographs, and (ii) we do not use any type of information other
than visual cues (i.e.\ no text).  The difference in the difficulty
is apparent by comparing the examples in~\cite{BergBergEdwa+2004}
with those used for evaluation in Section~\ref{Section: Evaluation
and Results}.  For example, in~\cite{BergBergEdwa+2004} the face
image size is restricted to be at least $86 \times 86$ pixels,
whilst a significant number of faces we use are of lower resolution.

Everingham and Zisserman~\cite{EverZiss2004} consider face
recognition in situation comedies. However, rather than using facial
feature detection, a quasi-3D model of the head is used to correct
for varying pose. Temporal information via shot tracking is
exploited for enriching the training corpus. In contrast, we do not
use any temporal information, and the use of local features
(Section~\ref{SubSubSection: Facial Feature Detection}) allows us to
compare two face images in spite of partial occlusions
(Section~\ref{SubSection: Comparing Signature Images}).

\begin{figure}[!t]
    \centering
    \begin{tabular}{l}
          \begin{tabular}{ll}
              \textbf{Input}:  & novel face image $\mathbf{I}$,\\
                               & training signature image $\mathbf{S}_r$.\\
              \textbf{Output}: & distance $d(\mathbf{I}, \mathbf{S}_r)$.\\
          \end{tabular}\vspace{5pt}
          \\ \hline \\
          \begin{tabular}{l}
            \textbf{1: Facial feature localization}\\
            \hspace{10pt}$\{\mathbf{x}_i\}  = \text{features}(\mathbf{I})$\\\\

            \textbf{2: Pose effects -- registration by affine warping}\\
            \hspace{10pt}$\mathbf{I}_R = \text{affine\_warp}\left( \mathbf{I}, \{\mathbf{x}_i\}\right)$\\\\

            \textbf{3: Background clutter -- face outline detection}\\
            \hspace{10pt}$\mathbf{I}_F = \mathbf{I}_R .* \text{mask}(\mathbf{I}_R) $\\\\

            \textbf{4: Illumination effects -- band-pass filtering}\\
            \hspace{10pt}$\mathbf{S} = \mathbf{I}_F \ast \mathbf{B}$\\\\

            \textbf{5: Pose effects -- registration refinement}\\
            \hspace{10pt}$\mathbf{S}_f = \text{warp\_using\_appearance}(\mathbf{I}_F, \mathbf{S}_r)$\\\\

            \textbf{6: Occlusion effects -- robust distance measure}\\
            \hspace{10pt}$d(\mathbf{I}, \mathbf{S}_r) = \text{distance}(\mathbf{S_r}, \mathbf{S_f})$\\\\
          \end{tabular}\\\hline
    \end{tabular}
    \caption[Proposed face matching algorithm.]{\it A summary of the main
                    steps of the proposed algorithm. A novel, `input' image $\mathbf{I}$ is
                    first preprocessed to produce a signature image $\mathbf{S_f}$,
                    which is then compared with signatures of each `reference' image
                    $\mathbf{S_r}$. The intermediate results of preprocessing are also
                    shown in Figure~\ref{Figure: Face Normalization}. }
    \vspace{6pt}\hrule
    \label{Alg: Overview}
\end{figure}

\begin{figure}[t]
  \centering
  \includegraphics[width=0.88\textwidth]{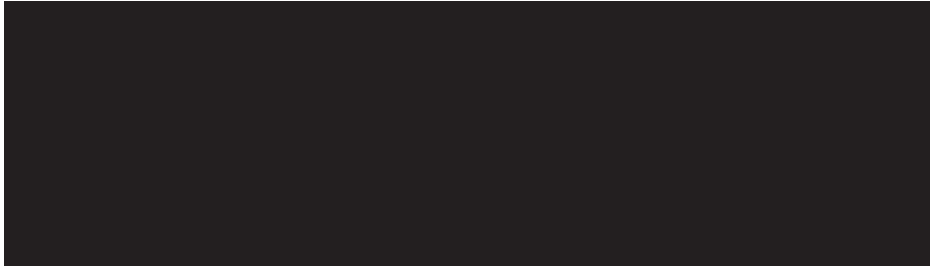}
  \caption[Signature image cascade.]{ \it Each step in the proposed
            preprocessing cascade produces a result invariant to a specific
            extrinsic imaging factor. The result is a `signature' image of
            a person's face. }

  \index{face!segmentation}
  \index{face!features}
  \index{face!registration}
  \index{illumination!normalization}
  \index{image!signature}
  \index{face!pose}

  \label{Figure: Face Normalization}
  \vspace{6pt}\hrule
\end{figure}

\section{Method details}\label{Section: Method Details}

    \index{face!pose}

In the proposed framework, the first step in processing a face image
is the normalization of the subject's pose i.e.\ registration. After
the face detection stage, faces are only roughly localized and
aligned -- more sophisticated registration methods are needed to
correct for the appearance effects of varying pose. One way of doing
this is to ``lock onto'' characteristic facial points and warp
images to align them. In our method, these facial points are the
locations of the mouth and the eyes.

\subsection{Facial feature detection}\label{SubSubSection: Facial
Feature Detection}

    \index{face!features}
    \index{detection!eye}
    \index{detection!mouth}
    \index{Support Vector Machines}

In the proposed algorithm Support Vector Machines\footnote{We used
the LibSVM implementation freely available at
\url{http://www.csie.ntu.edu.tw/~cjlin/libsvm/}}
(SVMs)~\cite{Burg1998, SchoSmol2002} are used for facial feature
detection. A related approach was described
in~\cite{BergBergEdwa+2004}; alternative methods include pictorial
structures~\cite{FelzHutt2005}, shape+appearance cascaded
classifiers~\cite{FukuYama1998} and the method of Cristinacce
\textit{et al.}~\cite{CrisCootSco2004}.

We represent each facial feature, i.e.\ the image patch
surrounding it, by a feature vector. An SVM with a set of
parameters (kernel type, its bandwidth and a regularization
constant) is then trained on a part of the training data and its
performance iteratively optimized on the remainder. The final
detector is evaluated by a one-time run on unseen data.

\subsubsection{Training}\label{SubSubSec: Training} 
\begin{figure}
  \centering
  \includegraphics[width=0.95\textwidth]{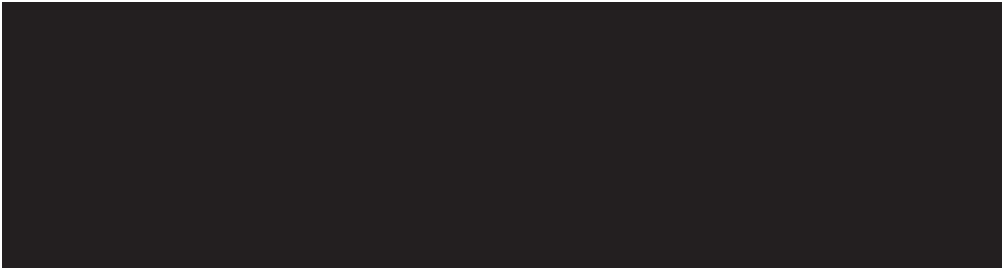}
  \caption[Appearance context importance for feature detection.]{ \it
            Without surrounding image context, distinguishing features in low
            resolution and severe illumination conditions is a hard
            task even for a human. Shown are a mouth and an eye
            that although easily recognized within the context of
            the whole image, are very similar in isolation.}

    \index{face!features}
    \index{detection!eye}
    \index{detection!mouth}
  \label{Figure: Facial Feature Detection Difficulties}
  \vspace{6pt}\hrule
\end{figure}

\begin{figure}
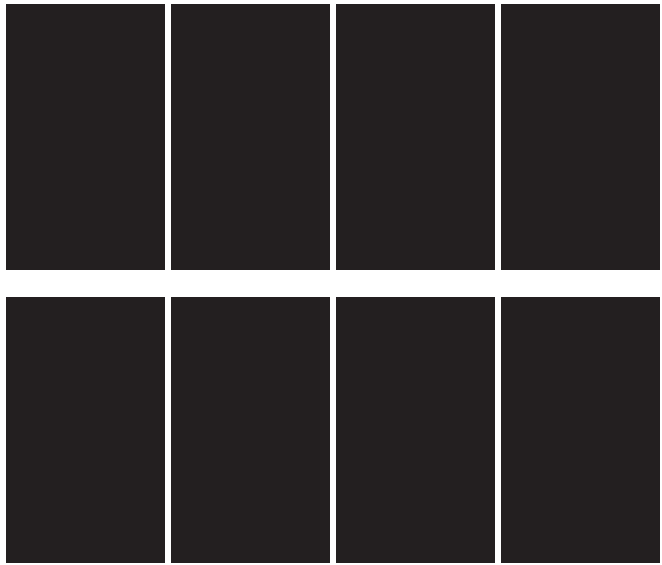


  \footnotesize
  \centering
  \includegraphics[width=0.15\textwidth]{CVPR2005_training_eye1.ps}
  \includegraphics[width=0.15\textwidth]{CVPR2005_training_eye2.ps}
  \includegraphics[width=0.15\textwidth]{CVPR2005_training_eye3.ps}
  \includegraphics[width=0.15\textwidth]{CVPR2005_training_eye4.ps}\\
  \vspace{10pt}
  \includegraphics[width=0.15\textwidth]{CVPR2005_training_eye5.ps}
  \includegraphics[width=0.15\textwidth]{CVPR2005_training_eye6.ps}
  \includegraphics[width=0.15\textwidth]{CVPR2005_training_eye7.ps}
  \includegraphics[width=0.15\textwidth]{CVPR2005_training_eye8.ps}\\
  \caption[SVM training data.]{ \it
                A subset of the data (1800 features were used in total)
                used to train the SVM-based eye detector. Notice the low
                resolution and the importance of the surrounding
                image context for precise localization
                (see Figure~\ref{Figure: Facial Feature Detection Difficulties}). }
            \label{Figure: Eye Training Data}
  \vspace{6pt}\hrule
\end{figure}

For training we
use manually localized facial features in a set of 300 randomly
chosen faces from the feature-length film ``Groundhog day'' and the
situation comedy ``Fawlty Towers''. Examples are extracted by taking
rectangular image patches centred at feature locations (see
Figures~\ref{Figure: Facial Feature Detection Difficulties}
and~\ref{Figure: Eye Training Data}). We represent each patch
$\mathbf{I} \in \mathbb{R}^{H \times W}$ with a feature vector
$\mathbf{v} \in \mathbb{R}^{2H \times W}$ containing appearance and
gradient information (we used $H=17$ and $W=21$ for a face image of
the size $81 \times 81$ - the units being pixels):
\begin{align}
    &v_A(Wy+x) = I(x, y)\\
    &v_G(Wy+x) = \left|\nabla I(x, y)\right|\\
    &\mathbf{v} = \begin{bmatrix}
                    \mathbf{v}_A\\
                    \mathbf{v}_G\\
                  \end{bmatrix}
    \label{Equation: Facial Feature Representation}
    \index{Support Vector Machines}
    \index{detection!eye}
    \index{detection!mouth}
\end{align}

\begin{figure}[t]
  \footnotesize
  \centering

  \includegraphics[width=0.7\textwidth]{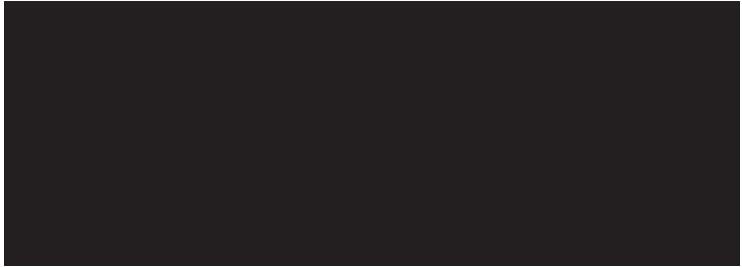}
  \caption[Fast, coarse-to-fine SVM-based feature detection.]
      { \it A summary of the efficient SVM-based eye detection:
            1: Prior on feature location restricts the search region.
            2: Only $\sim 25$\% of the locations are initially classified.
            3: Morphological dilation is used to approximate the dense
            classification result from a sparse output. 4: The largest
            prior-weighted cluster is chosen as containing the feature
            of interest. }
  \label{Figure: SVM_speedup}
  \vspace{6pt}\hrule
\end{figure}

\paragraph{Local information.}
In the proposed method, implicit local information is included for
increased robustness. This is done by complementing the image
appearance vector $\mathbf{v}_A$ with the greyscale intensity
gradient vector $\mathbf{v}_G$, as in equation~\eqref{Equation:
Facial Feature Representation}.

\paragraph{Synthetic data.}

\index{synthetic!data augmentation}

For robust classification, it is important that training data sets
are representative of the whole spaces that are discriminated
between. In uncontrolled imaging conditions, the appearance of
facial features exhibits a lot of variation, requiring an
appropriately large training corpus. This makes the approach with
manual feature extraction impractical. In our method, a large
portion of training data (1500 out of 1800 training examples) was
synthetically generated. Seeing that the surface of the face is
smooth and roughly fronto-parallel, its 3D motion produces locally
affine-like effects in the image plane. Therefore, we synthesize
training examples by applying random affine perturbations to the
manually detected set (for similar approaches to generalization from
a small amount of training data see
\cite{AranCipo2006,Mart2002,SungPogg1998}).

\subsubsection{SVM-based feature detector}
SVMs only provide classification decision for individual feature
vectors, but no associated probabilistic information. Therefore,
performing classification on all image patches produces as a result
a binary image (a feature is either present or not in a particular
location) from which only a single feature location is to be
selected.

Our method is based on the observation that due to the robustness to
noise of SVMs, the binary image output consists of connected
components of positive classifications (we will refer to these as
\emph{clusters}), see Figure~\ref{Figure: SVM_speedup}. We use a
prior on feature locations to focus on the cluster of interest.
Priors corresponding to the three features are assumed to be
independent and Gaussian (2D, with full covariance matrices) and are
learnt from the training corpus of 300 manually localized features
described in Section~\ref{SubSubSec: Training}. We then consider the
total `evidence' for a feature within each cluster:
\begin{equation}\label{Equation: SVM Pseudo-Posterior}
    e(\mathcal{S}) = \int_{\mathbf{x} \in \mathcal{S}} P(\mathbf{x})d\mathbf{x}
\end{equation}
where $\mathcal{S}$ is a cluster and $P(\mathbf{x})$ the Gaussian
prior on the facial feature location. An unbiased feature location
estimate with $\sigma \approx 1.5$ pixels was obtained by choosing
the mean of the cluster with largest evidence as the final feature
location. Intermediate results of the method are shown in
Figure~\ref{Figure: SVM_speedup}, while Figure~\ref{Figure: Feature
Detection Results} shows examples of detected features.

\begin{figure}[t]
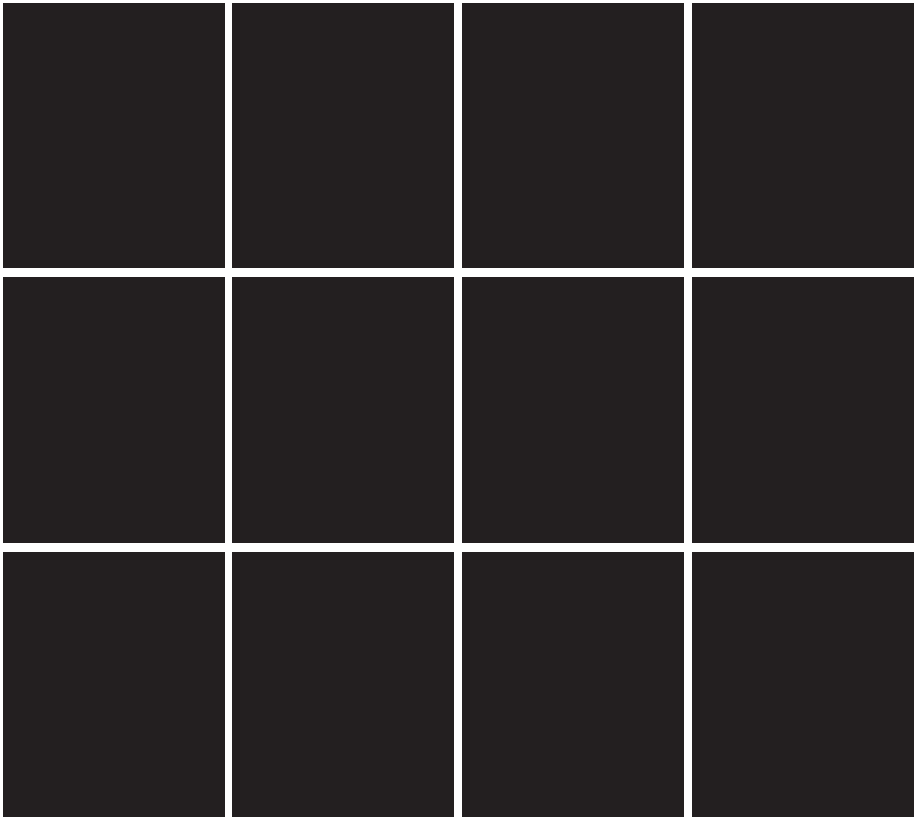

  \centering
  \includegraphics[width=0.21\textwidth]{CVPR2005_svm_final_detection_01.ps}
  \includegraphics[width=0.21\textwidth]{CVPR2005_svm_final_detection_02.ps}
  \includegraphics[width=0.21\textwidth]{CVPR2005_svm_final_detection_03.ps}
  \includegraphics[width=0.21\textwidth]{CVPR2005_svm_final_detection_04.ps}\\ \vspace{3pt}
  \includegraphics[width=0.21\textwidth]{CVPR2005_svm_final_detection_05.ps}
  \includegraphics[width=0.21\textwidth]{CVPR2005_svm_final_detection_06.ps}
  \includegraphics[width=0.21\textwidth]{CVPR2005_svm_final_detection_07.ps}
  \includegraphics[width=0.21\textwidth]{CVPR2005_svm_final_detection_08.ps}\\ \vspace{3pt}
  \includegraphics[width=0.21\textwidth]{CVPR2005_svm_final_detection_09.ps}
  \includegraphics[width=0.21\textwidth]{CVPR2005_svm_final_detection_10.ps}
  \includegraphics[width=0.21\textwidth]{CVPR2005_svm_final_detection_11.ps}
  \includegraphics[width=0.21\textwidth]{CVPR2005_svm_final_detection_12.ps}
  \caption[Feature detection examples.]
          { \it High accuracy in automatic detection of facial features is achieved
                in spite of wide variation in facial expression, pose, illumination
                and the presence of facial wear (e.g.\ glasses and makeup). }

  \index{face!features}
  \index{face!expression}
  \index{detection!eye}
  \index{detection!mouth}

  \label{Figure: Feature Detection Results}
  \vspace{6pt}\hrule
\end{figure}

\subsection{Registration}\label{SubSection: Registration}

    \index{face!registration}
    \index{face!pose}

In the proposed method dense point correspondences are implicitly or
explicitly used in several stages: for background clutter removal,
partial occlusion detection and signature image comparison
(Section~\ref{SubSection: Background Removal}--\ref{SubSection:
Comparing Signature Images}). To this end, images of faces are
affine warped to have salient facial features aligned with their
mean, canonical locations. The six transformation parameters are
uniquely determined from three pairs of point correspondences --
between detected facial features (the eyes and the mouth) and this
canonical frame. In contrast to global appearance-based methods
(e.g.~\cite{BlanVett1999, EdwaTaylCoot1998}) our approach is more
robust to partial occlusion. It is summarized in Figure~\ref{Alg:
Face Registration} with typical results shown in Figure~\ref{Figure:
Registration Results}.

\begin{figure}
    \centering
    \begin{tabular}{l}
          \begin{tabular}{ll}
              \textbf{Input}:   & canonical facial feature locations $\mathbf{x}_{can}$,\\
                                & face image $\mathbf{I}$, \\
                                & facial feature locations $\mathbf{x}_{in}$.\\
              \textbf{Output}:  & registered image $\mathbf{I}_{reg}$.\\
          \end{tabular}\vspace{5pt}
          \\ \hline \\
          \begin{tabular}{l}
            \textbf{1: Estimate the affine warp matrix}\\
            \hspace{10pt}$\mathbf{A} = \text{affine\_from\_correspondences}(\mathbf{x}_{can}, \mathbf{x}_{in})$\\\\

            \textbf{2: Compute eigenvalues of A}\\
            \hspace{10pt}$\{\lambda_1, \lambda_2\} = \text{eig}(\mathbf{A})$\\\\

            \textbf{3: Impose prior on shear and rescaling by A}\\
            \hspace{10pt}\textbf{if} $\left(|\mathbf{A}| \in [0.9, 1.1] \wedge \lambda_1
                    / \lambda_2 \in [0.6, 1.3]\right)$
                    \textbf{then}\\\\

            \hspace{15pt}\textbf{4: Warp the image}\\
            \hspace{25pt}$\mathbf{I}_{reg} = \text{affine\_warp}(\mathbf{I}; \mathbf{A})$\\\\

            \textbf{else}\\\\

            \hspace{15pt}\textbf{5: Face detector false +ve}\\
            \hspace{25pt}\\\\\hline
          \end{tabular}\\
    \end{tabular}
    \caption[Coarse feature-based face registration and false positive detection removal.]{\it A summary of the proposed
        facial feature-based registration of faces and removal of
        face detector false positives. }
    \label{Alg: Face Registration}
\end{figure}

\begin{figure}
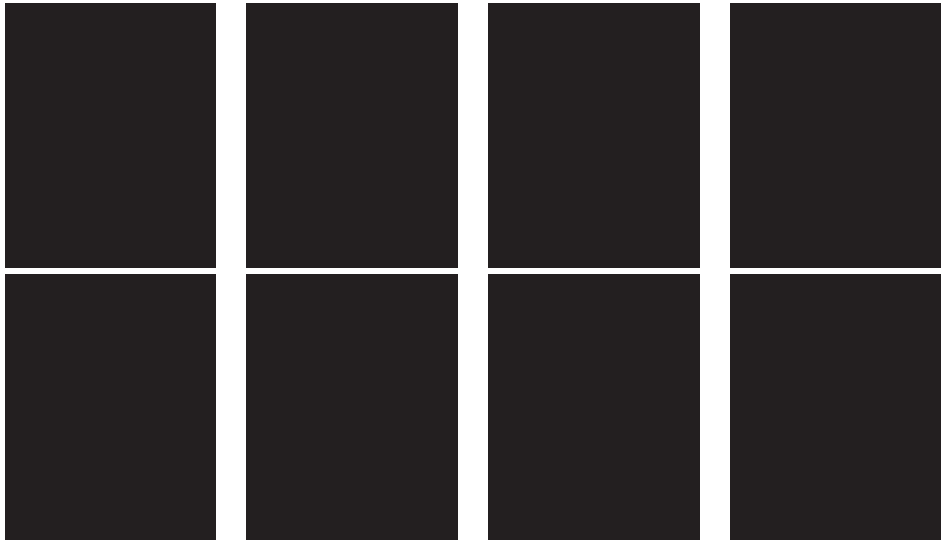

  \footnotesize
  \centering
  \begin{tabular}{cccc}
    \includegraphics[width=0.20\textwidth]{CVPR2005_registration_01.ps}&
    \includegraphics[width=0.20\textwidth]{CVPR2005_registration_02.ps}&
    \includegraphics[width=0.20\textwidth]{CVPR2005_registration_03.ps}&
    \includegraphics[width=0.20\textwidth]{CVPR2005_registration_04.ps} \\
    \includegraphics[width=0.20\textwidth]{CVPR2005_registration_05.ps}&
    \includegraphics[width=0.20\textwidth]{CVPR2005_registration_06.ps}&
    \includegraphics[width=0.20\textwidth]{CVPR2005_registration_07.ps}&
    \includegraphics[width=0.20\textwidth]{CVPR2005_registration_08.ps}
  \end{tabular}
  \caption[Affine registration examples.]
          { \it Original (top) and corresponding registered
                images (bottom).  The eyes and the mouth in all registered
                images are at the same, canonical locations.
                The effects of affine transformations are significant. }
  \label{Figure: Registration Results}
  \index{face!pose}
\end{figure}

\subsection{Background removal}\label{SubSection: Background
Removal} The bounding box of a face, supplied by the face detector,
typically contains significant background clutter and affine
registration boundary artefacts, see Figure~\ref{Figure:
Registration Results}. To realize a reliable comparison of two
faces, segmentation\index{face!segmentation} to foreground (i.e.\
face) and background regions has to be performed. We show that the
face outline can be robustly detected by combining a prior on the
face shape, learnt offline, and a set of measurements of intensity
discontinuity in an image of a face. The proposed method requires
only grey level information, performing equally well for colour and
greyscale input, unlike previous approaches which typically use skin
colour for segmentation (e.g.\ \cite{AranShakFish+2005}).

\begin{figure}
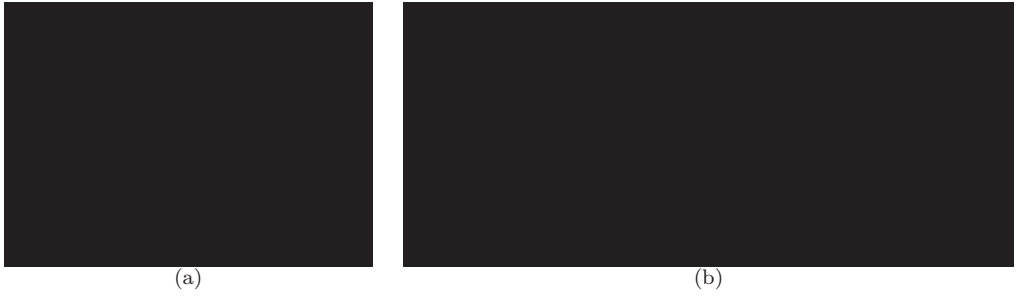

  \footnotesize
  \centering
  \begin{tabular}{VV}
    \includegraphics[width=0.35\textwidth]{CVPR2005_mesh.eps} &
    \includegraphics[width=0.58\textwidth]{CVPR2005_boundary_cs.eps}\\
    (a) & (b)\\
  \end{tabular}
  \caption[Markov chain-based face outline fitting.]
          { \it Markov chain observations:
                (a) A discrete mesh in radial coordinates (only 10\%
                of the points are shown for clarity) to which
                the boundary is confined. Also shown is a single
                measurement of image intensity in the radial
                direction and the detected high probability points.
                The plot of image intensity along this direction is
                shown in (b) along with the gradient magnitude
                used to select the high probability locations. }

  \index{face!segmentation}
  \label{Figure: Boundary}
  \vspace{6pt}\hrule
\end{figure}

In detecting the face outline, we only consider points confined to a
discrete mesh corresponding to angles equally spaced at $\Delta
\alpha$ and radii at $\Delta r$, see Figure~\ref{Figure:
Boundary}~(a); in our implementation we use $\Delta \alpha = 2\pi /
100$ and $\Delta r = 1$. At each mesh point we measure the image
intensity gradient in the radial direction -- if its magnitude is
locally maximal and greater than a threshold $t$, we assign it a
constant high-probability and a constant low probability otherwise,
see Figure~\ref{Figure: Boundary}~(a,b). Let $\mathbf{m}_i$ be a
vector of probabilities corresponding to discrete radius values at
angle $\alpha_i = i \Delta \alpha$, and $r_i$ the boundary location
at the same angle. We seek the maximum \textit{a posteriori}
estimate of the boundary radii:
\begin{align}
    \{ r_i \} =& \label{Equation: Boundary Model with Bayes}
    \arg \max_{\{ r_i \}} P(r_1, .., r_N | \mathbf{m}_1, .., \mathbf{m}_N) =
    \\
    &\arg \max_{\{ r_i \}} P(\mathbf{m}_1, .., \mathbf{m}_N | r_1, ..,
    r_N)P(r_1, .., r_N) \text{,  where } N = 2\pi / \Delta \alpha.
\end{align}

We make the Na\"{\i}ve Bayes assumption for the first term in
equation~\eqref{Equation: Boundary Model with Bayes}, whereas,
exploiting the observation that surfaces of faces are mostly smooth,
for the second term we assume to be a first-order Markov chain.
Formally:
\begin{align}\label{Equation: Boundary Model with Independence 1}
    &P(\mathbf{m}_1, .., \mathbf{m}_N | r_1, .., r_N) = \prod_{i=1}^N P(\mathbf{m}_i|r_i)
    = \prod_{i=1}^N m_i(r_j) \\
    &P(r_1, .., r_N) = P(r_1) \prod_{i=2}^N P(r_i | r_{i-1})
\end{align}

In our method model parameters (priors and likelihoods) are learnt
from 500 manually delineated face outlines. The application of the
model by maximizing expression in \eqref{Equation: Boundary Model
with Bayes} is efficiently realized using dynamic programming i.e.\
the well-known Viterbi algorithm~\cite{GrimStir1992}.

\paragraph{Feathering.} The described method of segmentation of face images
to foreground and background produces as a result a binary mask
image $\mathbf{M}$. As well as masking the corresponding registered
face image $\mathbf{I}_R$ (see Figure~\ref{Figure: Background
Removal Process}), we smoothly suppress image information around the
boundary to achieve robustness to small errors in its localization.
This is often referred to as \emph{feathering}:
\begin{align}\label{Equation: Feathering}
    &\mathbf{M}_F = \mathbf{M} \ast \exp {-\left(\frac {r(x, y)} {4} \right)^2}\\
    &I_F(x, y) = I_R(x, y) M_F(x, y)
\end{align}

Examples of segmented and feathered faces are shown in
Figure~\ref{Figure: Background Removal Examples}.

\begin{figure}[t]
  \footnotesize \centering
  \includegraphics[width=0.95\textwidth]{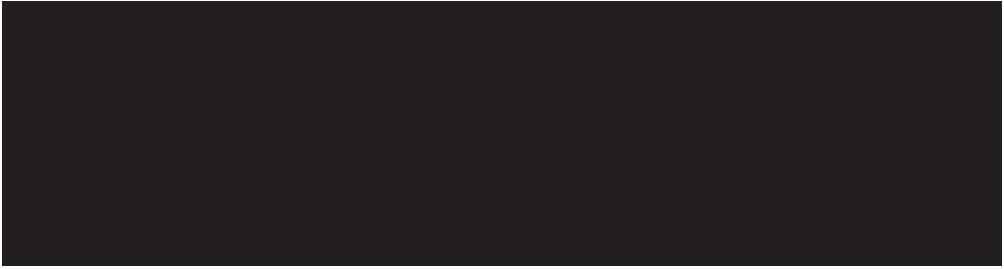}
  \caption[Background removal stages.]{ \it Original image, image with detected face outline, and
                the resulting image with the background masked. }

  \index{face!segmentation}
  \label{Figure: Background Removal Process}
\end{figure}

\begin{figure}[t]
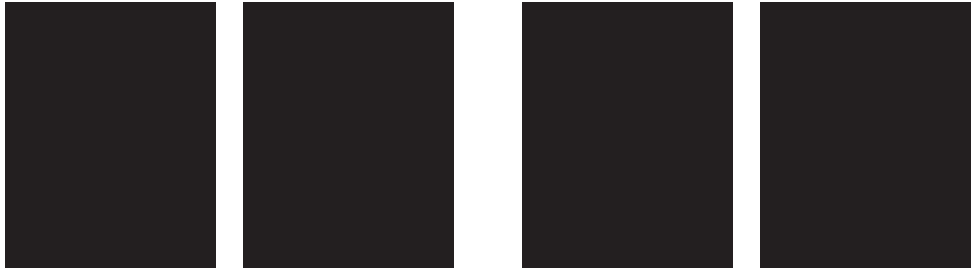

  \footnotesize \centering
  \includegraphics[width=0.2\textwidth]{IVAT2005_registered_01.ps} \hspace{5pt}
  \includegraphics[width=0.2\textwidth]{IVAT2005_segmented_01.ps}
  \hspace{20pt}
  \includegraphics[width=0.2\textwidth]{IVAT2005_registered_02.ps} \hspace{5pt}
  \includegraphics[width=0.2\textwidth]{IVAT2005_segmented_02.ps}
  \caption[Background clutter removal example.]{ \it Original images of detected and affine-registered faces
                and the result of the proposed segmentation
                algorithm. Subtle variations of the face
                outline caused by different poses and head
                shapes are handled with high precision. }

  \index{face!segmentation}
  \label{Figure: Background Removal Examples}
  \vspace{6pt}\hrule
\end{figure}

\subsection{Compensating for changes in
illumination}\label{SubSection: Illumination Compensation}

    \index{illumination!normalization}
    \index{image!signature}

The last step in processing of a face image to produce its signature
is the removal of illumination effects. As the most significant
modes of illumination changes are rather coarse -- ambient light
varies in intensity, while the dominant illumination source is
either frontal, illuminating from the left, right, top or bottom
(seldom) -- and noting that these produce mostly slowly varying, low
spatial frequency variations~\cite{FitzZiss2002} (also see
Section~\ref{CH2: Models}), we normalize for their effects by
band-pass filtering, see Figure~\ref{Figure: Face Normalization}:
\begin{equation}\label{Equation: Band Passing}
    \mathbf{S} = \mathbf{I}_F \ast \mathbf{G}_{\sigma = 0.5} - \mathbf{I}_F \ast \mathbf{G}_{\sigma = 8}
\end{equation}
This defines the signature image $\mathbf{S}$.

\subsection{Comparing signature images}\label{SubSection: Comparing
Signature Images} In Section~\ref{SubSubSection: Facial Feature
Detection}--\ref{SubSection: Illumination Compensation} a cascade of
transformations applied to face images was described, producing a
signature image insensitive to illumination, pose and background
clutter. We now show how the accuracy of facial feature alignment
and the robustness to partial occlusion can be increased further
when two signature images are compared.

\subsubsection{Improving registration}\label{SubSection: Improving
Registration}

    \index{face!registration}
    \index{face!features}

In the registration method proposed in Section~\ref{SubSection:
Registration}, the optimal affine warp parameters were estimated
from three point correspondences in 2D. Therefore, the 6 degrees of
freedom of the affine transformation were uniquely determined,
making the estimate sensitive to facial feature localization errors.
To increase the accuracy of registration, we propose a dense,
appearance-based affine correction to the already computed feature
correspondence-based registration.

\begin{figure*}[!t]
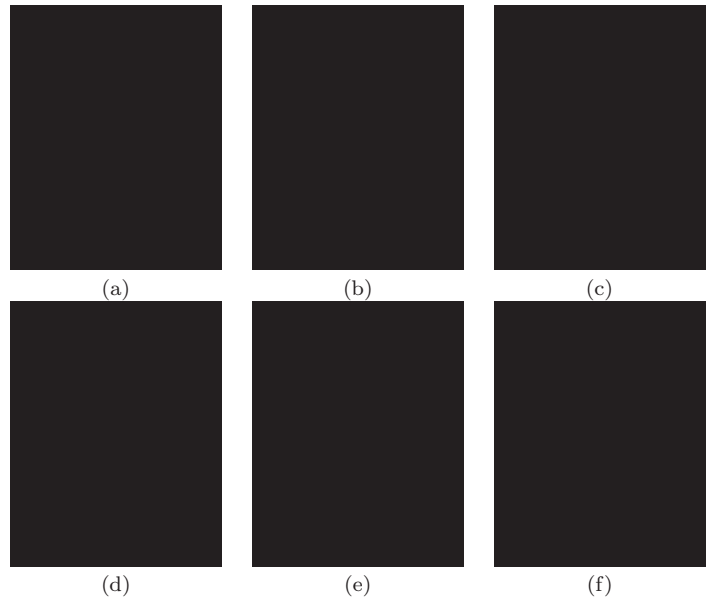

  \footnotesize
  \centering
  \begin{tabular}{ccc}
    \includegraphics[width=0.20\textwidth]{CVPR2005_salient_regions.ps} &
    \includegraphics[width=0.20\textwidth]{CVPR2005_fine_align_o1.ps} &
    \includegraphics[width=0.20\textwidth]{CVPR2005_fine_align_o2.ps}
    \\
    (a) & (b) & (c) \\
    \includegraphics[width=0.20\textwidth]{CVPR2005_fine_align_n1.ps} &
    \includegraphics[width=0.20\textwidth]{CVPR2005_fine_align_d1.ps} &
    \includegraphics[width=0.20\textwidth]{CVPR2005_fine_align_d2.ps}\\
    (d) & (e) & (f)\\
  \end{tabular}

  \caption[Appearance-based pose refinement.]
      { \it Pose refinement summary: (a) Salient regions of the face
            used for appearance-based computation of the residual affine registration.
            (b)(c) Images aligned using feature correspondences alone. (d) The salient
            regions shown in (a) are used to refine the pose of (b) so that it
            is more closely aligned with (c). The residual rotation between (b) and
            (c) is removed. This correction can be seen clearly in the difference
            images: (e) is $|\mathbf{S}_c - \mathbf{S}_b|$, and (f) is
            $|\mathbf{S}_c - \mathbf{S}_d|$. }

  \index{face!features}

  \label{Figure: Fine Alignment}
  \vspace{6pt}\hrule
\end{figure*}

In our algorithm, the corresponding characteristic regions of two
faces, see Figure~\ref{Figure: Fine Alignment}~(a), are perturbed by
small translations to find the optimal \emph{residual shift} (i.e.\
that which gives the highest normalized cross-correlation score
between the two overlapping regions). These new point
correspondences now overdetermine the residual affine transformation
(which we estimate in the least $L_2$ norm of the error sense) that
is applied to the image. Some results are shown in
Figure~\ref{Figure: Fine Alignment}.

\subsubsection{Distance}

    \index{distance!robust}
    \index{occlusion}

\paragraph{Single query image.}  Given two signature images in
precise correspondence (see above), $\mathbf{S}_1$ and
$\mathbf{S}_2$, we compute the following distance between them:
\begin{equation}\label{Equation: Signature Image Distance}
    d_S(\mathbf{S}_1, \mathbf{S}_2) = \sum_x \sum_y h(S_1(x, y) - S_2(x,
    y))
\end{equation}
where $h(\Delta S) = (\Delta S)^2$ if the probability of occlusion
at $(x, y)$ is low and a constant value $k$ otherwise. This is
effectively the $L_2$ norm with added outlier (e.g.\ occlusion)
robustness, similar to~\cite{BlacJeps1998}. We now describe how
this threshold is determined.

\paragraph{Partial occlusions.}\label{SubSection:
Partial Occlusions}

    \index{occlusion}

Occlusions of imaged faces in films are common. Whilst some research
has addressed detecting and removing specific artefacts only, such
as glasses~\cite{JingMari2000}, here we give an alternative
non-parametric approach, and use a simple appearance-based
statistical method for occlusion detection. Given that the error
contribution at $(x, y)$ is $\varepsilon = \Delta S(x, y)$, we
detect occlusion if the probability $P_s(\varepsilon)$ that
$\varepsilon$ is due to inter- or intra- personal differences is
less than 0.05. Pixels are classified as occluded or not on an
independent basis. $P_s(\varepsilon)$ is learnt in a non-parametric
fashion from a face corpus with no occlusion.

The proposed approach achieved a reduction of 33\% in the expected
within-class signature image distance, while the effect on
between-class distances was found to be statistically
insignificant.

\paragraph{Multiple query images.}\label{SubSection:
Multiple Reference Images} The distance introduced in
equation~\eqref{Equation: Signature Image Distance} gives the
confidence measure that two signature images correspond to the same
person. Often, however, more than a single image of a person is
available as a query: these may be supplied by the user or can be
automatically added to the query corpus as the highest ranking
matches of a single image-based retrieval. In either case we want to
be able to quantify the confidence that the person in the novel
image is the same as in the query set.

\index{face!expression}

Seeing that the processing stages described so far greatly normalize
for the effects of changing pose, illumination and background
clutter, the dominant mode of variation across a query corpus of
signature images $\{\mathbf{S}_i\}$ can be expected to be due to
facial expression. We assume that the corresponding manifold of
expression is linear, making the problem that of point-to-subspace
matching~\cite{BlacJeps1998}. Given a novel signature image
$\mathbf{S}_N$ we compute a robust distance:
\begin{equation}\label{Equation: Set Distance}
    d_G\left(\{\mathbf{S}_i\}, \mathbf{S}_N\right) = d_S(\mathbf{F} \mathbf{F}^T
    \mathbf{S}_N, \mathbf{S}_N)
\end{equation}
where $\mathbf{F}$ is orthonormal basis matrix corresponding
to the linear subspace that explains 95\% of energy of variation
within the set $\{\mathbf{S}_i\}$.

\section{Empirical evaluation}\label{Section: Evaluation and
Results}

    \index{retrieval!content-based}
    \index{films!feature length}
    \index{face!recognition}

The proposed algorithm was evaluated on automatically detected faces
from the situation comedy ``Fawlty Towers'' (``A touch of class''
episode), and feature-length films ``Groundhog Day'' and ``Pretty
Woman''\footnote{Available at
\url{http://www.robots.ox.ac.uk/~vgg/data/}}. Detection was
performed on every 10th frame, producing respectively 330, 5500, and
3000 detected faces (including incorrect detections). Face images
(frame regions within bounding boxes determined by the face
detector) were automatically resized to $80 \times 80$ pixels, see
Figure~\ref{Figure: PW Matches}~(a).

\subsection{Evaluation methodology}

    \index{ordering!rank score}

Empirical evaluation consisted of querying the algorithm with each
image in turn (or image set for multiple query images) and ranking
the data in order of similarity to it. Two ways of assessing the
results were employed -- using Receiver Operator Characteristics
(ROC) and the rank ordering score introduced in Section~\ref{CH2:
Perf}.

\subsection{Results}

    \index{retrieval!content-based}

\begin{figure}[t]
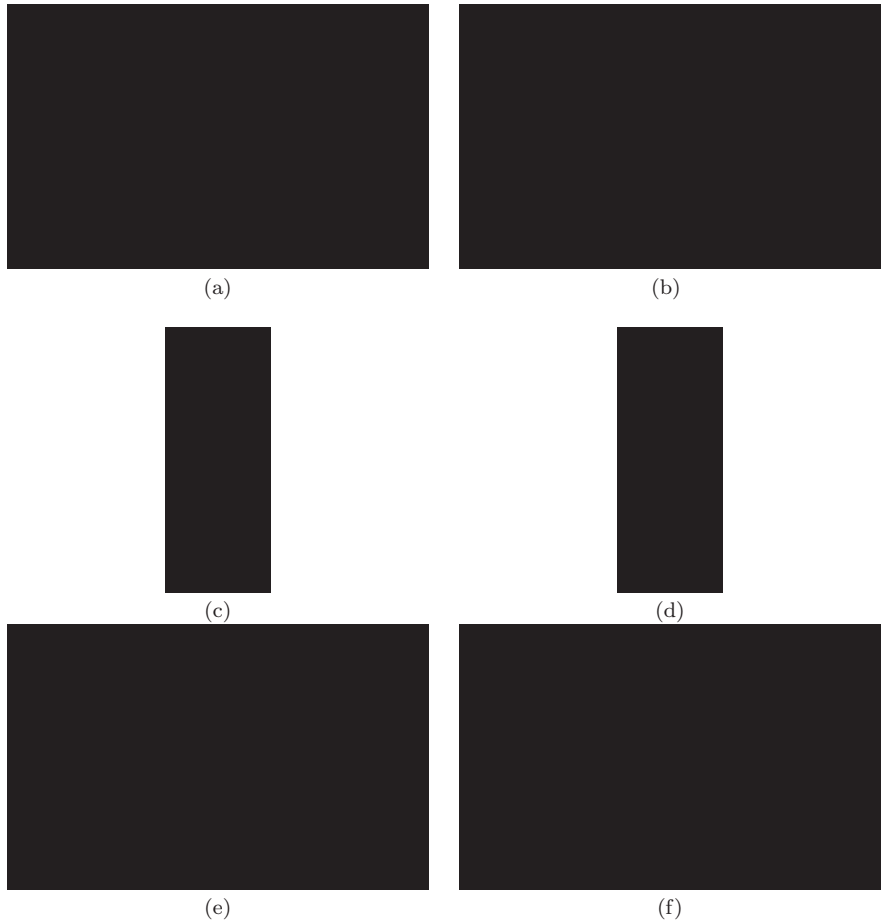

  \footnotesize
  \centering
  \begin{tabular}{cc}
    \includegraphics[width=0.4\textwidth]{CVPR2005_roc1.eps} &
    \includegraphics[width=0.4\textwidth]{CVPR2005_roc2.eps}\\
    (a) & (b) \vspace{10pt} \\
    \includegraphics[width=0.1\textwidth]{basil.eps} &
    \includegraphics[width=0.1\textwidth]{sybil.eps}\\
    (c) & (d) \\
    \includegraphics[width=0.4\textwidth]{CVPR2005_rank_score1.eps} &
    \includegraphics[width=0.4\textwidth]{CVPR2005_rank_score2.eps}\\
    (e) & (f) \\
  \end{tabular}
  \caption[``Fawlty Towers'' ROC curves.]{ \it (a, b) ROC curves for the retrieval
                of Basil (c) and Sybil (d) in ``Fawlty Towers''. The corresponding rank
                ordering scores across 35 retrievals are shown in (e) and (f), sorted for
                the ease of visualization.}

  \index{retrieval!content-based}
  \index{films!feature length}
  \index{face!recognition}

  \label{Figure: Basil and Sybil ROCs}
\end{figure}

\begin{figure}[t]
  \centering
  \includegraphics[width=0.6\textwidth]{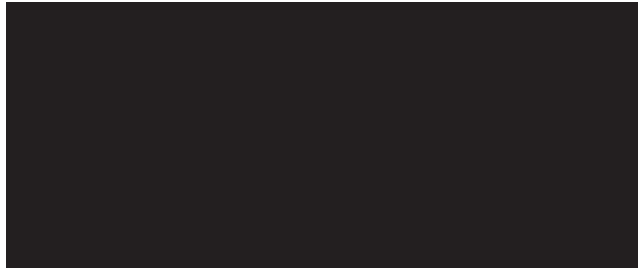}
  \caption[``Fawlty Towers'' rank ordering scores.]{ \it The average rank ordering score of the baseline algorithm
                and its improvement as each of the proposed processing
                stages is added. The improvement is demonstrated both in
                the increase of the average score, and also in the decrease
                of its standard deviation averaged over
                different queries. Finally, note that
                the averages are brought down by few very difficult queries,
                which is illustrated well in
                Figure~\ref{Figure: Basil and Sybil ROCs}~(e,f).  }

  \index{retrieval!content-based}
  \index{films!feature length}
  \index{face!recognition}

  \label{Figure: Stages Improvement}
\end{figure}

\begin{figure}[t]
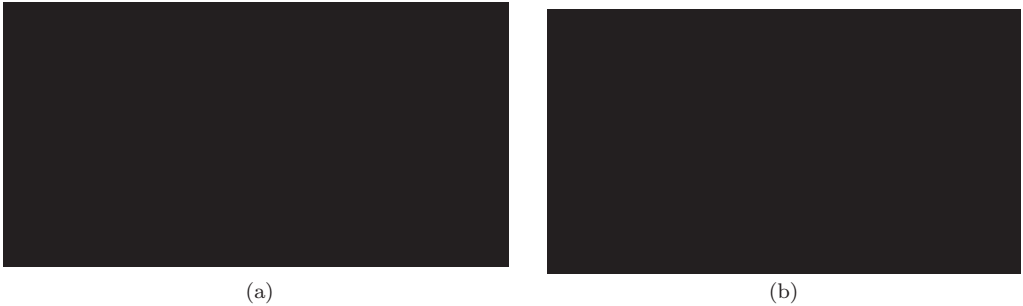

  \footnotesize
  \centering
  \begin{tabular}{VV}
    \includegraphics[width=0.48\textwidth]{PW_data.ps} \vspace{5pt} &
    \includegraphics[width=0.45\textwidth]{CVPR2005_pw_match.ps}\\
    (a) & (b)\\
  \end{tabular}
  \caption[``Pretty Woman'' data set.]{ \it
            (a) The ``Pretty Woman'' data set -- every 50th detected face
            is shown for compactness. Typical retrieval result is shown in
            (b) -- query images are outlined by a solid line, the incorrectly
            retrieved face by a dashed line. The performance of our algorithm
            is very good in spite of the small number of
            query images used and the extremely difficult data set --
            this character frequently changes wigs, makeup and facial
            expressions. }

  \index{retrieval!content-based}
  \index{films!feature length}
  \index{face!recognition}

  \label{Figure: PW Matches}
\end{figure}

Typical Receiver Operator Characteristic curves obtained with the
proposed method are shown in Figure~\ref{Figure: Basil and Sybil
ROCs}~(a, b). These show that excellent results are obtained using
as little as 1-2 query images, typically correctly recalling 92\% of
the faces of the query person with only 7\% of false retrievals. As
expected, more query images produced better retrieval accuracy, also
illustrated in Figure~\ref{Figure: Basil and Sybil ROCs}~(e, f).
Note that as the number of query images is increased, not only is
the ranking better on average but also more robust, as demonstrated
by a decreased standard deviation of rank order scores. This is very
important in practice, as it implies that less care needs to be
taken by the user in the choice of query images. For the case of
multiple query images, we compared the proposed subspace-based
matching with the k-nearest neighbours approach, which was found to
consistently produce worse results. The improvement of recognition
with each stage of the proposed algorithm is shown
in~Figure~\ref{Figure: Stages Improvement}.

Example retrievals are shown in Figures~\ref{Figure: PW
Matches}-\ref{Figure: Groundhog Day Matches}. Only a single
incorrect face is retrieved in the first 50, and this is with a low
matching confidence (i.e.\ ranked amongst the last in the retrieved
set). Notice the robustness of our method to pose, expression,
illumination and background clutter.

\begin{figure}[t]
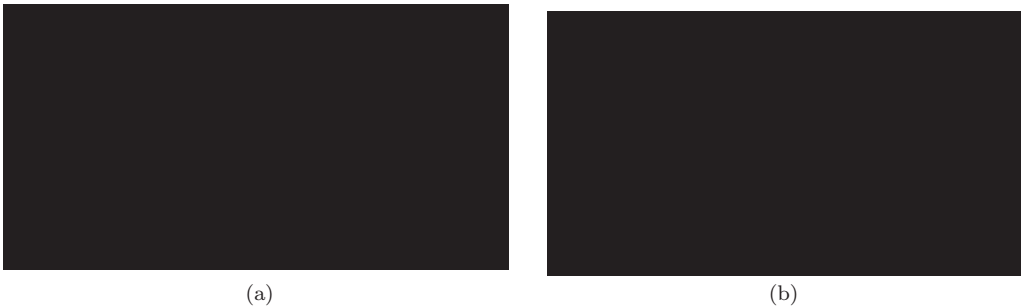

  \footnotesize
  \centering
  \begin{tabular}{VV}
    \includegraphics[width=0.48\textwidth]{FT_data.eps} \vspace{5pt} &
    \includegraphics[width=0.45\textwidth]{CVPR2005_ft_match.ps}\\
    (a) & (b)\\
  \end{tabular}
  \caption[``Fawlty Towers'' data set.]{ \it
            (a) The ``Fawlty Towers'' data set
            -- every 30th detected face is shown for compactness.
            Typical retrieval result is shown in (b) -- query images
            are outlined. There are no incorrectly retrieved faces in
            the top 50. }

  \index{retrieval!content-based}
  \index{films!feature length}
  \index{face!recognition}

  \label{Figure: FT Matches}
\end{figure}

\begin{figure}[t]
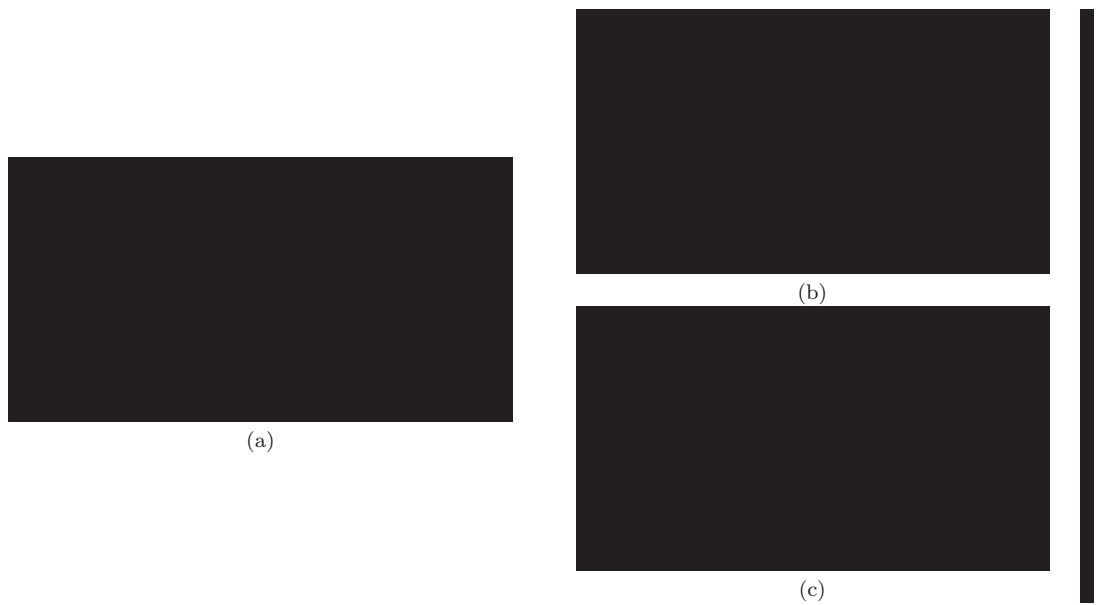

  \footnotesize
  \centering
  \begin{tabular}{VV}
    \begin{tabular}{c}
      \includegraphics[width=0.48\textwidth]{CVPR2005_ghday_data.eps}\\
      (a)
    \end{tabular} &

    \begin{tabular}{c}
      \includegraphics[width=0.45\textwidth]{CVPR2005_ghday_match1.ps}\\
      (b)\\
      \includegraphics[width=0.45\textwidth]{CVPR2005_ghday_match2.ps}\\
      (c)
    \end{tabular}\\
  \end{tabular}

  \caption[``Groundhog Day'' data set.]{ \it
            (a) The ``Groundhog Day'' data set
            -- every 30th detected face is shown for compactness.
            Typical retrieval results are shown in (b) and (c) --
            query images are outlined.
            There are no incorrectly retrieved faces in the top 50. }

  \index{retrieval!content-based}
  \index{films!feature length}
  \index{face!recognition}

  \label{Figure: Groundhog Day Matches}
\end{figure}

\section{Summary and conclusions}\label{Section: Conclusions} In
this chapter we introduced a content-based film-shot retrieval
system driven by a novel face recognition algorithm. The proposed
approach of systematically removing particular imaging distortions
-- pose, background clutter, illumination and partial occlusion has
been demonstrated to consistently achieve high recall and precision
rates on several well-known feature-length films and situation
comedies.

\section*{Related publications}

The following publications resulted from the work presented in this
chapter:

\begin{itemize}
  \item O. Arandjelovi{\'c} and A. Zisserman. Automatic face recognition for film character
                  retrieval in feature-length films. In
                  \textit{Proc. IEEE Conference on Computer Vision and Pattern Recognition},
                  \textbf{1}:pages 860--867, June 2005.
                  \cite{AranZiss2005}

  \item O. Arandjelovi\'c and A. Zisserman. \textit{Interactive video: Algorithms and Technologies}.,
                  chapter On Film Character Rretrieval in Feature-Length
                  Films., pages 89--103.
                  Springer-Verlag,
                  2006. ISBN 978-3-540-33214-5. \cite{AranZiss2006}
\end{itemize}

\graphicspath{{./11anisotropic/}}
\chapter{Automatic Cast Listing in Films}
\label{Chp: Clustering}
\newcommand{\mix}[0]{\textrm{mix}}

\begin{center}
  \footnotesize
  \vspace{-20pt}
  \framebox{\includegraphics[width=0.80\textwidth]{title_img.eps}}\\
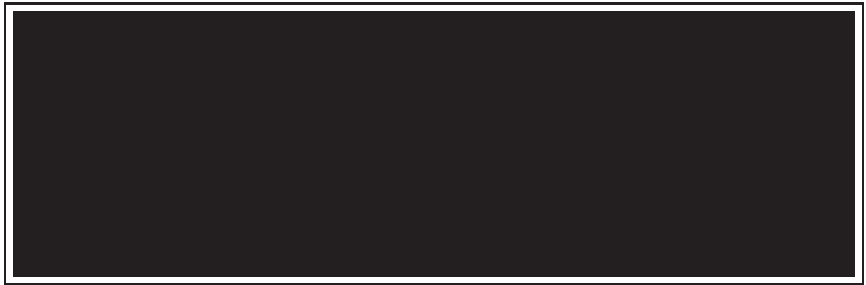
  Auguste Renoir. \textit{Moulin de la Galette}\\
  1876, Oil on canvas, 131 x 175 cm\\
  Mus\'{e}e d'Orsay, París
\end{center}

\cleardoublepage

In this chapter we continue looking at faces in feature-length
films. We consider the most difficult recognition setting of all --
fully automatic (i.e.\ without any dataset-specific training
information) listing of the individuals present in a video. In other
words, our goal is to automatically determine the cast of a
feature-length film without any user intervention.

The main contribution of this  chapter is an algorithm for
clustering over face appearance manifolds themselves. Specifically:
(i) we develop a novel algorithm for exploiting \emph{coherence} of
dissimilarities between manifolds, (ii) we show how to estimate the
optimal \emph{dataset-specific} discriminant manifold starting from
a \emph{generic} one, and (iii) we describe a fully automatic,
practical system based on the proposed algorithm.

We present the results of preliminary empirical evaluation which
show a vast improvement over traditional pattern clustering methods
in the literature.

\section{Introduction}
The problem that we address in this chapter is that of automatically
determining the cast of a feature-length film. This is a far more
difficult problem than that of character retrieval. However, it is
also more appealing from a practical standpoint: the method we will
describe can be used to pre-compute and compactly store identity
information for an entire film, rendering \emph{any} retrieval
equivalent to indexing a look-up table and, consequently, extremely
fast.

The first idea of this chapter concerns the observation that some
people are inherently more similar looking to each other than
others. As an example from our data set, in certain imaging
conditions Sir Hacker may be difficult to distinguish from his
secretary, Sir Humphrey, see respectively Figures~\ref{Fig:
Difficulties} and~\ref{Fig: 2 Classes in Manifold Space} in
Section~\ref{SubSec: Natural Manifold Space}). However, regardless
of the imaging setup he is unlikely to be mistaken, say, for his
wife, see Figure~\ref{Fig: Typical Track} in Section~\ref{SubSec:
Automatic Data Acquisition}. The question is then how to
automatically extract and represent the structure of these
inter-personal similarities from unlabelled sets of video sequences.
We show that this can be done by working in what we term the
\emph{manifold space} -- a vector space in which \emph{each point}
is an appearance manifold.

\begin{figure*}
  \centering
  \includegraphics[width=0.95\textwidth]{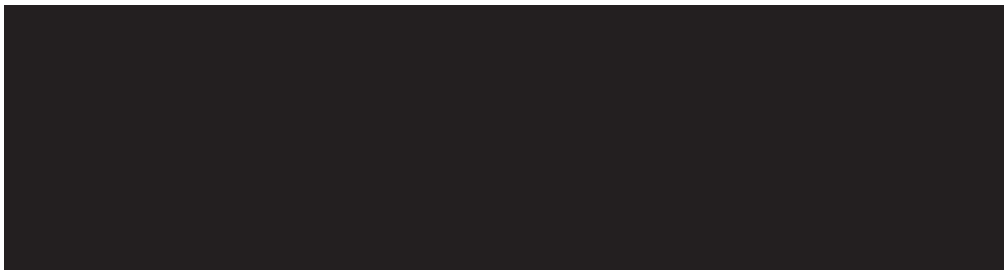}
  \caption[Sources and extent of appearance variations in ``Yes, Minister'' data set.]
      { \it The appearance of faces in films exhibits great
            variability depending on the extrinsic imaging conditions. Shown are the most
            common sources of intra-personal appearance variations (all faces are from
            the same episode of the situation comedy ``Yes, Minister''). }
            \label{Fig: Difficulties}
  \vspace{6pt}\hrule
\end{figure*}

The second major contribution of this chapter is a method for
unsupervised extraction of inter-class data for discriminative
learning on an unlabelled set of video sequences. In spirit, this
approach is similar to the work of Lee and Kriegman
\cite{LeeKrie2005} in which a generic appearance manifold is
progressively updated with new data to converge to a person-specific
one. In contrast, we start from a generic \emph{discriminative}
manifold and converge to a data-specific one, \emph{automatically}
collecting within-class data.

An overview of the entire system is shown in Figure~\ref{Alg:
Overview}.

\begin{figure}[!t]
    \centering
    \begin{tabular}{l}
          \begin{tabular}{ll}
              \textbf{Input}:  & film frames $\{ f_t \}$,\\
                               & generic discrimination subspace $\mathbf{B}_G$.\\
              \textbf{Output}: & cast classes $\mathbb{C}$.\\
          \end{tabular}\vspace{5pt}
          \\ \hline \\
          \begin{tabular}{l}
            \textbf{1: Data extraction -- face manifolds}\\
            \hspace{10pt}$\mathbf{T} = \text{get\_manifolds}(\{ f_t \})$\\\\

            \textbf{2: Synthetically repopulate manifolds}\\
            \hspace{10pt}$\mathbf{T} = \text{repopulate}(\mathbf{T})$\\\\

            \textbf{3: Adaptive discriminative learning -- distance matrix}\\
            \hspace{10pt}$\mathbf{D}_S = \text{distance}(\mathbf{T}, \mathbf{B}_G)$\\\\

            \textbf{4: Manifold space}\\
            \hspace{10pt}$\mathbb{M} = \text{MDS}(\mathbf{D}_S)$\\\\

            \textbf{5: Initial classes}\\
            \hspace{10pt}$\mathbb{C} = \text{classes}(\mathbf{D}_S)$\\\\

            \textbf{6: Anisotropic boundaries in manifold space}\\
            \hspace{10pt}\textbf{for} $\mathbf{C}_i,\mathbf{C}_j \in \mathbb{C}$\\\\

            \hspace{15pt}\textbf{7: PPCA models}\\
            \hspace{25pt}$(\mathcal{G}_i, \mathcal{G}_j) = \text{PPCA}(\mathbf{C}_i,\mathbf{C}_j, \mathbb{M})$\\\\

            \hspace{15pt}\textbf{8: Merge clusters using Weighted Description Length}\\
            \hspace{25pt}$\Delta \text{DL}(i, j) < threshold~?~\text{merge}(i, j, \mathbb{C})$\\\\
          \end{tabular}\\\hline
    \end{tabular}
    \caption[Cast clustering algorithm overview.]{\it A summary of the main
                    steps of the proposed algorithm. }
    \vspace{6pt}\hrule
    \label{Alg: Overview}
\end{figure}

\section{Method details} In this section we describe each of the
steps in the algorithmic cascade of the proposed method: (i)
automatic data acquisition and preprocessing, (ii) unsupervised
discriminative learning and (iii) clustering over appearance
manifolds.

\subsection{Automatic data acquisition}\label{SubSec: Automatic Data
Acquisition} Our cast clustering algorithm is based on pair-wise
comparisons of face manifolds \cite{AranShakFish+2005,
LeeHoYangKrie2003, MoghPent2002} that correspond to sequences of
moving faces. Hence, the first stage of the proposed method is
automatic acquisition of face data from a continuous feature-length
film. We (i) temporally segment the video into \emph{shots}, (ii)
detect faces in each and, finally, (iii) collect detections through
time by tracking in the $(X, Y, \text{scale})$ space.

\begin{figure}
  \centering
  \includegraphics[width=0.85\textwidth]{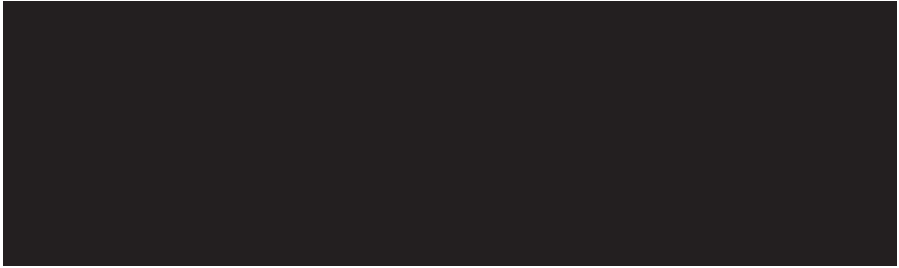}
  \caption[Edge Change Ratio response.]
      { \it The unsmoothed Edge Change Ratio for a 20s
            segment from the situation comedy ``Yes, Minister''. }
            \label{Fig: ECR}
\end{figure}

\paragraph{Shot transition detection.} A number of
reliable methods for shot transition detection have been proposed in
the literature \cite{HampJainWeym1995, OtsuTono1993,
ZabiMillMai1995, ZhanKankSmol1993}. We used the Edge Change Ratio
(ECR) \cite{ZabiMillMai1995} algorithm as it is able in a unified
manner to detect all 3 standard types of shot transitions: (i) hard
cuts, (ii) fades and (iii) dissolves. The ECR is defined as: {
\begin{align} ECR_n = \max ( X_n^{in} / \sigma_n, X_{n-1}^{out} /
\sigma_{n-1})
\end{align}}
where $\sigma_n$ is the number of edge pixels computed using the
Canny edge detector \cite{Cann1986}, and $X_n^{in}$ and $X_n^{out}$
the number of entering and existing edge pixels in frame $n$. Shot
changes are then recognized by considering local peaks of $ECR_n$,
exceeding a threshold, see \cite{Lien1998, ZabiMillMai1995} for
details and Figure~\ref{Fig: ECR} for an example.

\begin{figure}
  \centering
  \includegraphics[width=0.95\textwidth]{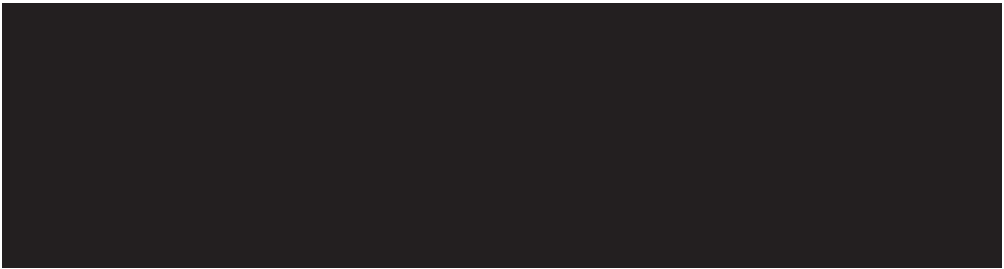}
  \caption[Connecting face detections with tracks.]{ \it The X coordinate of detected faces (red dots) through time in
            a single shot and the resulting tracks connecting them (blue lines) as
            determined by our algorithm. }
            \label{Fig: Tracking}
\end{figure}

\begin{figure*}[!t]
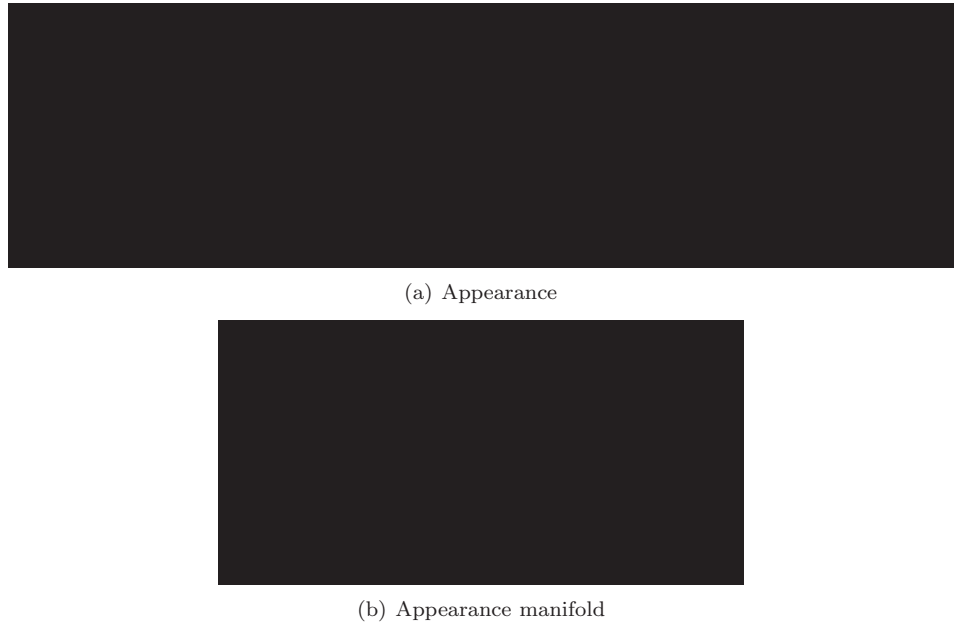

  \centering
  \subfigure[Appearance]
  {\includegraphics[width=0.9\textwidth]{CVPR2006_track.eps}}
  \subfigure[Appearance manifold]
  {\includegraphics[width=0.5\textwidth]{CVPR2006_track_3d.eps}}
  \caption[Typical face track.]{ \it A typical face track obtained using our algorithm. Shown
            are (a) the original images are detected by the face detector (rescaled
            to the uniform scale of $50 \times 50$ pixels) and (b) as points in the
            3D principal component space with temporal connections.  }
            \label{Fig: Typical Track}
  \vspace{6pt}\hrule
\end{figure*}

\paragraph{Face tracking through shots.} We detect
faces in cluttered scenes on an independent, frame-by-frame basis
with the Viola-Jones \cite{ViolJone2004} cascaded
algorithm\footnote{We used the freely available code, part of the
Intel$^\circledR$ OpenCV library.}. For each detected face, the
detector provides a triplet of the $X$ and $Y$ locations of its
centre and scale $s$. In the proposed method, face detections are
connected by tracks using a simple tracking algorithm in the 3D
space $\mathbf{x} = (X, Y, s)$. We employ a form of the Kalman
filter in which observations are deemed completely reliable (i.e.\
noise-free) and the dynamic model is that of zero mean velocity
$\left[ \dot{\mathbf{x}} \right] = 0$ with a diagonal noise
covariance matrix. A typical tracking result is illustrated in
Figure~\ref{Fig: Tracking} with a single face track obtained shown
in Figure~\ref{Fig: Typical Track}.

\subsection{Appearance manifold discrimination}
Having collected face tracks from a film, we turn to the problem of
clustering these sub-sequences by corresponding identity. Due to the
smoothness of faces, each track corresponds to an appearance
manifold \cite{AranShakFish+2005, LeeHoYangKrie2003, MoghPent2002},
as illustrated in Figure~\ref{Fig: Typical Track}. We want to
compare these manifolds and use the structure of the variation of
dissimilarity between them to deduce which ones describe the same
person.

\begin{figure}[!t]
    \centering
    \begin{tabular}{l}
          \begin{tabular}{ll}
              \textbf{Input}:  & manifolds $\mathbf{T} = \{ T_i \}$.\\
                               & generic discrimination subspace $\mathbf{B}_G$.\\
              \textbf{Output}: & distance matrix $\mathbf{D}_S$.\\
          \end{tabular}\vspace{5pt}
          \\ \hline \\
          \begin{tabular}{l}
            \textbf{1: Distance matrix using generic discrimination}\\
            \hspace{10pt}$\mathbf{D}_G = \text{distance}(T, \mathbf{B}_G)$\\\\

            \textbf{2: Provisional classes}\\
            \hspace{10pt}$\mathbb{C}_T = \text{classes}(\mathbf{D}_G)$\\\\

            \textbf{3: Data-specific discrimination space}\\
            \hspace{10pt}$\mathbf{B}_S = \text{constraint\_sspace}(\mathbb{C}_T)$\\\\

            \textbf{4: Mixed discrimination space}\\
            \hspace{10pt}$\mathbf{B}_C = \text{combine\_eigenspaces}(\mathbf{B}_S, \mathbf{B}_G)$\\\\

            \textbf{5: Distance matrix using data-specific discrimination}\\
            \hspace{10pt}$\mathbf{D}_S = \text{distance}(\mathbf{T}, \mathbf{B}_G)$\\\\
          \end{tabular}\\\hline
    \end{tabular}
    \caption[Data-specific discrimination algorithm summary.]
       {\it From generic to data-specific discrimination -- algorithm summary. }
    \vspace{6pt}\hrule
    \label{Alg: Discrimination}
\end{figure}

\paragraph{Data preprocessing.}
As in the previous chapter, as the first step in the comparison of
two appearance manifolds we employ simple preprocessing on a
frame-by-frame basis that normalizes for the majority of
illumination effects and suppresses the background. If $\mathbf{X}$
is an image of a face, in the usual form of a raster-ordered pixel
vector, we first normalize for the effects of illumination using a
high-pass filter (previously used in \cite{AranZiss2005,
FitzZiss2002}) scaled by local image intensity:
\begin{align}
  &\mathbf{X}_L = \mathbf{X} \ast \mathbf{G}_{\sigma = 1.5} \\
  &\mathbf{X}_H = \mathbf{X} - \mathbf{X}_L \\
  &X_I(x, y) = X_H(x, y) / X_L(x, y).
\end{align}
This is similar to the Self-Quotient Image of Wang \textit{et al.}
\cite{WangLiWang2004}. The purpose of local scaling is to equalize
edge strengths in shadowed (weak) and well-illuminated (strong)
regions of the face.

Background is suppressed with a weighting mask $\mathbf{M}_F$,
produced by feathering (similar to \cite{AranZiss2005}) the mean
face outline $\mathbf{M}$, as shown in Figure~\ref{Fig: Feathered
Mask}: {\begin{align}\label{Equation: Feathering}
    &\mathbf{M}_F = \mathbf{M} \ast \exp {-\left(\frac {r(x, y)} {4} \right)^2}\\
    &X_F(x, y) = X_I(x, y) M_F(x, y).
\end{align}}

\begin{figure}
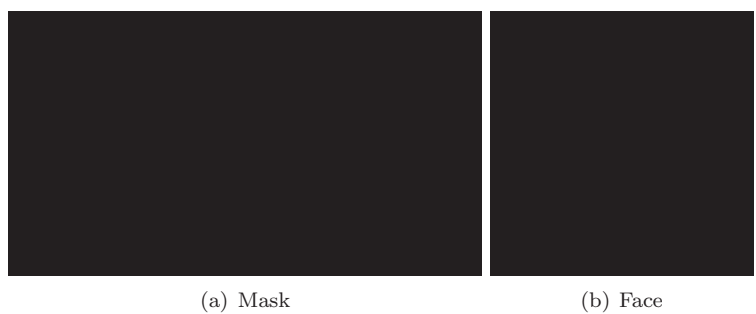

  \centering
  \subfigure[Mask]
  {\includegraphics[width=0.45\textwidth]{CVPR2006_mask.eps}}
  \subfigure[Face]{\includegraphics[width=0.25\textwidth]{CVPR2006_face_before_after.eps}}
  \caption[Illumination normalization and background suppression.]
      { \it (a) The mask used to suppress cluttered background in images
            of automatically detected faces, and (b) an example of a detected, unprocessed
            face and the result of illumination normalization and background suppression. }
            \label{Fig: Feathered Mask}
  \vspace{6pt}\hrule
\end{figure}

\paragraph{Synthetic data augmentation.} Many of the
collected face tracks in films are short and contain little pose
variation. For this reason, we automatically enrich the training
data corpus by stochastically repopulating geodesic neighbourhoods
of randomly drawn manifold samples. This is the same approach we
used in Chapter~\ref{Chp: KRAD} so here we only briefly summarize it
for continuity.

Under the assumption that the face to image space embedding function
is smooth, geodesically close images correspond to small changes in
imaging parameters (e.g.\ yaw or pitch). Hence, using the
first-order Taylor approximation of the effects of a projective
camera, the face motion manifold is locally topologically similar to
the affine warp manifold. The proposed algorithm then consists of
random draws of a face image $\mathbf{x}$ from the data, stochastic
perturbation of $\mathbf{x}$ by a set of affine warps
$\{\mathbf{A}_j\}$ and finally, the augmentation of data by the
warped images.

\subsubsection{Comparing normalized appearance manifolds}\label{SubSubSec: Comparing Normalized Appearance Manifolds} For
pair-wise comparisons of manifolds we employ the Constraint Mutual
Subspace method (CMSM) \cite{FukuYama2003}, based on principal
angles between subspaces \cite{Hote1936,Oja1983}. This choice is
motivated by: (i) CMSM's good performance reports in the literature
\cite{AranShakFish+2005, FukuYama2003}, (ii) its computational
efficiency \cite{BjorGolu1973} and compact data representation, and
(iii) its ability to extract the most similar modes of variation
between two subspaces, see Chapter~\ref{Chp: Thermal} for more
detail.

\begin{figure}
  \centering
  \includegraphics[width=0.9\textwidth]{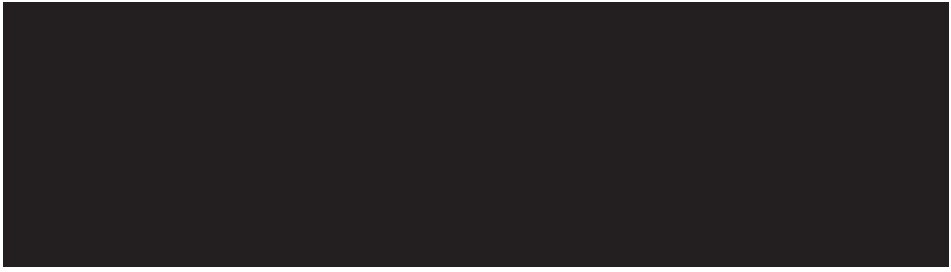}
  \caption[PCA, LDA and Constraint subspaces.]{ \it A visualization of the basis of the linear constraint
            subspace, the most \emph{descriptive} linear subspace (eigenspace
            using PCA \cite{TurkPent1991}) and the most \emph{discriminative}
            linear subspace in terms of within and between class scatter (LDA
            \cite{BelhHespKrie1997}). }
            \label{Fig: Basis}
  \vspace{6pt}\hrule
\end{figure}

As in \cite{FukuYama2003}, we represent each appearance manifold by
a minimal linear subspace it is embedded in -- estimated using
Probabilistic PCA \cite{TippBish1999}. The similarity of two such
subspaces is then computed as the mean of their first 3 canonical
correlations, after the projection onto the \emph{constraint
subspace} -- a linear manifold that attempts to maximize the
separation (in terms of canonical correlations) between different
class subspaces, see Figure~\ref{Fig: Basis}.

\paragraph{Computing the constraint subspace.} Let $\{
\mathbf{B}_i \} = \mathbf{B}_1, \dots, \mathbf{B}_N$ be orthonormal
basis matrices representing subspaces corresponding to $N$ different
classes (cast members, in our case). Fukui and Yamaguchi
\cite{FukuYama2003} compute the orthonormal basis matrix
$\mathbf{B}_C$ corresponding to the constraint subspace using PCA
from: {
\begin{align}
  (\mathbf{B}_R ~ \mathbf{B}_C) \begin{pmatrix}
                                  \Lambda_L & \mathbf{0} \\
                                  \mathbf{0} & \Lambda_S \\
                                \end{pmatrix}
  \begin{pmatrix} \mathbf{B}_R^T \\ \mathbf{B}_C^T \end{pmatrix} = \sum_{i=1}^N \mathbf{B}_i
  \mathbf{B}_i^T \text{, } \mathbf{B}_R^T \mathbf{B}_C = \mathbf{0}
  \label{Eqn: Constraint Subspace}
\end{align}}
where $\Lambda_L$ and $\Lambda_S$ are diagonal matrices with
diagonal entries, respectively, greater or equal than 1 and less
than 1. We modify this approach by weighting the contribution of the
projection matrix $\mathbf{B}_i$ by the number of samples used to
compute it. This way, a more robust estimate is obtained as
subspaces computed from smaller amounts of data (i.e.\ with lower
Signal-to-Noise Ratio) are de-emphasized: {
\begin{align}
  (\mathbf{B}_R ~ \mathbf{B}_C) \begin{pmatrix}
                                  \Lambda_L & \mathbf{0} \\
                                  \mathbf{0} & \Lambda_S \\
                                \end{pmatrix}
  \begin{pmatrix} \mathbf{B}_R^T \\ \mathbf{B}_C^T \end{pmatrix} = N \sum_{i=1}^N N_i \mathbf{B}_i
  \mathbf{B}_i^T / \sum_{i=1}^N N_i
  \label{Eqn: Constraint Subspace1}
\end{align}}

\paragraph{From generic to data-specific
discrimination.} The problem of estimating $\mathbf{B}_C$ lies in
the fact that we do not know which appearance manifolds belong to
the same class and which to different classes i.e.\ $\{ \mathbf{B}_i
\}$ are unknown. We therefore start from a \emph{generic} constraint
subspace $\mathbf{B}_C^g$, computed offline from a large data
corpus. For example, for the evaluation reported in
Section~\ref{Sec: Evaluation and Results} we estimated
$\mathbf{B}_i,~i=1,\dots,100$ using the \textit{CamFace} data set
(see Appendix~\ref{App: CamFace}).

\begin{figure}
  \centering
  \includegraphics[width=0.7\textwidth]{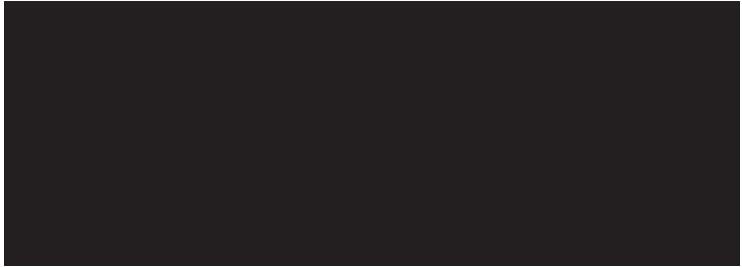}
  \caption[MSM ROC, estimated offline.]{ \it The ROC curve of the Constraint Mutual Subspace Method,
            estimated offline. Shown are two salient points of the
            curve, corresponding to high precision and high recall. }
            \label{Fig: Typical ROC}
  \vspace{6pt}\hrule
\end{figure}

Now, consider the Receiver-Operator Characteristic (ROC) curve of
CMSM in Figure~\ref{Fig: Typical ROC}, also estimated offline. The
inherent tradeoff between recall and precision is clear, making it
impossible to immediately draw class boundaries using the
inter-manifold distance only. Instead, we propose to exploit the two
marked salient points of the curve merely to collect data for the
construction of the constraint subspace. Starting from an arbitrary
manifold, the ``high recall'' point allows to confidently partition
a \emph{part} of the data into different classes. Then, using
manifolds in each of the classes we can gather intra-class data
using the ``high precision'' point. The collected class information
can then be used to compute the basis $\mathbf{B}_C^s$ of the
\emph{data-specific} constraint subspace.

The problem in using the above defined data-specific constraint
subspace $\mathbf{B}_C^s$ is that it is constructed using only the
easiest to classify data. Hence, it cannot be expected to
discriminate well in difficult cases, corresponding to the points on
the ROC curve between ``high precision'' and ``high recall''. To
solve this problem, we do not \emph{substitute} the data-specific
for the generic constraint subspace, but iteratively \emph{combine}
the two based on our confidence $0.0 \leq \alpha \leq 1.0$ in the
former: {
\begin{align}
  \mathbf{B}_C = \mix(\alpha, 1-\alpha, \mathbf{B}_C^s,
  \mathbf{B}_C^g)
\end{align}}
where $\alpha$ and $(1-\alpha)$ are mixing weights. We used an
eigenspace mixing algorithm similar to Hall \textit{et al.}
\cite{HallMarsMart2000}. The mixing confidence parameter $\alpha$ is
determined as follows. Consider clustering appearance manifolds
using each of the two salient points. The ``high precision'' point
will give an overestimate $N_h \geq N$ of the number of classes $N$,
while the ``high recall'' one an underestimate $N_l \leq N$. The
closer $N_h$ and $N_l$ are, the more confident we can be that the
constraint subspace estimate is good. Hence, we compute $\alpha$ as
their normalized difference (which ensures that the condition $0.0
\leq \alpha \leq 1.0$ is satisfied): {
\begin{align}
  \alpha = 1 - \frac {N_h - N_l} { M - 1}
\end{align}}
where $M$ is the number of appearance manifolds.

\subsection{The manifold space}\label{SubSec: Natural Manifold
Space} In Section~\ref{SubSubSec: Comparing Normalized Appearance
Manifolds} we described how to preprocess and pairwise compare
appearance manifolds, optimally exploiting generic information for
discriminating between human faces and automatically extracted
data-specific information. One of the main premises of the proposed
clustering method is that there is a structure to inter- and
intra-personal distances between appearance manifolds. To discover
and exploit this structure, we consider a \emph{manifold space} -- a
vector space in which each \emph{point} represents an appearance
manifold. In the proposed method, manifold representations in this
space are constructed implicitly.

\begin{figure}
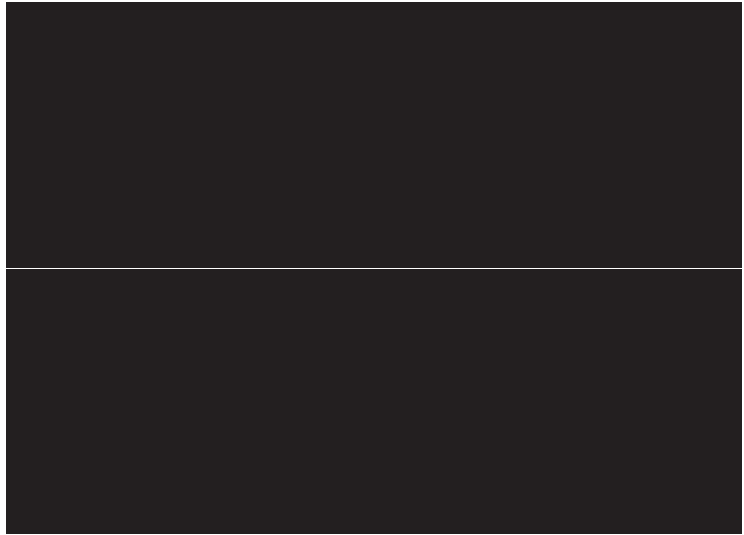

  \centering
  \includegraphics[width=0.7\textwidth]{CVPR2006_manifold_space1.eps}\\
  \includegraphics[width=0.7\textwidth]{CVPR2005_processed_manifolds.eps}
  \caption[Distributions of appearance manifolds in the ``manifold space''.]
      { \it Manifolds in the manifold space (shown are its first 3
            principal components), corresponding to preprocessed tracks of
            faces of the two main characters in the situation comedy
            ``Yes, Minister''. Each red dot corresponds to a single
            appearance manifold of Jim Hacker and black star to a manifold
            of Sir Humphrey (samples from two typical manifolds are shown below the
            plot). The distribution of manifolds in the space shows a
            clear structure. In particular, note that intra-class manifold
            distances are often greater than inter-manifold ones.
            Learning distributions of \emph{manifolds} provides a much more
            accurate way of classification. }
            \label{Fig: 2 Classes in Manifold Space}
  \vspace{6pt}\hrule
\end{figure}

We start by computing a symmetric $N \times N$ distance matrix
$\mathbf{D}$ between all pairs of appearance manifolds using the
method described in the previous section:
\begin{align}
  D(i, j) = \text{CMSM\_dist}(i, j).
  \label{Eqn: DMatrix}
\end{align}
Note that the entries of $\mathbf{D}$ do not obey the triangle
inequality, i.e.\ in general: $D(i, j) \nleq D(i, k) + D(i, j)$. For
this reason, we next compute the normalized distance matrix
$\hat{\mathbf{D}}$ using Floyd's algorithm \cite{CormLeisRive1990}:
{
\begin{align}
  \forall k.~\hat{D}(i, j) = \min [D(i, j), \hat{D}(i, k)+\hat{D}(k, j)].
\end{align}}
Finally, we employ a Multi-Dimensional Scaling (MDS) algorithm
(similarly as Tenenbaum \textit{et al.} \cite{TeneSilvLang2000}) on
$\hat{\mathbf{D}}$ to compute the natural embedding of appearance
manifolds under the derived metric. A typical result of embedding is
shown in Figure~\ref{Fig: 2 Classes in Manifold Space}.

\begin{figure}
  \centering
  \includegraphics[width=0.55\textwidth]{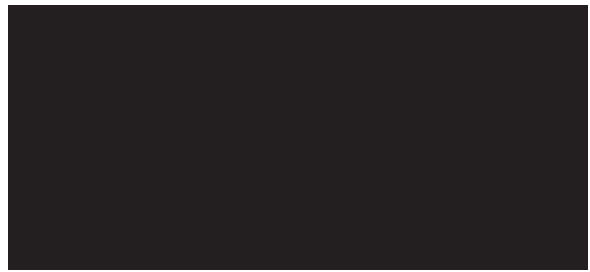}
  \caption[Distance threshold-based clustering: manifold space interpretation.]
      { \it In the manifold space, the usual form of
            clustering -- where manifolds within a certain distance
            (chosen from the ROC curve) from each other are grouped under
            the same class -- corresponds to placing a hyper-spherical
            kernel at each manifold. }
            \label{Fig: Isotropic}
  \vspace{6pt}\hrule
\end{figure}

\paragraph{Anisotropically evolving class boundaries.}
Consider previously mentioned clustering of appearance manifolds
using a particular point on the ROC curve, corresponding to a
distance threshold $d_t$. It is now easy to see that in the
constructed manifold space this corresponds to hyper-spherical class
boundaries of radius $d_t$ centred at each manifold, see
Figure~\ref{Fig: Isotropic}. We now show how to construct
anisotropic class boundaries by considering the distributions of
manifolds. First, (i) simple, isotropic clustering in the manifold
space is performed using the ``high precision'' point on the ROC
curve, then (ii) a single parametric, Gaussian model is fit to each
provisional same-class cluster of manifolds, and finally (iii)
Gaussian models corresponding to the provisional classes are merged
in a pair-wise manner, using a criterion based on the model+data
Description Length \cite{DudaHartStor2001}. The criterion for
class-cluster merging is explained in detail next.


\paragraph{Class-cluster merging.} In the proposed
method, classes are represented by Gaussian clusters in the
implicitly computed manifold space. Initially, the number of
clusters is overestimated, each including only those appearance
manifolds for which the same-class confidence is very high, using
the manifold distance corresponding to the ``high precision'' point
on the CMSM's ROC curve. Then, clusters are pair-wise merged.
Intuitively, if two Gaussian components are quite distant and have
little overlap, not much evidence for each is needed to decide they
represent different classes. The closer they get and the more they
overlap, more supporting manifolds are needed to prevent merging. We
quantify this using what we call the \emph{weighted Description
Length} $DL_w$ and merge tentative classes if $\Delta DL_w <
threshold$ (we used $threshold=-20$).

Let $j$-th of $C$ appearance manifolds be $\mathbf{m}_j$ and let it
consist of $n(j)$ face images. Then we compute the log-likelihood of
$\mathbf{m}_j$ given the Gaussian model $\mathcal{G}(\mathbf{m};
\mathbf{\Theta})$ in the manifold space, weighted by the number of
supporting-samples $n(j)$: {
\begin{align}
  C \sum_{j=1}^{C} n(j) \log P(\mathbf{m}_i|\mathbf{\Theta}) / \sum_{j=1}^{C} n(j)
  \label{Eqn: Weighted Likelihood}
\end{align}}
The weighted Description Length of class data under the same model
then becomes: {
\begin{align}
  DL_w(\mathbf{\Theta}, \{\mathbf{m}_j \}) = &\frac{1}{2} N_E \log_2(n(j))
  - \left[\prod_{j=1}^{C}
     P(\mathbf{m}_i|\mathbf{\Theta})^{n(j)}\right]^{C / \sum n(j)}
  \label{Eqn: Weighted DL}
\end{align}}

\section{Empirical evaluation}
\label{Sec: Evaluation and Results} In this section we report the
empirical results of evaluating the proposed algorithm on the ``Open
Government'' episode of the situation comedy ``Yes,
Minister''\footnote{Available at
\url{http://mi.eng.cam.ac.uk/~oa214/academic/}}. Face detection was
performed on every 5th out of 42,800 frames, producing 7,965
detections, see Figure~\ref{Fig: Data}~(a). A large number of
non-face images is included in this number, see Figure~\ref{Fig:
Data}~(b). Using the method for collecting face motion sequences
described in Section~\ref{SubSec: Automatic Data Acquisition} and
discarding all tracks that contain less than 10 samples removes most
of these. We end up with approximately 300 appearance manifolds to
cluster. The primary and secondary cast consisted of 7 characters:
Sir Hacker, Miss Hacker, Frank, Sir Humphrey, Bernard, a BBC
employee and the PM's secretary.

Baseline clustering performance was established using the CMSM-based
isotropic method with thresholds corresponding to the ``high
recall'' and ``high precision'' points on the ROC curve. Formally,
two manifolds are classified to the same class if the distance $D(i,
j)$ between them is less than the chosen threshold, see \eqref{Eqn:
DMatrix} and Figure~\ref{Fig: Isotropic}. Note that the converse is
not true due to the transitivity of the in-class relation.

\begin{figure}[!t]
  \centering
  \subfigure[]
    {\includegraphics[width=0.75\textwidth]{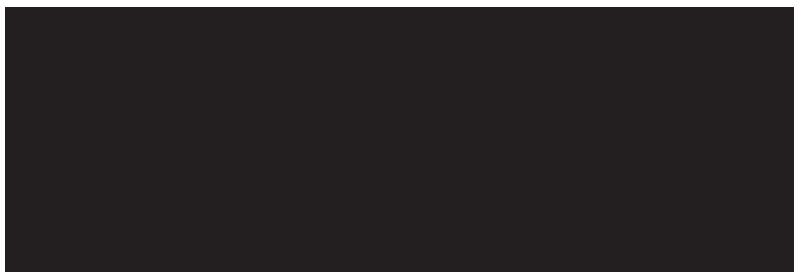}}
  \subfigure[]
    {\includegraphics[width=0.8\textwidth]{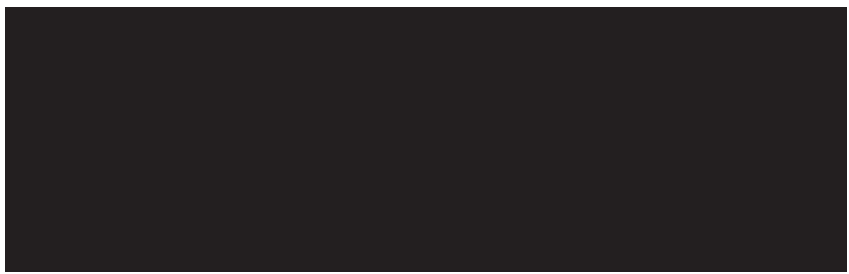}}
  \caption["Yes, Minister" data set.]{ \it (a) The "Yes, Minister" data set -- every 70th
            detection is shown for compactness. A large number of
            non-faces is present, typical of which are shown in (b). }
            \label{Fig: Data}
  \vspace{6pt}\hrule
\end{figure}

\subsection{Results}
The cast listing results using the two baseline isotropic algorithms
are shown in Figure~\ref{Fig: Results Isotropic}~(a) and~\ref{Fig:
Results Isotropic}~(b) -- for each class we displayed a 10 image
sample from its most likely manifold (under the assumption of normal
distribution, see Section~\ref{SubSubSec: Comparing Normalized
Appearance Manifolds}). As expected, the ``high precision'' method
produced a gross overestimate of the number of different individuals
e.g.\ suggesting three classes both for Sir Hacker and Sir Humphrey,
and two for Bernard. Conversely, the ``high recall'' method
underestimates the true number of classes. However, rather more
interestingly, while grouping different individuals under the same
class, this result still contains two classes for Sir Hacker. This
is a good illustration of the main premise of this chapter, showing
that the in-class distance threshold has to be chosen \emph{locally}
in the manifold space, if high clustering accuracy is to be
achieved. That is what the proposed method implicitly does.

\begin{figure}
  \centering
  \small
  \subfigure[]{
  \begin{tabular}{VV}
    Class 01: & \includegraphics[width=0.62\textwidth]{ha_1109__01.eps}\\

    Class 02: & \includegraphics[width=0.62\textwidth]{ha_1109__02.eps}\\

    Class 03: & \includegraphics[width=0.62\textwidth]{ha_1109__03.eps}\\

    Class 04: & \includegraphics[width=0.62\textwidth]{ha_1109__04.eps}\\

    Class 05: & \includegraphics[width=0.62\textwidth]{ha_1109__05.eps}\\

    Class 06: & \includegraphics[width=0.62\textwidth]{ha_1109__06.eps}\\

    Class 07: & \includegraphics[width=0.62\textwidth]{ha_1109__07.eps}\\

    Class 08: & \includegraphics[width=0.62\textwidth]{ha_1109__08.eps}\\

    Class 09: & \includegraphics[width=0.62\textwidth]{ha_1109__09.eps}\\

    Class 10: & \includegraphics[width=0.62\textwidth]{ha_1109__10.eps}\\

    Class 11: & \includegraphics[width=0.62\textwidth]{ha_1109__11.eps}\\

    Class 12: & \includegraphics[width=0.62\textwidth]{hp_1110__01.eps}\\

    Class 13: & \includegraphics[width=0.62\textwidth]{ha_1109__12.eps}\\
    \vspace{-5pt}
  \end{tabular}}

  \subfigure[]{
  \begin{tabular}{VV}
    Class 01: & \includegraphics[width=0.62\textwidth]{hr_1109__01.eps}\\

    Class 02: & \includegraphics[width=0.62\textwidth]{hr_1109__02.eps}\\

    Class 03: & \includegraphics[width=0.62\textwidth]{hr_1109__03.eps}\\

    Class 04: & \includegraphics[width=0.62\textwidth]{hr_1109__04.eps}\\
    \vspace{-5pt}
  \end{tabular}}

  \caption[Isotropic clustering results.]
      { \it (a) ``High precision'' and (b) ``high recall'' point
            isotropic clustering results. The former vastly overestimates
            the number of cast members (e.g.\ classes 01, 03 and 13 correspond
            to the same individual), while the latter underestimates
            it. Both methods fail to distinguish between inter- and intra-personal
            changes of appearance manifolds. }
            \label{Fig: Results Isotropic}
\end{figure}

The cast listing obtained with anisotropic clustering is shown in
Figure~\ref{Fig: Results}. For each class we displayed 10 images
from the highest likelihood sequence. It can be seen that the our
method correctly identified the main cast of the film. No characters
are `repeated', unlike in both Figure~\ref{Fig: Results
Isotropic}~(a) and Figure~\ref{Fig: Results Isotropic}~(b). This
shows that the proposed algorithm for growing class boundaries in
the manifold space has implicitly learnt to distinguish between
intrinsic and extrinsic variations \emph{between appearance
manifolds}. Figure~\ref{Fig: One Class} corroborates this
conclusion.

\begin{figure}
  \centering
  \small
  \begin{tabular}{VV}
    Sir Hacker:\hspace{6pt} & \includegraphics[width=0.70\textwidth]{msm_1111__01.eps}\\

    Miss Hacker: & \includegraphics[width=0.70\textwidth]{msm_1111__02.eps}\\

    Humphrey:\hspace{7pt} & \includegraphics[width=0.70\textwidth]{msm_1111__03.eps}\\

    Secretary:\hspace{11pt} & \includegraphics[width=0.70\textwidth]{msm_1111__04.eps}\\

    Bernard:\hspace{16pt} & \includegraphics[width=0.70\textwidth]{msm_1111__05.eps}\\

    Frank:\hspace{23pt} & \includegraphics[width=0.70\textwidth]{amsm_1111__01.eps}
  \end{tabular}\\\vspace{8pt}
  \caption[Anisotropic manifold space clustering results.]
      { \it Anisotropic clustering results -- shown are 10
            frame sequences from appearance manifolds most ``representative''
            of the obtained classes  (i.e.\ the highest likelihood ones in the
            manifold space). Our method has correctly identified
            6 out of 7 primary and secondary cast members, without
            suffering from the problems of the two isotropic algorithms
            see Figure~\ref{Fig: Results Isotropic} and Figure~\ref{Fig: One Class}.
            }
            \label{Fig: Results}
  \vspace{6pt}\hrule
\end{figure}

\begin{figure}
  \centering
  \includegraphics[width=0.70\textwidth]{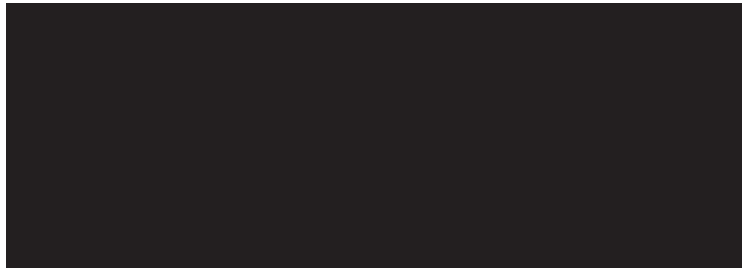}
  \caption[Examples of ``Sir Humphrey'' cluster tracks.]
      { \it Examples from the ``Sir Humphrey'' cluster --
            each horizontal strip is a 10 frame sample from a single
            face track. Notice a wide range of appearance changes between
            different tracks:
            extreme illumination conditions, pose and facial expression
            variation. The bottom-most strip corresponds to an incorrectly
            clustered track of ``BBC employee''. }
            \label{Fig: One Class}
\end{figure}

An inspection of the results revealed a particular failure mode of
the algorithm, also predicted from the theory presented in previous
sections. Appearance manifolds corresponding to the ``BBC employee''
were classified to the class dominated by Sir Humphrey, see
Figure~\ref{Fig: One Class}. The reason for this is a relatively
short appearance of this character, producing a small number of
corresponding face tracks. Consequently, with reference to
\eqref{Eqn: Weighted Likelihood} and \eqref{Eqn: Weighted DL}, not
enough evidence was present to maintain them as a separate class. It
is important to note, however, that qualitatively speaking this is a
tradeoff inherent to the problem in question. Under an assumption of
isotropic noise in image space, \emph{any} class in the film's cast
can generate \emph{any} possible appearance manifold -- it is enough
evidence for each class that makes good clustering possible.

\section{Summary and conclusions}
A novel clustering algorithm was proposed to automatically determine
the cast of a feature-length film, without any dataset-specific
training information. The coherence of inter- and intra-personal
dissimilarities between appearance manifolds was exploited by
mapping each manifold into a single point in the manifold space.
Hence, clustering was performed on actual appearance manifolds. A
mixture-based generative model was used to anisotropically grow
class boundaries corresponding to different individuals. Preliminary
evaluation results showed a dramatic improvement over traditional
clustering approaches.

\section*{Related publications}

The following publications resulted from the work presented in this
chapter:

\begin{itemize}
  \item O. Arandjelovi\'c and R. Cipolla. Automatic cast listing in feature-length films with
                  anisotropic manifold space. In \textit{Proc. IEEE Conference on Computer Vision
                  Pattern Recognition (CVPR)}, \textbf{2}:pages 1513--1520, June
                  2006. \cite{AranCipo2006c}
\end{itemize}

\part{Conclusions, Appendices and Bibliography}

\graphicspath{{./12conc/}}
\chapter{Conclusion}
\label{Chp: Conc}
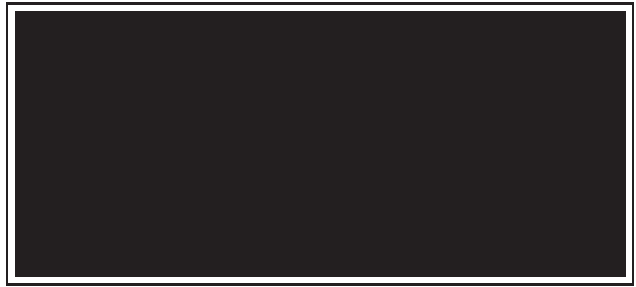
\begin{center}
  \vspace{-20pt}
  \footnotesize
  \framebox{\includegraphics[width=0.58\textwidth]{title_img.eps}}\\
  Pieter P. Rubens. \textit{The Last Judgement}\\
  1617, Oil on canvas, 606 x 460 cm\\
  Alte Pinakothek, Munich\\
\end{center}

\cleardoublepage

This chapter summarizes the thesis. Firstly, we briefly highlight
the main contributions of the presented research. We then focus on
the two major conceptual and algorithmic novelties -- the
\textit{Generic Shape-Illumination Manifold} recognition and the
\textit{Anisotropic Manifold Space Clustering} method. The
significance of the contributions of these two algorithms to the
face recognition field are considered in more detail. Finally, we
discuss the limitations of the proposed methods and conclude the
chapter with an outline of promising directions for future research.

\section{Contributions}
Each of the chapters~\ref{Chp: MDD} to~\ref{Chp: Clustering} and
appendices~\ref{App: TCGMM} to~\ref{App: CamFace} was the topic of a
particular contribution to the field of face recognition. For
clarity, these are briefly summarized in Figure~\ref{Fig: Conts}.

\begin{figure}[!h]
  \Large
  \begin{tabularx}{1.00\textwidth}{rX}
    \normalsize \bf ~Chapter~\ref{Chp: MDD} & \normalsize
          Statistical recognition algorithm suitable in the case when
          training data contains typical appearance variations.\\
    \normalsize \bf ~Chapter~\ref{Chp: KRAD} & \normalsize
          Appearance matching by nonlinear manifold unfolding, in the
          presence of varying pose, noise contamination, face detector
          outliers and mild illumination changes.\\
    \normalsize \bf ~Chapter~\ref{Chp: Thermal} & \normalsize
          Illumination invariant recognition by decision-level fusion
          of optical and infrared thermal imagery.\\
    \normalsize \bf ~Chapter~\ref{Chp: Filters} & \normalsize
          Illumination invariant recognition by decision-level fusion
          of raw grayscale and image filter preprocessed visual data.\\
    \normalsize \bf ~Chapter~\ref{Chp: BoMPA} & \normalsize
          Derivation of a local appearance manifold illumination
          invariant, exploited in the proposed learning-based nonlinear
          extension of canonical correlations between subspaces.\\
    \normalsize \bf ~Chapter~\ref{Chp: Auth} & \normalsize
          Person identification system based on combining appearance
          manifolds with a simple illumination and pose model.\\
    \normalsize \bf ~Chapter~\ref{Chp: gSIM} & \normalsize
          Unified framework for data-driven learning and model-based
          appearance manifold matching in the presence of large pose,
          illumination and motion pattern variations.\\
    \normalsize \bf ~Chapter~\ref{Chp: Char} & \normalsize
          Content-based video retrieval based on face recognition;
          fine-tuned facial registration, accurate background clutter
          removal and robust distance for partial face occlusion.\\
    \normalsize \bf ~Chapter~\ref{Chp: Clustering} & \normalsize
          Automatic identity-based clustering of tracked people in
          feature-length videos; the manifold space concept.\\
    \normalsize \bf ~Appendix~\ref{App: TCGMM} & \normalsize
          Concept of Temporally-Coherent Gaussian mixtures and algorithm
          for their incremental fitting.\\
    \normalsize \bf ~Appendix~\ref{App: MPMM} & \normalsize
          Probabilistic extension of canonical correlation-based
          pattern recognition by subspace matching.\\
    \normalsize \bf ~Appendix~\ref{App: CamFace} & \normalsize
          Algorithm for automatic extraction of faces and background
          removal from cluttered video scenes.\\
          & \\
  \end{tabularx}
  \caption[Thesis contributions.]
    { \it A summary of the contributions of this thesis.}
  \label{Fig: Conts}
  \vspace{6pt}\hrule
\end{figure}

We now describe the two main contributions of the thesis in more
detail, namely the \textit{Generic Shape-Illumination Manifold}
method of Chapter~\ref{Chp: gSIM} and the \textit{Anisotropic
Manifold Space} clustering algorithm of Chapter~\ref{Chp:
Clustering}.

\subsection*{Generic Shape-Illumination Manifold algorithm} Starting with
Chapter~\ref{Chp: MDD} and concluding with Chapter~\ref{Chp: Auth}
we considered the problem of matching video sequences of faces,
gradually decreasing restrictions on the data acquisition process
and recognizing using less training data. This ended in our
proposing the \textit{Generic Shape-Illumination Manifold}
algorithm, in detail described in Chapter~\ref{Chp: gSIM}. The
algorithm was shown to be extremely successful (nearly perfectly
recognizing all individuals) on a large data set of over 1300 video
sequences in realistic imaging conditions. Repeated explicitly, by
this we mean that recognition is performed in the presence of: (i)
large pose variations, (ii) extreme illumination conditions
(significant non-Lambertian effects), (iii) large illumination
changes, (iv) uncontrolled head motion pattern, and (v) low video
resolution.

Our algorithm was shown to greatly outperform current
state-of-the-art face recognition methods in the literature and the
best performing commercial software. This is the result of the
following main novel features:
\begin{enumerate}
  \item
    Combination of data-driven machine learning and prior
    knowledge-based photometric model,
  \item
    Concept of the Generic Shape-Illumination Manifold as a way
    of compactly representing complex illumination effects across all
    human faces (illumination robustness),
  \item
    Video sequence re-illumination algorithm, used to learn the
    Generic Shape-Illumination Manifold (low resolution robustness),
    and
  \item
    Automatic selection of the most reliable faces on which to base
    the recognition decision (pose and outlier robustness).
\end{enumerate}

\subsection*{Anisotropic Manifold Space clustering}
The last two chapters of this thesis considered face recognition in
feature-length films, for the purpose of content-based retrieval and
organization. The \textit{Anisotropic Manifold Space} clustering
algorithm was proposed to automatically determine the cast of a
feature-length film, without any dataset-specific training
information.

Preliminary evaluation results on an episode of the situation comedy
``Yes, Minister'' were vastly superior to those of conventional
clustering methods. The power of the proposed approach was
demonstrated by showing that the correct cast list was produced even
using a very simple algorithm for normalizing images of faces and
comparing individual manifolds. The key novelties are:
\begin{enumerate}
  \item
    Clustering over appearance manifolds themselves, which were
    automatically extracted from a continuous video stream,
  \item
    Concept of the \textit{manifold space} -- a vector space in which
    each point is an appearance manifold,
  \item
    Iterative algorithm for estimating the optimal discriminative
    subspace for an unlabelled dataset, given the \emph{generic}
    discriminative subspace,
    and
  \item
    A hierarchial manifold space clustering algorithm based on
    the proposed appearance manifold-driven \emph{weighted
    description length} and an underlying generative mixture model.
\end{enumerate}

\section{Future work}
We conclude the thesis with a discussion on the most promising
avenues for further research that the work presented has opened up.
We will again focus on the two major contributions of this work, the
\textit{Generic Shape-Illumination Manifold} method of
Chapter~\ref{Chp: gSIM} and the \textit{Anisotropic Manifold Space}
clustering algorithm of Chapter~\ref{Chp: Clustering}.

\subsection*{Generic Shape-Illumination Manifold algorithm}
The proposed \textit{Generic Shape-Illumination Manifold} method has
immediate potential for improvement in the following three areas:
\begin{enumerate}
  \item[i]   Computational efficiency,
  \item[ii]  Manifold representation, and
  \item[iii] Partial occlusion and facial expression changes.
\end{enumerate}

In Section~\ref{gSIM: SubSec: Results} we analyzed the computational
complexity and the running time of our implementation of the
algorithm. Empirical results show a computational increase that is
dominated by a term roughly quadratic in the number of detected
faces in a video sequence. The most expensive stages of the method
are the computation of geodesic distances and $K$-nearest
neighbours. While neither of these can be made more asymptotically
efficient (they correspond to the all-pairs shortest path problem
\cite{CormLeisRive1990}), they can be potentially avoided if a
different manifold representation is employed. Possible candidates
are some of the representations used in this thesis:
Chapters~\ref{Chp: MDD} and~\ref{Chp: Auth} showed that Gaussian
mixtures are suitable for modelling face appearance manifolds, while
piece-wise linear models were employed in Chapter~\ref{Chp: MDD}.
Either of these would have the benefit of (i) constant storage
requirements (in our current method, memory needed to represent a
manifold is linear in the number of faces) and (ii) avoidance of the
two most computationally expensive stages in the proposed method.
Additionally, a novel incremental learning approach of such
representations is described in Appendix~\ref{App: TCGMM}.

A more fundamental limitation of the \textit{Generic
Shape-Illumination Manifold} algorithm is its sensitivity to partial
occlusions and facial expression changes. The former is likely an
easier problem to tackle. Specifically, several recent methods for
partial face occlusion detection (e.g.\ \cite{LeeKrie2005,
WillBlakCipo2004}) may prove useful in this regard: by detecting the
occluded region of the face, pose matching and then robust
likelihood estimation can be performed using only the non-occluded
regions by marginalization of the density corresponding to the
Generic SIM. Extending the algorithm to successfully deal with
expression changes is a more challenging problem and a worthwhile
aim for future research.

\subsection*{Anisotropic Manifold Space clustering}
The \textit{Anisotropic Manifold Space} algorithm for clustering of
face appearance manifolds can be extended in the following
directions:
\begin{enumerate}
  \item[i]   More sophisticated appearance matching,
  \item[ii]  The use of local manifold space projection, and
  \item[iii] Discriminative model fitting.
\end{enumerate}
We now summarize these.

With the purpose of decreasing the computational load of empirical
evaluation, as well as demonstrating the power of the introduced
Manifold Space clustering, our implementation of the algorithm in
Chapter~\ref{Chp: Clustering} used a very simple, linear manifold
model with per-frame image filtering-based illumination
normalization. The limitations of both the linear manifold model and
the filtering approach to achieving illumination robustness were
discussed throughout the thesis (e.g.\ see Chapter~\ref{Chp:
Review}). A more sophisticated approach, such as one based on the
proposed \textit{Generic Shape-Illumination Manifold} of
Chapter~\ref{Chp: gSIM} would be the most immediate direction for
improvement.

The proposed \textit{Anisotropic Manifold Space} algorithm applies
MDS to construct an embedding of all appearance manifolds in a
feature-length video. This has the unappealing consequences of (i)
rapidly growing computational load and (ii) decreased accuracy of
the embedding with the increase in the number of manifolds. Both of
these limitations can be overcome by recognizing that very distant
manifolds should not affect mutual clustering membership. Hence, in
the future we intend to investigate ways of automatically \textit{a
priori} partitioning the Manifold Space and unfolding it only a part
at a time i.e.\ locally.

Finally, the clustering algorithm in the Manifold Space is based on
a generative approach with the underlying Gaussian model of class
data. Clustering methods better tuned for discrimination are likely
to prove as more suitable for the task at hand.

\appendix

\graphicspath{{./13igmm/}}
\chapter{Incremental Learning of Temporally-Coherent GMMs }
\label{App: TCGMM}
\begin{center}
  \vspace{-20pt}
  \footnotesize
  \framebox{\includegraphics[width=0.7575\textwidth]{title_img.eps}}\\
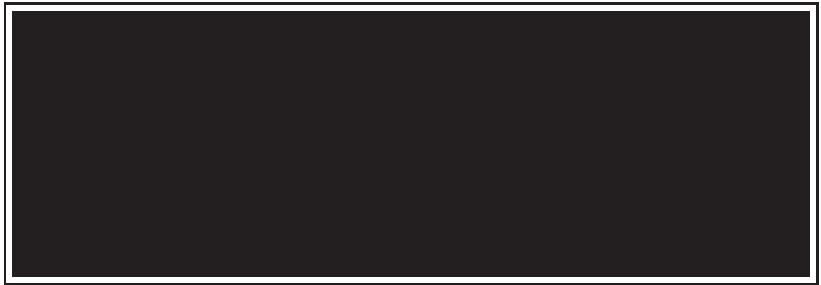
  Vincent Van Gogh. \textit{Basket of Potatoes}\\
  1885, Oil on canvas, 45.0 x 60.5 cm\\
  Van Gogh Museum, Amsterdam
\end{center}

\cleardoublepage

In this appendix we address the problem of learning Gaussian Mixture
Models (GMMs) \emph{incrementally}. Unlike previous approaches which
universally assume that new data comes in blocks representable by
GMMs which are then merged with the current model estimate, our
method works for the case when novel data points arrive
\emph{one-by-one}, while requiring little additional memory. We keep
only two GMMs in memory and no historical data. The current fit is
updated with the assumption that the number of components is fixed,
which is increased (or reduced) when enough evidence for a new
component is seen. This is deduced from the change from the oldest
fit of the same complexity, termed the \emph{Historical GMM}, the
concept of which is central to our method. The performance of the
proposed method is demonstrated qualitatively and quantitatively on
several synthetic data sets and video sequences of faces acquired in
realistic imaging conditions.

\section{Introduction}\label{Sec: Introduction}
The Gaussian Mixture Model (GMM) is a semi-parametric method for
high-dimensional density estimation. It is used widely across
different research fields, with applications to computer vision
ranging from object recognition \cite{DahmKeysNey+2001}, shape
\cite{CootTayl1999} and face appearance modelling
\cite{GrosYangWaib2000} to colour-based tracking and segmentation
\cite{RajaMcKeGong1998}, to name just a few. It is worth
emphasizing the key reasons for its practical appeal: (i) its
flexibility allows for the modelling of complex and nonlinear
pattern variations \cite{GrosYangWaib2000}, (ii) it is simple and
efficient in terms of memory, (iii) a principled model complexity
selection is possible, and (iv) there are theoretically guaranteed
to converge algorithms for model parameter estimation.

Virtually all previous work with GMMs has concentrated on non time
critical applications, typically in which model fitting (i.e.\
model parameter estimation) is performed offline, or using a
relatively small training corpus. On the other hand, the recent
trend in computer vision is oriented towards real-time
applications (for example for human-computer interaction and
on-the-fly model building) and modelling of increasingly complex
patterns which inherently involves large amounts of data. In both
cases, the usual batch fitting becomes impractical and an
incremental learning approach is necessary.

\paragraph{Problem challenges.} Incremental learning of GMMs is a
surprisingly difficult task. One of the main challenges of this
problem is the model complexity selection which is required to be
dynamic by the very nature of the incremental learning framework.
Intuitively, if all information that is available at any time is
the current GMM estimate, a single novel point \emph{never}
carries enough information to cause an increase in the number of
Gaussian components. Another closely related difficulty lies in
the \emph{order} in which new data arrives \cite{HallHick2004}. If
successive data points are always badly correlated, then a large
amount of data has to be kept in memory if accurate model order
update is to be achieved.

\subsection{Related previous work}\label{SubSec: Related Previous Work}
The most common way of fitting a GMM is using the
Expectation-Maximization (EM) algorithm \cite{DempLairRubi1977}.
Starting from an estimate of model parameters, soft membership of
data is computed (the \emph{Expectation} step) which is then used
to update the parameters in the maximal likelihood (ML) manner
(the \emph{Maximization} step). This is repeated until
convergence, which is theoretically guaranteed. In practice,
initialization is frequently performed using the $K$-means
clustering algorithm \cite{Bish1995, DudaHartStor2001}.

\paragraph{Incremental approaches.} Incremental fitting of GMMs has
already been addressed in the machine learning literature. Unlike
the proposed method, most of the existing methods assume that
novel data arrives in \emph{blocks} as opposed to a single datum
at a time. Hall \textit{et al.} \cite{HallMarsMart2000} merge
Gaussian components in a pair-wise manner by considering volumes
of the corresponding hyperellipsoids. A more principled method was
recently proposed by Song and Wang \cite{SongWang2005} who use the
$W$ statistic for covariance and the Hotelling's $T^2$ statistic
for mean equivalence. However, they do not fully exploit the
available probabilistic information by failing to take into
account the \emph{evidence} for each component at the time of
merging. Common to both \cite{HallMarsMart2000} and
\cite{SongWang2005} is the failure to make use of the existing
model when the GMM corresponding to new data is fitted. What this
means is that even if some of the new data is already explained
well by the current model, the EM fitting will try to explain it
in the context of other novel data, affecting the accuracy of the
fit as well as the subsequent component merging. The method of
Hicks \textit{et al.} \cite{HickHallMars2003} (also see
\cite{HallHick2004}) does not suffer from the same drawback. The
authors propose to first ``concatenate'' two GMMs and then
determine the optimal model order by considering models of
\emph{all} low complexities and choosing the one that gives the
largest penalized log-likelihood. A similar approach of combining
Gaussian components was also described by Vasconcelos and Lippman
\cite{VascLipp1998}.

\paragraph{Model order selection.}
Broadly speaking, there are three classes of approaches for GMM
model order selection: (i) EM-based using validation data, (ii)
EM-based using model validity criteria, and (iii) dynamic
algorithms. The first approach involves random partitioning of the
data to training and validation sets. Model parameters are then
iteratively estimated from training data and the complexity that
maximizes the posterior of the validation set is sought. This method
is typically less preferred than methods of the other two groups,
being wasteful both of the data and computation time. The most
popular group of methods is EM-based and uses the posterior of all
data, penalized with model complexity. Amongst the most popular are
the Minimal Description Length (MDL) \cite{Riss1978}, Bayesian
Information (BIC) \cite{Schw1978} and Minimal Message Length (MML)
\cite{WallDowe1999} criteria. Finally, there are methods which
combine the fitting procedure with dynamic model order selection.
Briefly, Zwolinski and Yang \cite{ZwolYang2001}, and Figueredo and
Jain \cite{FiguJain2003} overestimate the complexity of the model
and reduce it by discarding ``improbable'' components. Vlassis and
Likas \cite{VlasLika1999} use a weighted sample kurtoses of Gaussian
kernels, while Verbeek \textit{et al.} introduce a heuristic greedy
approach in which mixture components are added one at the time
\cite{VerbVlasKros2003}.

\section{Incremental GMM estimation}\label{Sec: Incremental GMM Estimation}

A GMM with $M$ components in a $D$-dimensional embedding space is
defined as:
\begin{equation}\label{Eqn: GMM Definition}
    \mathcal{G}\left(\mathbf{x}; \mathbf{\theta}\right) =
    \sum_{j=1}^M \alpha_j \mathcal{N}(\mathbf{x}; \mathbf{\mu}_j, \mathbf{C}_j )
\end{equation}
where $\theta = \left(\{\alpha_i\},
\{\mathbf{\mu}_i\}\{\mathbf{C}_i\}\right)$ is the set of model
parameters, $\alpha_i$ being the prior of the $i$-th Gaussian component
with the mean $\mathbf{\mu}_i$ and covariance $\mathbf{C}_i$:
\begin{equation}\label{Eqn: Gaussian Definition}
    \mathcal{N}(\mathbf{x}; \mathbf{\mu}, \mathbf{C}) =
    \frac {1} { (2\pi)^{D/2} \sqrt{|\mathbf{C}|}}
    \exp{\left( - \frac{1}{2} (\mathbf{x} - \mathbf{\mu})^T \mathbf{C}^{-1}(\mathbf{x} - \mathbf{\mu}) \right) }
\end{equation}

\subsection{Temporally-coherent GMMs}\label{SubSec: Temporally-Coherent GMMs}
\begin{figure}[!h]
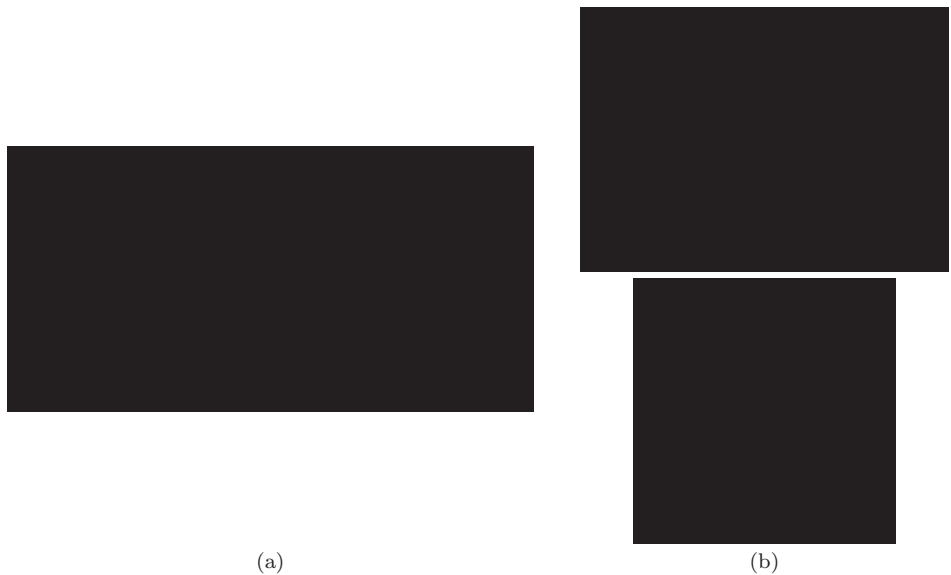

  \footnotesize
  \centering
  \begin{tabular}{VV}
    \includegraphics[width=0.5\textwidth]{BMVC2005_nn_dist_prob.eps} &
    \begin{tabular}{c}
      \includegraphics[width=0.35\textwidth]{BMVC2005_fmm_gmm.eps} \\
      \includegraphics[width=0.25\textwidth]{BMVC2005_faces1.eps} \\
    \end{tabular}\\
    (a) & (b) \\
  \end{tabular}

  \caption[Temporally-coherent GMMs.]
  {\it                  (a) Average distribution of Euclidean distance
                        between temporally consecutive faces across
                        video sequences of faces in unconstrained
                        motion. The distribution peaks at a low,
                        but greater-than-zero distance, which is typical
                        of Temporally-Coherent GMMs analyzed in this
                        appendix. Both too low and too large distances are
                        infrequent, in this case the former due to the
                        time gap between the acquisition of consecutive
                        video frames, the latter due to the smoothness
                        of face shape and texture. (b) A typical sequence
                        projected to the first three principal components
                        estimated from the data, the corresponding MDL
                        EM fit
                        and the component centres visualized as
                        images. On average, we found that over 80\% of
                        pairs of successive faces have the highest
                        likelihood of having been generated by the same Gaussian
                        component. }
            \label{Fig: TC-GMM}
            \vspace{6pt}\hrule
\end{figure}

We assume \emph{temporal coherence} on the order in which data
points are seen. Let $\{\mathbf{x}_t\} \equiv \{\mathbf{x}_0,
\ldots, \mathbf{x}_T \}$ be a stream of data, its temporal
ordering implied by the subscript. The assumption of an underlying
Temporally-Coherent GMM (TC-GMM) on $\{\mathbf{x}_t\}$ is:
\begin{eqnarray*}
  \mathbf{x}_0     &\sim& \mathcal{G}\left(\mathbf{x}; \mathbf{\theta}\right) \\
  \mathbf{x}_{t+1} &\sim& p_S\left(\|\mathbf{x}_{t+1} - \mathbf{x}_t\|\right) \cdot
                          \mathcal{G}\left(\mathbf{x}; \mathbf{\theta}\right)
                          \normalfont
\end{eqnarray*}
where $p_S$ is a unimodal density. Intuitively,
while data is distributed according to an underlying Gaussian
mixture, it is also expected to vary smoothly with time, see
Figure~\ref{Fig: TC-GMM}.

\subsection{Method overview}
The proposed method consists of a three-stage model update each time
a new data point becomes available, see Figure~\ref{Alg: Overview}.
At each time step: (i) model parameters are updated under the
constraint of fixed complexity, (ii) new Gaussian components are
postulated by model splitting and (iii) components are merged to
minimize the expected model description length. We keep in memory
only two GMMs and \emph{no historical data}. One is the current GMM
estimate, while the other is the oldest model od the same complexity
after which no permanent new cluster creation took place -- we term
this the \emph{Historical GMM}.

\begin{figure}[!t]
    \centering
    \begin{tabular}{l}
          \begin{tabular}{ll}
              \textbf{Input}:  & set of observations $\{\mathbf{x}_i\}$,\\
                               & KPCA space dimensionality $D$.\\
              \textbf{Output}: & kernel principal components $\{\mathbf{u}_i\}$.\\
          \end{tabular}\vspace{5pt}
          \\ \hline \\
          \begin{tabular}{l}
            \textbf{1: Fixed-complexity update:}\\
            \hspace{10pt}update$(\mathcal{G}_N, \mathbf{x})$\\\\

            \textbf{2: Model splitting:}\\
            \hspace{10pt}$\mathcal{G}_M$ = split-all$(\mathcal{G}_N, \mathcal{G}_N^{(h)})$\\\\

            \textbf{3: Pair-wise component merging:}\\
            \hspace{10pt}\textbf{for all} $(i,j) \in (1..N, 1..N)$\\\\

            \hspace{15pt}\textbf{4: Expected description length:}\\
            \hspace{25pt}$[L_1, L_2] = $DL$\bigl\{$merge$(\mathcal{G}_M, i, j)$,~split$(\mathcal{G}_M, i, j)\bigr\}$\\\\

            \hspace{15pt}\textbf{5: Complexity update}\\
            \hspace{25pt}$\mathcal{G}_M = L_1 < L_2$~?~merge$(\mathcal{G}_M, i, j)$~:~split$(\mathcal{G}_M, i, j)$\\\\
          \end{tabular}\\\hline
    \end{tabular}
    \caption[Incremental TC-GMM algorithm.]
        {\it A summary of the proposed Incremental TC-GMM algorithm. }
    \vspace{6pt}\hrule
    \label{Alg: Overview}
\end{figure}

\begin{figure}[!t]
  \centering
  \includegraphics[width=0.55\textwidth]{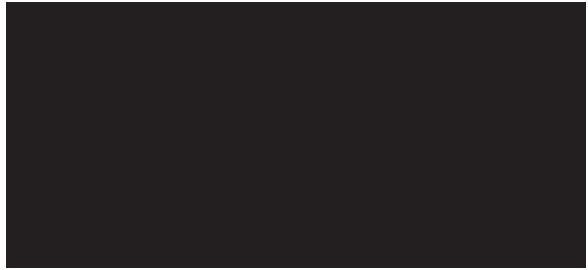}
  \caption[Fixed complexity update.]
    { \it               Fixed complexity update: the mean and the covariance
                        of each Gaussian component are updated according to the
                        probability that it generated the novel observation (red circle).
                        Old covariances are shown as dashed, the updated ones
                        as solid ellipses corresponding to component
                        parameters, while historical data points are displayed as blue dots. }
  \label{Fig: GMM Update}
  \vspace{6pt}\hrule
\end{figure}

\subsection{GMM update for fixed complexity}\label{SubSec: GMM Update for Fixed Complexity}

In the first stage of our algorithm, the current GMM
$\mathcal{G}\left(\mathbf{x}; \mathbf{\theta}\right)$ is updated
under the \emph{constraint of fixed model complexity}, i.e.\ fixed
number of Gaussian components. We start with the assumption that the
current model parameters are estimated in the ML fashion in a local
minimum of the EM algorithm: {
\begin{align}\label{Eqn: Starting Params}
    \alpha_i = \frac {\sum_j p(i|\mathbf{x}_j) } {N} &&
    \mathbf{\mu}_i = \frac {\sum_j \mathbf{x}_j p(i|\mathbf{x}_j) }
    { \sum_j p(i|\mathbf{x}_j) } &&
    \mathbf{C}_i = \frac {\sum_j (\mathbf{x}_j - \mathbf{\mu}_i)(\mathbf{x}_j - \mathbf{\mu}_i)^T p(i|\mathbf{x}_j) }
    { \sum_j p(i|\mathbf{x}_j) }
\end{align}}
where $p(i|\mathbf{x}_j)$ is the probability of the $i$-th component
conditioned on data point $\mathbf{x}_j$. Similarly, for the updated
set of GMM parameters $\mathbf{\theta}^{*}$ it holds:
{\begin{align}\label{Eqn: Update Params1}
    &\alpha_i^{*} = \frac {\sum_j p^{*}(i|\mathbf{x}_j) + p^{*}(i|\mathbf{x}) } {N+1}
    \quad \quad
    \mathbf{\mu}_i^{*} = \frac {\sum_j \mathbf{x}_j p^{*}(i|\mathbf{x}_j) + \mathbf{x} p^{*}(i|\mathbf{x})}
    { \sum_j p^{*}(i|\mathbf{x}_j) + p^{*}(i|\mathbf{x}) } \\
    &\mathbf{C}_i^{*} = \frac {\sum_j (\mathbf{x}_j - \mathbf{\mu}_i^{*})(\mathbf{x}_j - \mathbf{\mu}_i^{*})^T p^{*}(i|\mathbf{x}_j) +
     (\mathbf{x} - \mathbf{\mu}_i^{*})(\mathbf{x} - \mathbf{\mu}_i^{*})^T p^{*}(i|\mathbf{x})}
    { \sum_j p^{*}(i|\mathbf{x}_j) + p^{*}(i|\mathbf{x}) }
\end{align}}

The key problem is that the probability of each component
conditioned on the data changes \emph{even for historical data}
$\{ \mathbf{x}_j \}$. In general, the change in conditional
probabilities can be arbitrarily large as the novel observation
$\mathbf{x}$ can lie anywhere in the $\mathbb{R}^D$ space.
However, the expected correlation between temporally close points,
governed by the underlying TC-GMM model allows us to make the
assumption that component likelihoods do not change much with the
inclusion of novel information in the model:
\begin{equation}\label{Eqn: Component Prob}
     p^{*}(i|\mathbf{x}_j) =  p(i|\mathbf{x}_j)
\end{equation}
This assumption is further justified by the two stages of our
algorithm that follow (Sections~\ref{SubSec: Model Splitting}
and~\ref{SubSec: Theoretically Rigourous Merging Framework}) -- a
large change in probabilities $p(i|\mathbf{x}_j)$ occurs only when
novel data is not well explained by the current model. When enough
evidence for a new Gaussian components is seen, model complexity is
increased, while old component parameters switch back to their
original value. Using \eqref{Eqn: Component Prob}, a simple
algebraic manipulation of \eqref{Eqn: Starting Params}-\eqref{Eqn:
Update Params1}, omitted for clarity, and writing $\sum_j
p(i|\mathbf{x}_j) \equiv E_i$, leads to the following: {
\begin{align}
    &\alpha_i^{*} = \frac {E_i + p(i|\mathbf{x}) } {N+1}
    \quad \quad
    \mathbf{\mu}_i^{*} = \frac { \mathbf{\mu}_i E_i + \mathbf{x} p(i|\mathbf{x}) }
    { E_i + p(i|\mathbf{x}) } \\
    &\mathbf{C}_i^{*} = \frac { (\mathbf{C}_i + \mathbf{\mu}_i \mathbf{\mu}_i^T -
            \mathbf{\mu}_i \mathbf{\mu}_i^{*T} - \mathbf{\mu}_i^{*} \mathbf{\mu}_i^T +
            \mathbf{\mu}_i^{*} \mathbf{\mu}_i^{*T}) E_i +
     (\mathbf{x} - \mathbf{\mu}_i^{*})(\mathbf{x} - \mathbf{\mu}_i^{*})^T
    p(i|\mathbf{x})} { E_i + p(i|\mathbf{x}) }
    \label{Eqn: Update Params2}
\end{align}}
It can be seen that the update equations depend only on the
parameters of the old model and the sum of component likelihoods,
but \emph{no historical data}. Therefore the additional memory
requirements are of the order $O(M)$, where $M$ is the number of
Gaussian components. Constant-complexity model parameter update is
illustrated in Figure~\ref{Fig: GMM Update}.

\subsection{Model splitting}\label{SubSec: Model Splitting}
\begin{figure}[t]
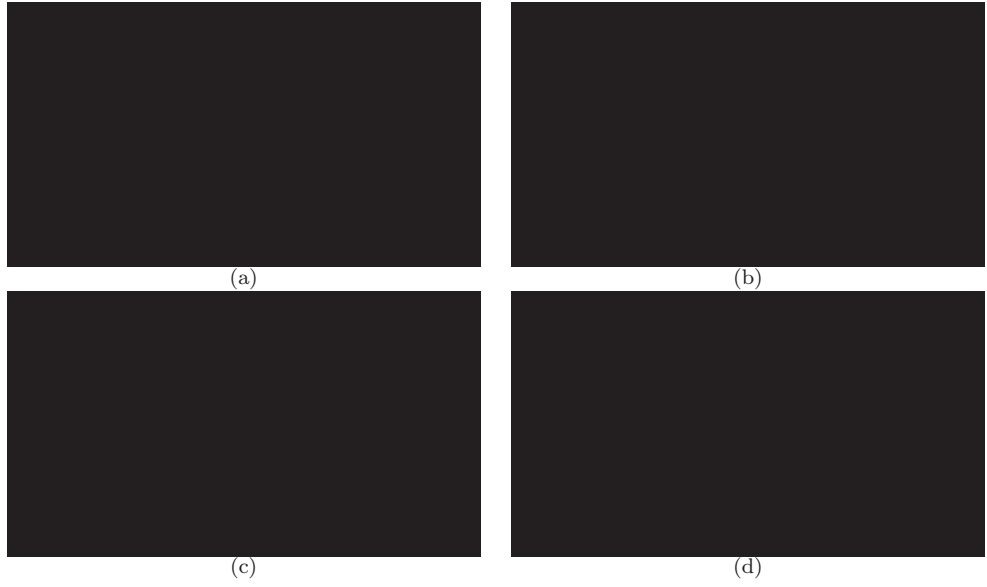

  \footnotesize
  \centering
  \begin{tabular}{VV}
    \includegraphics[width=0.45\textwidth]{BMVC2005_GMM_init2_new.eps} &
    \includegraphics[width=0.45\textwidth]{BMVC2005_split_before2_new.eps} \\
    (a) & (b)  \\
    \includegraphics[width=0.45\textwidth]{BMVC2005_split2_new.eps} &
    \includegraphics[width=0.45\textwidth]{BMVC2005_split_DLs2.eps}\\
    (c) & (d) \\
  \end{tabular}

  \caption[Dynamic model order selection.]
    { \it   Dynamic model order selection: (a)
                        Historical GMM. (b) Current GMM before the arrival
                        of novel data. (c) New data point (red circle)
                        causes the splitting of a Gaussian component,
                        resulting in a 3-component mixture. (d) The
                        contribution to the expected model description
                        length for merging and splitting of the component,
                        as the number of novel data points is increased.
                        }
            \label{Fig: GMM Splitting}
            \vspace{6pt}\hrule
\end{figure}

One of the greatest challenges of incremental GMM learning is the
dynamic model order selection. In the second stage of our algorithm,
new Gaussian clusters are postulated based on the parameters of the
current parameter model estimate $\mathcal{G}$ and the
\emph{Historical GMM} $\mathcal{G}^{(h)}$, which is central to our
idea. As, by definition, no permanent model order changes occurred
between the Historical and the current GMMs, they have the same
number of components and, importantly, the 1-1 correspondence
between them is known (the current GMM is merely the Historical GMM
that was updated under the constraint of fixed model complexity).
Therefore, for each pair of corresponding components
$(\mathbf{\mu}_i,\mathbf{C}_i)$ and
$(\mathbf{\mu}_i^{(h)},\mathbf{C}_i^{(h)})$ we compute the
`difference' component, see Figure~\ref{Fig: GMM Splitting}~(a-c).
Writing \eqref{Eqn: Starting Params} for the Historical and the
current GMMs, and using the assumption in \eqref{Eqn: Component
Prob} the $i$-th difference component parameters become: {
\begin{align}
    &\alpha_i^{(n)} = \frac {E_i - E_i^{(h)}} { N - N^{(h)} }
    \quad \quad
    \mathbf{\mu}_i^{(n)} = \frac { \mathbf{\mu}_i E_i -
    \mathbf{\mu}_i^{(h)} E_i^{(h)} }
    { E_i - E_i^{(h)} } \label{Eqn: Mean Difference}\\
    &\mathbf{C}_i^{(n)} = \frac { \mathbf{C}_i E_i -
    ( \mathbf{C}_i^{(h)} + \mathbf{\mu}_i^{(h)}{\mathbf{\mu}_i^{(h)}}^T )
      E_i^{(h)} +
    ( \mathbf{\mu}_i^{(h)}\mathbf{\mu}_i^T + \mathbf{\mu}_i{\mathbf{\mu}_i^{(h)}}^T)
      E_i^{(h)} -
    \mathbf{\mu}_i \mathbf{\mu}_i^T E_i
    }
    { E_i - E_i^{(h)} } + \label{Eqn: Cov Difference} \\ \notag
    & \quad \quad
      \mathbf{\mu}_i^{(n)}\mathbf{\mu}_i^T + \mathbf{\mu}_i{\mathbf{\mu}_i^{(n)}}^T
    - \mathbf{\mu}_i^{(n)} {\mathbf{\mu}_i^{(n)}}^T
\end{align}}

\subsection{Component merging}
\label{SubSec: Theoretically Rigourous Merging Framework}

In the proposed method, dynamic model complexity estimation is based
on the MDL criterion. Briefly, MDL assigns to a model a cost related
to the amount of information necessary to encode the model and the
data \emph{given} the model. This cost, known as the description
length $L(\mathbf{\theta}|\{\mathbf{x}_i\})$, is equal to the
log-likelihood of the data under that model penalized by the model
complexity, measured as the number of free parameters $N_E$:
\begin{equation}\label{Eqn: DL for GMMs}
    L\left(\mathbf{\theta}|\{\mathbf{x}_i\}\right) = \frac{1}{2} N_E \log_2(N) -
    \log_2 P\left(\{\mathbf{x}_i\}|\mathbf{\theta}\right)
\end{equation}
In the case of an $M$-component GMM with full
covariance matrices in $\mathbb{R}^D$ space, free parameters are
$(M-1)$ for priors, $MD$ for means and $MD(D+1)/2$ for covariances:
\begin{equation}
    N_E = M-1 + MD + M \frac {D(D+1)}{2}
\end{equation}

The problem is that for the computation of
$P\left(\{\mathbf{x}_i\}|\mathbf{\theta}\right)$ historical data
$\{\mathbf{x}_i\}$ is needed -- which is unavailable. Instead of
$P\left(\{\mathbf{x}_i\}|\mathbf{\theta}\right)$, we propose to
compute the \emph{expected} likelihood of the same number of data
points and, hence, use the expected description length as the model
order selection criterion. Consider two components with the
corresponding multivariate Gaussian densities $p_1(\mathbf{x}) \sim
\mathcal{N}(\mathbf{x}; \mathbf{\mu}_1, \mathbf{C}_1)$ and
$p_2(\mathbf{x}) \sim \mathcal{N}(\mathbf{x}; \mathbf{\mu}_2,
\mathbf{C}_2)$. The expected likelihood of $N_1$ points drawn from
the former and $N_2$ from the latter given model $\alpha_1
p_1(\mathbf{x}) + \alpha_2 p_2(\mathbf{x})$ is: {
\begin{align}\label{Eqn: Expected Evidence S}
    E\left[P(\{\textbf{x}_j\}|\mathbf{\theta}_S)\right] = \left(\int p_1(\textbf{x}) (\alpha_1
    p_1(\textbf{x}) + \alpha_2
    p_2(\textbf{x}))d\textbf{x}\right)^{N_1} \notag \\
    \left(\int p_2(\textbf{x}) (\alpha_1
    p_1(\textbf{x}) + \alpha_2 p_2(\textbf{x}))d\textbf{x}\right)^{N_2}
\end{align}}
where integrals of the type $\int p_i(\mathbf{x})p_j(\mathbf{x})
d\mathbf{x}$ are recognized as related to the Bhattacharyya
distance, and are for Gaussian distributions easily computed as: {
\begin{align}\label{Eqn: Bhattacharyya 1}
    d_B(p_i, p_j) = \int
    p_i(\mathbf{x})p_j(\mathbf{x}) d\mathbf{x} =
    \frac {\exp (-K/2)} {(2\pi)^{D/2}|\textbf{C}_i \textbf{C}_j \textbf{C}|^{1/2}}
\end{align}}
where:
\begin{align}\label{Eqn: Bhattacharyya 2}
    &\textbf{C} = \left(\textbf{C}_i^{-1} + \textbf{C}_j^{-1}\right)^{-1} \\
    &\mathbf{\mu} = \textbf{C} (\textbf{C}_i^{-1} \mathbf{\mu}_i + \textbf{C}_j^{-1} \mathbf{\mu}_j) \\
    &K = \mathbf{\mu}_i \mathbf{C}_i^{-1} \mathbf{\mu}_i^T +
    \mathbf{\mu}_j \mathbf{C}_j^{-1} \mathbf{\mu}_j^T - \mathbf{\mu} \mathbf{C}^{-1} \mathbf{\mu}^T
\end{align}

On the other hand, consider the case when the two components are
merged i.e.\ replaced by a single Gaussian component with the
corresponding density $p(\mathbf{x})$. Then we compute the
expected likelihood of $N_1$ points drawn from $p_1(\mathbf{x})$
and $N_2$ points drawn from $p_2(\mathbf{x})$, given model
$p(\mathbf{x})$:
\begin{equation}
    E\left[P(\{\mathbf{x}_j\}|\mathbf{\theta}_M)\right] = \left(\int p(\mathbf{x}) p_1(\mathbf{x})
    d\mathbf{x}\right)^{N_1} \cdot \left(\int p(\mathbf{x}) p_2(\mathbf{x})
    d\mathbf{x}\right)^{N_2}
    \label{Eqn: Expected Evidence M}
\end{equation}
Substituting the expected evidence and model complexity in
\eqref{Eqn: DL for GMMs} we get:
\begin{align}\label{Eqn: Expected DLs}
    \Delta E[L] = E[L_S] - E[L_M] = \frac{1}{4} D(D+1) \log_2(N_1+N_2)
    - \notag \\
    \log_2 E[P(\{\mathbf{x}_j\}|\mathbf{\theta}_S)] + \log_2 E[P(\{\mathbf{x}_j\}|\mathbf{\theta}_M)]
\end{align}
Then the condition for merging is simply $\Delta E[L] > 0$, see
Figure~\ref{Fig: GMM Splitting}~(d). Merging equations are virtually
the same as \eqref{Eqn: Mean Difference} and \eqref{Eqn: Cov
Difference} for model splitting, so we do not repeat them.

\section{Empirical evaluation}
The proposed method was evaluated on several synthetic data sets
and video sequences of faces in unconstrained motion, acquired in
realistic imaging conditions and localized using the Viola-Jones
face detector \cite{ViolJone2004}, see Figure~\ref{Fig:
TC-GMM}~(b). Two synthetic data sets that we illustrate its
performance on are:

\begin{enumerate}
    \item 100 points generated from a Gaussian with a diagonal
    covariance matrix in radial coordinates: $r \sim
    \mathcal{N}(\overline{r} = 5, \sigma_r = 0.1)$, $\phi \sim
    \mathcal{N}(\overline{\phi} = 0, \sigma_{\phi} = 0.7)$
    \item 80 points generated from a uniform distribution in $x$
    and a Gaussian noise perturbed sinusoid in $y$ coordinate :
    $x \sim \mathcal{U}(\min x = 0, \max x = 10)$, $y \sim
    \mathcal{N}(\overline{y} = \sin x, \sigma_{y} = 0.1)$
\end{enumerate}
Temporal ordering was imposed by starting from the data point with
the minimal $x$ coordinate and then iteratively choosing as the
successor the nearest neighbour out of yet unused points. The
initial GMM parameters, the final fitting results and the
comparison with the MDL-EM fitting are shown in Figure~\ref{Fig:
Synthetic Sets}. In the case of face motion video sequences,
temporal ordering of data is inherent in the acquisition process.
An interesting fitting example is shown and compared with the
MDL-EM batch approach in Figure~\ref{Fig: Face Set}.

Qualitatively, both in the case of synthetic and face data it can be
seen that our algorithm consistently produces meaningful GMM estimates.
Quantitatively, the results are comparable with the widely accepted EM
fitting with the underlying MDL criterion, as witnessed by the description
lengths of the obtained models.

\paragraph{Failure modes.} On our data sets two types
phenomena in data sometimes caused unsatisfactory fitting results.
The first, one inherently problematic to our algorithm, is when
newly available data is well explained by the Historical GMM.
Referring back to Section~\ref{SubSec: Model Splitting}, it can be
seen in \eqref{Eqn: Mean Difference} and \eqref{Eqn: Cov
Difference} that this data contributes to the confidence of
creating a \emph{new} GMM component whereas it should not. The
second failure mode was observed when the assumption of temporal
coherence (Section~\ref{SubSec: Temporally-Coherent GMMs}) was
violated, e.g.\ when our face detector failed to detect faces in
several consecutive video frames. While this cannot be considered
an inherent fault of our algorithm, it does point out that
ensuring temporal coherence of data is not always a trivial task
in practice.

In  conclusion, while promising, a more comprehensive evaluation
on different sets of real data is needed to fully understand the
behaviour of the proposed method.

\begin{figure}
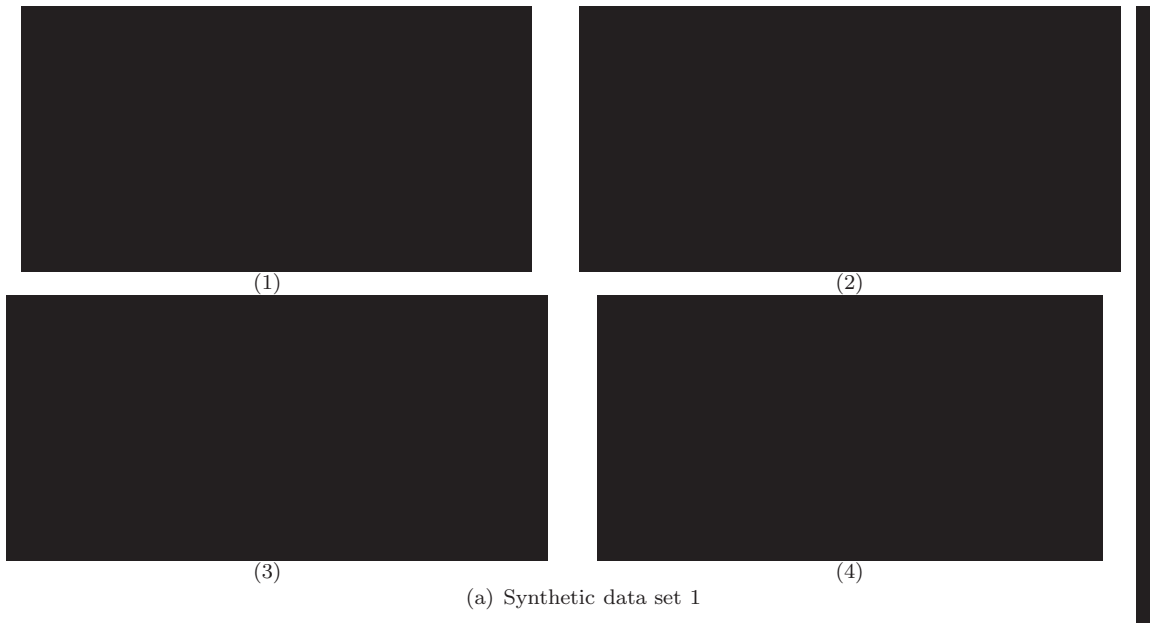

  \footnotesize
  \centering

  \subfigure[Synthetic data set 1]{
  \begin{tabular}{VV}
    \hspace{-18pt}\includegraphics[width=.485\textwidth]{BMVC2005_GMM_init2_new.eps} &
    \includegraphics[width=.515\textwidth]{BMVC2005_GMM_em2.eps} \\
    \hspace{-18pt}(1) & (2) \\
    \hspace{-18pt}\includegraphics[width=.515\textwidth]{BMVC2005_GMM_inc2.eps} &
    \includegraphics[width=.48\textwidth]{BMVC2005_DLs2.eps} \\
    \hspace{-18pt}(3) & (4) \\
  \end{tabular}}
  \caption[Evaluation: synthetic data.]{ \it    Synthetic data: (1) data (dots) and
                         the initial model (visualized as ellipses corresponding
                         to the parameters of the Gaussian components). (2)
                         MDL-EM GMM fit. (3) Incremental GMM fit. (4) Description
                         length of GMMs fitted using EM and the proposed
                         incremental algorithm (shown is the description
                         length of the final GMM estimate). Our method produces
                         qualitatively meaningful results which are also
                         qualitatively comparable with the best fits obtained using
                         the usual batch method. }
            \label{Fig: Synthetic Sets}
\end{figure}

\begin{figure}
  \ContinuedFloat
  \footnotesize
  \centering

  \subfigure[Synthetic data set 2]{
  \begin{tabular}{VV}
    \hspace{-18pt}\includegraphics[width=.515\textwidth]{BMVC2005_GMM_init3.eps} &
    \includegraphics[width=.492\textwidth]{BMVC2005_GMM_em3.eps} \\
    \hspace{-18pt}(1) & (2) \\
    \hspace{-18pt}\includegraphics[width=.515\textwidth]{BMVC2005_GMM_inc3.eps} &
    \includegraphics[width=.48\textwidth]{BMVC2005_DLs3.eps} \\
    \hspace{-18pt}(3) & (4) \\
  \end{tabular}}

  \caption[Evaluation: synthetic data.]{ \it    Synthetic data: (1) data (dots) and
                         the initial model (visualized as ellipses corresponding
                         to the parameters of the Gaussian components). (2)
                         MDL-EM GMM fit. (3) Incremental GMM fit. (4) Description
                         length of GMMs fitted using EM and the proposed
                         incremental algorithm (shown is the description
                         length of the final GMM estimate). Our method produces
                         qualitatively meaningful results which are also
                         qualitatively comparable with the best fits obtained using
                         the usual batch method. }
            \label{Fig: Synthetic Sets}
\end{figure}

\begin{figure}
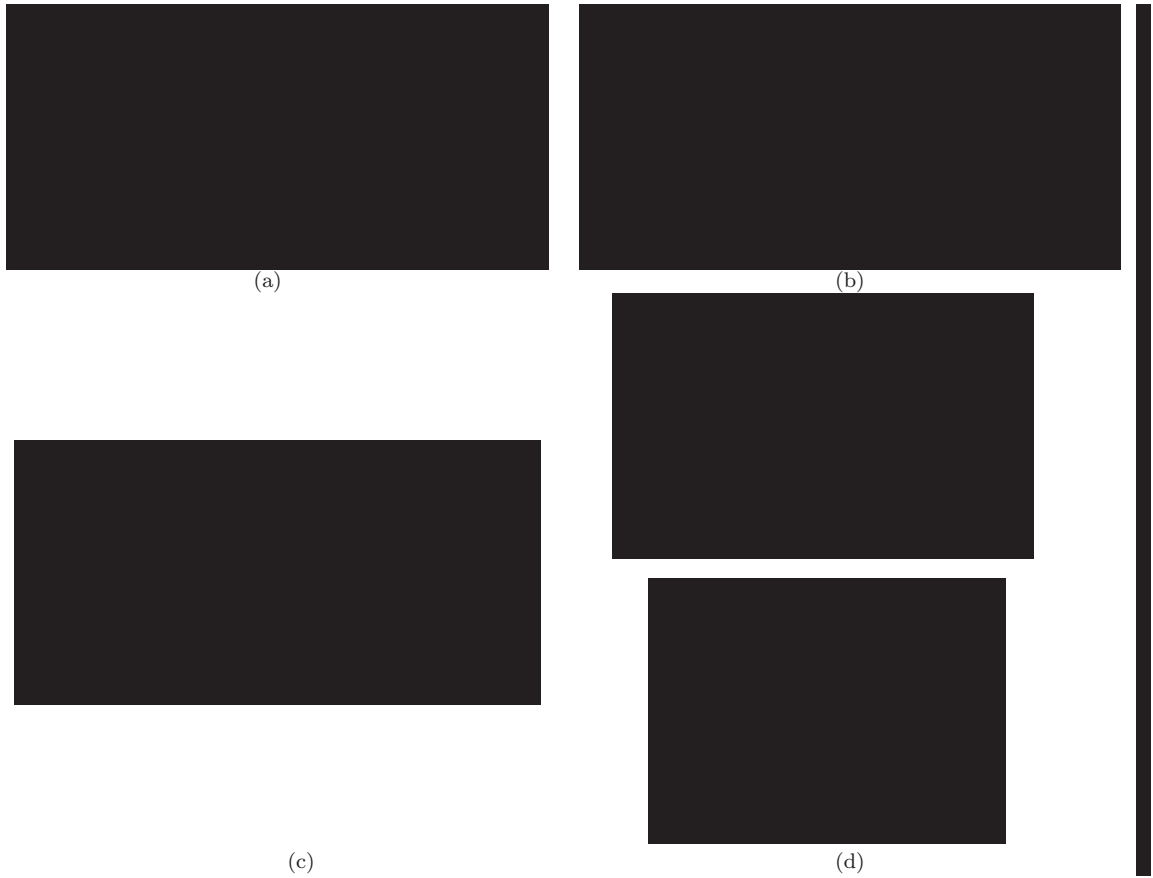

  \footnotesize
  \centering

  \begin{tabular}{VV}
    \hspace{-18pt}\includegraphics[width=.515\textwidth]{BMVC2005_GMM_em_faces.eps} &
    \includegraphics[width=.515\textwidth]{BMVC2005_GMM_inc_faces.eps} \\
    \hspace{-18pt}(a) & (b) \\
    \hspace{-18pt}\includegraphics[width=.5\textwidth]{BMVC2005_DLs_faces.eps} &
    \hspace{-18pt}\begin{tabular}{cc}
      \includegraphics[width=.4\textwidth]{BMVC2005_faces_em1.eps}
      \vspace{5pt} \\
      \includegraphics[width=.34\textwidth]{BMVC2005_faces_inc1.eps}\\
    \end{tabular}\\
    (c) & (d) \\
  \end{tabular}

  \caption[Evaluation: face motion data.]{ \it    Face motion data: data (dots) and (a)
                         MDL-EM GMM fit. (b) Incremental GMM fit. (c) Description
                         length of GMMs fitted using EM and the proposed
                         incremental algorithm (shown is the description
                         length of the final GMM estimate). (d) GMM component
                         centres visualized as images for the MDL-EM fit (top)
                         and the incremental algorithm (bottom). }
            \label{Fig: Face Set}
\end{figure}

\section{Summary and conclusions}
A novel algorithm for incremental learning of Temporally-Coherent
Gaussian mixtures was introduced. Promising performance was
empirically demonstrated on synthetic data and face appearance
streams extracted from realistic video, and qualitatively and
quantitatively compared with the standard EM-based fitting.

\section*{Related publications}

The following publications resulted from the work presented in this
appendix:

\begin{itemize}
  \item O. Arandjelovi\'c and R. Cipolla. Incremental learning of temporally-coherent Gaussian
                  mixture models. In \textit{Proc. IAPR British Machine Vision Conference (BMVC)},
                  \textbf{2}:pages 759--768, September 2005.
                  \cite{AranCipo2005a}

  \item O. Arandjelovic' and R. Cipolla. Incremental learning of temporally-coherent Gaussian
                  mixture models. \textit{Society of Manufacturing Engineers (SME) Technical Papers},
                  May 2006. \cite{AranCipo2006d}
\end{itemize}

\graphicspath{{./14mpmm/}}
\chapter{Maximally Probable Mutual Modes}
\label{App: MPMM}
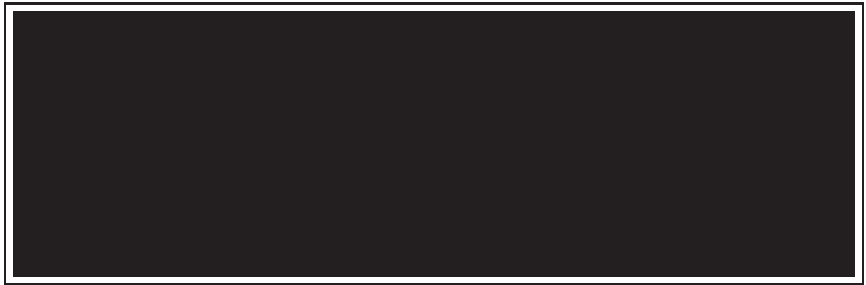
\begin{center}
  \vspace{-20pt}
  \footnotesize
  \framebox{\includegraphics[width=0.80\textwidth]{title_img.eps}}\\
  Salvador Dali. \textit{Archeological Reminiscence of Millet's Angelus}\\
  1933-5, Oil on panel, 31.7 x 39.3 cm\\
  Salvador Dali Museum, St. Petersburg, Florida
\end{center}

\cleardoublepage

In this appendix we consider discrimination between linear patches
corresponding to local appearance variations within face image sets.
We propose the \textit{Maximally Probable Mutual Modes} (MMPM)
algorithm, a probabilistic extension of the Mutual Subspace Method
(MSM). Specifically we show how the local manifold illumination
invariant introduced in Section~\ref{Sec: Manifold Illumination
Invariants} naturally leads to a formulation of ``common modes'' of
two face appearance distributions. Recognition is then performed by
finding the most probable mode, which is shown to be an eigenvalue
problem. The effectiveness of the proposed method is demonstrated
empirically on the \textit{CamFace} dataset.

\section{Introduction} In Section~\ref{Sec: Manifold
Illumination Invariants} we proposed a piece-wise linear
representation of face appearance variation as suitable for
exploiting the identified local manifold illumination invariant.
Recognition by comparing nonlinear appearance manifolds was thus
reduced to the problem of comparing linear patches, which was
performed using canonical correlations. Here we address the problem
of comparing linear patches in more detail and propose a
probabilistic extension to the concept of canonical correlations.

\subsection{Maximally probably mutual modes}
\begin{figure}
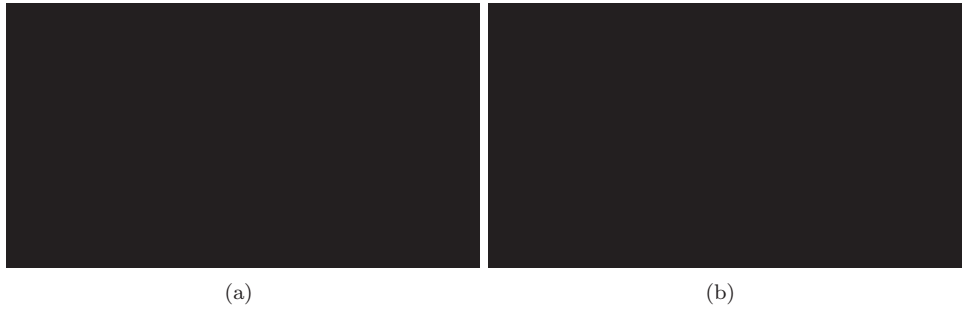

  \centering
  \subfigure[]{\includegraphics[width=0.45\textwidth]{mfold_1.eps}}
  \subfigure[]{\includegraphics[width=0.45\textwidth]{mfold_2.eps}}
  \caption[Piece-wise manifolds.]
    {\it Piece-wise representations of nonlinear manifolds:
        as a collection of (a) infinite-extent linear subspaces vs.
        (b) Gaussian densities. }
  \label{Fig: Mfolds}
  \vspace{6pt}\hrule
\end{figure}

In Chapter~\ref{Chp: BoMPA}, linear patches used to piece-wise
approximate an appearance manifold were represented by linear
subspaces, much like in the Mutual Subspace Method (MSM) of Fukui
and Yamaguchi \cite{FukuYama2003}. The patches themselves, however,
are finite in extent and are hence better characterized by
probability density functions, such as Gaussian densities. This is
the approach we adopt here, see Figure~\ref{Fig: Mfolds}.

Unlike in the case when dealing with subspaces, in general both of
the compared distributions can generate \emph{any} point in the
$D$-dimensional embedding space. Hence, the concept of the
most-correlated patterns (c.f.\ canonical correlations) from the two
classes is not meaningful in this context. Instead, we are looking
for a mode -- i.e.\ a linear direction in the pattern space -- along
which both distributions corresponding to the two classes are most
likely to ``generate'' observations.

We define the mutual probability $p_m(\mathbf{x})$ to be the product
of two densities at $\mathbf{x}$:
\begin{align}
  p_m(\mathbf{x}) = p_1(\mathbf{x}) p_2(\mathbf{x}).
\end{align}
Generalizing this, the mutual probability $S_{\mathbf{v}}$ of an
entire linear mode $\mathbf{v}$ is then:
\begin{align}
  S_{\mathbf{v}} = \int_{-\infty}^{+\infty} p_1(x \mathbf{v}) p_2(x \mathbf{v}) dx.
\end{align}

Substituting $\frac {1} {(2\pi)^{D/2} |\mathbf{C}_i|^{1/2}}
    \exp \left[ - \frac {1}{2} \mathbf{x}^T \mathbf{C}_i^{-1} \mathbf{x} \right]$ for $p_i(\mathbf{x})$, we obtain:
\begin{align}\label{Eqn: obj 1}
    S_{\mathbf{v}} &= \int_{-\infty}^{+\infty} \frac {1} {(2\pi)^{D/2} |\mathbf{C}_1|^{1/2}}
    \exp \left[ - \frac {1}{2} x \mathbf{v}^T \mathbf{C}_1^{-1} \mathbf{v} x \right]  \notag \\
    &\frac {1} {(2\pi)^{D/2} |\mathbf{C}_2|^{1/2}} \exp \left[ - \frac {1}{2} x \mathbf{v}^T \mathbf{C}_2^{-1} \mathbf{v} x \right] dx = \\
    & \frac {1} {(2\pi)^D |\mathbf{C}_1 \mathbf{C}_2|^{1/2}}
    \int_{-\infty}^{+\infty} \exp \left[ - \frac {1}{2} x^2 \mathbf{v}^T \left(\mathbf{C}_1^{-1} + \mathbf{C}_2^{-1} \right) \mathbf{v} \right]
    dx.
\end{align}
Noting that the integral is now over a 1D Gaussian distribution (up
to a constant):
\begin{align}\label{Eqn: obj 1}
    S_{\mathbf{v}} = &\frac {1} {(2\pi)^D |\mathbf{C}_1 \mathbf{C}_2|^{1/2}}
    (2\pi)^{1/2} \left[ \mathbf{v}^T \left(\mathbf{C}_1^{-1} + \mathbf{C}_2^{-1} \right) \mathbf{v}
    \right]^{-1/2} = \\
    &(2\pi)^{1/2-D} |\mathbf{C}_1 \mathbf{C}_2|^{-1/2}
    \left[ \mathbf{v}^T \left(\mathbf{C}_1^{-1} + \mathbf{C}_2^{-1} \right) \mathbf{v} \right]^{-1/2}
    \label{Eqn: ObjFn Final}
\end{align}

The expression above favours directions in which both densities have
large variances, i.e.\ in which Signal-to-Noise ratio is the
highest, as one may intuitively expect, see Figure~\ref{Fig: MPMM}.

\begin{figure}
  \centering
  \includegraphics[width=0.75\textwidth]{mpmm.eps}
  \caption[Maximally Probable Mutual Mode illustration.]
    {\it Conceptual drawing of the Maximally Probable Mutual Mode
         concept for 2D Gaussian densities. }
  \label{Fig: MPMM}
  \vspace{6pt}\hrule
\end{figure}

The mode that maximizes the mutual probability $S_{\mathbf{v}}$ can
be found by considering eigenvalue decomposition of
$\mathbf{C}_1^{-1} + \mathbf{C}_2^{-1}$. Writing:
\begin{align}
  &\mathbf{C}_1^{-1} + \mathbf{C}_2^{-1} = \sum_{i=1}^D \lambda_i \mathbf{u}_i
  \mathbf{u}_i^T,
\end{align}
where $0 \leq \lambda_1 \leq \lambda_2 \leq \ldots \leq \lambda_D$
and
\begin{align}
  &\mathbf{u}_i \cdot \mathbf{u}_j = 0, i \neq j \notag \\
  &\mathbf{u}_i \cdot \mathbf{u}_i = 1.
\end{align}
and since $\{\mathbf{u}_i\}$ span $\mathbb{R}^D$:
\begin{align}
  &\mathbf{v} = \sum_{i=1}^D \alpha_i \mathbf{u}_i,
\end{align}
it is then easy to show that the maximal value of \eqref{Eqn: ObjFn
Final} is:
\begin{align}
  \max_{\mathbf{v}} S_{\mathbf{v}} = (2\pi)^{1/2-D} |\mathbf{C}_1
  \mathbf{C}_2|^{-1/2} \lambda_D^{-1/2}.
  \label{Eqn: MPMM}
\end{align}
This defines the class similarity score $\nu$. It is achieved for
$\alpha_{1,\dots,D-1} = 0,~\alpha_D=1$ or $\mathbf{v} =
\mathbf{u}_D$, i.e.\ the direction of the eigenvector corresponding
to the smallest eigenvalue of $\mathbf{C}_1^{-1} +
\mathbf{C}_2^{-1}$. A visualization of the most probable mode
between two face sets of Figure~\ref{Fig: Data} is shown in
Figure~\ref{Fig: MPMM}.

\begin{figure}[!t]
  \centering
  \includegraphics[width=0.75\textwidth]{ICPR2006_people_data.eps}
  \vspace{-20pt}
  \caption[Example face sequences.]{\it Examples of detected faces from the CamFace database.
                A wide range of illumination and pose changes is present. }
  \label{Fig: Data}
\end{figure}

\begin{figure}
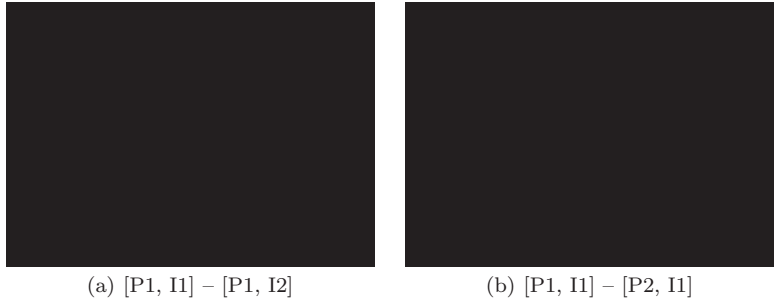

  \centering
  \footnotesize
  \begin{tabular}{cc}
    \includegraphics[width=0.35\textwidth]{same_classes.eps} &
    \includegraphics[width=0.35\textwidth]{different_classes.eps}\\
    (a) [P1, I1] -- [P1, I2] & (b) [P1, I1] -- [P2, I1] \\
  \end{tabular}

  \caption[Maximally probable mutual mode as image.]{ \it The maximally probable mutual mode,
            shown as an image, when two compared face sets belong to the (a) same and (b)
            different individuals (also see Figure~\ref{Fig: Data}). }
  \label{Fig: MPMM}
  \vspace{6pt}\hrule
\end{figure}

\subsection{Numerical and implementation issues}
The expression for the similarity score $\nu = \max_{\mathbf{v}}
S_{\mathbf{v}}$ in \eqref{Eqn: MPMM} involves the computation of
$|\mathbf{C}_1 \mathbf{C}_2|^{-1/2}$. This is problematic as
$\mathbf{C}_1 \mathbf{C}_2$ may be a singular, or a nearly singular,
matrix (e.g.\ because the number of face images is much lower than
the image space dimensionality $D$).

We solve this problem by assuming that the dimensionality of the
principal linear subspaces corresponding to $\mathbf{C}_1$ and
$\mathbf{C}_2$ is $M \ll D$, and that data is perturbed by isotropic
Gaussian noise. If $\lambda_1^{(i)} \leq \lambda_2^{(i)} \leq \dots
\leq \lambda_D^{(i)}$ are the eigenvalues of $\mathbf{C}_i$:
\begin{align} \forall j > M.~\lambda_{D-j}^{(1)} = \lambda_{D-j}^{(2)}.
\end{align}
Then, writing
\begin{align}
    |\mathbf{C}_i| = \prod_{j=1}^D \lambda_j^{(i)},
\end{align}
we get:
\begin{align}
  \nu &= (2\pi)^{1/2-D} |\mathbf{C}_1
  \mathbf{C}_2|^{-1/2} \lambda_D^{-1/2} = \\
  &= const \times \left( \lambda_D \prod_{i=D-M+1}^D \lambda_i^{(1)} \lambda_i^{(2)} \right)^{-1/2}.
\end{align}

\section{Experimental evaluation}\label{Sec: Empirical Evaluation}
We demonstrate the superiority of the Maximally Probable Mutual
Modes to the Mutual Subspace Method \cite{FukuYama2003} on the
\textit{CamFace} data set using the Manifold Principal Angles
algorithm of Chapter~\ref{Chp: BoMPA}. With the purpose of focusing
on the underlying comparison of linear subspaces we omit the
contribution of global appearance in the overall manifold similarity
score by setting $\alpha = 0$ in \eqref{Eqn: Similarity}.

A summary of the results is shown in Table~\ref{Tab:MPMM} with
the Receiver-Operator Characteristics (ROC) curve for the MPMM
method in Figure~\ref{Fig: ROC}. The proposed method achieved a
significantly higher average recognition rate than the original MSM
algorithm.

\begin{table}
  \centering
 \Large
  \caption[MPMM and MSM evaluation results.]{ \it Recognition performance statistics (\%).\vspace{10pt} }
    \begin{tabular*}{1.00\textwidth}{@{\extracolsep{\fill}}ll|ccc}
      \Hline
      \bf \normalsize ~Method && \normalsize MPMM  & \normalsize MSM &\\
      \hline
      \multirow{2}{*}{\bf  \normalsize ~Recognition rate} & \normalsize ~average      & \small 92.0  & \small 58.3 &\\
                                                          & \normalsize ~std          & \small 7.8   & \small 24.3 &\\
      \Hline
    \end{tabular*}
    \label{Tab:MPMM}
\end{table}

\begin{figure}
  \centering
  \includegraphics[width=0.60\textwidth]{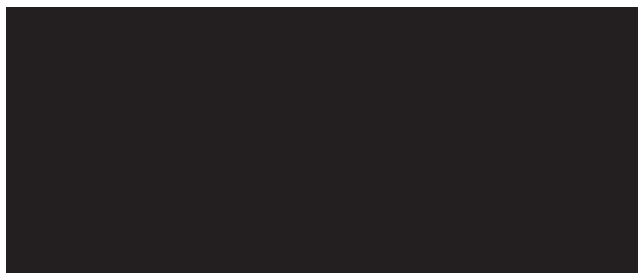}
  \caption[MPMM Receiver-Operator Characteristics curve.]
    {\it Receiver-Operator Characteristic of MPMM. }
  \label{Fig: ROC}
  \vspace{6pt}\hrule
\end{figure}

\section{Summary and conclusions} \label{Sec: Summary} We described
a probabilistic extension to the concept of canonical correlations
which has been widely used in the pattern recognition literature.
The resulting method was demonstrated suitable for matching local
appearance variations between face sets, exploiting a manifold
illumination invariant.

\section*{Related publications}

The following publications resulted from the work presented in this
appendix:

\begin{itemize}
  \item O. Arandjelovi\'c and R. Cipolla. Face set classification using maximally probable mutual
                  modes.  In \textit{Proc. IEEE International Conference on Pattern Recognition (ICPR)},
                  pages 511--514,
                  August 2006. \cite{AranCipo2006e}
\end{itemize}

\graphicspath{{./15data/}}
\chapter{The \textit{CamFace} data set}
\label{App: CamFace}
\begin{center}
  \footnotesize
  \framebox{\includegraphics[trim=0cm 0.3cm 0cm 1cm, clip=true, width=0.85\textwidth]{title_img.eps}}\\
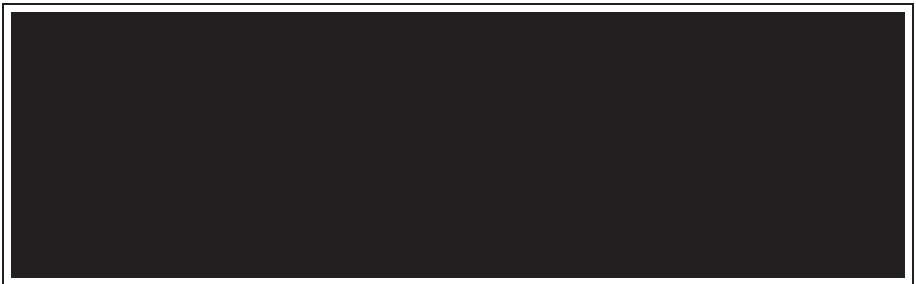
  Camille Pissarro. \textit{Boulevard Montmartre}\\
  1897, Oil on canvas, 74 x 92.8 cm\\
  The State Hermitage Museum, Leningrad
\end{center}

\cleardoublepage

The University of Cambridge Face database (\textit{CamFace}
database) is a collection of video sequences of largely
unconstrained, random head movement in different illumination
conditions, acquired for the purpose of developing and evaluating
face recognition algorithms. This appendix describes (i) the
database and its acquisition, and (ii) a novel method for automatic
extraction of face images from videos of head motion in a cluttered
environment, suitable as a preprocessing step to recognition
algorithms. The database and the preprocessing described are used
extensively in this thesis.

\section{Description}
\begin{figure*}
  \centering
  \includegraphics[width=0.7\textwidth]{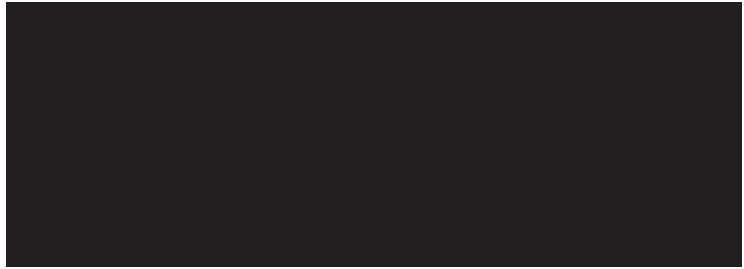}
  \caption[\textit{CamFace} ages distribution.]{ \it The distribution of people's
            ages across the CamFace data set. }
  \label{Fig: Ages}
\end{figure*}

The \textit{CamFace} data set is a database of face motion video
sequences acquired in the Department of Engineering, University of
Cambridge. It contains 100 individuals of varying age, ethnicity and
gender, see Figure~\ref{Fig: Ages} and Table~\ref{Tab: Gender}.

\begin{table}
  \Large
  \caption[\textit{CamFace} gender distribution.]
   { \it The proportion of the two genders in the CamFace dataset. \vspace{10pt}}
  \begin{tabularx}{1.00\textwidth}{l|XcX|XcX}
    \Hline
    \normalsize \bf ~Gender && \bf \normalsize Male  &&& \bf \normalsize Female &\\
    \hline
    \normalsize \bf ~Number && \small  67            &&& \small  33 &\\
    \Hline
  \end{tabularx}
  \label{Tab: Gender}
\end{table}

\begin{figure}
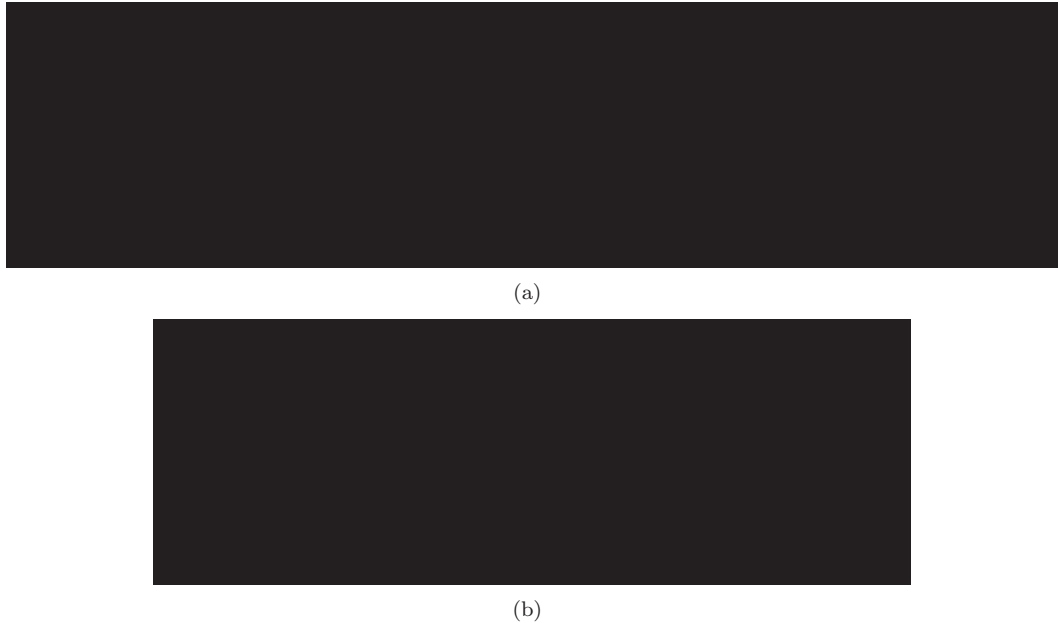

  \centering
  \subfigure[]
   {\includegraphics[width=1.0\textwidth]{ICCV2005_illuminations.eps}}
  \subfigure[]
   {\includegraphics[width=0.72\textwidth]{PAMI2005_illumination_effects2.eps}}
  \caption[\textit{CamFace} illuminations]{ \it (a) Illuminations 1--7 from the CamFace data set.
            (b) Five different individuals in the illumination setting number 6. In spite of the
            same spatial arrangement of light sources, their effect on the appearance of
            faces changes significantly due to variations in people's heights and the
            \textit{ad lib} chosen position relative to the camera. }
            \label{Fig: Illums}
\end{figure}

For each person in the database we collected 14 video sequences of
the person in quasi-random motion. We used 7 different illumination
configurations and acquired 2 sequences with each for a given
person, see Figure~\ref{Fig: Illums}. The individuals were
instructed to approach the camera and move freely, with the loosely
enforced constraint of being able to see their eyes on the screen
providing visual feedback in front of them, see Figure~\ref{Fig:
Acquisition1}~(a). Most sequences contain significant yaw and pitch
variation, some translatory motion and negligible roll. Mild facial
expression changes are present in some sequences (e.g.\ when the
user was smiling or talking to the person supervising the
acquisition), see Figure~\ref{Fig: Example}.

\begin{figure*}
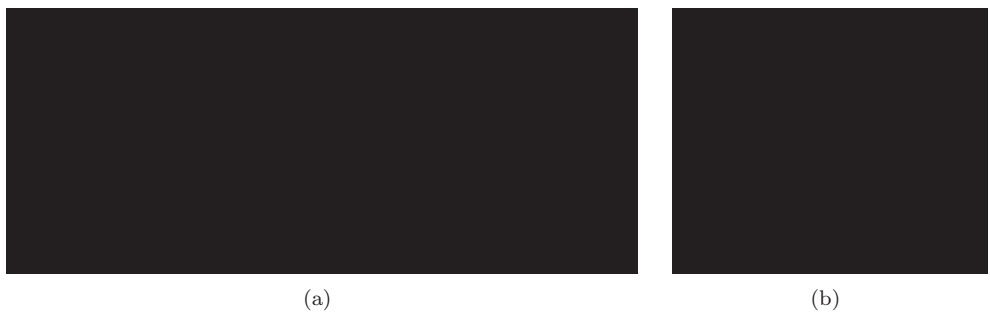

  \centering
  \subfigure[]{\includegraphics[width=0.6\textwidth]{software.eps}}\hspace{10pt}
  \subfigure[]{\includegraphics[width=0.3\textwidth]{camera.eps}}
  \caption[\textit{CamFace} acquisition setup.]{ \it (a) Visual feedback
            displayed to the user during data acquisition. (b) The pin-hole
            camera used to collect the CamFace data set.  }
  \label{Fig: Acquisition1}
  \vspace{6pt}\hrule
\end{figure*}

\begin{figure*}
  \centering
  \includegraphics[width=0.88\textwidth]{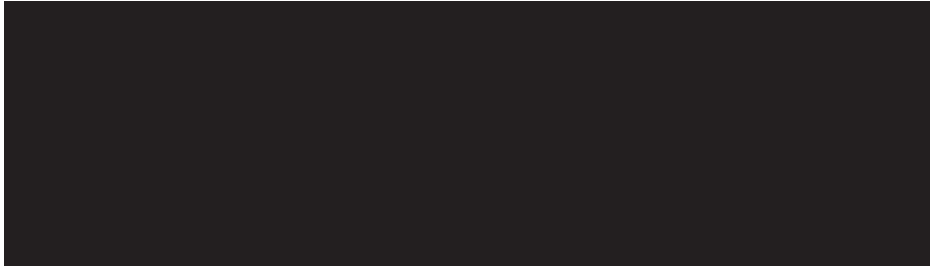}
  \caption[Typical \textit{CamFace} sequence.]
    { \it A 100 frame, 10 fps video sequence typical for the CamFace data set.
          The user positions himself ad lib and performs quasi-random head motion.
          Although instructed to keep head pose variations within
          the range in which the eyes are clearly visible, note that a significant
          number of poses does not meet this requirement. }
  \label{Fig: Example}
\end{figure*}

\paragraph{Acquisition hardware.} Video sequences were acquired using a
simple pin-hole camera with automatic gain control, mounted at 1.2m
above the ground and pointing upwards at 30 degrees to the
horizontal, see Figure~\ref{Fig: Acquisition1}. Data was acquired at
10fps, giving 100 frames for each 10s sequence, in 320 by 240 pixel
resolution, see Figure~\ref{Fig: Example} for an example and
Table~\ref{Tab: CamFace1} for a summary. On average, the face
occupies an area of 60 by 60 pixels.

\begin{table}
  \Large
  \caption[\textit{CamFace} statistics.]{ \it An overview of CamFace data set statistics.\vspace{10pt} }
  \begin{tabularx}{1.00\textwidth}{l|X|X|X|X}
    \Hline
    \normalsize & \bf \normalsize \vspace{1pt} Individuals & \bf \normalsize \vspace{1pt} Illuminations & \bf \normalsize Sequences per illumination per person & \bf \normalsize Frames per second (fps) \\
    \hline
    \normalsize \bf ~Number   & \small 100         & \small 7             & \small 2 & \small 10 \\
    \Hline
  \end{tabularx}
  \label{Tab: CamFace1}
\end{table}

\section{Automatic extraction of faces}\label{AppC:AutomaticExtraction} 
\begin{figure*}
  \footnotesize
  \centering
  \includegraphics[width=0.8\textwidth]{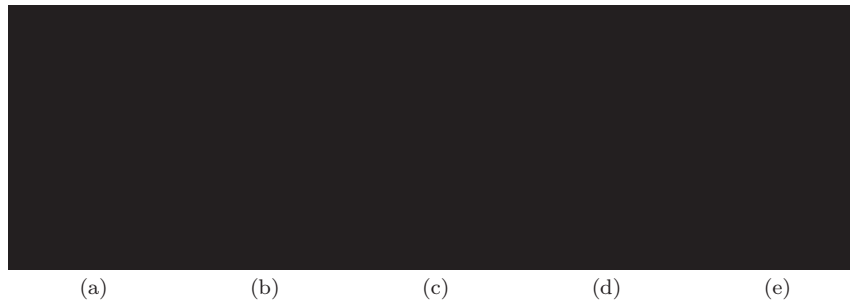}
  \begin{tabular*}{0.72\textwidth}{@{\extracolsep{\fill}}ccccc}
    \hspace{0.02\textwidth} (a) & (b) & (c) & (d) & (e) \\
  \end{tabular*}
  \caption[Face preprocessing pipeline.]
        { \it Illustration of the described preprocessing pipeline.
            (a) Original input frame with resolution of  $320 \times 240$ pixels.
            (b) Face detection with average bounding box size of $75 \times 75$ pixels.
            (c) Resizing to the uniform scale of $40 \times 40$ pixels.
            (d) Background removal and feathering.
            (e) The final image after histogram equalization.}
            \label{Fig: Loc}
  \vspace{6pt}\hrule
\end{figure*}

\begin{figure*}
  \centering
  \includegraphics[width=1.0\textwidth]{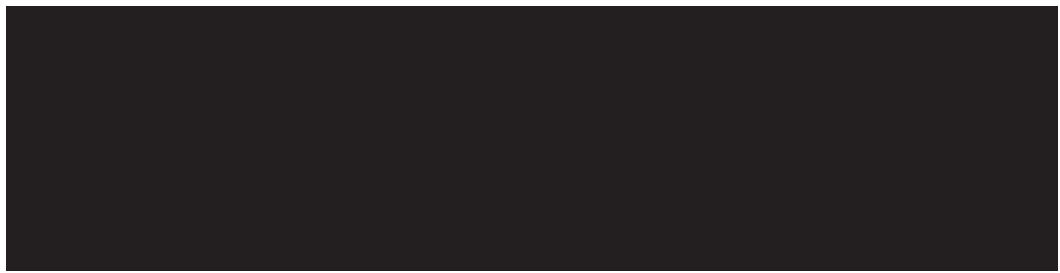}
  \caption[Typical \textit{CamFace} detections.]
    { \it Per-frame face detector output from a typical 100 frame,
          10 fps video sequence (also see Figure~\ref{Fig: Example}).
          The detector is robust to a rather large range of pose
          changes.
     }
  \label{Fig: ExampleF}
\end{figure*}

\begin{figure}[!t]
  \centering
  \includegraphics[width=0.6\textwidth]{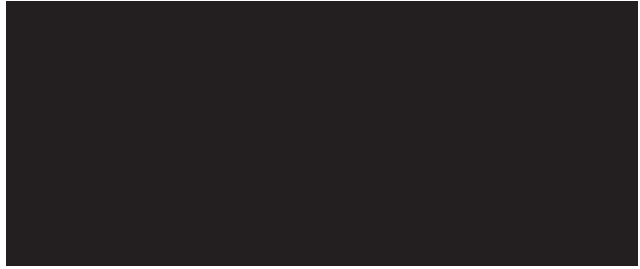}\\
  \caption[Number of face detections across the \textit{CamFace} data set.]
          { \it A histogram of the number of face detections per sequence across
                the \textit{CamFace} data set.}
            \label{Fig: NFaces}
\end{figure}

We used the Viola-Jones
cascaded detector \cite{ViolJone2004} in order to localize faces in
cluttered images. Figure~\ref{Fig: Example} shows examples of input
frames, Figure~\ref{Fig: Loc}~(b) an example of a correctly detected
face and Figure~\ref{Fig: ExampleF} all detected faces in a typical
sequence. A histogram of the number of detections we get across the
entire data set is shown in Figure~\ref{Fig: NFaces}.

\begin{figure}
  \centering
  \includegraphics[width=0.8\textwidth]{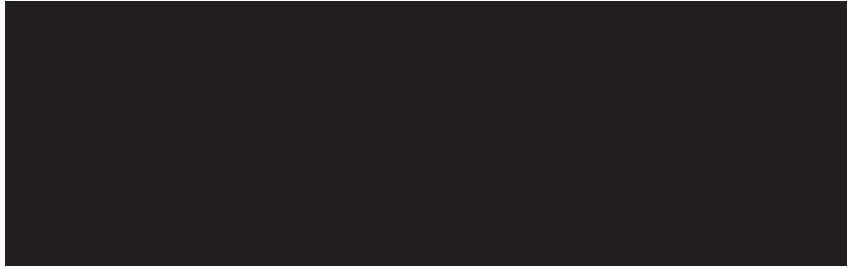}\\
  \caption[False positive face detections.]
    { \it Typical false face detections identified by our algorithm.}
  \label{Fig: Falses}
\end{figure}

\paragraph{Rejection of false positives.} The face
detector achieves high true positive rates for our database. A
larger problem is caused by false alarms, even a small number of
which can affect the density estimates. We use a coarse skin colour
classifier to reject many of the false detections. The classifier is
based on 3-dimensional colour histograms built for two classes: skin
and non-skin pixels \cite{JoneRehg1999}. A pixel can then be
classified by applying the likelihood ratio test. We apply this
classifier and reject detections in which too few ($<60\%$) or too
many ($>99\%$) pixels are labelled as skin. This step removes the
vast majority of non-faces as well as faces with grossly incorrect
scales -- see Figure~\ref{Fig: Falses} for examples of successfully
removed false positives.

\paragraph{Background clutter removal.}  The bounding box of
a detected face typically contains a portion of the background. The
removal of the background is beneficial because it can contain
significant clutter and also because of the danger of learning to
discriminate based on the background, rather than face appearance.
This is achieved by \emph{set-specific} skin colour segmentation:
Given a set of images from the same subject, we construct colour
histograms for that subject's face pixels and for the near-face
background pixels in that set. Note that the classifier here is
tuned for the given subject \emph{and} the given background
environment, and thus is more ``refined'' than the coarse classifier
used to remove false positives. The face pixels are collected by
taking the central portion of the few most symmetric images in the
set (assumed to correspond to frontal face images); the background
pixels are collected from the 10 pixel-wide strip around the face
bounding box provided by the face detector, see Figure~\ref{Fig:
Hists}. After classifying each pixel within the bounding box
independently, we smooth the result using a simple 2-pass algorithm
that enforces the connectivity constraint on the face and boundary
regions, see Figure~\ref{Fig: Loc}~(d). A summary of the cascade in
its entirety is shown in Figure~\ref{Fig: Preprocess}.

\begin{figure*}[!t]
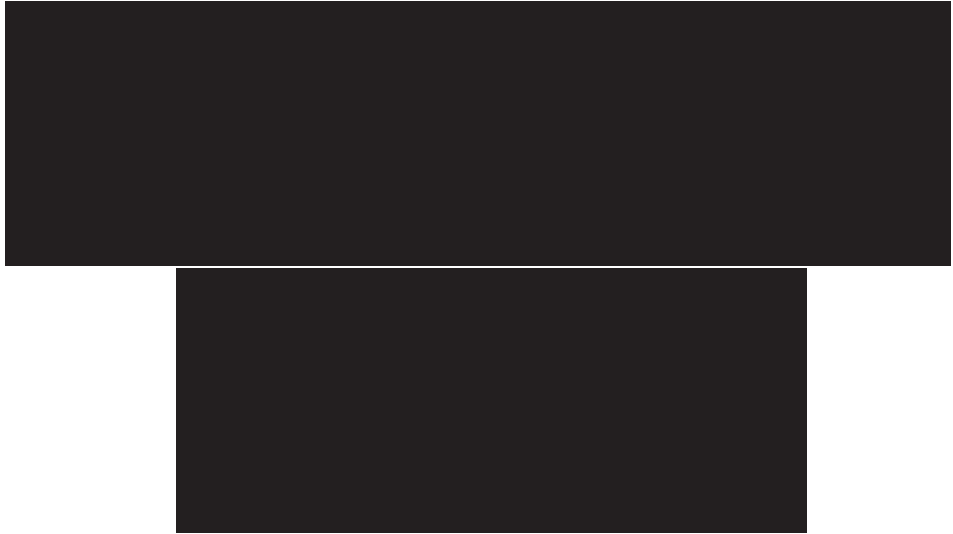

  \centering
  \includegraphics[width=0.9\textwidth]{sym_score.eps}\\
  \hspace{10pt}\includegraphics[width=0.6\textwidth]{sym_poses.eps}
  \caption[Face ``frontality'' measure.]
    { \it The response of our vertical symmetry-based measure of
          the ``frontality'' of a face, used to select the most
          reliable faces for extraction of background and foreground
          colour models. Also see Figures~\ref{Fig: Hists}
          and~\ref{Fig: Preprocess}.
     }
  \label{Fig: Sym}
\end{figure*}

\begin{figure}
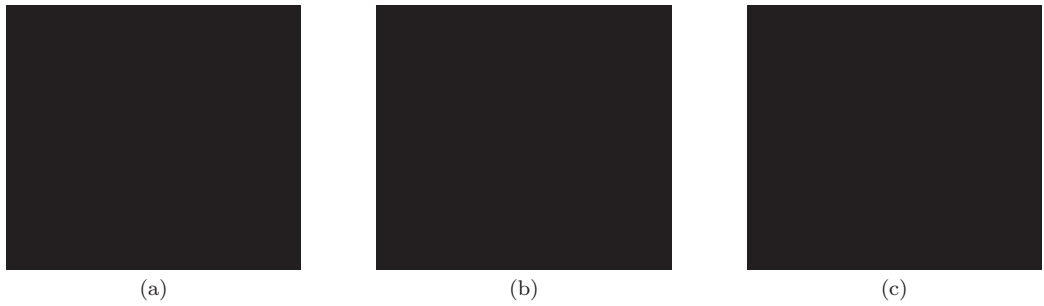

  \centering
  \footnotesize
  \begin{tabular*}{1.00\textwidth}{@{\extracolsep{\fill}}ccc}
    \includegraphics[width=0.28\textwidth]{sampling.eps} &
    \includegraphics[width=0.28\textwidth]{faces_hist_01.eps} &
    \includegraphics[width=0.28\textwidth]{background_hist_01.eps} \\
    (a) & (b) & (c) \\
  \end{tabular*}
  \caption[Face and background colour models.]
        { \it (a) Areas used to sample face and background colours, and the
              corresponding (b) face and (b) background histograms
              in RGB space used for ML skin-colour detection. Larger
              blobs correspond to higher densities and are colour-coded. }
             \label{Fig: Hists}
\end{figure}

\begin{figure}[!t]
  \centering
  \rotatebox{90}{\includegraphics[width=1.25\textwidth]{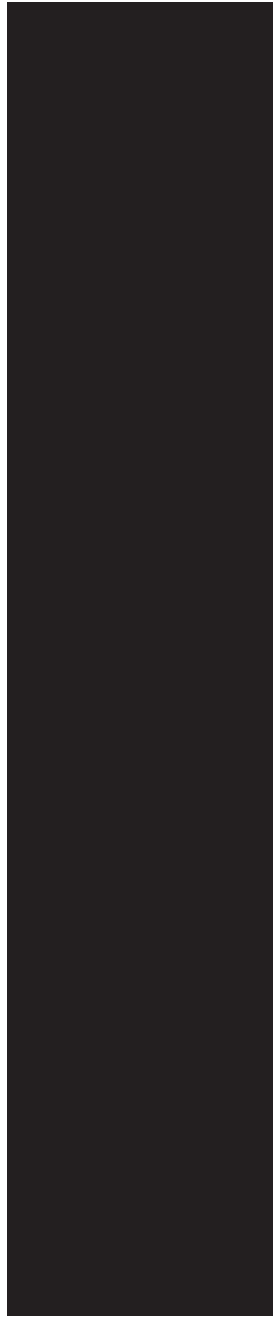}}
  \caption[Preprocessing cascade.]
          { \it A schematic representation of the face localization
                and normalization cascade.}
            \label{Fig: Preprocess}
\end{figure}

\section*{Related publications}

The following publications contain portions of work presented in
this appendix:

\begin{itemize}
  \item O. Arandjelovi{\'c}, G. Shakhnarovich, J. Fisher, R. Cipolla, and T. Darrell.
                  Face recognition with image sets using manifold density divergence. In
                  \textit{Proc. IEEE Conference on Computer Vision and Pattern Recognition (CVPR)},
                  \textbf{1}:pages 581--588, June 2005.
                  \cite{AranShakFish+2005}

  \item O. Arandjelovi{\'c} and R. Cipolla. An information-theoretic approach to face recognition
                  from face motion manifolds. \textit{Image and Vision Computing (special issue
                  on Face Processing in Video Sequences)}, \textbf{24}(6):pages 639--647, June 2006.
                  \cite{AranCipo2006}
\end{itemize}

\fancyhead[RO,LE]{\textcolor{Gray}\leftmark}
\fancyhead[RE,LO]{\textcolor{Gray}{\S\thesection}}
\fancyfoot[LE,RO]{\hrule\vspace{10pt}~\thepage~} \fancyfoot[CE,CO]{}

\nocite{Bish1995, CoveThom1991, DudaHartStor2001, GonzWood1992,
GrimStir1992, LiJain2004, Mack2003, SnedCoch1989, Wass1989}

\nocite{AranCipo2013}
\nocite{AranHammCipo2010}
\nocite{AranCipo2009,AranCipo2009a}
\nocite{AranCipo2007,KimAranCipo2007,AranHammCipo2007}
\nocite{AranZiss2006,AranCipo2006,AranCipo2006a,JohnBrosShot+2006,AranCipo2006c,AranCipo2006d,BrosJohnShot+2006,AranCipo2006b,AranHammCipo2006,AranCipo2006e,AranHammCipo2006a}
\nocite{AranShakFish+2005,AranZiss2005,KimAranCipo2005,AranCipo2005a}
\nocite{Aran2004,AranCipo2004,AranCipo2004a}

\clearpage

\bibliographystyle{oa_thesis}

\fancyhead[RO,LE]{\textcolor{Gray}{Bibliography}}
\fancyhead[RE,LO]{}

\begin{spacing}{2.1}
\bibliography{./my_bibliography,./05therm/biometric}

\begin{thebibliography}{Wan04b}
\expandafter\ifx\csname urlstyle\endcsname\relax
  \providecommand{\doi}[1]{doi:\discretionary{}{}{}#1}\else
  \providecommand{\doi}{doi:\discretionary{}{}{}\begingroup
  \urlstyle{rm}\Url}\fi

\bibitem[Abd98]{AbdiValeEdel1998}
H.~Abdi, D.~Valentin and B.~E. Edelman.
\newblock Eigenfeatures as intermediate level representations: The case for
  {PCA} models.
\newblock \emph{Brain and Behavioral Sciences}, \textbf{21}:pages 17--18, 1998.

\bibitem[Adi97]{AdinMoseUllm1997}
Y.~Adini, Y.~Moses and S.~Ullman.
\newblock Face recognition: The problem of compensating for changes in
  illumination direction.
\newblock \emph{IEEE Transactions on Pattern Analysis and Machine Intelligence
  (TPAMI)}, \textbf{19}(7):pages 721--732, 1997.

\bibitem[Aka91]{AkamSasaFukuSuen1991}
S.~Akamatsu, T.~Sasaki, H.~Fukumachi and Y.~Suenaga.
\newblock A robust face identification scheme - {KL} expansion of an invariant
  feature space.
\newblock \emph{Intelligent Robots and Computer Vision},
  \textbf{1607}(10):pages 71--84, 1991.

\bibitem[Ara04a]{Aran2004}
O.~Arandjelovi{\'c}.
\newblock Face recognition from face motion manifolds.
\newblock First year {Ph.D.\ }report, University of Cambridge, Cambridge, UK,
  2004.

\bibitem[Ara04b]{AranCipo2004}
O.~Arandjelovi{\'c} and R.~Cipolla.
\newblock Face recognition from face motion manifolds using robust kernel
  resistor-average distance.
\newblock \emph{In Proc. IEEE Workshop on Face Processing in Video Sequence},
  \textbf{5}:page~88, 2004.

\bibitem[Ara04c]{AranCipo2004a}
O.~Arandjelovi{\'c} and R.~Cipolla.
\newblock An illumination invariant face recognition system for access control
  using video.
\newblock \emph{In Proc. British Machine Vision Conference (BMVC)}, pages
  537--546, 2004.

\bibitem[Ara05a]{AranCipo2005a}
O.~Arandjelovi{\'c} and R.~Cipolla.
\newblock Incremental learning of temporally-coherent {G}aussian mixture
  models.
\newblock \emph{In Proc. British Machine Vision Conference (BMVC)},
  \textbf{2}:pages 759--768, 2005.

\bibitem[Ara05b]{AranShakFish+2005}
O.~Arandjelovi{\'c}, G.~Shakhnarovich, J.~Fisher, R.~Cipolla and T.~Darrell.
\newblock Face recognition with image sets using manifold density divergence.
\newblock \emph{In Proc. IEEE Conference on Computer Vision and Pattern
  Recognition (CVPR)}, \textbf{1}:pages 581--588, 2005.

\bibitem[Ara05c]{AranZiss2005}
O.~Arandjelovi{\'c} and A.~Zisserman.
\newblock Automatic face recognition for film character retrieval in
  feature-length films.
\newblock \emph{In Proc. IEEE Conference on Computer Vision and Pattern
  Recognition (CVPR)}, \textbf{1}:pages 860--867, 2005.

\bibitem[Ara06a]{AranCipo2006c}
O.~Arandjelovi{\'c} and R.~Cipolla.
\newblock Automatic cast listing in feature-length films with anisotropic
  manifold space.
\newblock \emph{In Proc. IEEE Conference on Computer Vision and Pattern
  Recognition (CVPR)}, \textbf{2}:pages 1513--1520, 2006.

\bibitem[Ara06b]{AranCipo2006b}
O.~Arandjelovi{\'c} and R.~Cipolla.
\newblock Face recognition from video using the generic shape-illumination
  manifold.
\newblock \emph{In Proc. European Conference on Computer Vision (ECCV)},
  \textbf{4}:pages 27--40, 2006.

\bibitem[Ara06c]{AranCipo2006e}
O.~Arandjelovi{\'c} and R.~Cipolla.
\newblock Face set classification using maximally probable mutual modes.
\newblock \emph{In Proc. IAPR International Conference on Pattern Recognition
  (ICPR)}, pages 511--514, 2006.

\bibitem[Ara06d]{AranCipo2006d}
O.~Arandjelovi{\'c} and R.~Cipolla.
\newblock Incremental learning of temporally-coherent {G}aussian mixture
  models.
\newblock \emph{Society of Manufacturing Engineers (SME) Technical Papers},
  \textbf{2}, 2006.

\bibitem[Ara06e]{AranCipo2006}
O.~Arandjelovi{\'c} and R.~Cipolla.
\newblock An information-theoretic approach to face recognition from face
  motion manifolds.
\newblock \emph{Image and Vision Computing (special issue on Face Processing in
  Video)}, \textbf{24}(6):pages 639--647, 2006.

\bibitem[Ara06f]{AranCipo2006a}
O.~Arandjelovi{\'c} and R.~Cipolla.
\newblock A new look at filtering techniques for illumination invariance in
  automatic face recognition.
\newblock \emph{In Proc. IEEE International Conference on Automatic Face and
  Gesture Recognition (FG)}, pages 449--454, 2006.

\bibitem[Ara06g]{AranHammCipo2006a}
O.~Arandjelovi{\'c}, R.~Hammoud and R.~Cipolla.
\newblock On person authentication by fusing visual and thermal face
  biometrics.
\newblock \emph{In Proc. IEEE Conference on Advanced Video and Singal Based
  Surveillance (AVSS)}, pages 50--56, 2006.

\bibitem[Ara06h]{AranHammCipo2006}
O.~Arandjelovi{\'c}, R.~I. Hammoud and R.~Cipolla.
\newblock Multi-sensory face biometric fusion (for personal identification).
\newblock \emph{In Proc. IEEE International Workshop on Object Tracking and
  Classification Beyond the Visible Spectrum (OTCBVS)}, pages 128--135, 2006.

\bibitem[Ara06i]{AranZiss2006}
O.~Arandjelovi{\'c} and A.~Zisserman.
\newblock \emph{Interactive Video: Algorithms and Technologies.}, chapter On
  Film Character Retrieval in Feature-Length Films., pages 89--103.
\newblock Springer-Verlag, 2006.
\newblock ISBN 978-3-540-33214-5.

\bibitem[Ara07a]{AranCipo2007}
O.~Arandjelovi{\'c} and R.~Cipolla.
\newblock \emph{Face recognition}, chapter Achieving illumination invariance
  using image filters., pages 15--30.
\newblock I-Tech Education and Publishing, Vienna, Austria, 2007.
\newblock ISBN 978-3-902613-03-5.

\bibitem[Ara07b]{AranHammCipo2007}
O.~Arandjelovi{\'c}, R.~I. Hammoud and R.~Cipolla.
\newblock \emph{Face Biometrics for Personal Identification}, chapter Towards
  Person Authentication by Fusing Visual and Thermal Face Biometrics, pages
  75--90.
\newblock Springer-Verlag, 2007.
\newblock ISBN 978-3-540-49344-0.

\bibitem[Ara09a]{AranCipo2009a}
O.~Arandjelovi{\'c} and R.~Cipolla.
\newblock A methodology for rapid illumination-invariant face recognition using
  image processing filters.
\newblock \emph{Computer Vision and Image Understanding (CVIU)},
  \textbf{113}(2):pages 159--171, 2009.

\bibitem[Ara09b]{AranCipo2009}
O.~Arandjelovi{\'c} and R.~Cipolla.
\newblock A pose-wise linear illumination manifold model for face recognition
  using video.
\newblock \emph{Computer Vision and Image Understanding (CVIU)},
  \textbf{113}(1):pages 113--125, 2009.

\bibitem[Ara10]{AranHammCipo2010}
O.~Arandjelovi{\'c}, R.~I. Hammoud and R.~Cipolla.
\newblock Thermal and reflectance based personal identification methodology in
  challenging variable illuminations.
\newblock \emph{Pattern Recognition (PR)}, \textbf{43}(5):pages 1801--1813,
  2010.

\bibitem[Ara13]{AranCipo2013}
O.~Arandjelovi{\'c} and R.~Cipolla.
\newblock Achieving robust face recognition from video by combining a weak
  photometric model and a learnt generic face invariant.
\newblock \emph{Pattern Recognition (PR)}, \textbf{46}(1):pages 9--23, 2013.

\bibitem[Arc03]{ArcaCampLanz2003}
S.~Arca, P.~Campadelli and R.~Lanzarotti.
\newblock A face recognition system based on local feature analysis.
\newblock \emph{Lecture Notes in Computer Science (LNCS)}, \textbf{2688}:pages
  182--189, 2003.

\bibitem[Bae02]{BaekDrapBeveShe2002}
A.~D. Baek, B.~A. Draper, J.~R. Beveridge and K.~She.
\newblock {PCA} vs. {ICA}: A comparison on {FERET} data set.
\newblock \emph{In Proc. International Conference on Computer Vision, Pattern
  Recognition and Image Processing}, pages 824--827, 2002.

\bibitem[Bag96]{BaglCalvReic1996}
J.~Baglama, D.~Calvetti and L.~Reichel.
\newblock Iterative methods for the computation of a few eigenvalues of a large
  symmetric matrix.
\newblock \emph{BIT}, \textbf{36}(3):pages 400--440, 1996.

\bibitem[Bai05]{BaiYinShi+2005}
X.~Bai, B.~Yin, Q.~Shi and Y.~Sun.
\newblock Face recognition based on supervised locally linear embedding method.
\newblock \emph{Journal of Information and Computational Science},
  \textbf{2}(4):pages 641--–646, 2005.

\bibitem[Bar98a]{Barr1998}
W.~A. Barrett.
\newblock A survey of face recognition algorithms and testing results.
\newblock \emph{Systems and Computers}, \textbf{1}:pages 301--305, 1998.

\bibitem[Bar98b]{BarrRissYu1998}
A.~R. Barron, J.~Rissanen and B.~Yu.
\newblock The minimum description length principle in coding and modeling.
\newblock \emph{IEEE Transactions on Information Theory}, \textbf{44}(6):pages
  2743--2772, 1998.

\bibitem[Bar98c]{BartLadeSejn1998}
M.~S. Bartlett, H.~M. Lades and T.~J. Sejnowski.
\newblock Independent component representations for face recognition.
\newblock \emph{In Proc. SPIE Symposium on Electronic Imaging: Science and
  Technology; Conference on Human Vision and Electronic Imaging {III}},
  \textbf{3299}:pages 528--539, 1998.

\bibitem[Bar02]{BartMoveSejn2002}
M.~S. Bartlett, J.~R. Movellan and T.~J. Sejnowski.
\newblock Face recognition by independent component analysis.
\newblock \emph{IEEE Transactions on Neural Networks (TNN)},
  \textbf{13}(6):pages 1450--1464, 2002.

\bibitem[Bel96]{BelhKrie1996}
P.~N. Belhumeur and D.~J. Kriegman.
\newblock What is the set of images of an object under all possible lighting
  conditions?
\newblock \emph{In Proc. IEEE Conference on Computer Vision and Pattern
  Recognition (CVPR)}, pages 270--277, 1996.

\bibitem[Bel97]{BelhHespKrie1997}
P.~N. Belhumeur, J.~P. Hespanha and D.~J. Kriegman.
\newblock Eigenfaces vs. fisherfaces: Recognition using class specific linear
  projection.
\newblock \emph{IEEE Transactions on Pattern Analysis and Machine Intelligence
  (TPAMI)}, \textbf{19}(7):pages 711--720, 1997.

\bibitem[Bel98]{BelhKrie1998}
P.~N. Belhumeur and D.~J. Kriegman.
\newblock What is the set of images of an object under all possible
  illumination conditions?
\newblock \emph{International Journal of Computer Vision (IJCV)},
  \textbf{28}(3):pages 245--260, 1998.

\bibitem[Ber04]{BergBergEdwa+2004}
T.~L. Berg, A.~C. Berg, J.~Edwards, M.~Maire, R.~White, Y.~W. Teh,
  E.~Learned-Miller and D.~A. Forsyth.
\newblock Names and faces in the news.
\newblock \emph{In Proc. IEEE Conference on Computer Vision and Pattern
  Recognition (CVPR)}, \textbf{2}:pages 848--854, 2004.

\bibitem[Bev01]{BeveSheDrap2001}
J.~R. Beveridge, K.~She and B.~A. Draper.
\newblock A nonparametric statistical comparison of principal component and
  linear discriminant subspaces for face recognition.
\newblock \emph{In Proc. IEEE Conference on Computer Vision and Pattern
  Recognition (CVPR)}, \textbf{1}:pages 535--542, 2001.

\bibitem[Bic94]{BichPent1994}
M.~Bichsel and A.~P. Pentland.
\newblock Human face recognition and the face image set's topology.
\newblock \emph{Computer Vision, Graphics and Image Processing: Image
  Understanding}, \textbf{59}(2):pages 254--261, 1994.

\bibitem[Big97]{BiguBiguDuc+1997}
E.~S. Bigun, J.~Bigun, B.~Duc and S.~Fischer.
\newblock Expert conciliation for multimodal person authentication systems
  using {B}ayesian statistics.
\newblock \emph{In Proc. International Conference on Audio- and Video-based
  Biometric Person Authentication (AVBPA)}, pages 291--300, 1997.

\bibitem[Bis95]{Bish1995}
C.~M. Bishop.
\newblock \emph{Neural Networks for Pattern Recognition}.
\newblock Oxford University Press, Oxford, England, 1995.

\bibitem[Bj{\"o}73]{BjorGolu1973}
{\AA}.~Bj{\"o}rck and G.~H. Golub.
\newblock Numerical methods for computing angles between linear subspaces.
\newblock \emph{Mathematics of Computation}, \textbf{27}(123):pages 579–--594,
  1973.

\bibitem[Bla98]{BlacJeps1998}
M.~J. Black and A.~D. Jepson.
\newblock Recognizing facial expressions in image sequences using local
  parameterized models of image motion.
\newblock \emph{International Journal of Computer Vision (IJCV)},
  \textbf{26}(1):pages 63--84, 1998.

\bibitem[Bla99]{BlanVett1999}
V.~Blanz and T.~Vetter.
\newblock A morphable model for the synthesis of {3D} faces.
\newblock \emph{In Proc. Conference on Computer Graphics (SIGGRAPH)}, pages
  187--194, 1999.

\bibitem[Bla03]{BlanVett2003}
V.~Blanz and T.~Vetter.
\newblock Face recognition based on fitting a {3D} morphable model.
\newblock \emph{IEEE Transactions on Pattern Analysis and Machine Intelligence
  (TPAMI)}, \textbf{25}(9):pages 1063--1074, 2003.

\bibitem[Ble66]{Bled1966}
W.~W. Bledsoe.
\newblock Man-machine facial recognition.
\newblock \emph{Technical Report PRI:22, Panoramic Research Inc.}, 1966.

\bibitem[Bol03]{Bolm2003}
D.~S. Bolme.
\newblock Elastic bunch graph matching.
\newblock Master's thesis, Colorado State University, 2003.

\bibitem[{Bos}02]{Bost2002}
{Boston Globe}.
\newblock Face recognition fails in {Boston} airport.
\newblock July 2002.

\bibitem[{Bri}04]{BBC2004}
{British Broadcasting Corporation}.
\newblock Doubts over passport face scans.
\newblock \emph{{BBC News Online, UK Edition}}, October 21, 2004.
\newblock {\url{http://news.bbc.co.uk/1/hi/uk/3762398.stm}}.

\bibitem[Bro06]{BrosJohnShot+2006}
G.~Brostow, M.~Johnson, J.~Shotton, O.~Arandjelovi{\'c}, V.~Kwatra and
  R.~Cipolla.
\newblock Semantic photo synthesis.
\newblock \emph{In Proc. Eurographics}, 2006.

\bibitem[Bru93]{BrunPogg1993a}
R.~Brunelli and T.~Poggio.
\newblock Face recognition: Features vs. templates.
\newblock \emph{IEEE Transactions on Pattern Analysis and Machine Intelligence
  (TPAMI)}, \textbf{15}(10):pages 1042--1052, 1993.

\bibitem[Bru95a]{BrunFala1995}
R.~Brunelli and D.~Falavigna.
\newblock Person recognition using multiple cues.
\newblock \emph{IEEE Transactions on Pattern Analysis and Machine Intelligence
  (TPAMI)}, \textbf{17}(10):pages 955--966, 1995.

\bibitem[Bru95b]{BrunFalaPogg+1995}
R.~Brunelli, D.~Falavigna, T.~Poggio and L.~Stringa.
\newblock Automatic person recognition by using acoustic and geometric
  features.
\newblock \emph{Machine Vision and Applications}, \textbf{8}(5):pages 317--325,
  1995.

\bibitem[Bud04]{BuddPavlKaka2004}
P.~Buddharaju, I.~Pavlidis and I.~Kakadiaris.
\newblock Face recognition in the thermal infrared spectrum.
\newblock \emph{In Proc. IEEE International Workshop on Object Tracking and
  Classification Beyond the Visible Spectrum (OTCBVS)}, pages 133--138, 2004.

\bibitem[Bud05]{BuddPavlTsia2005}
P.~Buddharaju, I.~T. Pavlidis and P.~Tsiamyrtzis.
\newblock Physiology-based face recognition.
\newblock \emph{In Proc. IEEE Conference on Advanced Video and Singal Based
  Surveillance (AVSS)}, pages 354--359, 2005.

\bibitem[Buh94]{BuhmLadeEeck1994}
J.~M. Buhmann, M.~Lades and F.~Eeckmann.
\newblock Illumination-invariant face recognition with a contrast sensitive
  silicon retina.
\newblock \emph{Advances in Neural Information Processing Systems (NIPS)},
  pages 769--776, 1994.

\bibitem[Bur98]{Burg1998}
C.~J.~C. Burges.
\newblock A tutorial on support vector machines for pattern recognition.
\newblock \emph{Data Mining and Knowledge Discovery}, \textbf{2}(2):pages
  121--167, 1998.

\bibitem[BY98]{BenAbdeMayo1998}
S.~Ben-Yacoub, Y.~Abdeljaoued and E.~Mayoraz.
\newblock Fusion of face and speech data for person identity verification.
\newblock \emph{IEEE Transactions on Neural Networks (TNN)},
  \textbf{10}(5):pages 1065--1075, 1998.

\bibitem[Cam00]{CampFeriCesa2000}
T.~E. de~Campos, R.~S. Feris and R.~M. Cesar~Junior.
\newblock Eigenfaces versus eigeneyes: First steps toward performance
  assessment of representations for face recognition.
\newblock \emph{In Proc. Mexican International Conference on Artificial
  Intelligence}, pages 193--201, 2000.

\bibitem[Can86]{Cann1986}
J.~Canny.
\newblock A computational approach to edge detection.
\newblock \emph{IEEE Transactions on Pattern Analysis and Machine Intelligence
  (TPAMI)}, \textbf{8}(6):pages 679--698, 1986.

\bibitem[Cap00]{CappMaioMalt2000}
R.~Cappelli, D.~Maio and D.~Maltoni.
\newblock A computational approach to edge detection.
\newblock \emph{In Proc. International Workshop on Multiple Classifier
  Systems}, pages 351--361, 2000.

\bibitem[Cha65]{ChanBled1965}
H.~Chan and W.~W. Bledsoe.
\newblock A man-machine facial recognition system: Some preliminary results.
\newblock \emph{Technical Report, Panoramic Research Inc.}, 1965.

\bibitem[Cha99]{ChatBorsPita1999}
V.~Chatzis, A.~G. Bors and I.~Pitas.
\newblock Multimodal decision-level fusion for person authentication.
\newblock \emph{IEEE Transactions on Systems, Man, and Cybernetics, Part A:
  Systems and Humans}, \textbf{29}(6):pages 674--681, 1999.

\bibitem[Cha03]{ChanSeetBurt+2003}
H.~L. Chang, H.~Seetzen, A.~M. Burton and A.~Chaudhuri.
\newblock Face recognition is robust with incongruent image resolution:
  Relationship to security video images.
\newblock \emph{Journal of Experimental Psychology: Applied}, \textbf{9}:pages
  33--41, 2003.

\bibitem[Che97]{ChenDesaZhan1997}
Z.-Y. Chen, M.~Desai and X.-P. Zhang.
\newblock Feedforward neural networks with multilevel hidden neurons for
  remotely sensed image classification.
\newblock \emph{In Proc. IEEE International Conference on Image Processing
  (ICIP)}, \textbf{2}:pages 653--656, 1997.

\bibitem[Che03]{ChenFlynBowy2003}
X.~Chen, P.~Flynn and K.~Bowyer.
\newblock Visible-light and infrared face recognition.
\newblock \emph{In Proc. Workshop on Multimodal User Authentication}, pages
  48--55, 2003.

\bibitem[Che05]{ChenFlynBowy2005}
X.~Chen, P.~Flynn and K.~Bowyer.
\newblock {IR} and visible light face recognition.
\newblock \emph{Computer Vision and Image Understanding (CVIU)},
  \textbf{99}(3):pages 332--358, 2005.

\bibitem[Coo95]{CootTaylCoopGrah1995}
T.~Cootes, C.~Taylor, D.~Cooper and J.~Graham.
\newblock Active shape models - their training and application.
\newblock \emph{Computer Vision and Image Understanding}, \textbf{61}(1):pages
  38--59, 1995.

\bibitem[Coo98]{CootEdwaTayl1998}
T.~F. Cootes, G.~J. Edwards and C.~J. Taylor.
\newblock Active appearance models.
\newblock \emph{In Proc. European Conference on Computer Vision (ECCV)},
  \textbf{2}:pages 484--498, 1998.

\bibitem[Coo99a]{CootTayl1999}
T.~Cootes and C.~Taylor.
\newblock A mixture model for representing shape.
\newblock \emph{Image and Vision Computing (IVC)}, \textbf{17}(8):pages
  567--573, 1999.

\bibitem[Coo99b]{CootEdwaTayl1999}
T.~F. Cootes, G.~J. Edwards and C.~J. Taylor.
\newblock Comparing active shape models with active appearance models.
\newblock \emph{In Proc. British Machine Vision Conference (BMVC)}, pages
  173--182, 1999.

\bibitem[Coo02]{CootKitt2002}
T.~F. Cootes and P.~Kittipanya-ngam.
\newblock Comparing variations on the active appearance models.
\newblock \emph{In Proc. British Machine Vision Conference (BMVC)}, pages
  837--846, 2002.

\bibitem[Cor90]{CormLeisRive1990}
T.~H. Cormen, C.~E. Leiserson and R.~L. Rivest.
\newblock \emph{Introduction to Algorithms}.
\newblock MIT Press, 1990.
\newblock ISBN 978-0-262-03293-3.

\bibitem[Cov91]{CoveThom1991}
T.~M. Cover and J.~A. Thomas.
\newblock \emph{Elements of Information Theory.}
\newblock Wiley, New York, 1991.

\bibitem[Cra99]{CrawCostKatoAkam1999}
I.~Craw, N.~P. Costen, T.~Kato and S.~Akamatsu.
\newblock How should we represent faces for automatic recognition?
\newblock \emph{IEEE Transactions on Pattern Analysis and Machine Intelligence
  (TPAMI)}, \textbf{21}:pages 725--736, 1999.

\bibitem[Cri04]{CrisCootSco2004}
D.~Cristinacce, C.~T. F. and I.~Scott.
\newblock A multistage approach to facial feature detection.
\newblock \emph{In Proc. British Machine Vision Conference (BMVC)},
  \textbf{1}:pages 277--286, 2004.

\bibitem[Dah01]{DahmKeysNey+2001}
J.~Dahmen, D.~Keysers, H.~Ney and M.~O. G{\"{u}}ld.
\newblock Statistical image object recognition using mixture densities.
\newblock \emph{Journal of Mathematical Imaging and Vision},
  \textbf{14}(3):pages 285--296, 2001.

\bibitem[Dau80]{Daug1980}
J.~Daugman.
\newblock Two-dimensional spectral analysis of cortical receptive field
  profiles.
\newblock \emph{Vision Research}, \textbf{20}:pages 847--856, 1980.

\bibitem[Dau88]{Daug1988}
J.~Daugman.
\newblock Complete discrete {2-D Gabor} transforms by neural networks for image
  analysis and compression.
\newblock \emph{IEEE Transactions on Acoustics, Speech and Signal Processing},
  \textbf{36}:pages 1169--1179, 1988.

\bibitem[Dau92]{Daug1992}
J.~Daugman.
\newblock High confidence personal identification by rapid video analysis of
  iris texture.
\newblock \emph{In Proc. IEEE International Carnahan Conference on Security
  Technology}, pages 50--60, 1992.

\bibitem[DeC98]{deCaMetaSton1998}
D.~DeCarlos, D.~Metaxas and M.~Stone.
\newblock An anthropometric face model using variational techniques.
\newblock \emph{In Proc. Conference on Computer Graphics (SIGGRAPH)}, pages
  67–--74, 1998.

\bibitem[Dem77]{DempLairRubi1977}
A.~P. Dempster, N.~M. Laird and D.~B. Rubin.
\newblock Maximum likelihood from incomplete data via the em algorithm.
\newblock \emph{Journal of the Royal Statistical Society}, \textbf{39}:pages
  1--38, 1977.

\bibitem[deM93]{deMeCott1993}
D.~deMers and G.~Cottrell.
\newblock Non-linear dimensionality reduction.
\newblock \emph{Advances in Neural Information Processing Systems},
  \textbf{5}:pages 580--587, 1993.

\bibitem[DiP91]{DPaol1991}
S.~DiPaola.
\newblock Extending the range of facial types.
\newblock \emph{The Journal of Visualization and Computer Animation},
  \textbf{2}(4):pages 129--131, 1991.

\bibitem[Dor03]{DornAhlb2003}
F.~Dornaika and J.~Ahlberg.
\newblock Face model adaptation for tracking and active appearance model
  training.
\newblock \emph{In Proc. British Machine Vision Conference (BMVC)}, pages
  559–--568, 2003.

\bibitem[Dra03]{DrapBaekBartBeve2003}
B.~A. Draper, K.~Baek, M.~S. Bartlett and J.~R. Beveridge.
\newblock Recognizing faces with {PCA} and {ICA}.
\newblock \emph{Computer Vision and Image Understanding}, \textbf{91}:pages
  115--137, 2003.

\bibitem[Dud00]{DudaHartStor2001}
R.~O. Duda, P.~E. Hart and D.~G. Stork.
\newblock \emph{Pattern Classification}.
\newblock John Wily \& Sons, Inc., New York, 2nd edition, 2000.
\newblock ISBN 0-471-05669-3.

\bibitem[Edw97]{EdwaTaylCoot1997}
G.~J. Edwards, C.~J. Taylor and T.~F. Cootes.
\newblock Learning to identify and track faces in image sequences.
\newblock \emph{In Proc. British Machine Vision Conference (BMVC)}, pages
  130--139, 1997.

\bibitem[Edw98a]{EdwaTaylCoot1998}
G.~Edwards, C.~Taylor and T.~Cootes.
\newblock Interpreting face images using active appearance models.
\newblock \emph{In Proc. IEEE International Conference on Automatic Face and
  Gesture Recognition (FG)}, pages 300--305, 1998.

\bibitem[Edw98b]{EdwaCootTayl1998}
G.~J. Edwards, T.~F. Cootes and C.~J. Taylor.
\newblock Face recognition using active appearance models.
\newblock \emph{In Proc. European Conference on Computer Vision (ECCV)},
  \textbf{2}:pages 581--595, 1998.

\bibitem[Edw99]{EdwaTaylCoot1999}
G.~Edwards, C.~Taylor and T.~Cootes.
\newblock Improving identification performance by integrating evidence from
  sequences.
\newblock \emph{In Proc. IEEE Conference on Computer Vision and Pattern
  Recognition (CVPR)}, \textbf{1}:pages 486--491, 1999.

\bibitem[Eve04]{EverZiss2004}
M.~Everingham and A.~Zisserman.
\newblock Automated person identification in video.
\newblock \emph{In Proc. IEEE International Conference on Image and Video
  Retrieval (CIVR)}, pages 289--298, 2004.

\bibitem[Fag06]{FaggPaplChin2006}
N.~Faggian, A.~Paplinski and T.-J. Chin.
\newblock Face recognition from video using active appearance model
  segmentation.
\newblock \emph{In Proc. IAPR International Conference on Pattern Recognition
  (ICPR)}, pages 287--290, 2006.

\bibitem[Fel05]{FelzHutt2005}
P.~F. Felzenszwalb and D.~Huttenlocher.
\newblock Pictorial structures for object recognition.
\newblock \emph{International Journal of Computer Vision (IJCV)},
  \textbf{61}(1):pages 55--79, 2005.

\bibitem[Fer01]{FeraBernVial+2001}
R.~Feraud, O.~Bernier, J.-E. Viallet and M.~Collobert.
\newblock A fast and accurate face detector based on neural networks.
\newblock \emph{International Journal of Computer Vision (IJCV)},
  \textbf{23}(1):pages 42--53, 2001.

\bibitem[Fig02]{FiguJain2003}
M.~Figueiredo and A.~Jain.
\newblock Unsupervised learning of finite mixture models.
\newblock \emph{IEEE Transactions on Pattern Analysis and Machine Intelligence
  (TPAMI)}, \textbf{24}(3):pages 381--396, 2002.

\bibitem[Fis81]{FiscBoll1981}
M.~A. Fischler and R.~C. Bolles.
\newblock Random sample consensus: A paradigm for model fitting with
  applications to image analysis and automated cartography.
\newblock \emph{IEEE Transactions on Computers}, \textbf{24}(6):pages 381--395,
  1981.

\bibitem[Fit02]{FitzZiss2002}
A.~Fitzgibbon and A.~Zisserman.
\newblock On affine invariant clustering and automatic cast listing in movies.
\newblock \emph{In Proc. European Conference on Computer Vision (ECCV)}, pages
  304--320, 2002.

\bibitem[Fre95]{FreuScha1995}
Y.~Freund and R.~E. Schapire.
\newblock A decision-theoretic generalization of on-line learning and an
  application to boosting.
\newblock \emph{In Proc. European Conference on Computational Learning Theory},
  pages 23--37, 1995.

\bibitem[Fri03]{FrieYesh2003}
G.~Friedrich and Y.~Yeshurun.
\newblock Seeing people in the dark: Face recognition in infrared images.
\newblock \emph{In Proc. British Machine Vision Conference (BMVC)}, pages
  348--359, 2003.

\bibitem[Fuk98]{FukuYama1998}
K.~Fukui and O.~Yamaguchi.
\newblock Facial feature point extraction method based on combination of shape
  extraction and pattern matching.
\newblock \emph{Systems and Computers in Japan}, \textbf{29}(6):pages
  2170--2177, 1998.

\bibitem[Fuk03]{FukuYama2003}
K.~Fukui and O.~Yamaguchi.
\newblock Face recognition using multi-viewpoint patterns for robot vision.
\newblock \emph{International Symposium of Robotics Research}, 2003.

\bibitem[Gab88]{Gabo1946}
D.~Gabor.
\newblock Theory of communication.
\newblock \emph{Journal of the Institute of Electrical Engineers},
  \textbf{93}(3):pages 429--457, 1888.

\bibitem[Gao02]{GaoLeun2002}
Y.~Gao and M.~K.~H. Leung.
\newblock Face recognition using line edge map.
\newblock \emph{IEEE Transactions on Pattern Analysis and Machine Intelligence
  (TPAMI)}, \textbf{24}(6):pages 764--779, 2002.

\bibitem[Gar04]{GarcDela2004}
C.~Garcia and M.~Delakis.
\newblock Convolutional face finder: A neural architecture for fast and robust
  face detection.
\newblock \emph{IEEE Transactions on Pattern Analysis and Machine Intelligence
  (TPAMI)}, \textbf{26}(11):pages 1408--1423, 2004.

\bibitem[Gav00]{Gavr2000}
D.~M. Gavrila.
\newblock Pedestrian detection from a moving vehicle.
\newblock \emph{In Proc. European Conference on Computer Vision (ECCV)},
  \textbf{2}:pages 37--49, 2000.

\bibitem[Geo98]{GeorKrieBelh1998}
A.~S. Georghiades, D.~J. Kriegman and P.~N. Belhumeur.
\newblock Illumination cones for recognition under variable lighting: Faces.
\newblock \emph{In Proc. IEEE Conference on Computer Vision and Pattern
  Recognition (CVPR)}, pages 52--58, 1998.

\bibitem[Git85]{Gitt1985}
R.~Gittins.
\newblock \emph{Canonical Analysis: A Review with Applications in Ecology.}
\newblock Springer-Verlag, 1985.

\bibitem[Gol72]{GoldHarmLesk1972}
A.~J. Goldstein, L.~D. Harmon and A.~B. Lesk.
\newblock Man-machine interaction in human-face identification.
\newblock \emph{Bell System Technical Journal}, \textbf{51}(2):pages 399--427,
  1972.

\bibitem[Gon92]{GonzWood1992}
R.~Gonzalez and R.~Woods.
\newblock \emph{Digital Image Processing.}
\newblock Addison-Wesley Publishing Company, 1992.

\bibitem[Gor05]{Goro2005a}
D.~O. Gorodnichy.
\newblock Associative neural networks as means for low-resolution video-based
  recognition.
\newblock \emph{In Proc. International Joint Conference on Neural Networks},
  2005.

\bibitem[Gra18]{Gray1918}
H.~Gray.
\newblock \emph{Anatomy of the Human Body.}
\newblock Philadelphia: Lea {\&} Febiger, 20th edition, 1918.

\bibitem[Gra03]{Grae2003}
G.~A. von Graevenitz.
\newblock Biometrics in access control.
\newblock \emph{{A\&S} International}, \textbf{50}:pages 102--104, 2003.

\bibitem[Gri92]{GrimStir1992}
G.~R. Grimmett and D.~R. Stirzaker.
\newblock \emph{Probability and Random Processes.}
\newblock Clarendon Press, Oxford, 2nd edition, 1992.

\bibitem[Gro00]{GrosYangWaib2000}
R.~Gross, J.~Yang and A.~Waibel.
\newblock Growing {G}aussian mixture models for pose invariant face
  recognition.
\newblock \emph{In Proc. IAPR International Conference on Pattern Recognition
  (ICPR)}, \textbf{1}:pages 1088--1091, 2000.

\bibitem[Gro01]{GrosShiCohn2001}
R.~Gross, J.~She and J.~F. Cohn.
\newblock Quo vadis face recognition.
\newblock \emph{In Proc. Workshop on Empirical Evaluation Methods in Computer
  Vision}, \textbf{1}:pages 119--132, 2001.

\bibitem[Gro04]{GrosMattBake2004a}
R.~Gross, I.~Matthews and S.~Baker.
\newblock Generic vs. person specific active appearance models.
\newblock \emph{In Proc. British Machine Vision Conference (BMVC)}, pages
  457–--466, 2004.

\bibitem[Gro06]{GrosMattBake2006}
R.~Gross, I.~Matthews and S.~Baker.
\newblock Active appearance models with occlusion.
\newblock \emph{Image and Vision Computing (special issue on Face Processing in
  Video)}, \textbf{1}(6):pages 593--604, 2006.

\bibitem[Gya04]{GyaoBebiPavl2004}
A.~Gyaourova, G.~Bebis and I.~Pavlidis.
\newblock Fusion of infrared and visible images for face recognition.
\newblock \emph{In Proc. European Conference on Computer Vision (ECCV)},
  \textbf{4}:pages 456--468, 2004.

\bibitem[Hal00]{HallMarsMart2000}
P.~Hall, D.~Marshall and R.~Martin.
\newblock Merging and splitting eigenspace models.
\newblock \emph{IEEE Transactions on Pattern Analysis and Machine Intelligence
  (TPAMI)}, \textbf{22}(9):pages 1042--1048, 2000.

\bibitem[Hal04]{HallHick2004}
P.~M. Hall and Y.~Hicks.
\newblock A method to add {G}aussian mixture models.
\newblock \emph{Tech. Report, University of Bath}, 2004.

\bibitem[Ham95]{HampJainWeym1995}
A.~Hampapur, R.~C. Jain and T.~Weymouth.
\newblock Production model based digital vieo segmentation.
\newblock \emph{Multimedia Tools and Applications}, \textbf{1}(1):pages 9--46,
  1995.

\bibitem[Ham98a]{HamaAbuGust1998}
G.~Hamarneh, R.~Abu-Gharbieh and T.~Gustavsson.
\newblock Active shape models - part {I}: Modeling shape and gray level
  variations.
\newblock \emph{In Proc. of the Swedish Symposium on Image Analysis}, pages
  125--128, 1998.

\bibitem[Ham98b]{HamaAbuGust1998a}
G.~Hamarneh, A.-G. R. and T.~Gustavsson.
\newblock Active shape models - part {II}: Image search and classification.
\newblock \emph{In Proc. of the Swedish Symposium on Image Analysis}, pages
  129--132, 1998.

\bibitem[Hei00]{HeisPoggPont2000}
B.~Heisele, T.~Poggio and M.~Pontil.
\newblock Face detection in still gray images.
\newblock \emph{{A.I. Memo No. 1687, C.B.C.L. Paper No. 187 Center for
  Biological and Computational Learning, M.I.T.}}, 2000.

\bibitem[Heo03a]{HeoAbiKong+2003}
J.~Heo, B.~Abidi, S.~G. Kong and M.~Abidi.
\newblock Performance comparison of visual and thermal signatures for face
  recognition.
\newblock \emph{Biometric Consortium Conference}, 2003.

\bibitem[Heo03b]{HeoAbidPaik+2003}
J.~Heo, B.~Abidi, J.~Paik and M.~A. Abidi.
\newblock Face recognition: Evaluation report for {FaceIt$^{\circledR}$}.
\newblock \emph{In Proc. International Conference on Quality Control by
  Artificial Vision}, \textbf{5132}:pages 551--558, 2003.

\bibitem[Heo04]{HeoKongAbidAbid2004}
J.~Heo, S.~G. Kong, B.~R. Abidi and M.~A. Abidi.
\newblock Fusion of visual and thermal signatures with eyeglass removal for
  robust face recognition.
\newblock \emph{In Proc. IEEE Conference on Computer Vision and Pattern
  Recognition Workshops (CVPRW)}, page 122, 2004.

\bibitem[Hic03]{HickHallMars2003}
Y.~A. Hicks, P.~M. Hall and A.~D. Marshall.
\newblock A method to add {H}idden {M}arkov {M}odels with application to
  learning articulated motion.
\newblock \emph{In Proc. British Machine Vision Conference (BMVC)}, pages
  489--498, 2003.

\bibitem[Hin06]{HintSala2006}
G.~E. Hinton and R.~R. Salakhutdinov.
\newblock Reducing the dimensionality of data with neural networks.
\newblock \emph{Science}, \textbf{313}(5786):pages 504--507, 2006.

\bibitem[Hje01]{Hjel2001}
E.~Hjelm{\aa}s.
\newblock Face detection: A survey.
\newblock \emph{Computer Vision and Image Understanding}, (83):pages 236--274,
  2001.

\bibitem[Hon04]{HongKaplSmit2004}
P.~S. Hong, L.~M. Kaplan and M.~J.~T. Smith.
\newblock A comparison of the octave-band directional filter bank and {G}abor
  filters for texture classification.
\newblock \emph{In Proc. IEEE International Conference on Image Processing
  (ICIP)}, \textbf{3}:pages 1541--1544, 2004.

\bibitem[Hot36]{Hote1936}
H.~Hotelling.
\newblock Relations between two sets of variates.
\newblock \emph{Biometrika}, \textbf{28}:pages 321--372, 1936.

\bibitem[Hua05]{HuanAiLiLao2005}
C.~Huang, H.~Ai, Y.~Li and S.~Lao.
\newblock Vector boosting for rotation invariant multi-view face detection.
\newblock \emph{In Proc. IEEE International Conference on Computer Vision
  (ICCV)}, \textbf{1}:pages 446--453, 2005.

\bibitem[{Ide}03]{Iden}
{Identix Ltd.}
\newblock Faceit.
\newblock \emph{{\url{http://www.FaceIt.com/}}}, 2003.

\bibitem[Im03]{ImChoiKim2003}
S.~K. Im, H.~S. Choi and S.~W. Kim.
\newblock A direction-based vascular pattern extraction algorithm for hand
  vascular pattern verification.
\newblock \emph{ETRI Journal}, \textbf{25}(2):pages 101--108, 2003.

\bibitem[{Int}]{Int_}
{International Biometric Group}.
\newblock \emph{{\url{http://www.biometricgroup.com/}}}.

\bibitem[Isa98]{IsarBlak1998}
M.~Isard and A.~Blake.
\newblock {CONDENSATION} -- conditional density propagation for visual
  tracking.
\newblock \emph{International Journal of Computer Vision (IJCV)},
  \textbf{29}(1):pages 5--28, 1998.

\bibitem[Jai97]{JainHongPank+1997}
A.~Jain, L.~Hong, S.~Pankanti and R.~Bolle.
\newblock An identity authentication system using fingerprints.
\newblock \emph{IEEE paper}, \textbf{85}(9):pages 1365--1388, 1997.

\bibitem[Jin00]{JingMari2000}
Z.~Jing and R.~Mariani.
\newblock Glasses detection and extraction by deformable contour.
\newblock \emph{In Proc. IAPR International Conference on Pattern Recognition
  (ICPR)}, \textbf{2}:pages 933--936, 2000.

\bibitem[Joh01]{JohnSina2001}
D.~H. Johnson and S.~Sinanovi{\'c}.
\newblock Symmetrizing the {Kullback-Leibler} distance.
\newblock \emph{Technical report, Rice University}, 2001.

\bibitem[Joh06]{JohnBrosShot+2006}
M.~Johnson, G.~Brostow, J.~Shotton, O.~Arandjelovi{\'c}, V.~Kwatra and
  R.~Cipolla.
\newblock Semantic photo synthesis.
\newblock \emph{Computer Graphics Forum}, \textbf{3}(25), 2006.

\bibitem[Jon87]{JonePalm1987}
J.~P. Jones and L.~A. Palmer.
\newblock An evaluation of the two-dimensional {Gabor} filter model of simple
  receptive fields in cat striate cortex.
\newblock \emph{Neurophysiology}, \textbf{58}(6):pages 1233--1258, 1987.

\bibitem[Jon99]{JoneRehg1999}
M.~J. Jones and J.~M. Rehg.
\newblock Statistical color models with application to skin detection.
\newblock In \emph{In Proc. IEEE Conference on Computer Vision and Pattern
  Recognition (CVPR)}, pages 274--280. 1999.

\bibitem[Kai74]{Kail1974}
T.~Kailath.
\newblock A view of three decades of linear filtering theory.
\newblock \emph{IEEE Transactions on Information Theory}, \textbf{20}(2):pages
  146--181, 1974.

\bibitem[Kan73]{Kana1973}
T.~Kanade.
\newblock Picture processing system by computer complex and recognition of
  human faces.
\newblock Ph.D. thesis, Kyoto University, 1973.

\bibitem[Kan02]{KangCootTayl2002}
H.~Kang, T.~F. Cootes and C.~J. Taylor.
\newblock A comparison of face verification algorithms using appearance models.
\newblock \emph{In Proc. British Machine Vision Conference (BMVC)}, pages
  477--486, 2002.

\bibitem[Kas87]{KassWitkTerz1987}
M.~Kass, A.~Witkin and D.~Terzopoulos.
\newblock Snakes: Active contour models.
\newblock \emph{In Proc. IEEE International Conference on Computer Vision
  (ICCV)}, pages 259--268, 1987.

\bibitem[Kay72]{KayaKoba1972}
Y.~Kaya and K.~Kobayashi.
\newblock A basic study on human face recognition.
\newblock \emph{Frontiers of Pattern Recognition}, pages 265--289, 1972.

\bibitem[Kel70]{Kell1970}
M.~Kelly.
\newblock Visual identification of people by computer.
\newblock \emph{Technical Report AI-130, Stanford AI Project}, 1970.

\bibitem[Kim03]{KimmEladShak+2003}
R.~Kimmel, M.~Elad, D.~Shaked, R.~Keshet and I.~Sobel.
\newblock A variational framework for retinex.
\newblock \emph{International Journal of Computer Vision (IJCV)},
  \textbf{52}(1):pages 7--23, 2003.

\bibitem[Kim05a]{KimAranCipo2005}
T.~Kim, O.~Arandjelovi{\'c} and R.~Cipolla.
\newblock Learning over sets using boosted manifold principal angles {(BoMPA)}.
\newblock \emph{In Proc. British Machine Vision Conference (BMVC)},
  \textbf{2}:pages 779--788, 2005.

\bibitem[Kim05b]{KimKitt2005}
T.-K. Kim and J.~V. Kittler.
\newblock Locally linear discriminant analysis for multimodally distributed
  classes for face recognition with a single model image.
\newblock \emph{IEEE Transactions on Pattern Analysis and Machine Intelligence
  (TPAMI)}, \textbf{27}(3):pages 318--327, 2005.

\bibitem[Kim06]{KimKittCipo2006}
T.-K. Kim, J.~V. Kittler and R.~Cipolla.
\newblock Learning discriminative canonical correlations for object recognition
  with image sets.
\newblock \emph{In Proc. European Conference on Computer Vision (ECCV)}, pages
  251--262, 2006.

\bibitem[Kim07]{KimAranCipo2007}
T.-K. Kim, O.~Arandjelovi{\'c} and R.~Cipolla.
\newblock Boosted manifold principal angles for image set-based recognition.
\newblock \emph{Pattern Recognition (PR)}, \textbf{40}(9):pages 2475--2484,
  2007.

\bibitem[Kin97]{KingXu1997}
I.~King and L.~Xu.
\newblock Localized principal component analysis learning for face feature
  extraction and recognition.
\newblock \emph{In Proc. Workshop on 3D Computer Vision}, pages 124--128, 1997.

\bibitem[Kir90]{KirbSiro1990}
M.~Kirby and L.~Sirovich.
\newblock Application of the {K}arhunen-{L}oève procedure for the
  characterization of human faces.
\newblock \emph{IEEE Transactions on Pattern Analysis and Machine
  Intelligence}, \textbf{12}(1):pages 103--108, 1990.

\bibitem[Kit98]{KittHateDuinMata1998}
J.~Kittler, M.~Hatef, R.~Duin and J.~Matas.
\newblock On combining classifiers.
\newblock \emph{IEEE Transactions on Pattern Analysis and Machine
  Intelligence}, \textbf{20}(3):pages 226--239, 1998.

\bibitem[Koh77]{Koho1977}
T.~Kohonen.
\newblock \emph{Associative Memory: A System Theoretical Approach.}
\newblock Springer-Verlag, 1977.

\bibitem[Kon05]{KongHeoAbidPaik+2005}
S.~Kong, J.~Heo, B.~Abidi, J.~Paik and M.~Abidi.
\newblock Recent advances in visual and infrared face recognition -- a review.
\newblock \emph{Computer Vision and Image Understanding (CVIU)},
  \textbf{97}(1):pages 103--135, 2005.

\bibitem[Kot98]{KotrTefaPita1998}
C.~Kotropoulos, A.~Tefas and I.~Pitas.
\newblock Face authentication using variants of elastic graph matching based on
  mathematical morphology that incorporate local discriminant coefficients.
\newblock \emph{In Proc. IEEE Conference on Computer Vision and Pattern
  Recognition (CVPR)}, \textbf{1}:pages 814--819, 1998.

\bibitem[Kot00a]{KotrTefaPita2000}
C.~Kotropoulos, A.~Tefas and I.~Pitas.
\newblock Frontal face authentication using discriminating grids with
  morphological feature vectors.
\newblock \emph{IEEE Transactions on Multimedia}, \textbf{2}(1):pages 14--26,
  2000.

\bibitem[Kot00b]{KotrTefaPita2000a}
C.~Kotropoulos, A.~Tefas and I.~Pitas.
\newblock Frontal face authentication using morphological elastic graph
  matching.
\newblock \emph{IEEE Transactions on Image Processing (TIP)},
  \textbf{9}(4):pages 555--560, 2000.

\bibitem[Kot00c]{KotrTefaPita2000b}
C.~Kotropoulos, A.~Tefas and I.~Pitas.
\newblock Morphological elastic graph matching applied to frontal face
  authentication under well-controlled and real conditions.
\newblock \emph{Pattern Recognition (PR)}, \textbf{33}(12):pages 31--43, 2000.

\bibitem[Kr{\"u}00]{KrugHappSomm2000}
V.~Kr{\"u}ger, A.~Happe and G.~Sommer.
\newblock Affine real-time face tracking using {Gabor} wavelet networks.
\newblock \emph{In Proc. IAPR International Conference on Pattern Recognition
  (ICPR)}, \textbf{1}:pages 127--130, 2000.

\bibitem[Kr{\"u}02]{KrugSomm2002}
V.~Kr{\"u}ger and G.~Sommer.
\newblock {Gabor} wavelet networks for efficient head pose estimation.
\newblock \emph{Journal of the Optical Society of America},
  \textbf{19}(6):pages 1112--1119, 2002.

\bibitem[Le06]{LeSatoHoul2006}
D.-D. Le, S.~Satoh and M.~E. Houle.
\newblock Face retrieval in broadcsting news video by fusing temporal and
  intensity information.
\newblock \emph{In Proc. IEEE International Conference on Image and Video
  Retrieval (CIVR)}, pages 391--400, 2006.

\bibitem[Lee96]{Lee1996}
S.~Y. Lee, Y.~K. Ham and R.-H. Park.
\newblock Image representation using {2D Gabor} wavelets.
\newblock \emph{IEEE Transactions on Pattern Analysis and Machine Intelligence
  (TPAMI)}, \textbf{18}(10):pages 1--13, 1996.

\bibitem[Lee03]{LeeHoYangKrie2003}
K.~Lee, J.~Ho and D.~Kriegman.
\newblock Nine points of light: Acquiring subspaces for face recognition under
  variable lighting.
\newblock \emph{In Proc. IEEE Conference on Computer Vision and Pattern
  Recognition (CVPR)}, \textbf{1}:pages 519--526, 2003.

\bibitem[Lee04]{LeeMoghPfisMach2004}
J.~Lee, B.~Moghaddam, H.~Pfister and R.~Machiraju.
\newblock Finding optimal views for {3D} face shape modeling.
\newblock \emph{In Proc. IEEE International Conference on Automatic Face and
  Gesture Recognition (FG)}, pages 31--36, 2004.

\bibitem[Lee05]{LeeKrie2005}
K.~Lee and D.~Kriegman.
\newblock Online learning of probabilistic appearance manifolds for video-based
  recognition and tracking.
\newblock \emph{In Proc. IEEE Conference on Computer Vision and Pattern
  Recognition (CVPR)}, \textbf{1}:pages 852--859, 2005.

\bibitem[Li99]{LiLu1999}
S.~Li and J.~Lu.
\newblock Face recognition using the nearest feature line method.
\newblock \emph{IEEE Transactions on Neural Networks (TNN)},
  \textbf{10}(2):pages 439--443, 1999.

\bibitem[Li02]{LiZhuZhan+2002}
S.~Z. Li, L.~Zhu, Z.~Zhang, A.~Blake, H.~Zhang and H.~Shum.
\newblock Face recognition using the nearest feature line method.
\newblock \emph{In Proc. European Conference on Computer Vision (ECCV)},
  \textbf{4}:pages 67--81, 2002.

\bibitem[Li04]{LiJain2004}
S.~Z. Li and A.~K. Jain, editors.
\newblock \emph{Handbook of Face Recognition.}
\newblock Springer-Verlag, 2004.
\newblock ISBN 0-387-40595-x.

\bibitem[Li07]{LiChuLiao+2007}
S.~Z. Li, R.~Chu, S.~Liao and L.~Zhang.
\newblock Illumination invariant face recognition using near-infrared images.
\newblock \emph{IEEE Transactions on Pattern Analysis and Machine Intelligence
  (TPAMI)}, \textbf{29}(4):pages 627--639, 2007.

\bibitem[Lie98]{Lien1998}
R.~Lienhart.
\newblock Comparison of automatic shot boundary detection algorithms.
\newblock \emph{SPIE}, \textbf{3656}:pages 290--301, 1998.

\bibitem[Lin04]{LinFan2004}
C.~L. Lin and K.~C. Fan.
\newblock Biometric verification using thermal images of palm-dorsa vein
  patterns.
\newblock \emph{IEEE Transactions on Circuits and Systems for Video
  Technology}, \textbf{14}(2):pages 199--213, 2004.

\bibitem[Mac03]{Mack2003}
D.~J.~C. MacKay.
\newblock \emph{Information Theory, Inference and Learning Algorithms.}
\newblock Cambridge University Press, Cambridge, 2003.

\bibitem[Mae04]{MaedYamaFuku2004}
K.~Maeda, O.~Yamaguchi and K.~Fukui.
\newblock Towards 3-dimensional pattern recognition.
\newblock \emph{Statistical Pattern Recognition}, \textbf{3138}:pages
  1061--1068, 2004.

\bibitem[Mal98]{Malt1998}
E.~C. Malthouse.
\newblock Limitations of nonlinear {PCA} as performed with generic neural
  networks.
\newblock \emph{IEEE Transactions on Neural Networks (TNN)},
  \textbf{9}(1):pages 165--173, 1998.

\bibitem[Mal03]{MaltMaioJain+2003}
D.~Maloni, D.~Maio, A.~K. Jain and S.~Prabhakar.
\newblock \emph{Handbook of Fingerprint Recognition.}
\newblock Springer-Verlag, 2003.

\bibitem[Mar80]{Marc1980}
S.~Marcelja.
\newblock Mathematical description of the response of simple cortical cells..
\newblock \emph{Journal of the American Optical Society}, \textbf{70}:pages
  1297--1300, 1980.

\bibitem[Mar02]{Mart2002}
A.~M. Martinez.
\newblock Recognizing imprecisely localized, partially occluded and expression
  variant faces from a single sample per class.
\newblock \emph{IEEE Transactions on Pattern Analysis and Machine Intelligence
  (TPAMI)}, \textbf{24}(6):pages 748--763, 2002.

\bibitem[Mik01]{MikoChouSchm2001}
K.~Mikolajczyk, R.~Choudhury and C.~Schmid.
\newblock Face detection in a video sequence -- a temporal approach.
\newblock \emph{In Proc. IEEE Conference on Computer Vision and Pattern
  Recognition (CVPR)}, \textbf{2}:pages 96--101, 2001.

\bibitem[Mit05]{MitaKaneHori2005}
T.~Mita, T.~Kaneko and O.~Hori.
\newblock Joint haar-like features for face detection.
\newblock \emph{In Proc. IEEE International Conference on Computer Vision
  (ICCV)}, \textbf{2}:pages 1619--1626, 2005.

\bibitem[Miu04]{MiurNagaMiya2003}
N.~Miura, A.~Nagasaka and T.~Miyatake.
\newblock Feature extraction of finger vein patterns based on iterative line
  tracking and its application to personal identification.
\newblock \emph{Systems and Computers in Japan}, \textbf{35}(7):pages 61--71,
  2004.

\bibitem[Mog95]{MoghPent1995}
B.~Moghaddam and A.~Pentland.
\newblock An automatic system for model-based coding of faces.
\newblock \emph{In Proc. IEEE Data Compression Conference}, pages 1--5, 1995.

\bibitem[Mog98]{MoghWahiPent1998}
B.~Moghaddam, W.~Wahid and A.~Pentland.
\newblock Beyond eigenfaces - probabilistic matching for face recognition.
\newblock \emph{In Proc. IEEE International Conference on Automatic Face and
  Gesture Recognition (FG)}, pages 30--35, 1998.

\bibitem[Mog02]{MoghPent2002}
B.~Moghaddam and A.~Pentland.
\newblock Principal manifolds and probabilistic subspaces for visual
  recognition.
\newblock \emph{IEEE Transactions on Pattern Analysis and Machine Intelligence
  (TPAMI)}, \textbf{24}(6):pages 780--788, 2002.

\bibitem[MT89]{MagnMinhAnge+1989}
N.~Magnenat-Thalmann, H.~Minh, M.~Angelis and D.~Thalmann.
\newblock Design, transformation and animation of human faces.
\newblock \emph{The Visual Computer}, \textbf{5}:pages 32–--39, 1989.

\bibitem[Mun05]{MuneGaneArum+2005}
K.~Muneeswaran, L.~Ganesan, S.~Arumugam and K.~R. Soundar.
\newblock Texture classification with combined rotation and scale invariant
  wavelet features.
\newblock \emph{Pattern Recognition (PR)}, \textbf{38}(10):pages 1495--1506,
  2005.

\bibitem[{Nat}]{NCSC_}
{National Center for State Courts (NCSC)}.
\newblock Biometrics comparison chart.
\newblock \emph{{\url{http://ctl.ncsc.dni.us/biometweb/BMCompare.html}}}.

\bibitem[Nef96]{Nefi1996}
A.~V. Nefian.
\newblock Statistical approaches to face recognition.
\newblock Ph.D. thesis, Georgia Institute of Technology, 1996.

\bibitem[Nis06]{NishYama2006}
M.~Nishiyama and O.~Yamaguchi.
\newblock Face recognition using the classified appearance-based quotient
  image.
\newblock \emph{In Proc. IEEE International Conference on Automatic Face and
  Gesture Recognition (FG)}, pages 49--54, 2006.

\bibitem[Oja83]{Oja1983}
E.~Oja.
\newblock \emph{Subspace Methods of Pattern Recognition.}
\newblock Research Studies Press and J. Wiley, 1983.

\bibitem[Ots93]{OtsuTono1993}
O.~Otsuji and Y.~Tonomura.
\newblock Projection detecting filter for video cut detection.
\newblock \emph{In Proc. ACM International Conference on Multimedia}, pages
  251--257, 1993.

\bibitem[Pan06]{PangLiuYu2006}
Y.~Pang, Z.~Liu and N.~Yu.
\newblock A new nonlinear feature extraction method for face recognition.
\newblock \emph{NeuroComputing}, \textbf{69}(7--9):pages 949–--953, 2006.

\bibitem[Par75]{Park1975}
F.~I. Parke.
\newblock A model for human faces that allows speech synchronized animation.
\newblock \emph{Computers and Graphics}, \textbf{1}:pages 3--4, 1975.

\bibitem[Par82]{Park1982}
F.~I. Parke.
\newblock Parameterized models for facial animation.
\newblock \emph{IEEE Transactions on Computer Graphics and Applications},
  \textbf{2}(9):pages 61--68., 1982.

\bibitem[Par96]{Park1996}
F.~I. Parke.
\newblock \emph{Computer Facial animation.}
\newblock AKPeters, Wellesley, Massachusetts, 1996.

\bibitem[Pen94]{PentMoghStar1994}
A.~Pentland, B.~Moghaddam and T.~Starner.
\newblock View-based and modular eigenspaces for face recognition.
\newblock \emph{In Proc. IEEE Conference on Computer Vision and Pattern
  Recognition (CVPR)}, \textbf{84--91}, 1994.

\bibitem[Pen96]{PeneAtic1996}
P.~S. Penev and J.~J. Atick.
\newblock Local feature analysis: {A} general statistical theory for object
  representation.
\newblock \emph{Network: Computation in Neural Systems}, \textbf{7}(3):pages
  477--500, 1996.

\bibitem[Phi95]{PhilVard1995}
P.~J. Phillips and Y.~Vardi.
\newblock Data driven methods in face recognition.
\newblock \emph{In Proc. International Workshop on Automatic Face and Gesture
  Recognition}, pages 65--70, 1995.

\bibitem[Phi03]{PhilGrotMichBlac+2003}
P.~J. Phillips, P.~Grother, R.~J. Micheals, D.~M. Blackburn, E.~Tabassi and
  J.~M. Bone.
\newblock {FRVT} 2002: Overview and summary.
\newblock \emph{Technical report, {National Institute of Justice}}, 2003.

\bibitem[Pre92]{PresTeukVettFlan1992}
W.~H. Press, S.~A. Teukolsky, W.~T. Vetterling and B.~P. Flannery.
\newblock \emph{Numerical Recipes in {C} : The Art of Scientific Computing.}
\newblock Cambridge University Press, 2nd edition, 1992.

\bibitem[Pri01]{PricGree2001}
J.~R. Price and T.~F. Gee.
\newblock Towards robust face recognition from video.
\newblock \emph{In Proc. Applied Image Pattern Recognition Workshop, Analysis
  and Understanding of Time Varying Imagery}, pages 94--102, 2001.

\bibitem[Pro98]{ProkRied1998}
F.~J. Prokoski and R.~Riedel.
\newblock \emph{BIOMETRICS: Personal Identification in Networked Society},
  chapter Infrared Identification of Faces and Body Parts.
\newblock Kluwer Academic Publishers, 1998.

\bibitem[Pro00]{Prok2000}
F.~Prokoski.
\newblock History, current status, and future of infrared identification.
\newblock \emph{In Proc. IEEE International Workshop on Object Tracking and
  Classification Beyond the Visible Spectrum (OTCBVS)}, pages 5--14, 2000.

\bibitem[Pun04]{PunLee2004}
C.~M. Pun and M.~C. Lee.
\newblock Extraction of shift invariant wavelet features for classification of
  images with different sizes.
\newblock \emph{IEEE Transactions on Pattern Analysis and Machine Intelligence
  (TPAMI)}, \textbf{26}(9):pages 1228--1233, 2004.

\bibitem[Raj98]{RajaMcKeGong1998}
Y.~Raja, S.~J. McKenna and S.~Gong.
\newblock Segmentation and tracking using colour mixture models.
\newblock \emph{In Proc. Asian Conference on Computer Vision (ACCV)}, pages
  607--614, 1998.

\bibitem[Ris78]{Riss1978}
J.~Rissanen.
\newblock Modeling by shortest data description.
\newblock \emph{Automatica}, \textbf{14}:pages 465--471, 1978.

\bibitem[Rom02]{RomdBlanVett2002}
S.~Romdhani, V.~Blanz and T.~Vetter.
\newblock Face identification by fitting a {3D} morphable model using linear
  shape and texture error functions.
\newblock \emph{In Proc. European Conference on Computer Vision (ECCV)}, pages
  3--19, 2002.

\bibitem[Rom03a]{RomdTorrSchoBlak2004}
S.~Romdhani, P.~H.~S. Torr, B.~{Sch\"{o}lkopf} and A.~Blake.
\newblock Efficient face detection by a cascaded reduced support vector
  expansion.
\newblock \emph{Proceedings of the Royal Society, Series A}, \textbf{460}:pages
  3283–--3297, 2003.

\bibitem[Rom03b]{RomdVett2003}
S.~Romdhani and T.~Vetter.
\newblock Efficient, robust and accurate fitting of a {3D} morphable model.
\newblock \emph{In Proc. IEEE International Conference on Computer Vision
  (ICCV)}, pages 59--66, 2003.

\bibitem[Ros03]{RossJain2003}
A.~Ross and A.~K. Jain.
\newblock Information fusion in biometrics.
\newblock \emph{Pattern Recognition Letters}, \textbf{24}(13):pages 2115--2125,
  2003.

\bibitem[Ros05]{RossGovi2005}
A.~Ross and R.~Govindarajan.
\newblock Feature level fusion using hand and face biometrics.
\newblock \emph{In Proc. SPIE Conference on Biometric Technology for Human
  Identification II}, \textbf{5779}:pages 196--204, 2005.

\bibitem[Ros06]{RossNandJain2006}
A.~Ross, K.~Nandakumar and A.~K. Jain.
\newblock \emph{Handbook of Multibiometrics}.
\newblock Springer, New York, USA, 2006.

\bibitem[Row98]{RowlBaluKana1998}
H.~A. Rowley, S.~Baluja and T.~Kanade.
\newblock Neural network-based face detection.
\newblock \emph{IEEE Transactions on Pattern Analysis and Machine
  Intelligence}, \textbf{20}(1):pages 23--38, 1998.

\bibitem[Row01]{RoweSaul2001}
S.~Roweis and L.~K. Saul.
\newblock Nonlinear dimensional reduction by locally linear embedding.
\newblock \emph{Science}, \textbf{290}, 2001.

\bibitem[RR01]{RiklShas2001}
T.~Riklin-Raviv and A.~Shashua.
\newblock The quotient image: Class based re-rendering and recognition with
  varying illuminations.
\newblock \emph{IEEE Transactions on Pattern Analysis and Machine Intelligence
  (TPAMI)}, \textbf{23}(2):pages 219--139, 2001.

\bibitem[Saa06]{SaatTown2006}
Y.~Saatci and C.~Town.
\newblock Cascaded classification of gender and facial expression using active
  appearance models.
\newblock \emph{In Proc. IEEE International Conference on Automatic Face and
  Gesture Recognition (FG)}, pages 393--400, 2006.

\bibitem[Sad04]{SadeKitt2004}
M.~T. Sadeghi and J.~V. Kittler.
\newblock Decision making in the lda space: Generalised gradient direction
  metric.
\newblock \emph{In Proc. IEEE International Conference on Automatic Face and
  Gesture Recognition (FG)}, pages 248--253, 2004.

\bibitem[Sal83]{SaltMcGi1983}
G.~Salton and M.~J. McGill.
\newblock \emph{Introduction to Modern Information Retrieval.}
\newblock McGraw Hill, New York, 1983.

\bibitem[Sch78]{Schw1978}
G.~Schwarz.
\newblock Estimating the dimension of a model.
\newblock \emph{Annals of Statistics}, \textbf{6}:pages 461--464, 1978.

\bibitem[{Sch}99]{SchoSmolMull1999}
B.~{Sch\"{o}lkopf}, A.~Smola and K.~{M\"{u}ller}.
\newblock \emph{Advances in Kernel Methods -- SV Learning}, chapter Kernel
  principal component analysis., pages 327--352.
\newblock MIT Press, Cambridge, MA, 1999.

\bibitem[Sch00]{Schneiderman2000}
H.~Schneiderman.
\newblock A statistical approach to {3D} object detection applied to faces and
  cars.
\newblock Ph.D. thesis, Robotics Institute, Carnegie Mellon University, 2000.

\bibitem[{Sch}02]{SchoSmol2002}
B.~{Sch\"{o}lkopf} and A.~Smola.
\newblock \emph{Learning with kernels.}
\newblock MIT Press, Cambridge, MA, 2002.

\bibitem[Scl98]{SclaIsid1998}
S.~Sclaroff and J.~Isidoro.
\newblock Active blobs.
\newblock \emph{In Proc. IEEE International Conference on Computer Vision
  (ICCV)}, pages 1146--1153, 1998.

\bibitem[Sco03]{ScotCootTayl2003}
I.~M. Scott, T.~F. Cootes and C.~J. Taylor.
\newblock Improving appearance model matching using local image structure.
\newblock \emph{In Proc. International Conference on Information Processing in
  Medical Imaging}, pages 258--269, 2003.

\bibitem[Sel02]{equinoxreport02}
A.~Selinger and D.~Socolinsky.
\newblock Appearance-based facial recognition using visible and thermal
  imagery: A comparative study.
\newblock Technical Report 02-01, Equinox Corporation, 2002.

\bibitem[Sen99]{Seni1999}
A.~W. Senior.
\newblock Face and feature finding for a face recognition system.
\newblock \emph{In Proc. International Conference on Audio and Video-based
  Biometric Person Authentication}, pages 154--159, 1999.

\bibitem[Sha02a]{ShakFishDarr2002}
G.~Shakhnarovich, J.~W. Fisher and T.~Darrel.
\newblock Face recognition from long-term observations.
\newblock \emph{In Proc. European Conference on Computer Vision (ECCV)},
  \textbf{3}:pages 851--868, 2002.

\bibitem[Sha02b]{ShasLeviAvid2002}
A.~Shashua, A.~Levin and S.~Avidan.
\newblock Manifold pursuit -- a new approach to appearance based recognition.
\newblock \emph{In Proc. IAPR International Conference on Pattern Recognition
  (ICPR)}, pages 590--594, 2002.

\bibitem[Sha03]{ShanGaoCaoZhao2003}
S.~Shan, W.~Gao, B.~Cao and D.~Zhao.
\newblock Illumination normalization for robust face recognition against
  varying lighting conditions.
\newblock \emph{In Proc. IEEE International Workshop on Analysis and Modeling
  of Faces and Gestures}, pages 157--164, 2003.

\bibitem[Shi04]{ShimShim2004}
T.~Shimooka and K.~Shimizu.
\newblock Artificial immune system for personal identification with finger vein
  pattern.
\newblock \emph{In Proc. International Conference on Knowledge-Based
  Intelligent Information and Engineering Systems}, pages 511--518, 2004.

\bibitem[Sim04]{SimZhang2004}
T.~Sim and S.~Zhang.
\newblock Exploring face space.
\newblock \emph{In Proc. IEEE Workshop on Face Processing in Video}, page~84,
  2004.

\bibitem[Sin04]{SingGyaoBebi+2004}
S.~Singha, A.~Gyaourovaa, G.~Bebisa and I.~Pavlidis.
\newblock Infrared and visible image fusion for face recognition.
\newblock \emph{In Proc. SPIE Defense and Security Symposium (Biometric
  Technology for Human Identification)}, 2004.

\bibitem[Sir87]{SiroKirb1987}
L.~Sirovich and M.~Kirby.
\newblock Low-dimensional procedure for the characterization of human faces.
\newblock \emph{Journal of Optical Society of America}, \textbf{4}(3):pages
  519--524, 1987.

\bibitem[Siv03]{SiviZiss2003}
J.~Sivic and A.~Zisserman.
\newblock Video {G}oogle: A text retrieval approach to object matching in
  videos.
\newblock \emph{In Proc. IEEE International Conference on Computer Vision
  (ICCV)}, \textbf{2}:pages 1470--1477, 2003.

\bibitem[Siv05]{SiviEverZiss2005}
J.~Sivic, M.~Everingham and A.~Zisserman.
\newblock Person spotting: video shot retrieval for face sets.
\newblock \emph{In Proc. IEEE International Conference on Image and Video
  Retrieval (CIVR)}, pages 226--236, 2005.

\bibitem[Sne89]{SnedCoch1989}
G.~W. Snedecor and W.~G. Cochran.
\newblock \emph{Statistical Methods.}
\newblock Iowa State University Press, Ames, Iowa, 8 edition, 1989.

\bibitem[Soc02]{SocoSeli2002}
D.~Socolinsky and A.~Selinger.
\newblock A comparative analysis of face recognition performance with visible
  and thermal infrared imagery.
\newblock \emph{In Proc. IAPR International Conference on Pattern Recognition
  (ICPR)}, \textbf{4}:pages 217--222, 2002.

\bibitem[Soc03]{SocoSeliNeuh2003}
D.~Socolinsky, A.~Selinger and J.~Neuheisel.
\newblock Face recognition with visible and thermal infrared imagery.
\newblock \emph{Computer Vision and Image Understanding (CVIU)},
  \textbf{91}(1--2):pages 72--114, 2003.

\bibitem[Soc04]{SocoSeli2004}
D.~A. Socolinsky and A.~Selinger.
\newblock Thermal face recognition over time.
\newblock \emph{In Proc. IAPR International Conference on Pattern Recognition
  (ICPR)}, \textbf{4}:pages 187--190, 2004.

\bibitem[Son05]{SongWang2005}
M.~Song and H.~Wang.
\newblock Highly efficient incremental estimation of {G}aussian mixture models
  for online data stream clustering.
\newblock \emph{In Proc. SPIE Conference on Intelligent Computing: Theory And
  Applications}, 2005.

\bibitem[Sri03]{SrivLiu2003}
A.~Srivastana and X.~Liu.
\newblock Statistical hypothesis pruning for recognizing faces from infrared
  images.
\newblock \emph{Image and Vision Computing (IVC)}, \textbf{21}(7):pages
  651--661, 2003.

\bibitem[Ste03]{StenThayTorrCipo2003}
B.~Stenger, A.~Thayananthan, P.~H.~S. Torr and R.~Cipolla.
\newblock Filtering using a tree-based estimator.
\newblock \emph{In Proc. IEEE International Conference on Computer Vision
  (ICCV)}, \textbf{2}:pages 1063--1070, 2003.

\bibitem[Ste06]{SterPnevPoly2006}
A.~Stergiou, A.~Pnevmatikakis and L.~Polymenakos.
\newblock {EBGM} vs. subspace projection for face recognition.
\newblock \emph{In Proc. International Conference on Computer Vision Theory and
  Applications}, 2006.

\bibitem[Sun98]{SungPogg1998}
K.~K. Sung and T.~Poggio.
\newblock Example-based learning for view-based human face detection.
\newblock \emph{IEEE Transactions on Pattern Analysis and Machine Intelligence
  (TPAMI)}, \textbf{20}(1):pages 39--51, 1998.

\bibitem[Sun05]{SunHengJin+2005}
Q.-S. Sun, P.-A. Heng, Z.~Jin and D.~Xia.
\newblock Face recognition based on generalized canonical correlation analysis.
\newblock \emph{In Proc. IEEE International Conference on Intelligent
  Computing}, \textbf{2}:pages 958--967, 2005.

\bibitem[Tak98]{Taka1998}
B.~Tak\'acs.
\newblock Comparing face images using the modified {H}ausdorff distance.
\newblock \emph{Pattern Recognition (PR)}, \textbf{31}(12):pages 1873--1881,
  1998.

\bibitem[Ten00]{TeneSilvLang2000}
J.~B. Tenenbaum, V.~d. Silva and J.~C. Langford.
\newblock A global geometric framework for nonlinear dimensionality reduction.
\newblock \emph{Science}, \textbf{290}(5500):pages 2319--2323, 2000.

\bibitem[Tip99a]{TippBish1999a}
M.~E. Tipping and C.~M. Bishop.
\newblock Mixtures of probabilistic principal component analyzers.
\newblock \emph{Neural Computation}, \textbf{11}(2):pages 443--482, 1999.

\bibitem[Tip99b]{TippBish1999}
M.~E. Tipping and C.~M. Bishop.
\newblock Probabilistic principal component analysis.
\newblock \emph{Journal of the Royal Statistical Society}, \textbf{61}(3):pages
  611--622, 1999.

\bibitem[Tor00]{TorrLoreVila2000}
L.~Torres, L.~Lorente and J.~Vila.
\newblock Automatic face recognition of video sequences using selfeigenfaces.
\newblock \emph{In Proc. IAPR International Symposium on Image/Video
  Communications over Fixed and Mobile Networks}, 2000.

\bibitem[Tos06]{Tosh}
Toshiba.
\newblock Facepass.
\newblock \emph{{\url{www.toshiba.co.jp/mmlab/tech/w31e.htm}}}, 2006.

\bibitem[Tru05]{TruOlagHamm+2005}
L.~Trujillo, G.~Olague, R.~Hammoud and B.~Hernandez.
\newblock Automatic feature localization in thermal images for facial
  expression recognition.
\newblock \emph{In Proc. IEEE International Workshop on Object Tracking and
  Classification Beyond the Visible Spectrum (OTCBVS)}, \textbf{3}:page~14,
  2005.

\bibitem[Tur91a]{TurkPent1991}
M.~Turk and A.~Pentland.
\newblock Eigenfaces for recognition.
\newblock \emph{Journal of Cognitive Neuroscience}, \textbf{3}(1):pages 71--86,
  1991.

\bibitem[Tur91b]{TurkPent1991a}
M.~Turk and A.~Pentland.
\newblock Face recognition using {E}igenfaces.
\newblock \emph{In Proc. IAPR International Conference on Pattern Recognition
  (ICPR)}, pages 586--591, 1991.

\bibitem[Vas98]{VascLipp1998}
N.~Vasconcelos and A.~Lippman.
\newblock Learning mixture hierarchies.
\newblock \emph{Advances in Neural Information Processing Systems (NIPS)},
  pages 606--612, 1998.

\bibitem[Ver99]{VerlChol1999}
P.~Verlinde and G.~Cholet.
\newblock Comparing decision fusion paradigms using {k-NN} based classifiers,
  decision trees and logistic regression in a multi-modal identity verification
  application.
\newblock \emph{In Proc. International Conference on Audio- and Video-Based
  Biometric Person Authentication (AVBPA)}, \textbf{5}(2):pages 188--193, 1999.

\bibitem[Ver03]{VerbVlasKros2003}
J.~J. Verbeek, N.~Vlassis and B.~Kr{\"{o}}se.
\newblock Efficient greedy learning of {G}aussian mixture models.
\newblock \emph{Neural Computation}, \textbf{5}(2):pages 469--485, 2003.

\bibitem[Vet04]{VettRomd2004}
T.~Vetter and S.~Romdhani.
\newblock Face modelling and recognition tutorial (part {II}).
\newblock \emph{Face Recognition Tutorial at IEEE European Conference on
  Computer Vision}, 2004.

\bibitem[Vio04]{ViolJone2004}
P.~Viola and M.~Jones.
\newblock Robust real-time face detection.
\newblock \emph{International Journal of Computer Vision (IJCV)},
  \textbf{57}(2):pages 137--154, 2004.

\bibitem[Vla99]{VlasLika1999}
N.~Vlassis and A.~Likas.
\newblock A kurtosis-based dynamic approach to {G}aussian mixture modeling.
\newblock \emph{IEEE Transactions on Systems, Man, and Cybernetics -- Part A:
  Systems and Humans}, \textbf{24}(9):pages 393--399, 1999.

\bibitem[Wal99a]{WallDowe1999}
C.~S. Wallace and D.~L. Dowe.
\newblock Minimum message length and kolmogorov complexity.
\newblock \emph{Computer Journal}, \textbf{42}(4):pages 270--283, 1999.

\bibitem[Wal99b]{WallBlanVett1999}
C.~Wallraven, V.~Blanz and T.~Vetter.
\newblock {3D} reconstruction of faces - combining stereo with class-based
  knowledge.
\newblock \emph{In Proc. Deutsche Arbeitsgemeinschaft f{\"{u}}r Mustererkennung
  ({DAGM}) Symposium}, pages 405--412, 1999.

\bibitem[Wal01]{WalkMill2001}
T.~C. Walker and R.~K. Miller.
\newblock \emph{Health Care Business Market Research Handbook.}
\newblock Norcross (GA): Richard K. Miller {\&} Associates, Inc., 5th edition,
  2001.

\bibitem[Wan03a]{WangTang2003}
X.~Wang and X.~Tang.
\newblock Unified subspace analysis for face recognition.
\newblock \emph{In Proc. IEEE International Conference on Computer Vision
  (ICCV)}, \textbf{1}:pages 679--686, 2003.

\bibitem[Wan03b]{WangTanJain2003}
Y.~Wang, T.~Tan and A.~K. Jain.
\newblock Combining face and iris biometrics for identity verification.
\newblock \emph{In Proc. International Conference on Audio- and Video-based
  Biometric Person Authentication (AVBPA)}, pages 805--813, 2003.

\bibitem[Wan04a]{WangLiWang2004}
H.~Wang, S.~Z. Li and Y.~Wang.
\newblock Face recognition under varying lighting conditions using self
  quotient image.
\newblock \emph{In Proc. IEEE International Conference on Automatic Face and
  Gesture Recognition (FG)}, pages 819--824, 2004.

\bibitem[Wan04b]{WangSungVenk2004}
J.~Wang, E.~Sung and R.~Venkateswarlu.
\newblock Registration of infrared and visible-spectrum imagery for face
  recognition.
\newblock \emph{In Proc. IEEE International Conference on Automatic Face and
  Gesture Recognition (FG)}, pages 638--644, 2004.

\bibitem[Wan04c]{WangTang2004}
X.~Wang and X.~Tang.
\newblock Random sampling {LDA} for face recognition.
\newblock \emph{In Proc. IEEE Conference on Computer Vision and Pattern
  Recognition (CVPR)}, pages 259--265, 2004.

\bibitem[Wan05]{WangJiaHu+2005}
Y.~Wang, Y.~Jia, C.~Hu and M.~Turk.
\newblock Non-negative matrix factorization framework for face recognition.
\newblock \emph{International Journal of Pattern Recognition and Artificial
  Intelligence}, \textbf{19}(4):pages 495--511, 2005.

\bibitem[Was89]{Wass1989}
A.~I. Wasserman.
\newblock \emph{Neural Computing: Theory and Practice.}
\newblock Van Nostrand Reinhold, New York, 1989.

\bibitem[Wen93]{Weng1993}
J.~Weng, N.~Ahuja and T.~S. Huang.
\newblock Learning recognition and segmentation of {3-D} objects from {2-D}
  images.
\newblock \emph{In Proc. IEEE International Conference on Computer Vision
  (ICCV)}, pages 121--128, 1993.

\bibitem[Wil94]{WildAsmuGreeHsu1994}
R.~P. Wildes, J.~C. Asmuth, G.~L. Green, S.~C. Hsu, R.~Kolczynski, J.~Matey and
  S.~McBride.
\newblock A system for automated iris recognition.
\newblock \emph{In Proc. IEEE Workshop on Applications of Computer Vision},
  pages 121--128, 1994.

\bibitem[Wil04]{WillBlakCipo2004}
O.~Williams, A.~Blake and R.~Cipolla.
\newblock The variational ising classifier {(VIC)} algorithm for coherently
  contaminated data.
\newblock \emph{Advances in Neural Information Processing Systems (NIPS)},
  pages 1497–--1504, 2004.

\bibitem[Wis97]{WiskFellKrug+1997}
L.~Wiskott, J.-M. Fellous, N.~Kr{\"u}ger and C.~von~der Malsburg.
\newblock Face recognition by elastic bunch graph matching.
\newblock \emph{IEEE Transactions on Pattern Analysis and Machine
  Intelligence}, \textbf{19}(7):pages 775--779, 1997.

\bibitem[Wis99a]{WiskFell1999}
L.~Wiskott and J.-M. Fellous.
\newblock \emph{Intelligent Biometric Techniques in Fingerprint and Face
  Recognition}, chapter Face Recognition by Elastic Bunch Graph Matching.,
  pages 355--396.
\newblock 1999.

\bibitem[Wis99b]{WiskFellKrug+1999}
L.~Wiskott, J.-M. Fellous, N.~Kr{\"u}ger and C.~von~der Malsburg.
\newblock Face recognition by elastic bunch graph matching.
\newblock \emph{Intelligent Biometric Techniques in Fingerprint and Face
  Recognition}, pages 355--396, 1999.

\bibitem[Wol01]{WolfSocoEvel2001}
L.~B. Wolff, D.~A. Socolinsky and C.~K. Eveland.
\newblock Quantitative measurement of illumination invariance for face
  recognition using thermal infrared imagery.
\newblock \emph{In Proc. IEEE International Workshop on Object Tracking and
  Classification Beyond the Visible Spectrum (OTCBVS)}, 2001.

\bibitem[Wol03]{WolfShas2003}
L.~Wolf and A.~Shashua.
\newblock Learning over sets using kernel principal angles.
\newblock \emph{Journal of Machine Learning Research (JMLR)},
  \textbf{4}(10):pages 913--931, 2003.

\bibitem[Wu04]{WuAiHuanLao2004}
B.~Wu, H.~Ai, C.~Huang and S.~Lao.
\newblock Fast rotation invariant multi-view face detection based on real
  adaboost.
\newblock \emph{In Proc. IEEE International Conference on Automatic Face and
  Gesture Recognition (FG)}, pages 79--84, 2004.

\bibitem[Yam98]{YamaFukuMaed1998}
O.~Yamaguchi, K.~Fukui and K.~Maeda.
\newblock Face recognition using temporal image sequence.
\newblock \emph{In Proc. IEEE International Conference on Automatic Face and
  Gesture Recognition (FG)}, (10):pages 318--323, 1998.

\bibitem[Yam00]{Yamb2000}
W.~S. Yambor.
\newblock Analysis of {PCA}-based and fisher discriminant-based image
  recognition algorithms.
\newblock Master's thesis, Colorado State University, 2000.

\bibitem[Yan00]{YangAhujKrieg2000}
M.-H. Yang, N.~Ahuja and D.~Kriegman.
\newblock Face recognition using kernel eigenfaces.
\newblock \emph{In Proc. IEEE International Conference on Image Processing
  (ICIP)}, \textbf{1}:pages 37--40, 2000.

\bibitem[Yan02a]{Yang2002a}
M.-H. Yang.
\newblock Face recognition using extended {Isomap}.
\newblock \emph{In Proc. IEEE International Conference on Image Processing
  (ICIP)}, \textbf{2}:pages 117--120, 2002.

\bibitem[Yan02b]{Yang2002}
M.-H. Yang.
\newblock Kernel eigenfaces vs. kernel fisherfaces: Face recognition using
  kernel methods.
\newblock \emph{In Proc. IEEE International Conference on Automatic Face and
  Gesture Recognition (FG)}, pages 215--220, 2002.

\bibitem[Yan02c]{YangAhujKrieg2002}
M.-H. Yang, N.~Ahuja and D.~Kriegman.
\newblock A survey on face detection methods.
\newblock \emph{IEEE Transactions on Pattern Analysis and Machine Intelligence
  (TPAMI)}, \textbf{24}(1):pages 34--58, 2002.

\bibitem[Yan05]{YangGaoZhan+2005}
J.~Yang, X.~Gao, D.~Zhang and J.~Yang.
\newblock Kernel {ICA}: An alternative formulation and its application to face
  recognition.
\newblock \emph{Pattern Recognition (PR)}, \textbf{38}(10):pages 1784--1787,
  2005.

\bibitem[Yos99]{YoshTana1999}
S.~Yoshizawa and K.~Tanabe.
\newblock Dual differential geometry associated with {K}ullback-{L}eibler
  information on the {G}aussian distributions and its 2-parameter deformations.
\newblock \emph{Science University of Tokyo Journal of Mathematics},
  \textbf{35}(1):pages 113--137, 1999.

\bibitem[Zab95]{ZabiMillMai1995}
R.~Zabih, J.~Miller and K.~Mai.
\newblock A feature-based algorithm for detecting and classifying scene breaks.
\newblock \emph{In Proc. {ACM} International Conference on Multimedia}, pages
  189--200, 1995.

\bibitem[Zha93]{ZhanKankSmol1993}
H.~J. Zhang, A.~Kankanhalli and S.~Smoliar.
\newblock Automatic partitioning of full-motion video.
\newblock \emph{Multimedia Systems}, \textbf{1}(1):pages 10--28, 1993.

\bibitem[Zha98]{ZhaoChelKris1998}
W.~Zhao, R.~Chellappa and A.~Krishnaswamy.
\newblock Discriminant analysis of principal components for face recognition.
\newblock \emph{In Proc. IEEE International Conference on Automatic Face and
  Gesture Recognition (FG)}, pages 336--341, 1998.

\bibitem[Zha00]{ZhaoChelRosePhil2000}
W.~Zhao, R.~Chellappa, A.~Rosenfeld and P.~J. Phillips.
\newblock Face recognition: A literature survey.
\newblock \emph{UMD CFAR Tech. Report CAR-TR-948}, 2000.

\bibitem[Zha04]{ZhaoChelPhilRose2004}
W.~Zhao, R.~Chellappa, P.~J. Phillips and A.~Rosenfeld.
\newblock Face recognition: A literature survey.
\newblock \emph{ACM Computing Surveys}, \textbf{35}(4):pages 399--458, 2004.

\bibitem[Zho03]{ZhouKrueChel2003}
S.~Zhou, V.~Krueger and R.~Chellappa.
\newblock Probabilistic recognition of human faces from video.
\newblock \emph{Computer Vision and Image Understanding}, \textbf{91}(1):pages
  214--245, 2003.

\bibitem[Zwo01]{ZwolYang2001}
M.~Zwolinski and Z.~R. Yang.
\newblock Mutual information theory for adaptive mixture models.
\newblock \emph{IEEE Transactions on Pattern Analysis and Machine Intelligence
  (TPAMI)}, \textbf{23}(4):pages 396--403, 2001.

\end{thebibliography}
\end{spacing}
\end{document}